\documentclass[10pt,twocolumn,letterpaper]{article}

\usepackage[wacv,algorithms]{wacv}      

\usepackage[utf8]{inputenc} 
\usepackage[T1]{fontenc}    
\usepackage{amsfonts}       
\usepackage{nicefrac}       
\usepackage{microtype}      
\usepackage{tabularx}
\usepackage{multicol}
\usepackage{multirow}
\usepackage{makecell}
\usepackage{longtable}
\usepackage{tabu}
\usepackage{supertabular}
\usepackage{enumitem}
\usepackage{etoolbox}
\usepackage{lipsum}
\usepackage{cuted} 
\usepackage{placeins} 

\usepackage[accsupp]{axessibility}

\newcommand{\Ours}{\textsc{BoxSplitGen}}
\newcommand{\Veps}{\boldsymbol{\epsilon}}

\newcommand{\B}{\mathbf}
\newcommand{\C}{\mathcal}
\newcolumntype{Y}{>{\centering\arraybackslash}X}
\newcolumntype{Z}{>{\scriptsize\centering\arraybackslash}X}

\definecolor{wacvblue}{rgb}{0.21,0.49,0.74}
\usepackage[pagebackref,breaklinks,colorlinks,allcolors=wacvblue]{hyperref}

\def\wacvPaperID{2785} 
\def\confName{WACV}
\def\confYear{2026}

\title{\Ours: A Generative Model for 3D Part Bounding Boxes in Varying Granularity}

\author{
Juil Koo\textsuperscript{$1\ast$} $\quad$
Wei-Tung Lin\textsuperscript{$2\ast$} $\quad$
Chanho Park\textsuperscript{1}$\quad$
Chanhyeok Park\textsuperscript{1} $\quad$
Minhyuk Sung\textsuperscript{1} \\
\textsuperscript{1}KAIST$\quad$
\textsuperscript{2}NVIDIA\\
{\tt\small \{63days,charlieppark,chpark1111,mhsung\}@kaist.ac.kr
$\quad$
weitungl@nvidia.com}
}

\begin{document}
\maketitle
\def\thefootnote{*}\footnotetext{Equal contribution.}\def\thefootnote{\arabic{footnote}}

\newif\ifarxiv
\arxivtrue

\newcommand{\projecturl}{\href{https://boxsplitgen.github.io}{Project page}}

\makeatletter
\newcommand{\manuallabel}[2]{\def\@currentlabel{#2}\label{#1}}
\makeatother
\manuallabel{sec:implementation_details}{S.7}
\manuallabel{sec:interactive_demo}{S.2}
\manuallabel{sec:evaluation_box_splitting}{S.4}
\manuallabel{sec:other_options}{S.6}
\manuallabel{sec:suppl_more_box_splitting_qualitative_results}{S.12}
\manuallabel{sec:evaluation_shape_generation}{S.5}
\manuallabel{sec:suppl_more_shape_generation_qualitative_results}{S.13}
\manuallabel{tab:datasets_stats}{S2}

\begingroup
\setlength{\abovecaptionskip}{4pt}
\setlength{\belowcaptionskip}{4pt}
\begin{strip}
  \centering
  \includegraphics[width=\textwidth,height=0.27\textheight,keepaspectratio]{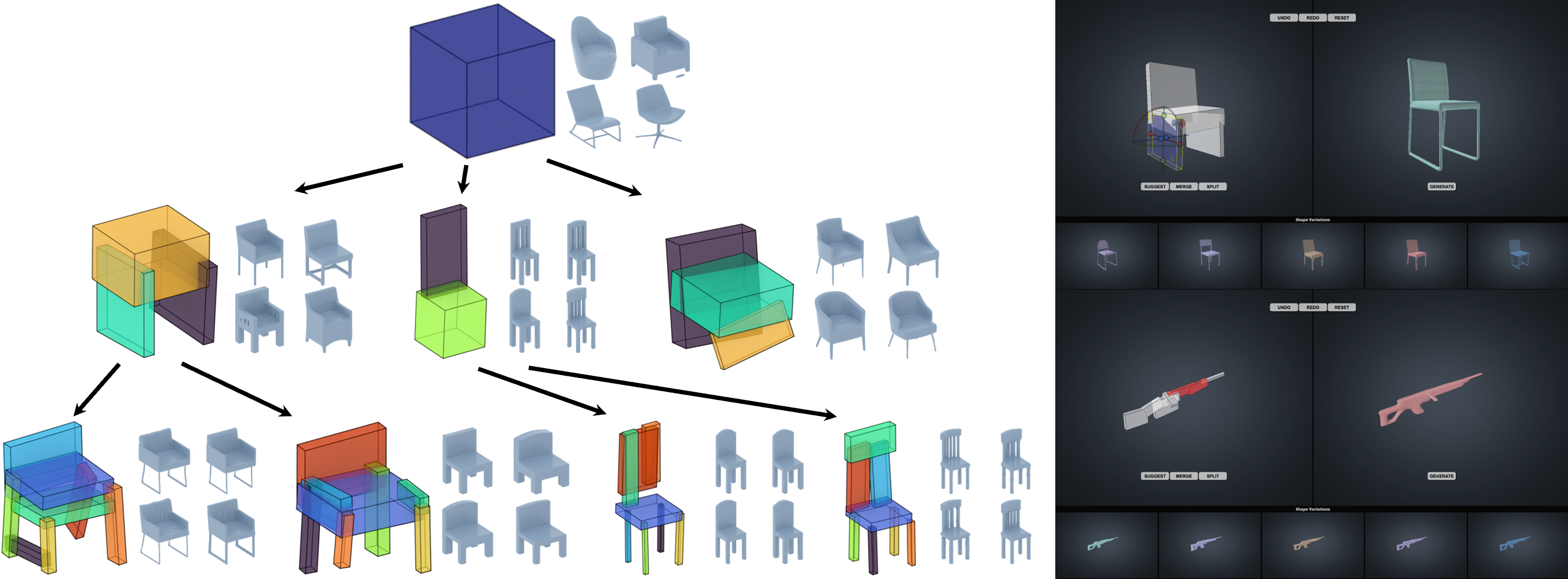}
  \captionof{figure}{\textbf{An overview of box-splitting-based 3D shape generative framework.}
  The left shows our iterative box splitting and box-to-shape generation, where diverse shapes at the top of the tree become increasingly specific deeper in the tree. The right showcases our user-interactive box and shape editing demo.}
  \label{fig:teaser}
  \vspace{0.1em} 
\end{strip}
\endgroup

\begin{abstract}
Human creativity follows a perceptual process, moving from abstract ideas to finer details during creation. While 3D generative models have advanced dramatically, models specifically designed to assist human imagination in 3D creation--particularly for detailing abstractions from coarse to fine--have not been explored. We propose a framework that enables intuitive and interactive 3D shape generation by iteratively splitting bounding boxes to refine the set of bounding boxes. The main technical components of our framework are two generative models: the box-splitting generative model and the box-to-shape generative model. The first model, named \Ours{}, generates a collection of 3D part bounding boxes with varying granularity by iteratively splitting coarse bounding boxes. It utilizes part bounding boxes created through agglomerative merging and learns the reverse of the merging process—the splitting sequences. The model consists of two main components: the first learns the categorical distribution of the box to be split, and the second learns the distribution of the two new boxes, given the set of boxes and the indication of which box to split. The second model, the box-to-shape generative model, is trained by leveraging the 3D shape priors learned by an existing 3D diffusion model while adapting the model to incorporate bounding box conditioning. In our experiments, we demonstrate that the box-splitting generative model outperforms token prediction models and the inpainting approach with an unconditional diffusion model. Also, we show that our box-to-shape model, based on a state-of-the-art 3D diffusion model, provides superior results compared to a previous model. Our project page is \href{https://boxsplitgen.github.io}{https://boxsplitgen.github.io}.
\end{abstract}

\section{Introduction}
\label{sec:intro}
In recent years, significant progress has been made in 3D generative models, promising production-level quality in the near future. This success is largely driven by the rise of diffusion models and their applications across various 3D shape representations, including latent representations~\cite{Jun:2023Shap-E, Chou:2022Diffusion-sdf, Zhang:2023Shape2Vec, Zhang:2024Clay}, point clouds~\cite{Nichol:2022PointE}, voxels~\cite{Li:2023DiffusionSDF, Xiong:2024OctFusion}, meshes~\cite{Liu:2023MeshDiffusion, Alliegro:2023PolyDiff}, B-reps~\cite{Xu:2024BrepGen}, and Gaussian splats~\cite{Roessle:2024L3DG}. As we witness these remarkable advancements in 3D generation, the next frontier lies in enhancing the \emph{controllability} of the generation process.

For conditional generation, text prompts have become the most popular conditioning input for both 2D image~\cite{Cao:2023MasaCtrl, Geyer:2023TokenFlow, Hertz:2022Prompt, Patashnik:2023Localizing, Tumanyan:2023PaP, Wu:2023Tune, Zhang:2023ControlNet} and 3D shape generation~\cite{Zibo:2023Michelangelo, Sella:2023SpicE, Jun:2023Shap-E, Nam:2022LDM}. However, they offer limited controllability, particularly when it comes to spatial guidance. As an alternative, bounding box grounding has been explored for both 2D~\cite{Li:2023GLIGEN, Lee:2024ReGround} and 3D guided generation~\cite{Sella:2023SpicE}, offering notable benefits, such as ease of manipulation by users and its effectiveness in abstracting shapes and providing spatial guidance.

Particularly for 3D shapes, the key benefit of primitive-based abstraction lies in its ability to represent part-level structures~\cite{Kai:2008BoundingBox, Gottschalk:1996OBBTreeAH, Li:2017GRASS, Sun:2019Abstraction, Yang:2021CubSeg, Kim:2023SegStruct, Park:2024SMART} while effectively encoding \emph{hierarchical} relationships across parts~\cite{Sun:2019Abstraction, Li:2017GRASS, Paschalidou:2020UnsupervisedHierarchicalPart, Park:2024SMART}. Notably, foundational studies from the 1970s and 1980s~\cite{Binford:1971,Marr:1978,Hove:1988} revealed that the human visual system perceives 3D objects as hierarchically organized sets of primitives, progressing systematically from coarse to fine granularity. This hierarchical framework resonates with human creativity, where the imaginative process similarly unfolds through structured, incremental levels of detail.

Building on these studies, we propose a user-interactive 3D generation framework that reflects the hierarchical nature of human imagination. Figure~\ref{fig:teaser} and our \href{https://boxsplitgen.github.io}{project page} showcase user-interactive generation examples using part bounding boxes. Our framework enables users to create 3D objects starting with a rough design represented by coarse bounding boxes, progressively adding more detail by splitting a bounding box into smaller, finer-grained boxes. This process allows users to explore diverse shapes using coarse bounding boxes (intermediate nodes in the tree shown in Figure~\ref{fig:teaser}) while specifying the design by increasing the level of granularity (bottom nodes). The iterative approach helps users efficiently create detailed and desired 3D shapes with minimal effort. In the interface shown on the right in Figure~\ref{fig:teaser} and our \href{https://boxsplitgen.github.io}{project page}, users can transform bounding boxes and adjust their granularity by splitting them into finer components.

Creating such a framework requires two generative models: a box-splitting generative model and a box-to-shape generative model. The purpose of the first model is to generate two finer-grained bounding boxes by splitting a coarse bounding box. Training this generative model necessitates a dataset containing sequences of box-splitting results. While extensive research has been conducted on extracting part-level primitives from raw 3D shapes~\cite{Tulsiani:2017VolumetricPrimitives, Chen:2020BspNet, Paschalidou:2021NeuralParts, Yang:2021CubSeg, Deng:2020Cvxnet, Paschalidou:2019SuperQuadrics}, most previous methods are unable to generate bounding boxes at various granularity or capture their hierarchical relationships. Some prior works have introduced hand-crafted datasets~\cite{Mo:2019structurenet}, but these are limited in scalability and diversity across different shape categories. To address the lack of training data, we focus on leveraging a recent method~\cite{Park:2024SMART} that generates bounding boxes for parts through hierarchical merging, starting from super-segments. This agglomerative merging approach naturally produces a set of bounding boxes with varying levels of granularity, down to a single bounding box, along with their two-to-one merging relationships. Our goal is to learn the reverse process of merging.

We model the reverse process of hierarchical merging, iterative splitting, using a generative model named \Ours{}. While iterative splitting can be viewed as a sequential process, it has unique characteristics that make it incompatible with typical sequence generation models, such as GPT~\cite{radford2019language,Brown:2020GPT3}. First, the generation of the next token (bounding box) is conditioned not only on the previous tokens but also on the selection of the box to be split. Second, and more importantly, the selected box is removed after splitting. Consequently, the set of bounding boxes at an intermediate step is not a subset of those at the final step, making GPT-like models unsuitable for this task.
To address this, we propose a two-step autoregressive model that leverages a classifier and a diffusion model. In the first step, we learn the distribution of boxes to be selected for splitting using a classification network. In the second step, we use a conditional diffusion model, where the given set of boxes, along with an indication of the box to split, is used as conditioning input to generate the two split boxes.

The second generative model in our framework is a bounding-box-conditioned 3D shape generative model. To achieve this, we finetune 3DShape2VecSet~\cite{Zhang:2023Shape2Vec} to incorporate bounding box conditioning using the ControlNet~\cite{Zhang:2023ControlNet} approach, while preserving the high-quality 3D shape priors learned by 3DShape2VecSet.
ControlNet requires the representations of the denoised data and the input condition to be in the same format. Previous work~\cite{Sella:2023SpicE}, which uses Shape-E~\cite{Jun:2023Shap-E} as a base 3D diffusion model, addresses this by converting bounding box representations into multi-view images and processing them with a pretrained encoder. In contrast, we propose a simpler yet effective approach: directly encoding the bounding boxes into the latent representation using a learnable encoding layer, which is jointly trained with the ControlNet branch.



In our experiments on the ShapeNet~\cite{shapenet2015} dataset, we compare our method with several baselines for both box-splitting generation and bounding-box-conditioned shape generation. For box-splitting, we evaluate against a token prediction model and an inpainting approach using an unconditional diffusion model that preserves existing boxes while filling two new ones. Our conditional diffusion model achieves the best performance, while the inpainting baseline shows comparable but slightly inferior results. For bounding-box-conditioned shape generation, we compare with Spice-E~\cite{Sella:2023SpicE}, which uses Shape-E~\cite{Sella:2023SpicE} instead of our 3DShape2VecSet, and with a finetuned version of 3DShape2VecSet~\cite{Zhang:2023Shape2Vec} that replaces ControlNet with a Gated Mechanism. Our model outperforms these alternatives in both the quality of generated shapes and their alignment with the input bounding boxes.

\section{Related Work}
\label{sec:related_work}

\paragraph{3D Shape Abstraction Methods}
\label{sec3DA}
There is a substantial body of prior work focused on abstracting raw 3D shapes into collections of basic shape representations, such as cuboids~\cite{Tulsiani:2017VolumetricPrimitives, Sun:2019HierarchicalAbstractions, Yang:2021CubSeg, Kim:2023SegStruct, Park:2024SMART}, superquadrics~\cite{Paschalidou:2019SuperQuadrics}, primitives~\cite{Li:2011GlobFit, Zhou:2015GCD, Li:2019:SPFN, Le:2021CPFN}, and implicit fields~\cite{Paschalidou:2021NeuralParts, Chen:2020BspNet, Deng:2020Cvxnet, Chen:2019BaeNet, Niu:2022RimNet}. While these methods effectively extract primitives that represent parts of a given 3D shape, most do not capture the hierarchical relationships across these parts. However, a few approaches have been proposed that jointly extract both the primitives and their hierarchical structures. Sun~\etal~\cite{Sun:2019HierarchicalAbstractions} present a method for extracting cuboids in a hierarchical structure while enforcing inclusion relationships across levels. However, the method is limited to three levels of hierarchy, which restricts the range of granularities from coarse to fine. In contrast, SMART~\cite{Park:2024SMART} offers an iterative approach for agglomeratively merging super-segments into more abstract bounding boxes. Since SMART starts with fine-level super-segments, it allows for a much wider range of granularities. Thus, as training data for our generative model, we use the outputs of SMART while progressively merging bounding boxes until a single box remains. Our generative framework learns the reverse process of this iterative merging: splitting a bounding box into two.

\paragraph{Shape Structure Generative Models}
\label{secSGM}
Generative models for 3D shape structures have been explored in previous work, but they often face limitations in scalability or in controlling the number of primitives and the granularity of the abstraction. StructureNet~\cite{Mo:2019structurenet} and GRAINS~\cite{Li:2018GRAINS} are methods for generating sets of part-level bounding boxes for individual objects and object-level bounding boxes for scenes, respectively. Both approaches use recursive neural networks to train variational autoencoders (VAE) for learning N-ary hierarchies. However, these methods rely heavily on annotated datasets such as PartNet~\cite{Mo:2019PartNet} and SUNCG~\cite{Song:2016SUNCG}, which limit their scalability. SPAGHETTI~\cite{Hertz:2022Spaghetti}, SALAD~\cite{Koo:2023Salad}, and DiffFacto~\cite{Kiyohiro:2023DiffFacto} represent another line of work that does not require hand-crafted datasets for training. Instead, these methods generate part-level structures by learning from data obtained in an unsupervised manner~\cite{Koo:2023Salad}, or even while jointly learning the part-level structure~\cite{Hertz:2022Spaghetti, Kiyohiro:2023DiffFacto}. However, these methods are limited in their ability to vary the number of parts. GRASS~\cite{Li:2017GRASS}, SAGNet~\cite{Wu:2019SAGNet}, DSG-Net~\cite{Yang:2022DSG-Net}, and PASTA~\cite{Li:2024PASTA} are notable examples that generate varying numbers of part bounding boxes without the need for hand-crafted datasets. However, these methods do not allow for adjusting the granularity of the part-level representations. To the best of our knowledge, we are the first to propose a generative model for 3D part bounding boxes that allows for controlling the granularity.

\paragraph{Structure-Conditioned Shape Generative Models}
\label{secSCS}
Structural 3D shape abstraction has been used as user-controllable guidance in 3D object generation. For instance, Neural Template~\cite{Hui:2022} utilizes part-wise latent codes as conditions for 3D shape generation. SALAD~\cite{Koo:2023Salad} is a conditional diffusion model that takes a set of Gaussian blobs representing parts as input conditions. Note that the aforementioned methods do not allow for changes in the granularity of the abstract representation, unlike our approach.
Spice-E~\cite{Sella:2023SpicE} is the most recent generative model similar to ours, using a set of bounding boxes as a condition. It employs Shape-E~\cite{Jun:2023Shap-E} as a base unconditional 3D diffusion model and leverages the ControlNet~\cite{Zhang:2023ControlNet} approach to transform it into a conditional model that incorporates bounding boxes. Similar to Spice-E, we propose a bounding-box-to-shape generative model but build it on a more advanced 3D diffusion model, 3DShape2VecSet~\cite{Zhang:2023Shape2Vec}. While Spice-E represents bounding boxes as multi-view images and encodes them using Shape-E’s pretrained encoder, we propose a simpler yet effective approach: introducing an encoder layer that directly maps input bounding boxes to the latent representation of 3DShape2VecSet, training it jointly with the ControlNet branch. Our experiments demonstrate that this novel box-to-shape generative model produces higher-quality 3D shapes than Spice-E, benefiting from the superior generative capabilities of 3DShape2VecSet.
\begin{figure*}[t!]
\centering
{
\includegraphics[width=\linewidth]{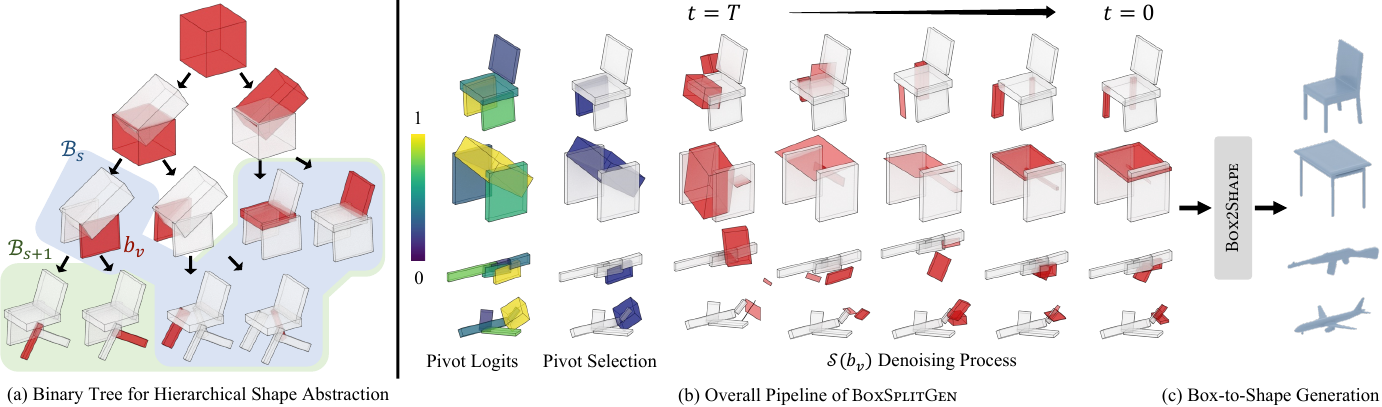}
}
\caption{\textbf{Overview of our hierarchical bounding box splitting framework.} On the left is a binary tree for 3D shape abstraction, where red-highlighted nodes $b_v$ are split into finer child nodes, with blue and green backgrounds showing split steps at $s$ and $s+1$. On the right, the framework performs pivot classification, samples two red-highlighted child boxes, and generates 3D shapes using our \textsc{Box2Shape} model.}
\label{fig:tree}
\end{figure*}

\section{\Ours{}: Box-Splitting Generative Model}
\label{sec:method}

\subsection{Problem Definition and Overview}
\label{sec:overview}
Our objective is to learn a generative model for sets of 3D bounding boxes as shape abstractions, which capture both a diverse collection of shapes and varying levels of granularity for each shape. To achieve this, we represent an arbitrary 3D shape using hierarchical shape abstractions~\cite{Park:2024SMART} structured as a \emph{binary tree}, as illustrated in Figure~\ref{fig:tree}. The root node of the binary tree is a single unit cube that completely encloses any arbitrary shape. As we traverse deeper into the tree, each internal node splits into two child nodes through \emph{splitting operation}. This operation refines the abstraction by subdividing a chosen box, called the pivot $b_v$, into two child boxes denoted by $\C{C}(b_v)$. Formally, at a given split step $s$ in the tree structure, we have a set of 3D bounding boxes $\C{B}_s = \{b_i\}_{i=1}^N$. Each split step can be expressed as $\C{B}_{s+1} := \C{B}_s \setminus \{b_v\} \cup \C{C}(b_v)$, 
where the coarser pivot box is replaced with two newly generated child boxes with finer details (See Figure~\ref{fig:tree}).
This split process provides progressively more detailed approximations of the input shape. At the finest level of detail, a collection of the leaf nodes in the tree represents the most detailed shape abstraction. This hierarchical representation thus can provide diverse shape abstractions at any level of granularity, from a single coarse cube at the root to a detailed collection of boxes at the leaves. 

Given this tree structure, we design the generative process of the bounding box sets $\C{B}_s$ at each split step $s$ as a Markov process with conditional probability distributions $p(\mathcal{B}_{s+1} | \mathcal{B}_s)$, which models the splitting operation. The conditional probability distribution $p(\mathcal{B}_{s+1} | \mathcal{B}_s)$ is further decomposed into two distributions: 1) the distribution of the pivot box, and 2) the distribution of the two child boxes from the selected pivot box:
\begin{align}
    p(\C{B}_{s+1} | \C{B}_s) = p(b_v | \mathcal{B}_s ) p(\C{C}(b_v) | b_v, \C{B}_s),
    \label{eq:conditional_distribution}
\end{align}
where $b_v \in \mathcal{B}_s$ is one of the bounding boxes in $\mathcal{B}_s$ selected as the pivot. In our method, named~\Ours{}, we learn the probability distribution of the pivot bounding box $p(b_v | \mathcal{B}_s )$ using a classification network, while learning the probability distribution of the two child boxes $p(\C{C}(b_v) | b_v, \C{B}_s)$ using a diffusion model.

Note that while our generative process is autoregressive, it cannot be modeled using typical sequence generation models, such as GPT-based architectures (as utilized in recent mesh generation works~\cite{Nash:2020Polygen,siddiqui2024meshgpt,tang2024edgerunner}), for the following reasons. First, the generation of the next token (bounding box) is conditioned on the selection of a pivot box, which necessitates the use of a pivot classifier. Second, the pivot box is removed in the subsequent step, resulting in a binary tree at intermediate granularity that is not a subtree of the binary tree at the finest granularity. As a result, binary trees at arbitrary granularity cannot be generated simply by sequentializing the finest granularity binary tree and feeding it into GPT.

Below, we first describe the data preparation process for training (Section~\ref{sec:data_preparation}). Next, we explain how the two models in our framework, pivot classifier and child-boxes diffusion model, are implemented (Sections~\ref{sec:pivot_classifer} and~\ref{sec:child_box_diffusion}).

\subsection{Data Preparation}
\label{sec:data_preparation}
Our training data consists of hierarchical shape abstractions generated using SMART~\cite{Park:2024SMART}. Given an initial set of over-segmented bounding boxes $\mathcal{B}_S = \{b_i\}_{i=1}^N$, SMART iteratively performs bottom-up merging. In each iteration, it selects two boxes and merges them into a single parent box that tightly encloses the combined region. This parent box serves as a pivot $b_v$ in our hierarchy, with the two merged boxes becoming its children $\mathcal{C}(b_v)$.
SMART continues the merging process until reaching to an appropriate number of bounding boxes that best describe the given shape. These boxes are used as the leaf nodes in our binary tree;
refer to Table~\ref{tab:datasets_stats} for the statistics on the number of leaf node boxes. Based on the leaf node boxes (SMART outputs), we further proceed with an iterative merging process until only a single box remains, building the binary tree up to the root. We set the root box to always be a unit cube; all raw shapes are normalized to fit within the unit cube.
Each 3D bounding box $b_i= \{c_i, s_i, o_i\} \in \mathbb{R}^{15}$ is parameterized by center $c_i \in \mathbb{R}^3$, side lengths $s_i \in \mathbb{R}^3$, and a flatten orientation matrix $o_i \in \mathbb{R}^{9}$.

\subsection{Pivot Classifier}
\label{sec:pivot_classifer}
The first component of \Ours{} is the pivot classifier, which models the categorical distribution of the pivot bounding box given set of the bounding boxes ($p(b_v | \mathcal{B}_s )$ in Equation~\ref{eq:conditional_distribution}).
We adopt a Transformer-based architecture~\cite{Peebles:2023DiT} to handle variable-sized input sets, where each box is encoded as a token in the latent space and processed through self-attention layers~\cite{Vaswani:2017Attention}.

\subsection{Child-Boxes Diffusion}
\label{sec:child_box_diffusion}
The second component of \Ours{} is the child-boxes diffusion model, a conditional diffusion model that learns the distribution of the two child bounding boxes $\C{C}(b_v)$ given the input bounding boxes $\C{B}_s$ and the selected pivot $b_v$ ($p(\C{C}(b_v) | b_v, \C{B}_s)$ in Equation~\ref{eq:conditional_distribution}). The noise prediction network $\Veps_\theta$ of the diffusion model is trained with the condition-output pairs extracted from consecutive box sets $\mathcal{B}_s$ and $\mathcal{B}_{s+1}$ in the training dataset:
\begin{align}
    \B{x}_0 &= \C{C}(b_v) \in \mathbb{R}^{2\times 15} \\
    \C{L} &= \mathbb{E} \Vert \Veps_\theta(\B{x}_t,t;b_v, \C{B}_s) - \Veps  \Vert, \quad \Veps \sim \C{N}(\B{0}, \B{I}),
    \label{eq:noise_prediction}
\end{align}
where $\B{x}_t$ is obtained through the forward process of diffusion models: $ \B{x}_t = \sqrt{\bar\alpha_t} \B{x}_0 + \sqrt{1 - \bar\alpha_t}\Veps$. The network of $\Veps_\theta$ consists of a Transformer encoder $\C{E}_\theta$ taking the input condition $\C{B}_s$ and $b_v$, and a decoder $\C{D}_\theta$ predicting the output noise. In the encoder $\C{E}_\theta$, 
the input bounding boxes $\C{B}_s$ are augmented with an indicator bit marking the pivot box $b_v$, and then processed through self-attention layers to produce a latent vector $\B{h} = \C{E}(\C{B}_s, b_v, | \C{B}_s |) \in \mathbb{R}^{| \C{B}_s | \times D}$; the number of input boxes $| \C{B}_s |$ is also fed as an additional input. The decoder $\C{D}_\theta$ then predicts the injected noise for the two noisy boxes $\B{x}_t \in \mathbb{R}^{2 \times 15}$ through the cross-attention between $\B{x}_t$ and $\B{h}$. See Figure~\ref{fig:second_stage_network} (a) for details of the architecture. More implementation details are provided in Appendix~\ref{sec:implementation_details} and our \href{https://boxsplitgen.github.io}{project page}.

\subsection{Alternative: Inpainting with Unconditional Model}
The conditional generation task performed by our Child-Boxes Diffusion can also be seen as a \emph{completion} task, where missing parts are generated while keeping the given parts fixed. Previous work~\cite{lugmayr2022repaint} has demonstrated that the completion task (inpainting for images) can also be achieved using an unconditional diffusion model by combining one-step denoising outputs for the missing parts with forward process outputs for the given parts at each denoising step. We also explore this option by training another diffusion model that uses only the number of bounding boxes as its only condition. The details of this alternative approach are discussed in Section~\ref{sec:stage_1_baselines}, where we evaluate it as one of the baselines. While this approach shows comparable performance, it yields slightly inferior results compared to the conditional model.

\begin{figure*}[t!]
    \centering
    \includegraphics[width=\linewidth]{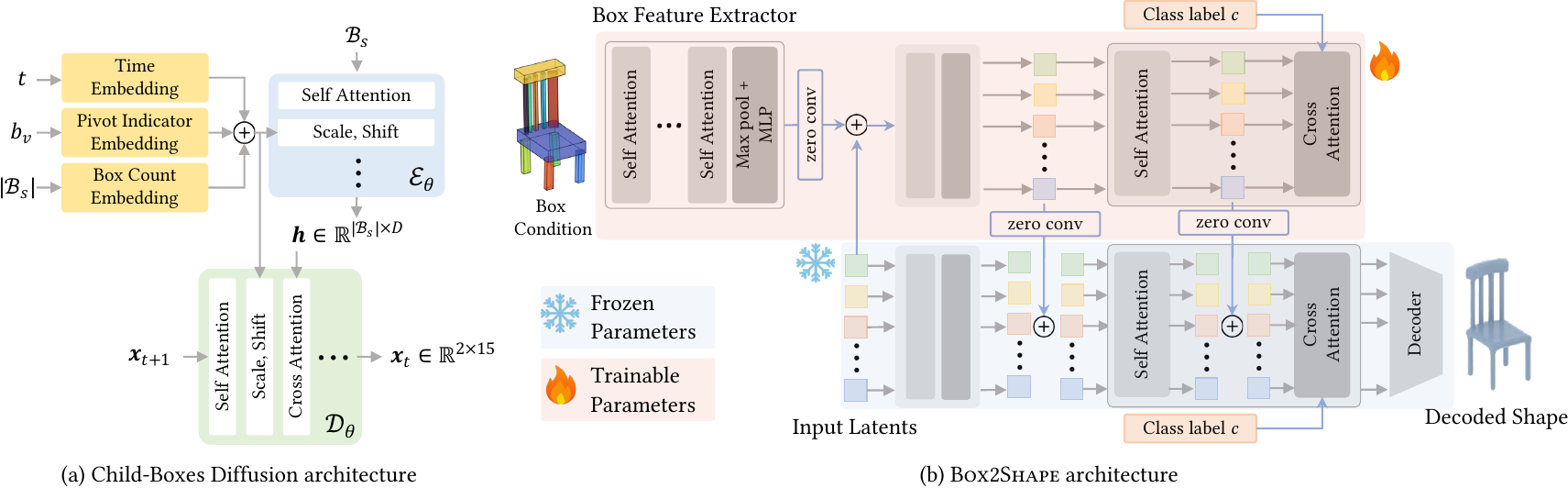}
    \vspace{-\baselineskip}
    \caption{\textbf{Diagram of network architectures.} (a) Child-Boxes Diffusion. (b) \textsc{Box2Shape}. Starting from a unit cube, we iteratively split boxes using Child-Boxes Diffusion to obtain the box condition with desired granularity, which then guides \textsc{Box2Shape} to generate aligned 3D shapes.}
    \vspace{-0.5\baselineskip}
    \label{fig:second_stage_network}
\end{figure*}

\section{\textsc{Box2Shape}: Box-to-Shape Generative Model}

Next, we propose a bounding-box-conditioned 3D shape generative model named \textsc{Box2Shape}. In contrast to the aforementioned bounding box generative model trained from scratch, here we aim to fully leverage the 3D shape priors learned by the state-of-the-art 3D diffusion model, 3DShape2VecSet~\cite{Zhang:2023Shape2Vec}. By finetuning this model, we ensure high fidelity in the generated shapes while effectively incorporating bounding box conditioning across varying levels of granularity.

A crucial requirement for finetuning pretrained networks to incorporate additional input conditions is smooth adaptation to conditional generation: the network should initially retain its original generation quality while gradually adapting to the specified condition. Two representative approaches for achieving this are the Gated Mechanism~\cite{Li:2023GLIGEN, Lee:2024ReGround, Sella:2023SpicE, Alayrac:2022Flamingo} and ControlNet~\cite{Zhang:2023ControlNet, Chen:2024Pixart-Delta}.

The Gated Mechanism introduces trainable layers gated by a gate parameter initialized to zero, ensuring unchanged output at the start of finetuning. However, it updates relatively few parameters, leading to slower convergence. In contrast, ControlNet duplicates existing layers, prepends a $1\times 1$ zero-convolution layer with weights and biases initialized to zero, and merges outputs via residual connections. This approach updates more parameters, enabling faster and more effective convergence compared to the Gated Mechanism. See Section~\ref{sec:shape_generation} for a detailed comparison of these two approaches. A notable example of adapting ControlNet to a 3D generative model is Spice-E~\cite{Sella:2023SpicE}, which uses Shape-E~\cite{Jun:2023Shap-E} as its backbone diffusion model whose generation quality is inferior to 3DShape2VecSet~\cite{Zhang:2023Shape2Vec}.

Adapting ControlNet~\cite{Zhang:2023ControlNet, Chen:2024Pixart-Delta} to our setup presents a challenge due to a mismatch in data representation between the input and the condition, as ControlNet requires these to be aligned. In our case, the backbone diffusion model, 3DShape2VecSet~\cite{Zhang:2023Shape2Vec}, represents 3D shapes as unordered latent sets \(\B{z} \in \mathbb{R}^{M \times D}\), while the input condition is a set of bounding boxes. Spice-E~\cite{Sella:2023SpicE} addresses this issue by representing input bounding boxes in the same format as their 3D shapes---multi-view images in their case---and encoding them using a pretrained encoder.
For 3DShape2VecSet, we find that a simpler approach performs effectively: incorporating a trainable encoding layer, \(\C{F}(\cdot)\), which directly processes the input bounding boxes and outputs latent sets without relying on a pretrained encoder. This module is trained jointly with the ControlNet branch. The architecture is illustrated in Figure~\ref{fig:second_stage_network} (b).
Our experiments detailed in Section~\ref{sec:experiment_results} demonstrate that our box-to-shape generative model based on 3DShape2VecSet outperforms Spice-E~\cite{Sella:2023SpicE} due to its strong priors for generating high-fidelity shapes.

\section{Experiment Results}
\label{sec:experiment_results}
We evaluate both components of our framework---the box-splitting model and the box-conditioned shape generation model---through qualitative and quantitative results using the ShapeNet~\cite{shapenet2015} dataset. Please see Appendix~\ref{sec:interactive_demo} and our \href{https://boxsplitgen.github.io}{project page} for user-interactive shape generation and editing demos. Additional experiment details are provided in Appendix~\ref{sec:evaluation_box_splitting} and~\ref{sec:evaluation_shape_generation}.


\subsection{Box-Splitting Generation}
\label{sec:box_generation}

\paragraph{Baselines.}
\label{sec:stage_1_baselines}
As discussed in Section~\ref{sec:overview}, our sequential generative process exhibits a unique characteristic. As outlined in Equation~\ref{eq:conditional_distribution}, each step can be decomposed into (1) selecting and removing one existing element ($p(b_v|\C{B}_s)$) and (2) producing two new elements at a time ($p(\C{C}(b_v) | b_v, \C{B}_s)$). Given the novelty of this problem, we extensively explore potential generative approaches. Specifically, we use the same pivot classifier to model $p(b_v | \C{B}_s)$, while exploring different methods for modeling $p(\C{C}(b_v) | b_v, \C{B}_s)$. This includes our Child-Boxes Diffusion, a conditional diffusion model, as well as two baselines:

\begin{itemize}[leftmargin=*]
    \item \textbf{Conditional Token Prediction Model}: We propose an approach based on sequence generation models that emulates the splitting process, departing from the typical sequential prediction of a single token at a time. The architecture is similar to the conditional diffusion model, taking $\C{B}_s$ and $b_v$ as conditions to generate two child boxes $\C{C}(b_v))$. However, the input and output representations are replaced with discretized representations, as is typical in GPT-like models. To enable the network to model a categorical distribution over tokens, we first train a VQ-VAE~\cite{Van:2017VQ-VAE} to encode continuous 15-dimensional box features into a discrete token space with a vocabulary size of $|V|=16K$. Unlike the conditional diffusion model, which predicts noise $\Veps \in \mathbb{R}^{2\times 15}$, this model outputs logits for two elements $\{l_1, l_2 \} \in \mathbb{R}^{2 \times |V|}$. The tokens sampled from the predicted categorical distributions are then decoded into the original continuous box parameters using the pre-trained VQ-VAE.
     
    \item \textbf{Unconditional Diffusion Model with Inpainting}: This model is trained to generate a varying number of complete bounding boxes without the pivot box as a condition but only the box count. During inference, we perform an inpainting technique. Specifically, given the set of input bounding boxes $\C{B}_s$ and the pivot box $b_v \in \C{B}_s$, we first duplicate $b_v$, increasing the total number of boxes to $|\C{B}_s|+1$, and then perform the DDIM inversion~\cite{dhariwal2021diffusion}, obtaining the standard normal sample $\B{x}_T$ from the input boxes. Next, we reset the portion of $\B{x}_T$ corresponding to the duplicates of $b_v$ to random standard normal samples. Then, we perform RePaint~\cite{lugmayr2022repaint} while treating $\C{B}_s \setminus \{b_v\}$ as the background to guide the inpainting process. Unlike our conditional diffusion model, which explicitly uses the pivot box as a condition, the limitation of this approach is that it cannot leverage information about the pivot box, as it performs inpainting after the pivot box is removed.
\end{itemize}
For further details on the baselines, please refer to Appendix~\ref{sec:other_options}.

\begin{table*}[t!]
\centering

\caption{\textbf{Quantitative comparison of shape abstraction generation with different pivot selection strategies.} MMD-CD scores and MMD-EMD scores are scaled by $10^3$ and $10^2$, respectively. The best results are highlighted in \textbf{bold}. The numbers are the averages across split levels $s=5$ and $s=8$.}
\label{tab:stage_1_uncond_cond}
\scriptsize
{
\setlength{\tabcolsep}{0.1em}
\renewcommand{\arraystretch}{1.0}
\definecolor{LightCyan}{rgb}{0.88,1,1}
\definecolor{Gray}{gray}{0.85}
\begin{tabularx}{\linewidth}{>{\centering\arraybackslash}m{1.5cm} >{\centering\arraybackslash}m{2.5cm}   Z Z Z Z Z Z Z Z Z Z Z Z}

\toprule

{ } &
  { } &
  \multicolumn{2}{c}{{ \textbf{COV ↑}}} &
  \multicolumn{2}{c}{{ \textbf{MMD ↓}}} &
  \multicolumn{2}{c}{{ \textbf{1-NNA ↓}}} &
  \multicolumn{2}{c}{{ \textbf{COV ↑}}} &
  \multicolumn{2}{c}{{ \textbf{MMD ↓}}} &
  \multicolumn{2}{c}{{ \textbf{1-NNA ↓}}} \\
  
  \cmidrule(lr){3-8} \cmidrule(lr){9-14}
  
  \multirow{-2}[2]{*}{{ \makecell{\textbf{Pivot}\\\textbf{Selection}}}} &
  \multirow{-2}[2]{*}{{ \textbf{Models}}} &
  { \textbf{CD}} &
  { \textbf{EMD}} &
  { \textbf{CD}} &
  { \textbf{EMD}} &
  { \textbf{CD}} &
  { \textbf{EMD}} &
  { \textbf{CD}} &
  { \textbf{EMD}} &
  { \textbf{CD}} &
  { \textbf{EMD}} &
  { \textbf{CD}} &
  { \textbf{EMD}} \\

  \midrule
  
  {} & {} & \multicolumn{6}{c}{{ \textbf{Chair}}} & \multicolumn{6}{c}{{ \textbf{Airplane}}} \\
  \midrule

\multirow{3}{*}{Random} & Token Pred. Model & 22.05 & 25.76 & 27.960 & 21.335 & 94.13 & 93.02 & 55.25 & 58.80 & 10.645 & 15.680 & 90.27 & 90.08 \\

{ } & Uncond. Diffusion & 29.70 & 30.97 & 18.784 & 18.128 & 88.68 & 87.93 & 63.82 & 68.21 & 7.207 & 12.953 & 88.50 & 87.90 \\

{ } & \textbf{Cond. Diffusion} & \textbf{32.84} & \textbf{33.97} & \textbf{16.896} & \textbf{16.910} & \textbf{85.28} & \textbf{83.43} & \textbf{77.75} & \textbf{77.38} & \textbf{6.842} & \textbf{12.599} & \textbf{86.35} & \textbf{85.64} \\
    
\midrule

\multirow{3}{*}{Classifier} & Token Pred. Model & 27.41 & 30.87 & 24.615 & 20.235 & 91.39 & 90.19 & 66.87 & 70.41 & 8.185 & 13.870 & 87.16 & 86.49 \\

{ } & { Uncond. Diffusion} & 33.03 & 35.68 & 18.174 & 17.598 & 88.53 & 85.77 & 71.76 & 77.75 & 6.811 & 12.505 & 86.15 & 85.55 \\

{ } &{\textbf{Cond. Diffusion}} & \textbf{46.08} & \textbf{47.08} & \textbf{14.166} & \textbf{15.580} & \textbf{75.87} & \textbf{73.33} & \textbf{82.89} & \textbf{80.56} & \textbf{6.478} & \textbf{12.188} & \textbf{85.62} & \textbf{84.81} \\

\midrule 

{ } & { } & \multicolumn{6}{c}{{ \textbf{Table}}} & \multicolumn{6}{c}{{ \textbf{Rifle}}} \\

\midrule

\multirow{3}{*}{Random} & Token Pred. Model & 18.77 & 17.24 & 32.175 & 22.580 & 90.77 & 92.59 & 57.42 & 58.67 & 3.400 & 9.475 & {87.18} & 86.72 \\

{ } & Uncond. Diffusion & 25.05 & 25.89 & 21.840 & 18.784 & 83.79 & 84.94 & \textbf{66.46} & \textbf{70.48} & 3.015 & 8.813 & \textbf{84.88} & \textbf{84.86} \\

{ } & \textbf{Cond. Diffusion} & \textbf{30.17} & \textbf{30.58} & \textbf{14.229} & \textbf{15.132} & \textbf{78.22} & \textbf{79.22} & {64.95} & {68.97} & \textbf{2.716} & \textbf{8.285} & 87.28 & {85.65} \\

\midrule

\multirow{3}{*}{Classifier} & Token Pred. Model & 25.45 & 24.47 & 25.895 & 19.680 & 87.47 & 88.72 & 68.72 & 71.86 & 3.495 & 9.230 & 84.28 & 82.67 \\

{ } & Uncond. Diffusion & 30.36 & 31.68 & 17.726 & 17.104 & 81.96 & 82.26 & \textbf{75.38} & \textbf{79.02} & 3.110 & 8.717 & \textbf{79.17} & \textbf{79.96} \\

{ } & \textbf{Cond. Diffusion} & \textbf{36.79} & \textbf{37.96} & \textbf{12.314} & \textbf{14.218} & \textbf{71.38} & \textbf{72.82} & {74.75} & {75.25} & \textbf{2.607} & \textbf{8.076} & {82.86} & {81.96} \\

\bottomrule

\end{tabularx}
}
\end{table*}

\begin{table*}[t!]
\centering

\caption{\textbf{Quantitative comparison of box-conditioned shape generation.} MMD-CD scores and MMD-EMD scores are scaled by $10^3$ and $10^2$, respectively. The best results are highlighted in \textbf{bold}.}

\label{tab:stage_2_comparison}
\scriptsize
{
\setlength{\tabcolsep}{0.0em}
\renewcommand{\arraystretch}{1.0}
\begin{tabularx}{\linewidth}{>{\centering\arraybackslash}m{2.5cm}@{} Z Z Z Z Z Z  >{\centering\scriptsize\arraybackslash}m{1.3cm}@{} >{\centering\scriptsize\arraybackslash}m{1.3cm}@{} >{\centering\scriptsize\arraybackslash}m{1cm}@{} >{\centering\scriptsize\arraybackslash}m{1cm}@{}}
\toprule
\multirow{2}[2]{*}{\textbf{Models}} & \multicolumn{2}{c}{\textbf{COV ↑}} & \multicolumn{2}{c}{\textbf{MMD ↓}} & \multicolumn{2}{c}{\textbf{1-NNA ↓}} & \multicolumn{4}{c}{\textbf{Box Alignment}} \\

\cmidrule(lr){2-7} \cmidrule(lr){8-11}

& \textbf{CD} & \textbf{EMD} & \textbf{CD} & \textbf{EMD} & \textbf{CD} & \textbf{EMD} & \textbf{Box-CD ↓} & \textbf{Box-EMD ↓} & \textbf{TOV ↓} & \textbf{VIoU ↑} \\

  \midrule

  Spice-E~\cite{Sella:2023SpicE} & 38.33 & 40.93 & 14.762 & 15.308 & 87.09 & 85.98 & 0.043 & 0.255 & 2.44 & 0.27 \\
  Gated 3DS2V~\cite{Zhang:2023Shape2Vec} & \textbf{54.55} & \textbf{57.42} & 13.067 & 14.991 & 78.93 & 77.14 & 0.012 & 0.143 & 1.39 & 0.17 \\
  \textbf{Box2Shape (Ours)} & 51.06 & 50.99 & \textbf{10.369} & \textbf{12.324} & \textbf{72.75} & \textbf{72.16} & \textbf{0.006} & \textbf{0.098} & \textbf{1.08} & \textbf{0.31} \\
\bottomrule
\end{tabularx}
}
\end{table*}

\paragraph{Quantitative Evaluation.}
To quantitatively evaluate the quality and diversity of the generated bounding boxes, we measure Coverage (COV), Minimum Matching Distance (MMD), and 1-Nearest Neighbor Accuracy (1-NNA) between the set of generated bounding boxes and the reference bounding box set. We use the training bounding box set as the reference set and generate 2,000 bounding boxes for each class by iteratively splitting them. 
We then compare the resulting boxes at various split levels to the reference set.

For a comprehensive analysis, we evaluate the metrics separately at two split steps, $s=5$ (coarser) and $s=8$ (finer), and then take their average, since these steps fall within the average bounding box count.
More detailed results for each split step and evaluation setups can be found in Appendix~\ref{sec:evaluation_box_splitting}.

Table~\ref{tab:stage_1_uncond_cond} shows the effectiveness of both our pivot classifier and our Child-Boxes Diffusion. Compared to random pivot selection, our pivot classifier improves overall box-splitting performance, enhancing both the quality and diversity of the produced abstractions. Under the same pivot selection strategy, \Ours{} consistently outperforms them in most metrics, demonstrating its robustness in modeling conditional distribution $p(\C{C}(b_v)|\C{B}_s, b_v)$.

\begin{figure*}[t!]
\centering
{
\scriptsize
\setlength{\tabcolsep}{0em}
\renewcommand\tabularxcolumn[1]{m{#1}}
\begin{tabularx}{\linewidth}{Z Z Z |  Z Z Z | Z Z Z | Z Z Z}
 \rotatebox{0}{\makecell{Token Pred.\\Model}} & \rotatebox{0}{\makecell{Uncond.\\Diffusion}} & \rotatebox{0}{\makecell{Cond.\\Diffusion}} & \rotatebox{0}{\makecell{Token Pred.\\Model}} & \rotatebox{0}{\makecell{Uncond.\\Diffusion}} & \rotatebox{0}{\makecell{Cond.\\Diffusion}} & \rotatebox{0}{\makecell{Token Pred.\\Model}} & \rotatebox{0}{\makecell{Uncond.\\Diffusion}} & \rotatebox{0}{\makecell{Cond.\\Diffusion}} & \rotatebox{0}{\makecell{Token Pred.\\Model}} & \rotatebox{0}{\makecell{Uncond.\\Diffusion}} & \rotatebox{0}{\makecell{Cond.\\Diffusion}} \\
\midrule
\multicolumn{3}{c|}{\includegraphics[width=0.25\textwidth]{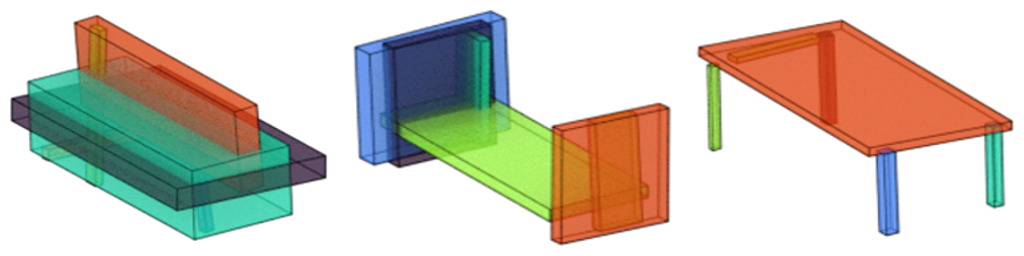}} &
\multicolumn{3}{c|}{\includegraphics[width=0.25\textwidth]{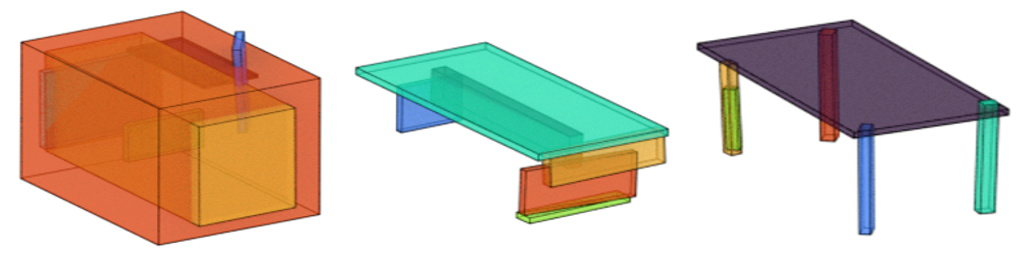}} &
\multicolumn{3}{c|}{\includegraphics[width=0.25\textwidth]{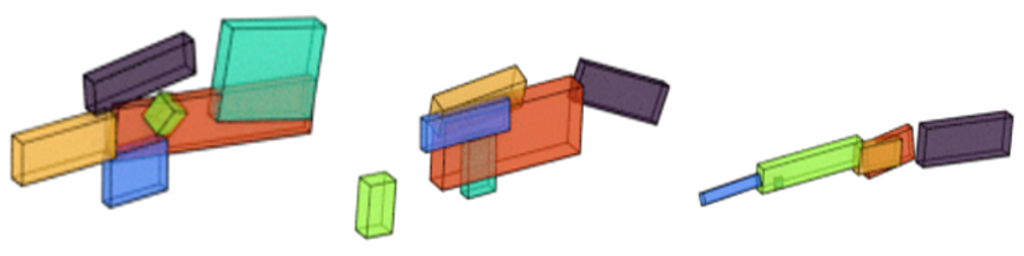}} &
\multicolumn{3}{c}{\includegraphics[width=0.25\textwidth]{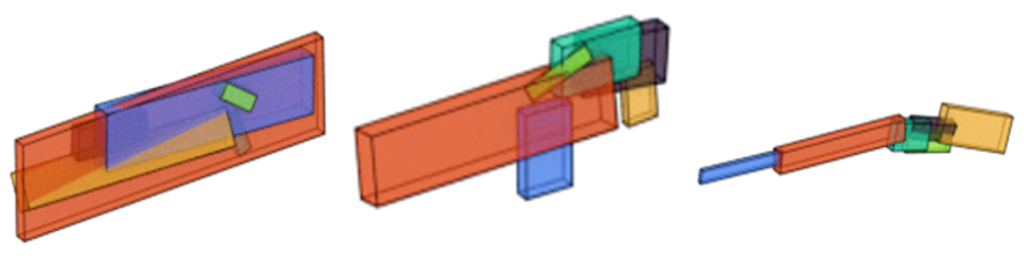}} \\

\multicolumn{3}{c|}{\includegraphics[width=0.25\textwidth]{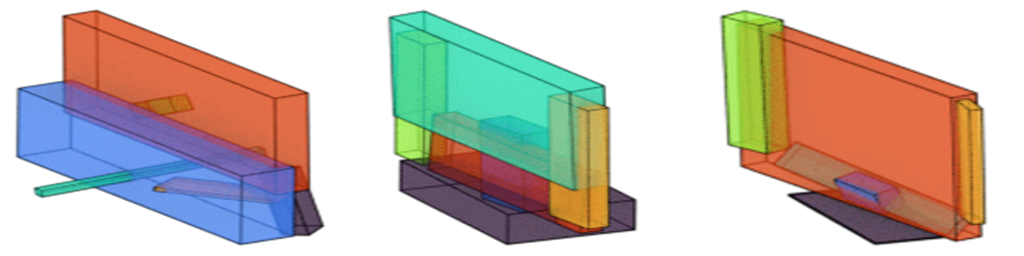}} &
\multicolumn{3}{c|}{\includegraphics[width=0.25\textwidth]{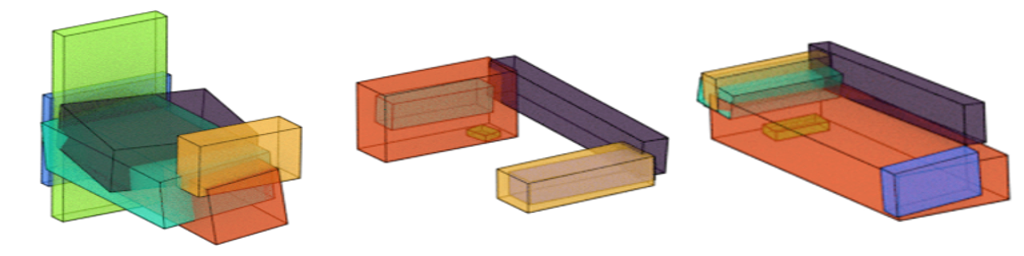}} &
\multicolumn{3}{c|}{\includegraphics[width=0.25\textwidth]{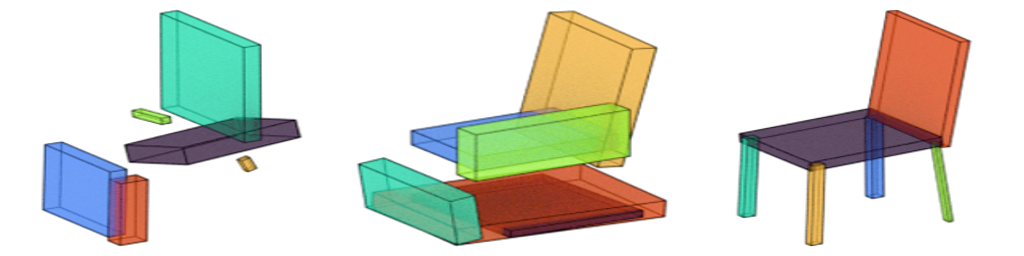}} &
\multicolumn{3}{c}{\includegraphics[width=0.25\textwidth]{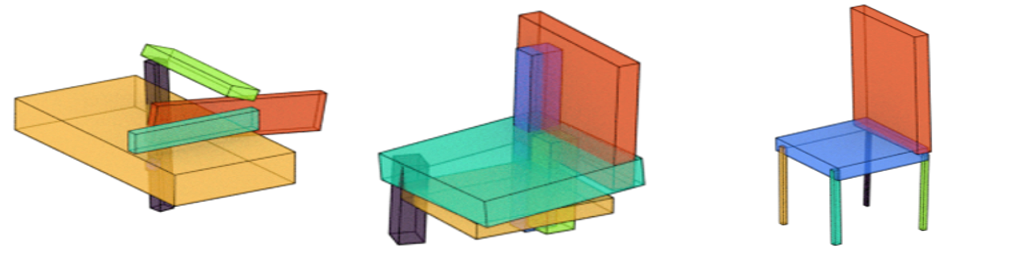}} \\

\multicolumn{3}{c|}{\includegraphics[width=0.25\textwidth]{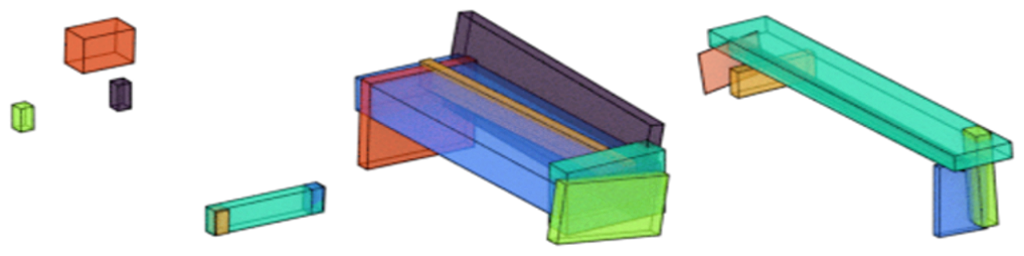}} &
\multicolumn{3}{c|}{\includegraphics[width=0.25\textwidth]{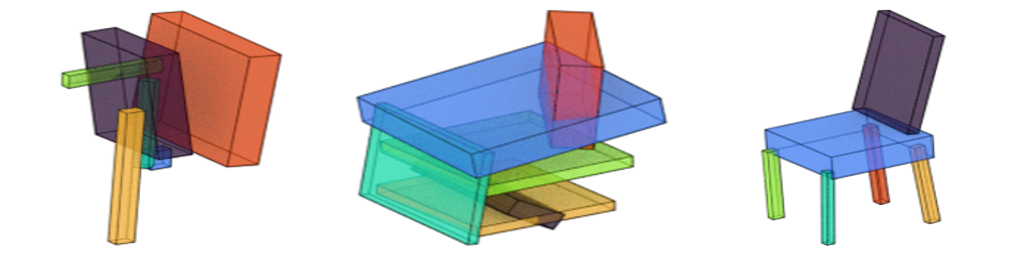}} &
\multicolumn{3}{c|}{\includegraphics[width=0.25\textwidth]{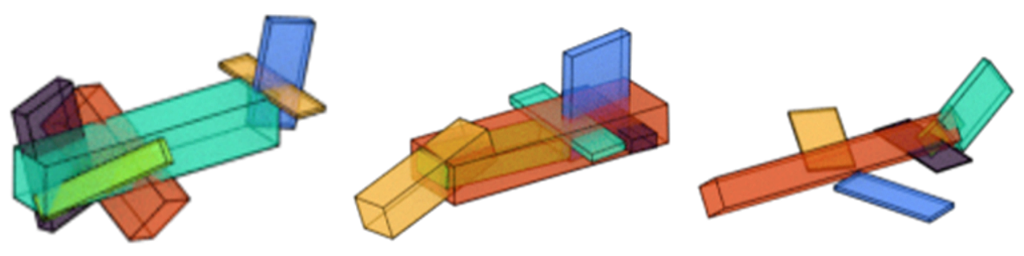}} &
\multicolumn{3}{c}{\includegraphics[width=0.25\textwidth]{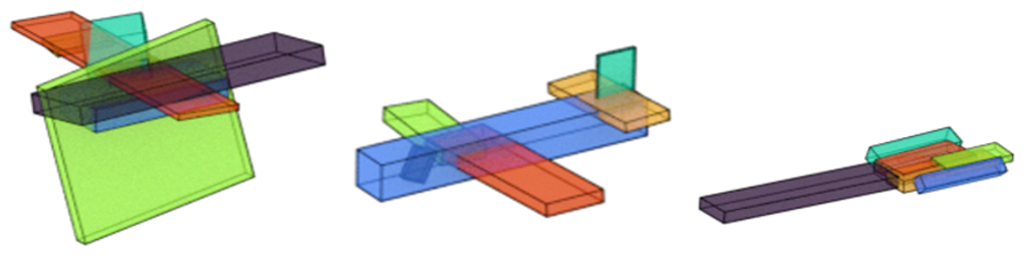}} \\

\end{tabularx}
\caption{\textbf{Qualitative comparison of shape abstraction generation. } For each pair of columns, we query the ground truth shape and retrieve the closest generated boxes measured with chamfer distance. Our method demonstrates higher-fidelity boxes.}
\label{fig:stage_1_comparison}
}
\end{figure*}

\begin{figure*}[t!]

\centering
\scriptsize
\setlength{\tabcolsep}{0.0em}
\begin{tabularx}{\linewidth}{ZZZZ | ZZZZ | ZZZZ}
\rotatebox{0}{\makecell{Input\\Boxes}} & \rotatebox{0}{\makecell{Spice-E\\\cite{Sella:2023SpicE}}} & \rotatebox{0}{\makecell{Gated\\3DS2V~\cite{Zhang:2023Shape2Vec}}} & \rotatebox{0}{Ours} &  \rotatebox{0}{\makecell{Input\\Boxes}} & \rotatebox{0}{\makecell{Spice-E\\\cite{Sella:2023SpicE}}} & \rotatebox{0}{\makecell{Gated\\3DS2V~\cite{Zhang:2023Shape2Vec}}} & \rotatebox{0}{Ours} & \rotatebox{0}{\makecell{Input\\Boxes}} & \rotatebox{0}{\makecell{Spice-E\\\cite{Sella:2023SpicE}}} & \rotatebox{0}{\makecell{Gated\\3DS2V~\cite{Zhang:2023Shape2Vec}}} & \rotatebox{0}{Ours}  \\ 

\midrule
\multicolumn{4}{c|}{\includegraphics[width=0.333\textwidth]{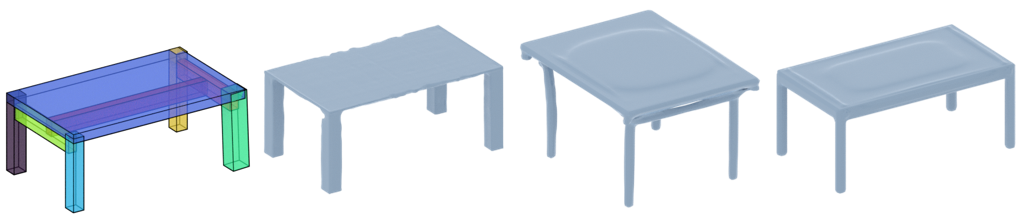}} &
\multicolumn{4}{c|}{\includegraphics[width=0.333\textwidth]{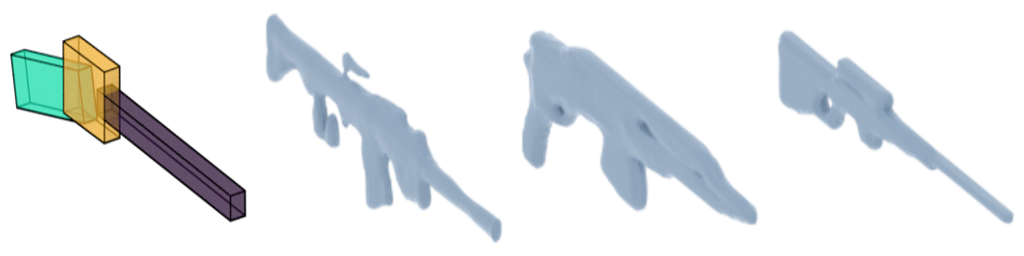}} &
\multicolumn{4}{c}{\includegraphics[width=0.333\textwidth]{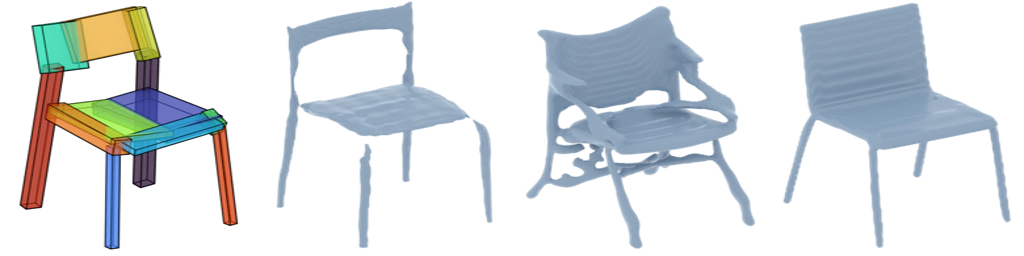}} \\

\multicolumn{4}{c|}{\includegraphics[width=0.333\textwidth]{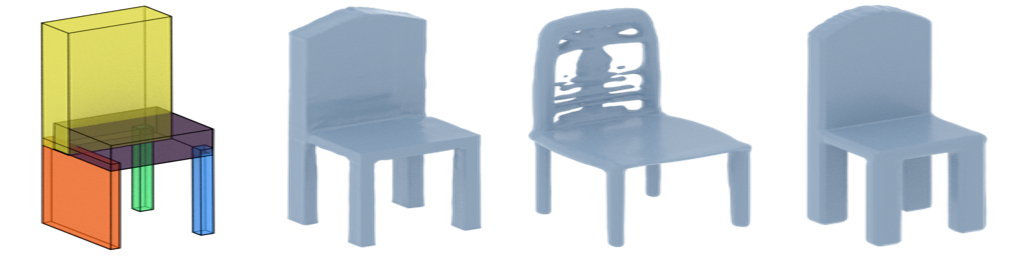}} &
\multicolumn{4}{c|}{\includegraphics[width=0.333\textwidth]{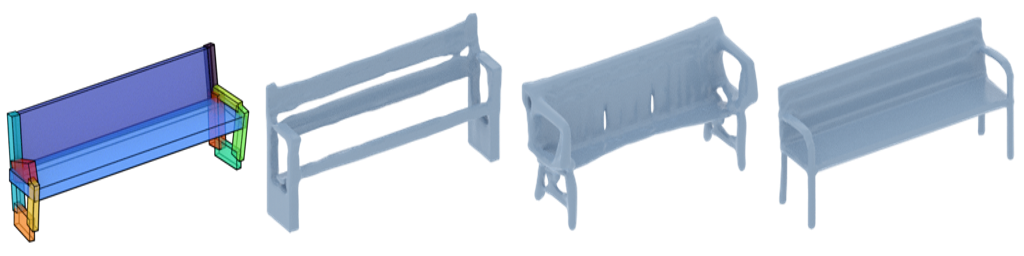}} &
\multicolumn{4}{c}{\includegraphics[width=0.333\textwidth]{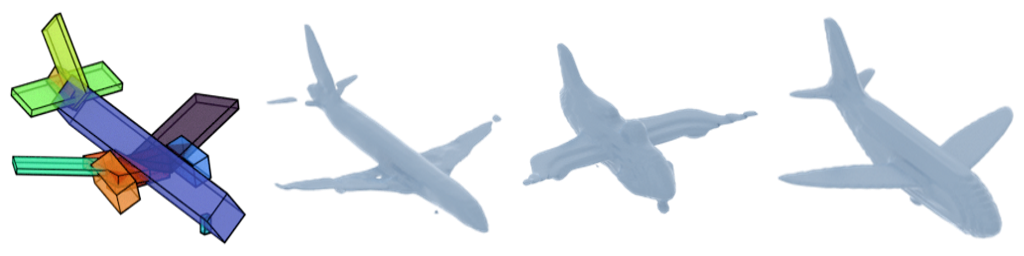}} 
\end{tabularx}
\caption{\textbf{Gallery of our generated bounding boxes and their final generated 3D shapes by box-conditioned shape generation models.} Each pair of columns shows the input bounding boxes (left) and their corresponding generated 3D shapes (right).}
\label{fig:shape_generation_qualitative_results}
\end{figure*}

\paragraph{Qualitative Comparison.}
Figure~\ref{fig:stage_1_comparison} presents a qualitative comparison across different classes. Refer to Appendix~\ref{sec:suppl_more_box_splitting_qualitative_results} for more qualitative results. As illustrated, our conditional diffusion model produces diverse bounding boxes that capture fine details of the 3D structure across various classes. On the other hand, the conditional token prediction model often struggles to produce plausible shape abstractions, highlighting the difficulty of modeling the splitting process using sequential token prediction models. Similarly, the inpainting approach with unconditional diffusion models often generates boxes that miss parts of the shape. 
It demonstrates that enforcing the other remaining boxes to remain fixed during the denoising process can easily cause deviations from the learned data manifold of diffusion models.


\subsection{Box-Conditioned Shape Generation}
\label{sec:shape_generation}
\paragraph{Baselines.} We compare our method with two baselines:
\begin{itemize}[leftmargin=*]
    \item \textbf{Spice-E}~\cite{Sella:2023SpicE}: A recent box-conditioned shape generative model. We finetune a pre-trained model on our dataset to process bounding boxes at varying granularity. 
    \item \textbf{Gated 3DShape2VecSet}~\cite{Zhang:2023Shape2Vec}: A variant of our model finetuned with the Gated Mechanism~\cite{Li:2023GLIGEN, Alayrac:2022Flamingo, Lee:2024ReGround} instead of ControlNet. Cross-attention layers gated by box features are injected into each Transformer block, while the rest of othe model is frozen.     
\end{itemize}

\paragraph{Quantitative Evaluation.}
For quantitative evaluation, we use the same metrics as in the box-splitting stage---COV, MMD, and 1-NNA---to assess the fidelity and the diversity of generated meshes, but using 3D shapes as the reference set. Additionally, following SMART~\cite{Park:2024SMART}, we evaluate the alignment between the input shape abstraction and its generated 3D shape using the following metrics: Total Outside Volume (TOV), Volumetric Intersection over Union (VIoU), Box-CD and Box-EMD. TOV and VIoU measure box alignment based on mesh volumes, while Box-CD and Box-EMD assess geometric alignment using CD and EMD.
See Appendix~\ref{sec:evaluation_shape_generation} for more details on the evaluation metrics and setups. 

Table~\ref{tab:stage_2_comparison} presents a quantitative comparison between our method and the baselines. Our method \textsc{Box2Shape} outperforms others in shape fidelity and diversity metrics---COV, MMD, and 1-NNA---with a significant advantage in 1-NNA. While Gated 3DShape2VecSet achieves higher COV, this is due to its lack of adaptation to box-conditioned generation, often producing shapes deviated from the input boxes. This limitation is further reflected in its lacking box alignment performance across all box alignment metrics: Box-CD, Box-EMD, TOV, VIoU. Spice-E~\cite{Sella:2023SpicE} demonstrates suboptimal performance compared to \textsc{Box2Shape}, largely due to its backbone model's limited shape prior. These results demonstrate that our conditioning approach achieves superior box alignment while preserving the original model's superior performance to generate high-fidelity and diverse shapes.
\paragraph{Qualitative Results.}
Figure~\ref{fig:shape_generation_qualitative_results} presents a qualitative comparison, where each pair of columns shows the input bounding boxes and their generated 3D shapes by different approaches. Refer to Appendix~\ref{sec:suppl_more_shape_generation_qualitative_results} for more results. Given an input set of bounding boxes, ours produces a plausible 3D shape while being well-aligned with the input boxes. In contrast, Spice-E~\cite{Sella:2023SpicE} often fails to capture fine-grained details in the 3D shapes due to its backbone's suboptimal performance. Compared to Gated 3DShape2VecSet, which is based on the Gated Mechanism, it often struggles to produce box-aligned 3D shapes, highlighting the limitations of the Gated Mechanism in modeling the conditional probability distribution with our diverse conditioning bounding boxes. 

\section{Conclusion}
\label{sec:conclusion}

We presented a box-splitting-based interactive 3D shape generation framework composed of two generative models. The first model, \Ours{}, is an autoregressive model that enables the progressive refinement of bounding boxes via splitting. We introduce a pivot classifier and a child-box diffusion model to select which box to split and to generate the two new boxes, respectively. The second model is a box-to-shape generative model that effectively adapts a pre-trained unconditional 3D diffusion model.
%
%
%
We demonstrate that the proposed framework facilitates intuitive 3D generation by mimicking the human imagination process from abstract concepts to detailed structures. Users can split and manipulate bounding boxes to generate aligned 3D shapes, with diversity naturally decreasing as the bounding boxes become fine-grained. 

As future work, we plan to improve our user-interactive 3D shape generation framework to incorporate additional spatial guidance, including other primitives~\cite{Koo:2023Salad, Hertz:2022Spaghetti,Tulsiani:2017VolumetricPrimitives} and 3D sketches~\cite{Zheng:2023LAS-Diffusion}.

\paragraph{Acknowledgements.} 

This work was supported by the IITP grants (RS-2022-00156435, RS-2024-00399817, RS-2025-25441313, RS-2025-25443318, RS-2025-02653113); and the Industrial Technology Innovation Program (RS-2025-02317326), all funded by the Korean government (MSIT and MOTIE), as well as by the DRB-KAIST SketchTheFuture Research Center.

{
    \small
    \bibliographystyle{ieeenat_fullname}
    \bibliography{main}
}
\renewcommand{\thesection}{A}
\renewcommand{\thetable}{A\arabic{table}}
\renewcommand{\thefigure}{A\arabic{figure}}
\clearpage 
\newpage
\section*{Appendix}
\ifarxiv
\newcommand{\refofpaper}[1]{\unskip}
\newcommand{\refinpaper}[1]{\unskip}
  
\else 
\tableofcontents
\clearpage
\newpage

\makeatletter
\newcommand{\manuallabel}[2]{\def\@currentlabel{#2}\label{#1}}

\makeatother
\manuallabel{fig:teaser}{1}
\manuallabel{tab:stage_1_uncond_cond}{1}
\manuallabel{sec:box_generation}{5.1}
\manuallabel{eq:noise_prediction}{3}
\manuallabel{sec:child_box_diffusion}{3.4}
\manuallabel{sec:overview}{3.1}
\manuallabel{sec:experiment_results}{5}
\newcommand{\refofpaper}[1]{of the main paper}
\newcommand{\refinpaper}[1]{in the main paper}
\renewcommand{\thesection}{S}
\renewcommand{\thetable}{S\arabic{table}}
\renewcommand{\thefigure}{S\arabic{figure}}

\fi

\ifarxiv
\else
\subsection{Overview}
In this supplementary document, we provide additional details and results that complement the main paper. We first describe our user-interactive box and shape editing demo and show example outputs in Section~\ref{sec:interactive_demo}. We then report additional results on box-splitting over PartNet in Section~\ref{sec:partnet_splitting}, followed by detailed descriptions of the data, evaluation metrics, and experimental setups for box-splitting generation in Section~\ref{sec:evaluation_box_splitting} and for box-conditioned shape generation in Section~\ref{sec:evaluation_shape_generation}. Next, we discuss alternative approaches for box-splitting generation in Section~\ref{sec:other_options} and present implementation details of all components in Section~\ref{sec:implementation_details}, along with more details on our user-interactive demo (Section~\ref{sec:demo_details}), a runtime analysis (Section~\ref{sec:runtime_analysis}), and further quantitative results of box splitting (Section~\ref{sec:more_box_splitting_quantitative_results}). Finally, we visualize full splitting sequences in Section~\ref{sec:sequence_visualization} and provide additional qualitative results for box-splitting generation in Section~\ref{sec:suppl_more_box_splitting_qualitative_results} and for box-conditioned shape generation in Section~\ref{sec:suppl_more_shape_generation_qualitative_results}.

\fi 


\subsection{User-Interactive Generation Demo and Examples}
\label{sec:interactive_demo}
\begin{figure*}[h!]
\centering
{
\scriptsize
\setlength{\tabcolsep}{0em}
\renewcommand\tabularxcolumn[1]{m{#1}}
\begin{tabularx}{\linewidth}{Y Y Y Y Y Y Y Y Y Y Y Y}

\multicolumn{12}{c}{Shape Variations with Coarse and Fine Abstractions} \\
\multicolumn{6}{c|}{Coarse Abstraction} & \multicolumn{6}{c}{Fine Abstraction} \\
\multicolumn{6}{c|}{
\includegraphics[width=.5\textwidth]{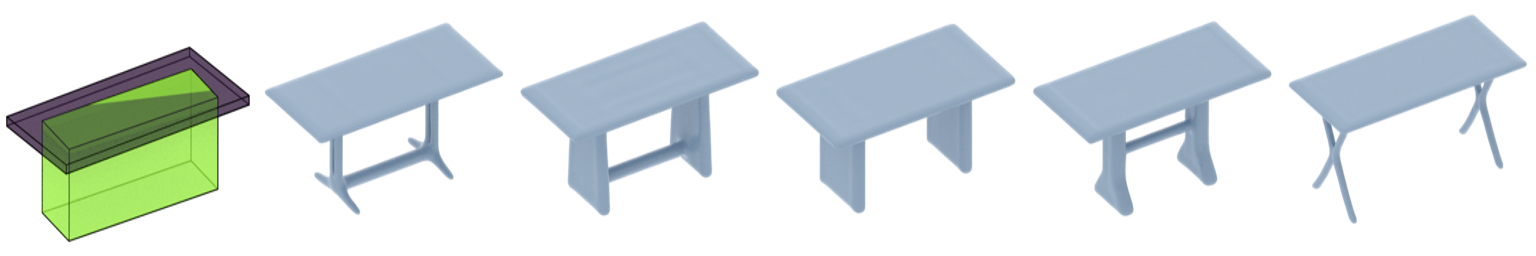}
} &
\multicolumn{6}{c}{
\includegraphics[width=.5\textwidth]{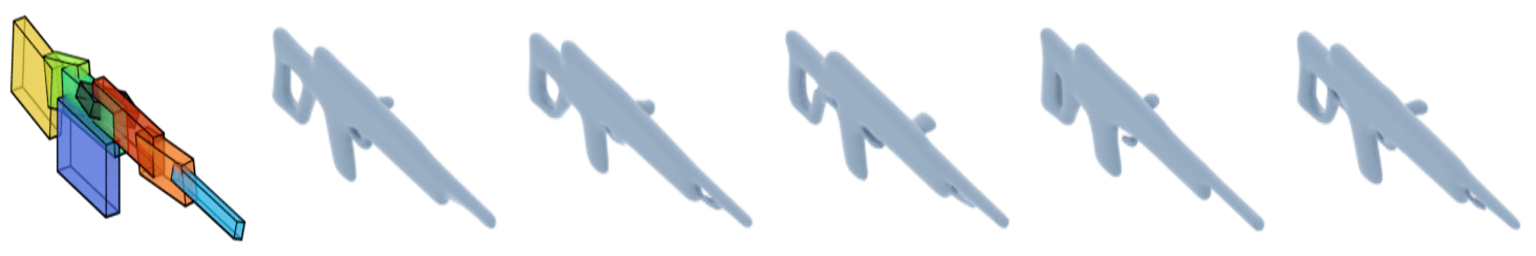}
} \\
\multicolumn{6}{c|}{
\includegraphics[width=.5\textwidth]{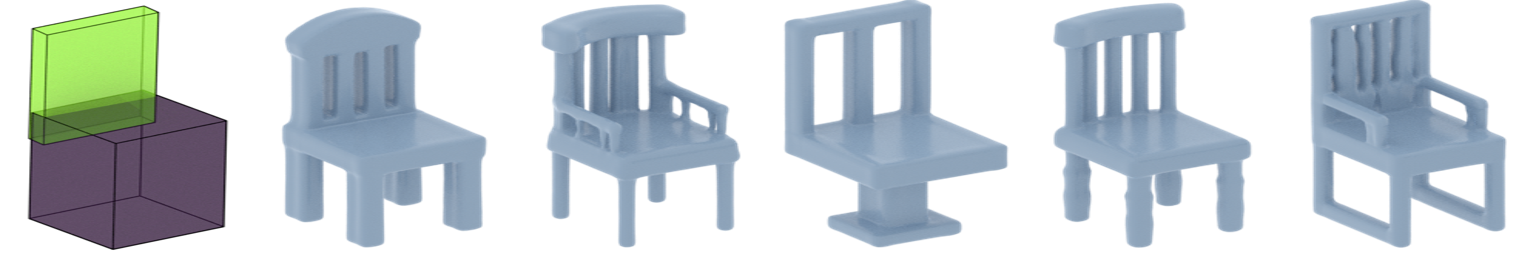}
} &
\multicolumn{6}{c}{
\includegraphics[width=.5\textwidth]{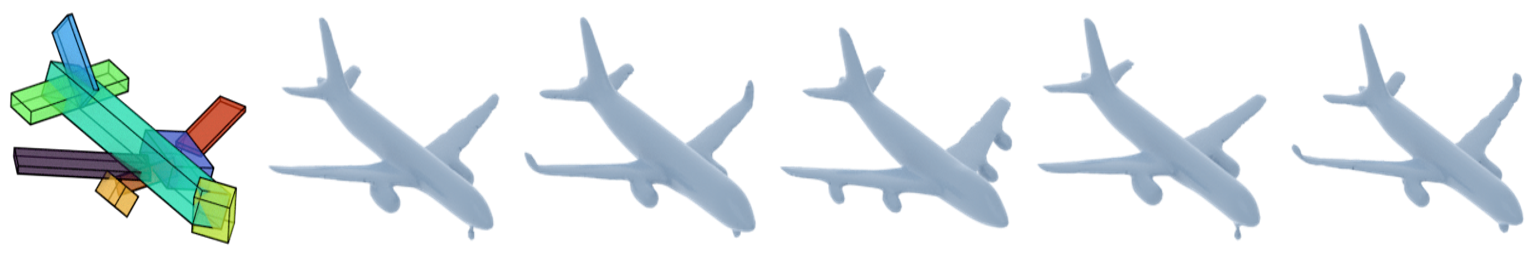}
} \\
\midrule 
\multicolumn{12}{c}{Shape Editing Results} \\
\multicolumn{4}{c|}{
\includegraphics[width=0.33\textwidth]{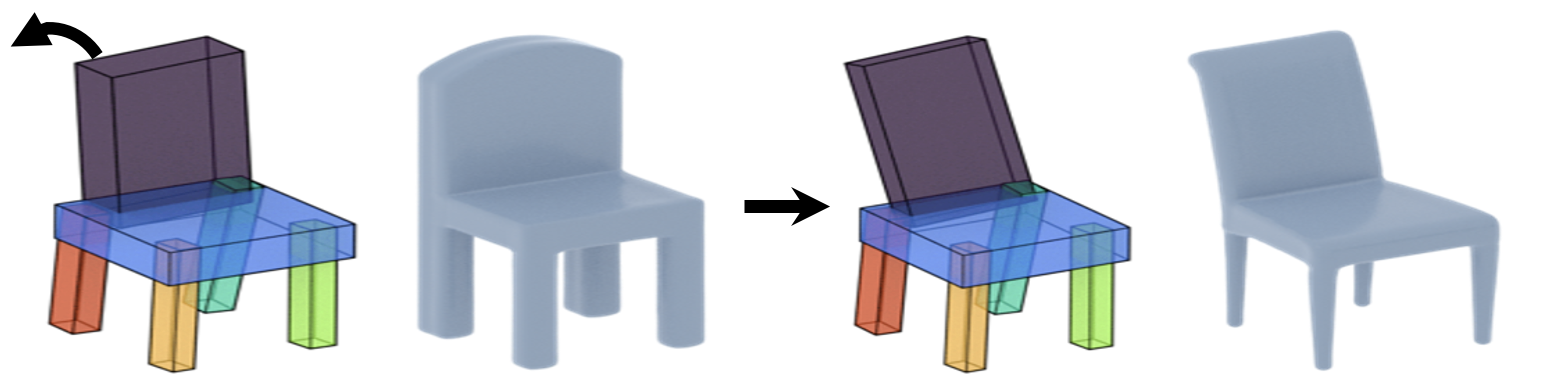}
} &
\multicolumn{4}{c|}{
\includegraphics[width=0.33\textwidth]{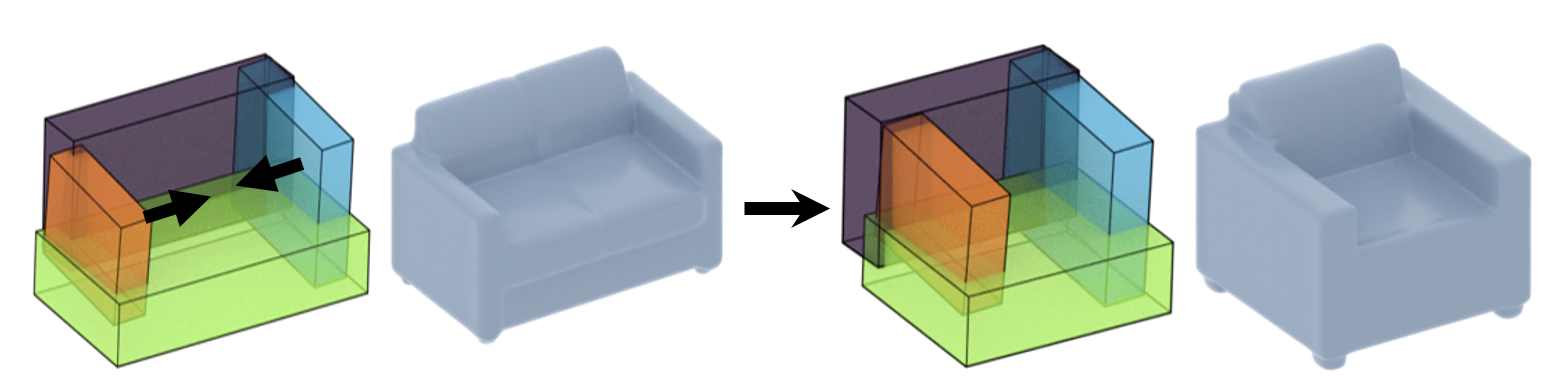}
} &
\multicolumn{4}{c}{
\includegraphics[width=0.33\textwidth]{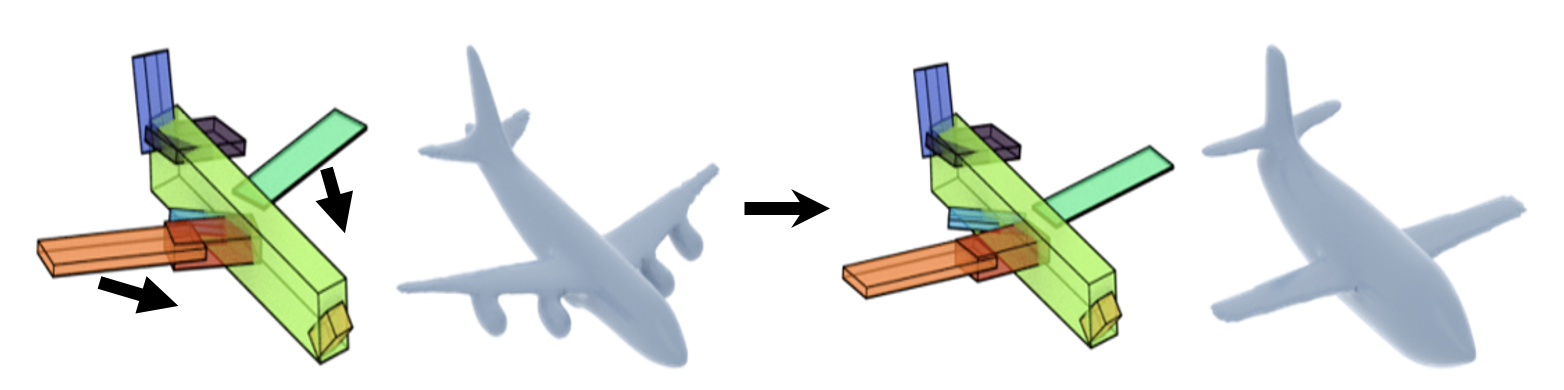}
} \\



\end{tabularx}}
\caption{\textbf{Shape variations and editing guided by bounding boxes.} The top row demonstrates shape variations with different granularity of the bounding boxes, while the bottom illustrates shape editing achieved by manipulating bounding boxes.}
\label{fig:shape_editing}
\end{figure*}

In Figure~\ref{fig:teaser}~\refofpaper{} and Figure~\ref{fig:shape_editing}, we showcase the interface and exemplary outputs of our user-interactive box and shape editing framework. The system allows users to create 3D shapes using bounding boxes as conditions, enabling not only the manipulation of boxes but also their splitting and merging, thereby controlling the granularity. This controllability of granularity is a crucial difference from prior work~\cite{Koo:2023Salad,Hertz:2022Spaghetti,Paschalidou:2021NeuralParts,Kiyohiro:2023DiffFacto,Hui:2022}, which operates at a fixed level of detail. It allows users to envision diverse shapes from coarser bounding boxes while generating specific shapes with detailed bounding boxes, as shown in Figure~\ref{fig:teaser}~\refofpaper{} and at the top of Figure~\ref{fig:shape_editing}.

The system suggests a pivot box to split using our pivot classifier and generates the resulting child boxes with our Child-Boxes Diffusion model. Given the set of bounding boxes, the bounding box-to-shape diffusion model generates a variety of aligned 3D shapes. For demo videos and interactive examples, please refer to our \href{https://boxsplitgen.github.io}{project page}.

Additionally, the system allows users to edit local parts of a 3D shape by transforming the corresponding bounding boxes. Figure~\ref{fig:shape_editing} bottom rows presents results achieved through box manipulation. When a selected box is modified, the decoded shape adjusts accordingly, highlighting our framework’s user-friendly and intuitive interface for shape editing.




\subsection{Box-Splitting on PartNet}
\label{sec:partnet_splitting}
\begin{table*}[ht!]
\centering
\caption{Comparison of SMART~\cite{Park:2024SMART} and PartNet~\cite{Mo:2019PartNet} across different metrics for Chair and Table categories. Bold indicates the best for each column.}

\setlength{\tabcolsep}{0.0em}
\scriptsize
\begin{tabularx}{\linewidth}{>{\centering\arraybackslash}m{1.3cm} Y Y Y Y Y Y Y Y}
\toprule
\multirow{2}[2]{*}{\textbf{Dataset}} & \multicolumn{4}{c}{\textbf{Chair}} & \multicolumn{4}{c}{\textbf{Table}} \\
\cmidrule(lr){2-5} \cmidrule(lr){6-9}
& \textbf{Tightness ↑} & \textbf{COV-EMD ↑} & \textbf{MMD-EMD ↓} & \textbf{1-NNA-EMD ↓} & \textbf{Tightness ↑} & \textbf{COV-EMD ↑} & \textbf{MMD-EMD ↓} & \textbf{1-NNA-EMD ↓} \\
\midrule
PartNet~\cite{Mo:2019PartNet} & \textbf{2.25} & 32.49 & \textbf{16.645} & 85.87 & \textbf{2.20} & 29.02 & \textbf{14.190} & \textbf{78.21} \\
SMART~\cite{Park:2024SMART} & 1.61 & \textbf{33.97} & 16.910 & \textbf{83.43} & 1.78 & \textbf{30.58} & 15.132 & 79.22
\\
\bottomrule
\end{tabularx}

\label{tab:smart_partnet}
\end{table*}
Table~\ref{tab:smart_partnet} shows the Tightness of bounding boxes introduced in SMART~\cite{Park:2024SMART} and results using same setup as Table~\ref{tab:stage_1_uncond_cond}, replacing SMART’s unsupervised over-segments with PartNet’s~\cite{Mo:2019PartNet} clean annotated leaf parts. As shown below, PartNet slightly improves MMD with cleaner decompositions but slightly worsens other metrics due to looser bounding boxes from semantic-level parts.

These comparable results show our method works with hand-crafted data when available, while SMART, despite being unsupervised, provides structurally meaningful hierarchical data, demonstrating its scalability to large-scale datasets like Objaverse~\cite{deitke2023objaverse}, where collecting part annotations is costly.

\subsection{Details on the Experiment Setups of Box Splitting Generation}
\label{sec:evaluation_box_splitting}
\paragraph{Data.}
\label{sec:dataset_stats}
We use 3D shapes from the ShapeNet~\cite{shapenet2015} dataset for our experiments. Each class in ShapeNet contains between 300 and 2,700 shapes. The average number of bounding boxes in the SMART outputs for each class ranges from 5 to 12. See Table~\ref{tab:datasets_stats} for detailed statistics. We construct a training set and a validation set by splitting the shapes within each class at an 8:2 ratio. For the box-splitting training, we train separate models for each class, whereas our box-to-shape generation model is trained jointly for all classes.

\paragraph{Evaluation metrics.}
The evaluation metrics used in Section~\ref{sec:box_generation}~\refofpaper{} are computed using two shape distance metrics: Chamfer Distance~\cite{ChamferDistance} (CD) and Earth Mover's Distance~\cite{EarthMoversDistance} (EMD). To measure CD and EMD between bounding boxes, we first convert a set of bounding boxes for an object into a watertight mesh~\cite{manifold}. The conversion process begins by representing the bounding boxes as a union of implicit surfaces, which is then converted into a surface mesh, effectively removing intersections across the bounding boxes. Next, we sample 2,048 points on the watertight mesh surfaces using Poisson disk sampling~\cite{Open3D}.

\subsection{Details on the Experiment Setups of Box-Conditioned Shape Generation}
\label{sec:evaluation_shape_generation}
To evaluate the diversity and fidelity of the 3D shapes generated from the input bounding boxes, we randomly sample 1,000 bounding boxes from the validation set for each class to construct the evaluation set for shape generation. For the reference set, we use the shape set from IM-Net~\cite{Chen:2019IMNet}. As in the evaluation of the box splitting stage, we sample 2,048 points over the surface of the shapes and measure geometric distances using Chamfer Distance (CD) and Earth Mover's Distance (EMD) for the COV, MMD, and 1-NNA metrics.

For assessing box alignment, we use a different evaluation bounding box set to ensure that the bounding boxes are sufficiently fine, such that the decoded shape tightly fits within the input boxes. This prevents input bounding boxes from being overly loose and simply enclosing the shapes. Specifically, we use the bounding boxes at the finest granularity level for each shape in the validation set to ensure a rigorous alignment evaluation between input bounding boxes and their decoded shapes.

Given a set of bounding boxes $\{b_i\}$ and a 3D shape $S$, TOV and VIoU are calculated as follows:
\begin{align}
\text{TOV} = \frac{\text{vol} (\bigcup_i b_i \setminus S) }{\text{vol}(S)},\quad \text{VIoU} = \frac{\text{vol} (S \cap \bigcup_i b_i)}{\text{vol} (S \cup \bigcup_i b_i)}.
\end{align}
In addition to the volume-based alignment metrics, we measure geometric alignment (Box-CD and Box-EMD) by computing CD and EMD between two point clouds: one sampled from the surface of the bounding boxes and the other from the decoded 3D shape. 

\subsection{Details on Alternative Approaches for Box Splitting Generation}
\label{sec:other_options}

We explore two alternative approaches for learning the conditional distribution of child boxes given a pivot box and its context. First, we investigate the use of unconditional diffusion models combined with inpainting techniques to generate child boxes while preserving the existing remaining boxes. Second, we explore a sequence generation approach inspired by their recent advances in 3D shape generation~\cite{siddiqui2024meshgpt, tang2024edgerunner, Mittal:2022Autosdf}. Below, we present more details on each alternative approach.





\paragraph{Unconditional Diffusion with Inpainting.}
Diffusion models have demonstrated great success in solving inverse problems~\cite{chung2022diffusion,chung2022mcg, song2023lgd, he2023mpgd, ye2024tfg, song2023pseudoinverseguided, kawar2022denoising, lugmayr2022repaint}, guiding the denoising process with some conditions in various applications (\eg background regions for image inpainting, blurred images for image deblurring, and low-resolution images for image super-resolution). The guided denoising process enables generating data that satisfies the given conditions with a \emph{unconditional} diffusion model. Inspired by this, we experimented with sampling the two child boxes using a unconditional diffusion model and diffusion-based image inpainting technique.

Specifically, we train the noise prediction network $\Veps_\theta$ in Equation~\ref{eq:noise_prediction}~\refofpaper{} without the condition encoder $\C{E}_\theta$ and cross-attention layers in the decoder $\C{D}_\theta$. Given the unconditional diffusion model, the set of input bounding boxes $\mathcal{B}_s$, and the sampled pivot box $b_v \in \mathcal{B}_s$, we first duplicate $b_v$, increasing the total number of boxes to $| \mathcal{B}_s | + 1$, and then perform the DDIM inversion~\cite{dhariwal2021diffusion}, obtaining the standard normal sample $\B{x}_T$ from the input boxes. Next, we reset the portion of $\B{x}_T$ corresponding to the duplicates of $b_v$ to random standard normal samples. Then, we perform inpainting while treating $\C{B}_s \setminus \{b_v\}$ as the background.


\paragraph{Conditional Token Prediction Model.}

While sequence generation models have shown remarkable capabilities in 3D shape generation~\cite{siddiqui2024meshgpt,tang2024edgerunner, Gao:2022Get3D, Nash:2020Polygen}, their application to hierarchical box-splitting poses unique challenges. Unlike traditional sequence generation tasks that generate 3D shapes, our goal is to learn the conditional probability of splitting one pivot node into two child nodes while removing the selected pivot. 

Since typical GPT-like models require a discretized representation to model categorical distributions, we quantize our continuous 15-dimensional box vector representation using VQ-VAE~\cite{Van:2017VQ-VAE}. The VQ-VAE~\cite{Van:2017VQ-VAE} is trained to encode input box vectors into a token space, with the encoded tokens being decoded back into the original vector. The VQ-VAE consists of simple MLPs. Following MeshGPT~\cite{siddiqui2024meshgpt}, we also incorporate a residual quantization technique~\cite{Juang:1982MultipleStageVQ,Martinez:2014Stacked}, where an input latent is discretized using a stack of $D$ ordered codes. We set $D=2$, and the number of codes, $|V|$, is set to 16,384.

For sequence modeling, we adapt the network of the conditional diffusion model to output the logits of two elements $[l_1, l_2] \in \mathbb{R}^{2\times |V|}$ in the token space instead of predicting their noise. Unlike traditional sequence generation tasks, our splitting process has unique characteristics: (1) it does not impose any order on the boxes—for instance, the selected pivot box can be an intermediate token in the input sequence, and (2) the splitting process generates two tokens simultaneously. To address these requirements, we introduce essential modifications to the standard sequence generation model training and inference approaches. For non-sequential autoregressive modeling, we forgo positional encoding techniques and use the Transformer~\cite{Vaswani:2017Attention} architecture to ensure order invariance. To predict two logits without considering their order, the training objective computes the cross-entropy loss for both possible orderings of the ground truth tokens and selects the minimum loss. The loss function is defined as:
$\C{L} = \min (\text{CE}([l_1, l_2], [v_1, v_2]), \text{CE}([l_1, l_2], [v_2, v_1]))$, where $\text{CE}$ represents the cross-entropy loss, and $v_1$ and $v_2$ are the ground truth token indices corresponding to $l_1$ and $l_2$, respectively. At inference time, we first sample the first token. When sampling the second token, we mask out the index of the first predicted token to ensure that the same box is not sampled again.



\begin{table*}[t!]
\centering
\caption{\textbf{Dataset statistics.} We report the number of shapes and the average number of bounding boxes for each class.}
\label{tab:datasets_stats}
\scriptsize
{
\setlength{\tabcolsep}{0.1em}
\renewcommand{\arraystretch}{1.0}

\begin{tabularx}{\linewidth}{>{\centering\arraybackslash}m{2.1cm} Y Y Y  >{\centering\arraybackslash}m{1cm} Y Y Y Y}
\toprule
\textbf{Class} & Table & Chair & Couch & Airplane  & Bench & Display & Rifle & Lamp \\
\midrule
\makecell{\# of shapes} & 2,725 & 1,976 & 1,108 & 420 & 409 & 407 & 398 & 322 \\
\midrule 
\makecell{Avg. \# of Boxes} & 6.96 & 9.54 & 5.37 & 11.46 & 8.92 & 3.98 & 8.05 & 7.46  \\
\bottomrule
\end{tabularx}

}
\end{table*}

\subsection{Implementation Details}
\label{sec:implementation_details}

\paragraph{Child-Boxes Diffusion.}
As discussed in Section~\ref{sec:child_box_diffusion}~\refofpaper{}, the noise prediction network of Child-Boxes Diffusion $\Veps_\theta$ consists of a Transformer encoder $\C{E}_\theta$ and a decoder $\C{D}_\theta$. The encoder consists of $6$ self-attention layers with a hidden dimension of $512$. To indicate the pivot box $b_v \in \C{B}_s$, we use a class embedding $\B{e}_c \in \mathbb{R}^{|\C{B}_s| \times 512}$ encoded from an indicator highlighting the pivot box's index. Additionally, we also encode the number of input boxes $|\C{B}_s|$ into a cardinality embedding $\B{e}_d \in \mathbb{R}^{512}$. These two embeddings are added to the output of each self-attention layer, yielding the final encoder output: $\B{h} = \C{E}_\theta(\C{B}_s, b_v, |\C{B}_s|) \in \mathbb{R}^{|\C{B}_s| \times 512}$. The decoder $\C{D}_\theta$ has a similar architecture to the encoder, with each self-attention layer followed by a cross-attention layer. The condition latent $\B{h}$ is fed as the key and value in every cross-attention layer, while the noisy two child boxes $\B{x}_t \in \mathbb{R}^{2 \times 15}$ are fed as query. We set a learning rate and batch size to $8e^{-4}$ and $2048$, respectively. For sampling, we use the DDIM~\cite{Song:2021DDIM} deterministic sampling process with 50 steps.

\paragraph{Inpainting with Unconditional Diffusion Models.}
The unconditional diffusion model for the inpainting technique adopts a similar architecture to that of the conditional diffusion model but without the decoder part. The learning rate, batch size and number of training epochs are the same as those used in \Ours{}. We use 50 steps for both DDIM inversion~\cite{dhariwal2021diffusion} and denoising process with masking.

\paragraph{Conditional Token Prediction Model.}
The conditional token prediction model adopts a similar architecture to the conditional diffusion model, with key modifications. The timestep embedding from the original network is removed. To condition the network on $\C{B}_s$, its discretized representation is obtained by encoding the boxes into quantized vectors using the pre-trained VQ-VAE, which are then fed into the encoder $\C{E}_\theta$. Two learnable tokens are initialized and passed through the cross-attention layers in the decoder, where these tokens attend to the quantized vectors. The final linear layer of the network is modified to output logits of dimension $|V|$. The same learning rate, batch size, and number of training epochs as those used in the conditional diffusion model are used for training.

\paragraph{Pivot Classifier.}
Similar to the architecture of the diffusion models, we utilize a Transformer encoder~\cite{Vaswani:2017Attention} to model the categorical distribution $p(b_v | \C{B}_s)$ introduced in Section~\ref{sec:overview}~\refofpaper{}. The encoder contains $6$ self-attention layers, each with a hidden dimension of $512$ and $4$ attention heads. An MLP layer follows the final self-attention layer, mapping the $512$-dimensional latent of each box to a scalar logit. To process inputs with a varying number of bounding boxes, we condition the network on the current number of boxes $| \C{B}_s |$ via adaLN-layers~\cite{Peebles:2023DiT}. We set the learning rate and batch size to $8e^{-4}$ and $2048$, respectively, and train the network for 100 epochs. To choose the pivot at the inference time, we sample the index of the pivot $v$ from a learned categorical distribution $p(b_v | \C{B}_s)$.

\begin{table*}[t!]
\centering
\caption{\textbf{Quantitative comparison of shape abstraction generation with different pivot selection strategies.} MMD-CD scores and MMD-EMD scores are scaled by $10^3$ and $10^2$, respectively. The best results are highlighted in \textbf{bold}. The numbers are the averages across split levels $s=5$ and $s=8$.}
\label{tab:stage_1_avg_5_8}
\scriptsize
{
\setlength{\tabcolsep}{0.1em}
\renewcommand{\arraystretch}{1.0}
\definecolor{LightCyan}{rgb}{0.88,1,1}
\definecolor{Gray}{gray}{0.85}
\begin{tabularx}{\linewidth}{>{\centering\arraybackslash}m{1.5cm} >{\centering\arraybackslash}m{2.5cm}  Z Z Z Z Z Z Z Z Z Z Z Z}

\toprule

{ } &
  { } &
  \multicolumn{2}{c}{{ \textbf{COV ↑}}} &
  \multicolumn{2}{c}{{ \textbf{MMD ↓}}} &
  \multicolumn{2}{c}{{ \textbf{1-NNA ↓}}} &
  \multicolumn{2}{c}{{ \textbf{COV ↑}}} &
  \multicolumn{2}{c}{{ \textbf{MMD ↓}}} &
  \multicolumn{2}{c}{{ \textbf{1-NNA ↓}}} \\

  \cmidrule(lr){3-8} \cmidrule(lr){9-14}

  \multirow{-2}[2]{*}{{ \makecell{\textbf{Pivot}\\\textbf{Selection}}}} &
  \multirow{-2}[2]{*}{{ \textbf{Models}}} &
  { \textbf{CD}} &
  { \textbf{EMD}} &
  { \textbf{CD}} &
  { \textbf{EMD}} &
  { \textbf{CD}} &
  { \textbf{EMD}} &
  { \textbf{CD}} &
  { \textbf{EMD}} &
  { \textbf{CD}} &
  { \textbf{EMD}} &
  { \textbf{CD}} &
  { \textbf{EMD}} \\

  \midrule
  
  {} & {} & \multicolumn{6}{c}{{ \textbf{Chair}}} & \multicolumn{6}{c}{{ \textbf{Airplane}}} \\
  \midrule

\multirow{3}{*}{Random} & Token Prediction Model & 22.05 & 25.76 & 27.960 & 21.335 & 94.13 & 93.02 & 55.25 & 58.80 & 10.645 & 15.680 & 90.27 & 90.08 \\

{ } & Uncond. Diffusion & 29.70 & 30.97 & 18.784 & 18.128 & 88.68 & 87.93 & 63.82 & 68.21 & 7.207 & 12.953 & 88.50 & 87.90 \\

{ } & \textbf{Cond. Diffusion (Ours)} & \textbf{32.84} & \textbf{33.97} & \textbf{16.896} & \textbf{16.910} & \textbf{85.28} & \textbf{83.43} & \textbf{77.75} & \textbf{77.38} & \textbf{6.842} & \textbf{12.599} & \textbf{86.35} & \textbf{85.64} \\
    
\midrule

\multirow{3}{*}{Classifier} & Token Prediction Model & 27.41 & 30.87 & 24.615 & 20.235 & 91.39 & 90.19 & 66.87 & 70.41 & 8.185 & 13.870 & 87.16 & 86.49 \\

{ } & { Uncond. Diffusion} & 33.03 & 35.68 & 18.174 & 17.598 & 88.53 & 85.77 & 71.76 & 77.75 & 6.811 & 12.505 & 86.15 & 85.55 \\

{ } &{\textbf{Cond. Diffusion (Ours)}} & \textbf{46.08} & \textbf{47.08} & \textbf{14.166} & \textbf{15.580} & \textbf{75.87} & \textbf{73.33} & \textbf{82.89} & \textbf{80.56} & \textbf{6.478} & \textbf{12.188} & \textbf{85.62} & \textbf{84.81} \\

\midrule 

{ } & { } & \multicolumn{6}{c}{{ \textbf{Table}}} & \multicolumn{6}{c}{{ \textbf{Rifle}}} \\

\midrule

\multirow{3}{*}{Random} & Token Prediction Model & 18.77 & 17.24 & 32.175 & 22.580 & 90.77 & 92.59 & 57.42 & 58.67 & 3.400 & 9.475 & {87.18} & 86.72 \\

{ } & Uncond. Diffusion & 25.05 & 25.89 & 21.840 & 18.784 & 83.79 & 84.94 & \textbf{66.46} & \textbf{70.48} & 3.015 & 8.813 & \textbf{84.88} & \textbf{84.86} \\

{ } & \textbf{Cond. Diffusion (Ours)} & \textbf{30.17} & \textbf{30.58} & \textbf{14.229} & \textbf{15.132} & \textbf{78.22} & \textbf{79.22} & {64.95} & {68.97} & \textbf{2.716} & \textbf{8.285} & 87.28 & {85.65} \\

\midrule

\multirow{3}{*}{Classifier} & Token Prediction Model & 25.45 & 24.47 & 25.895 & 19.680 & 87.47 & 88.72 & 68.72 & 71.86 & 3.495 & 9.230 & 84.28 & 82.67 \\

{ } & Uncond. Diffusion & 30.36 & 31.68 & 17.726 & 17.104 & 81.96 & 82.26 & \textbf{75.38} & \textbf{79.02} & 3.110 & 8.717 & \textbf{79.17} & \textbf{79.96} \\

{ } & \textbf{Cond. Diffusion (Ours)} & \textbf{36.79} & \textbf{37.96} & \textbf{12.314} & \textbf{14.218} & \textbf{71.38} & \textbf{72.82} & {74.75} & {75.25} & \textbf{2.607} & \textbf{8.076} & {82.86} & {81.96} \\

\midrule 

{ } & { } & \multicolumn{6}{c}{{ \textbf{Couch}}} & \multicolumn{6}{c}{{ \textbf{Bench}}} \\

\midrule

\multirow{3}{*}{Random} & Token Prediction Model & 44.80 & 40.73 & 12.810 & 13.750 & 87.41 & 89.30 & \textbf{67.45} & 60.11 & 11.155 & 14.520 & \textbf{87.15} & 92.40 \\

{ } & Uncond. Diffusion & 56.56 & 55.42 & 9.127 & 11.296 & \textbf{74.66} & \textbf{77.95} & 64.69 & 62.33 & \textbf{9.318} & \textbf{12.660} & 88.45 & \textbf{89.73} \\

{ } & \textbf{Cond. Diffusion (Ours)} & \textbf{58.18} & \textbf{56.28} & \textbf{8.677} & \textbf{11.220} & {77.68} & {79.70} & 64.43 & \textbf{69.68} & {10.630} & {13.167} & 87.23 & {90.43} \\
  
\midrule

\multirow{3}{*}{Classifier} & Token Prediction Model & 48.91 & 42.00 & 12.600 & 13.545 & 85.19 & 88.12 & \textbf{78.22} & 70.08 & 10.535 & 13.450 & 85.53 & 89.84 \\

{ } & Uncond. Diffusion & 56.51 & 54.71 & \textbf{9.527} & \textbf{11.497} & \textbf{75.19} & \textbf{77.77} & 67.85 & 69.82 & \textbf{9.195} & \textbf{12.266} & 86.12 & 87.29 \\

{ } & \textbf{Cond. Diffusion (Ours)} & \textbf{59.40} & \textbf{57.91} & {9.724} & {12.743} & {84.40} & {85.00} & 76.12 & \textbf{78.35} & {9.764} & {13.222} & \textbf{85.47} & \textbf{86.54} \\
    
\midrule

{ } & { } & \multicolumn{6}{c}{{ \textbf{Lamp}}} & \multicolumn{6}{c}{{ \textbf{Display}}} \\

\midrule

\multirow{3}{*}{Random} & Token Prediction Model & 65.09 & 64.75 & {25.340} & 22.940 & 87.78 & 87.08 & 50.00 & 58.84 & 14.010 & 15.605 & 92.52 & 92.64 \\

{ } & Uncond. Diffusion & 71.36 & 71.69 & \textbf{18.849} & \textbf{19.921} & \textbf{84.10} & 84.93 & 73.46 & 75.43 & \textbf{10.042} & \textbf{12.287} & \textbf{77.88} & \textbf{78.81} \\
  
{ } & \textbf{Cond. Diffusion (Ours)} & \textbf{80.00} & \textbf{79.83} & 25.810 & 21.988 & {87.01} & \textbf{83.94} & \textbf{74.57} & \textbf{80.47} & {11.492} & {13.368} & {84.34} & {84.36} \\

\midrule

\multirow{3}{*}{Classifier} & Token Prediction Model & 73.73 & 76.95 & 24.640 & 22.500 & \textbf{87.54} & \textbf{86.14} & 51.23 & 58.60 & 14.860 & 16.175 & 92.06 & 92.09 \\ 

{ } & Uncond. Diffusion & 67.79 & 71.36 & \textbf{22.328} & \textbf{21.207} & 87.80 & 86.60 & 72.48 & \textbf{74.44} & \textbf{12.702} & \textbf{13.345} & \textbf{80.21} & \textbf{81.05} \\

{ } & \textbf{Cond. Diffusion (Ours)} & \textbf{79.49} & \textbf{82.20} & 27.892 & 23.571 & 89.02 & 86.93 & \textbf{76.17} & {72.97} & 15.325 & 15.688 & {85.71} & {88.41} \\

\bottomrule

\end{tabularx}
}
\end{table*}
\begin{table*}[t!]
\centering
\caption{\textbf{Quantitative comparison of shape abstraction generation using the pivot classifier for pivot selection. Results are evaluated at split levels $s$=5 and $s$=8.} MMD-CD scores and MMD-EMD scores are scaled by $10^3$ and $10^2$, respectively. The best results are highlighted in \textbf{bold}.}
\label{tab:stage_1_pivot_cls_5_8}
\scriptsize
{
\setlength{\tabcolsep}{0.1em}
\renewcommand{\arraystretch}{1.0}
\definecolor{LightCyan}{rgb}{0.88,1,1}
\definecolor{Gray}{gray}{0.85}
\begin{tabularx}{\linewidth}{>{\centering\arraybackslash}m{0.8cm}@{}>{\centering\arraybackslash}m{2.5cm}@{}@{} Z Z Z Z Z Z Z Z Z Z Z Z}

\toprule
  
{} &
{} &
  \multicolumn{2}{c}{\textbf{COV ↑}} &
  \multicolumn{2}{c}{\textbf{MMD ↓}} &
  \multicolumn{2}{c}{\textbf{1-NNA ↓}} &
  \multicolumn{2}{c}{\textbf{COV ↑}} &
  \multicolumn{2}{c}{\textbf{MMD ↓}} &
  \multicolumn{2}{c}{\textbf{1-NNA ↓}} \\

  \cmidrule(lr){3-8} \cmidrule(lr){9-14}
  
  \multirow{-2}[2]{*}{{$s$}} &
  \multirow{-2}[2]{*}{{\textbf{Models}}} &
  \textbf{CD} &
  \textbf{EMD} &
  \textbf{CD} &
  \textbf{EMD} &
  \textbf{CD} &
  \textbf{EMD} &
  \textbf{CD} &
  \textbf{EMD} &
  \textbf{CD} &
  \textbf{EMD} &
  \textbf{CD} &
  \textbf{EMD} \\

\midrule

  {} & 
  {} &
  \multicolumn{6}{c}{\textbf{Chair}} &
  \multicolumn{6}{c}{\textbf{Airplane}} \\

\midrule

\multirow{3}{*}{5} &
  {Token Prediction Model} &
    29.24 & 31.73 & 23.008 & 19.648 & 90.83 & 89.64 & 65.04 & 69.44 & 7.722 & 13.552 & 87.59 & 86.47 \\ &

  Uncond. Diff. &
    33.62 & 36.49 & 18.802 & 17.843 & 89.48 & 87.45 & 74.33 & 78.73 & 6.631 & 12.464 & 86.01 & 85.55 \\ &

  \textbf{Ours} &
    \textbf{45.08} & \textbf{45.14} & \textbf{14.408} & \textbf{15.690} & \textbf{78.23} & \textbf{75.71} & \textbf{82.40} & \textbf{79.95} & \textbf{6.35} & \textbf{12.10} & \textbf{85.51} & \textbf{85.47} \\

\midrule
  
\multirow{3}{*}{8} &
  {Token Prediction Model} &
    25.57 & 30.00 & 26.224 & 20.817 & 91.95 & 90.75 & 68.70 & 71.39 & 8.647 & 14.186 & 86.72 & 86.51 \\ & 

  Uncond. Diff. &
    32.43 & 34.86 & 17.547 & 17.353 & 87.58 & 84.10 & 69.19 & 76.77 & 6.990 & 12.546 & 86.30 & 85.55 \\ &

  \textbf{Ours} &
    \textbf{47.08} & \textbf{49.03} & \textbf{13.923} & \textbf{15.471} & \textbf{73.51} & \textbf{70.96} & \textbf{83.37} & \textbf{81.17} & \textbf{6.610} & \textbf{12.277} & \textbf{85.72} & \textbf{84.14} \\

\midrule
  
{} &
{} &
  \multicolumn{6}{c}{\textbf{Table}} &
  \multicolumn{6}{c}{\textbf{Rifle}} \\

  \midrule
  
\multirow{3}{*}{5} &

    {Token Prediction Model} &
    26.31 & 24.64 & 23.917 & 18.890 & 86.25 & 88.28 & 72.11 & 73.37 & 3.237 & 8.647 & 83.19 & 81.44 \\ &

  Uncond. Diff. &
    32.00 & 32.03 & 16.720 & 16.761 & 81.77 & 82.50 & \textbf{78.64} & \textbf{82.91} & 2.855 & 8.487 & \textbf{79.15} & \textbf{79.77} \\ &
  \textbf{Ours} &
    \textbf{36.34} & \textbf{37.50} & \textbf{12.203} & \textbf{14.087} & \textbf{70.56} & \textbf{73.18} & {73.62} & {75.13} & \textbf{2.476} & \textbf{7.912} & {83.90} & {82.49} \\

\midrule
  
\multirow{3}{*}{8} &

    {Token Prediction Model} &
    24.60 & 24.30 & 27.872 & 20.473 & 88.68 & 89.15 & 65.33 & 70.35 & 3.753 & 9.622 & 85.36 & 83.90 \\ &

  Uncond. Diff. &
    28.73 & 31.33 & 18.733 & 17.447 & 82.14 & 82.01 & 72.11 & 75.13 & 3.365 & 8.947 & \textbf{79.19} & \textbf{80.15} \\ &
  \textbf{Ours} &
    \textbf{37.24} & \textbf{38.42} & \textbf{12.426} & \textbf{14.349} & \textbf{72.20} & \textbf{72.46} & \textbf{75.88} & \textbf{75.38} & \textbf{2.738} & \textbf{8.239} & {81.82} & {81.44} \\

\midrule
  
{} &
{} &
  \multicolumn{6}{c}{\textbf{Couch}} &
  \multicolumn{6}{c}{\textbf{Bench}} \\

\midrule

\multirow{3}{*}{5} &

    {Token Prediction Model} &
    52.26 & 45.57 & 11.194 & 12.683 & 83.90 & 86.64 & \textbf{78.74} & 68.77 & 10.453 & 13.628 & 86.94 & 90.63 \\ &

  Uncond. Diff. &
    \textbf{59.95} & \textbf{58.23} & \textbf{8.979} & 11.150 & \textbf{73.47} & \textbf{76.05} & 70.34 & 71.65 & \textbf{8.734} & \textbf{11.957} & \textbf{84.71} & 87.23 \\ &
  \textbf{Ours} &
    {55.15} & {54.25} & {10.344} & \textbf{13.890} & {93.21} & {92.08} & {78.48} & \textbf{81.10} & {9.296} & {13.276} & {85.89} & \textbf{86.85} \\

\midrule

\multirow{3}{*}{8} &

    {Token Prediction Model} &
    45.57 & 38.43 & 14.007 & 14.405 & 86.48 & 89.60 & \textbf{77.69} & 71.39 & 10.623 & 13.269 & \textbf{84.12} & 89.04 \\ &

  Uncond. Diff. &
    53.07 & 51.18 & 10.075 & 11.845 & 76.92 & 79.49 & 65.35 & 67.98 & \textbf{9.657} & \textbf{12.574} & 87.53 & 87.36 \\ &
  \textbf{Ours} &
    \textbf{63.65} & \textbf{61.57} & \textbf{9.104} & \textbf{11.597} & \textbf{75.60} & \textbf{77.91} & {73.75} & \textbf{75.59} & {10.231} & {13.169} & {85.05} & \textbf{86.22} \\

\midrule
  
{} &
{} &
  \multicolumn{6}{c}{\textbf{Lamp}} &
  \multicolumn{6}{c}{\textbf{Display}} \\

\midrule

\multirow{3}{*}{5} &

    {Token Prediction Model} &
    \textbf{81.02} & 80.34 & 22.570 & {21.316} & \textbf{85.36} & \textbf{84.71} & 57.00 & 61.92 & 13.309 & 14.728 & 90.78 & 90.78 \\ &

  Uncond. Diff. &
    71.86 & 74.92 & \textbf{21.189} & \textbf{20.627} & 85.62 & 85.10 & 76.90 & \textbf{77.15} & \textbf{11.585} & \textbf{12.772} & \textbf{79.56} & \textbf{78.77} \\ &
  \textbf{Ours} &
    {79.32} & \textbf{81.69} & 26.600 & 23.504 & {90.28} & {88.85} & \textbf{78.38} & {71.74} & 14.300 & 15.668 & {85.71} & {89.78} \\

\midrule

\multirow{3}{*}{8} &

    {Token Prediction Model} &
    66.44 & 73.56 & 26.714 & 23.688 & 89.72 & 87.58 & 45.45 & 55.28 & 16.409 & 17.625 & 93.35 & 93.39 \\ &
    
  Uncond. Diff. &
    63.73 & 67.80 & \textbf{23.467} & \textbf{21.786} & 89.98 & 88.10 & 68.06 & 71.74 & \textbf{13.819} & \textbf{13.918} & \textbf{80.85} & \textbf{83.34} \\ &
  \textbf{Ours} &
    \textbf{79.66} & \textbf{82.71} & 29.185 & 23.638 & \textbf{87.76} & \textbf{85.01} & \textbf{73.96} &\textbf{74.20} & 16.349 & {15.707} & {85.71} & {87.04} \\

\bottomrule

\end{tabularx}
}
\end{table*}
\begin{table*}[t!]
\centering
\caption{\textbf{Quantitative comparison of shape abstraction generation with randomly selected pivots. Results are evaluated at $s$=5 and $s$=8.} MMD-CD scores and MMD-EMD scores are scaled by $10^3$ and $10^2$, respectively. The best results are highlighted in \textbf{bold}.}
\label{tab:stage_1_random_5_8}
\scriptsize
{
\setlength{\tabcolsep}{0.1em}
\renewcommand{\arraystretch}{1.0}
\definecolor{LightCyan}{rgb}{0.88,1,1}
\definecolor{Gray}{gray}{0.85}
\begin{tabularx}{\linewidth}{>{\centering\arraybackslash}m{0.8cm}@{}>{\centering\arraybackslash}m{2.5cm}@{}@{} Z Z Z Z Z Z Z Z Z Z Z Z}

\toprule
  
{} &
{} &
  \multicolumn{2}{c}{\textbf{COV ↑}} &
  \multicolumn{2}{c}{\textbf{MMD ↓}} &
  \multicolumn{2}{c}{\textbf{1-NNA ↓}} &
  \multicolumn{2}{c}{\textbf{COV ↑}} &
  \multicolumn{2}{c}{\textbf{MMD ↓}} &
  \multicolumn{2}{c}{\textbf{1-NNA ↓}} \\

  \cmidrule(lr){3-8} \cmidrule(lr){9-14}
  
  \multirow{-2}[2]{*}{{$s$}} &
  \multirow{-2}[2]{*}{{\textbf{Models}}} &
  \textbf{CD} &
  \textbf{EMD} &
  \textbf{CD} &
  \textbf{EMD} &
  \textbf{CD} &
  \textbf{EMD} &
  \textbf{CD} &
  \textbf{EMD} &
  \textbf{CD} &
  \textbf{EMD} &
  \textbf{CD} &
  \textbf{EMD} \\

\midrule

  {} & 
  {} &
  \multicolumn{6}{c}{\textbf{Chair}} &
  \multicolumn{6}{c}{\textbf{Airplane}} \\

\midrule

\multirow{3}{*}{5} &
  {Token Prediction Model} &

    23.84 & 27.89 & 25.074 & 20.215 & 93.74 & 92.26 & 55.50 & 57.21 & 9.049 & 14.620 & 90.62 & 90.20 \\ &

  Uncond. Diff. &
    29.30 & 30.81 & 19.471 & 18.489 & 88.75 & 87.97 & 64.30 & 67.73 & 7.236 & 13.011 & 87.84 & 88.17 \\ &

  \textbf{Ours} &
    \textbf{31.30} & \textbf{32.65} & \textbf{17.09} & \textbf{17.10} & \textbf{85.51} & \textbf{84.60} & \textbf{79.22} & \textbf{78.00} & \textbf{6.69} & \textbf{12.51} & \textbf{85.60} & \textbf{85.72} \\

\midrule
  
\multirow{3}{*}{8} &
  {Token Prediction Model} &

    20.27 & 23.62 & 30.850 & 22.453 & 94.52 & 93.77 & 55.01 & 60.39 & 12.238 & 16.745 & 89.91 & 89.95 \\ & 

  Uncond. Diff. &
    30.11 & 31.14 & 18.097 & 17.767 & 88.62 & 87.90 & 63.33 & 68.70 & 7.179 & 12.895 & 89.17 & 87.63 \\ &

  \textbf{Ours} &
    \textbf{34.38} & \textbf{35.30} & \textbf{16.70} & \textbf{16.72} & \textbf{85.04} & \textbf{82.26} & \textbf{76.28} & \textbf{76.77} & \textbf{7.00} & \textbf{12.68} & \textbf{87.09} & \textbf{85.55} \\

\midrule
  
{} &
{} &
  \multicolumn{6}{c}{\textbf{Table}} &
  \multicolumn{6}{c}{\textbf{Rifle}} \\

  \midrule
  
\multirow{3}{*}{5} &

    {Token Prediction Model} &

    20.44 & 18.02 & 29.948 & 21.385 & 90.00 & 91.94 & 61.56 & 61.81 & 3.140 & 9.087 & 86.95 & 86.28 \\ &

  Uncond. Diff. &
    26.42 & 26.57 & 20.893 & 18.473 & 82.24 & 83.95 & \textbf{68.09} & \textbf{71.61} & 2.901 & 8.661 & \textbf{84.95} & \textbf{84.82} \\ &

  \textbf{Ours} &
    \textbf{29.43} & \textbf{29.51} & \textbf{14.22} & \textbf{15.14} & \textbf{78.43} & \textbf{79.64} & {65.33} & {68.59} & \textbf{2.65} & \textbf{8.20} & {87.57} & {86.24} \\

\midrule
  
\multirow{3}{*}{8} &

    {Token Prediction Model} &

    17.09 & 16.46 & 34.404 & 23.779 & 91.54 & 93.24 & 53.27 & 55.53 & 3.658 & 9.855 & 87.41 & 87.16 \\ &

  Uncond. Diff. &
    23.67 & 25.20 & 22.786 & 19.095 & 85.35 & 85.93 & \textbf{64.82} & \textbf{69.35} & 3.129 & 8.966 & \textbf{84.82} & \textbf{84.90} \\ &

  \textbf{Ours} &
    \textbf{30.92} & \textbf{31.66} & \textbf{14.24} & \textbf{15.12} & \textbf{78.02} & \textbf{78.81} & {64.57} & \textbf{69.35} & \textbf{2.79} & \textbf{8.37} & 86.99 & 85.07 \\

\midrule
  
{} &
{} &
  \multicolumn{6}{c}{\textbf{Couch}} &
  \multicolumn{6}{c}{\textbf{Bench}} \\

\midrule

\multirow{3}{*}{5} &

    {Token Prediction Model} &

    46.75 & 43.85 & 11.463 & 12.907 & 85.90 & 87.86 & \textbf{68.50} & 61.94 & 10.223 & 13.839 & \textbf{87.19} & 92.48 \\ &

  Uncond. Diff. &
    57.69 & \textbf{57.23} & 8.617 & \textbf{10.975} & \textbf{74.69} & \textbf{77.66} & 66.40 & 61.94 & \textbf{8.958} & \textbf{12.478} & {87.40} & \textbf{89.54} \\ &

  \textbf{Ours} &
    \textbf{57.87} & 56.69 & \textbf{8.42} & 11.06 & 76.79 & 79.65 & 65.88 & \textbf{71.92} & 9.91 & 12.69 & 87.61 & 90.26 \\

\midrule

\multirow{3}{*}{8} &

    {Token Prediction Model} &

    42.86 & 37.61 & 14.165 & 14.588 & 88.92 & 90.73 & \textbf{66.40} & 58.27 & 12.086 & 15.199 & 87.11 & 92.31 \\ &

  Uncond. Diff. &
    55.42 & 53.62 & 9.638 & 11.617 & \textbf{74.63} & \textbf{78.24} & 62.99 & 62.73 & \textbf{9.677} & \textbf{12.841} & 89.50 & \textbf{89.92} \\ &

  \textbf{Ours} &
    \textbf{58.50} & \textbf{55.88} & \textbf{8.94} & \textbf{11.38} & 78.56 & 79.75 & 62.99 & \textbf{67.45} & 11.35 & 13.64 & \textbf{86.85} & 90.59 \\

\midrule
  
{} &
{} &
  \multicolumn{6}{c}{\textbf{Lamp}} &
  \multicolumn{6}{c}{\textbf{Display}} \\

\midrule

\multirow{3}{*}{5} &

    {Token Prediction Model} &

    70.51 & 70.51 & 21.714 & 21.384 & 86.19 & 85.66 & 54.79 & 64.62 & 12.061 & 14.487 & 90.36 & 90.57 \\ &

  Uncond. Diff. &
    74.92 & 74.92 & \textbf{17.812} & \textbf{19.154} & \textbf{82.75} & \textbf{83.31} & 76.17 & 79.85 & \textbf{9.535} & \textbf{11.862} & \textbf{75.99} & \textbf{77.52} \\ &

  \textbf{Ours} &
    \textbf{83.73} & \textbf{82.03} & 24.88 & 21.33 & 86.45 & 83.79 & \textbf{78.13} & \textbf{83.54} & 10.90 & 12.88 & 82.51 & 82.55 \\

\midrule

\multirow{3}{*}{8} &

    {Token Prediction Model} &

    59.66 & 58.98 & 28.974 & 24.502 & 89.37 & 88.50 & 45.21 & 53.07 & 15.964 & 16.718 & 94.68 & 94.72 \\ &
    
  Uncond. Diff. &
    67.80 & 68.47 & \textbf{19.887} & \textbf{20.689} & \textbf{85.45} & 86.54 & 70.76 & 71.01 & \textbf{10.550} & \textbf{12.712} & \textbf{79.77} & \textbf{80.10} \\ &

  \textbf{Ours} &
    \textbf{76.27} & \textbf{77.63} & 26.74 & 22.65 & 87.58 & \textbf{84.10} & \textbf{71.01} & \textbf{77.40} & 12.08 & 13.86 & 86.17 & 86.17 \\

\bottomrule

\end{tabularx}
}
\end{table*}

\subsection{Runtime Analysis}
\label{sec:runtime_analysis}
Our Child-Boxes Diffusion takes 0.33 seconds per split, and ~\Ours{} (1.8s) introduces marginal increase over 3DShape2VecSet~\cite{Zhang:2023Shape2Vec} (1.1s), ensuring an efficient, responsive design workflow. All measurements were taken on an RTX 3090 GPU using a $128^3$ resolution for Marching Cubes.

\subsection{Details on User-Interactive Generation Demo}
\label{sec:demo_details}
Our demonstration consists of two main components: a web-based viewer and an inference server. The viewer, implemented based on Three.js~\cite{Threejs} and Potree~\cite{schutz2015potree}, provides an intuitive interface to manipulate bounding boxes and generate 3D shapes directly in the web browser. The inference server processes API requests and serves results from three pre-trained models: the Pivot Classifier, Child-Boxes Diffusion, and \textsc{Box2Shape}. Guided by the Pivot Classifier, the viewer identifies the most suitable bounding box to split, and the Child-Boxes Diffusion subsequently performs the actual splitting. Finally, the resulting bounding boxes are used to generate complete 3D shapes via \textsc{Box2Shape}, whose predicted occupancy fields are converted into 3D meshes using Occupancy-Based Dual Contouring~\cite{hwang2024odc}. This integrated workflow enables an interactive and efficient approach to bounding box manipulation and high-fidelity 3D shape generation.

\subsection{More Quantitative Results of Box Splitting}
\label{sec:more_box_splitting_quantitative_results}
In this section, we present more quantitative results of box splitting generation, as discussed in Section~\ref{sec:experiment_results}~\refofpaper{}.
Table~\ref{tab:stage_1_avg_5_8} shows the average results across two split levels, $s=5$ and $s=8$, for all eight categories. Meanwhile, Tables~\ref{tab:stage_1_pivot_cls_5_8} and~\ref{tab:stage_1_random_5_8} provide the results for $s=5$ and $s=8$ separately, using the pivot classifier or random pivot selection, respectively.

\clearpage
\newpage 
\subsection{Qualitative Results of Split Sequence}
\label{sec:sequence_visualization}

\begin{figure*}[h!]
\centering
{
\includegraphics[width=\textwidth]{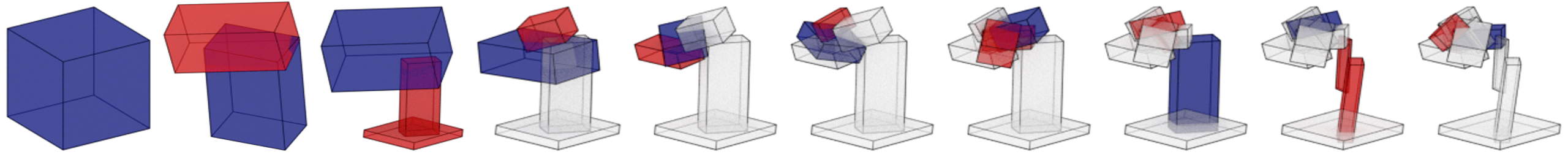} \\
\includegraphics[width=\textwidth]{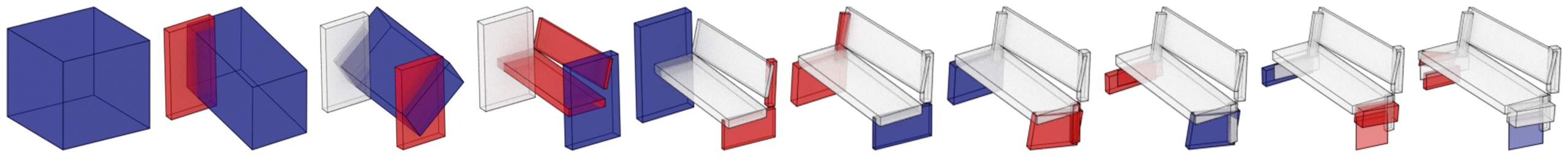} \\
\includegraphics[width=\textwidth]{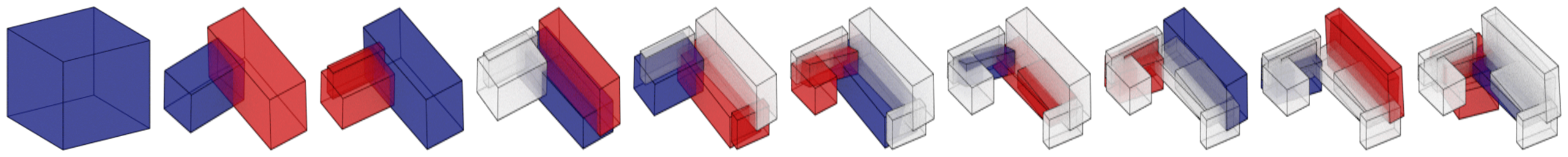} \\
\includegraphics[width=\textwidth]{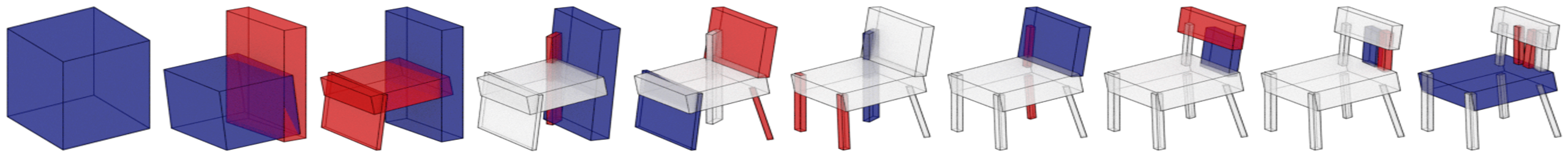} \\
\includegraphics[width=\textwidth]{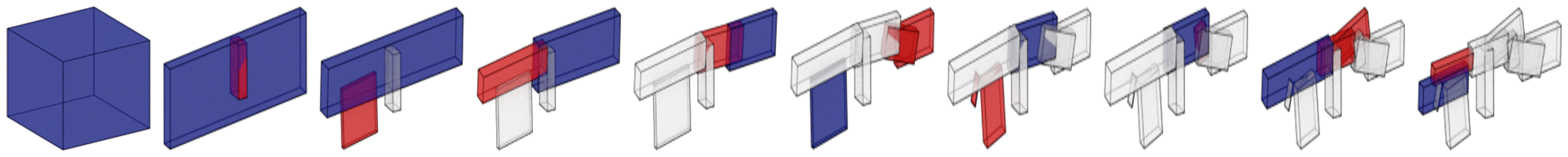} \\
\includegraphics[width=\textwidth]{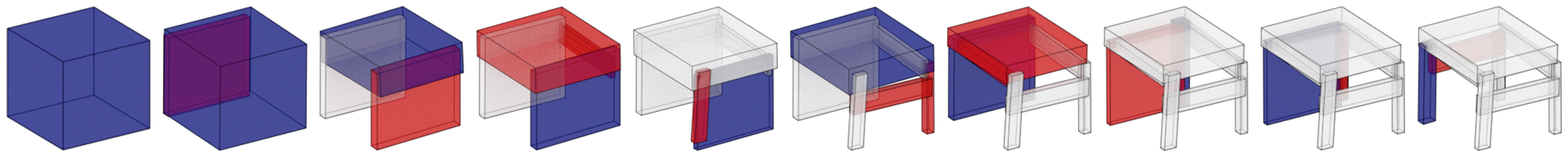} \\
\includegraphics[width=\textwidth]{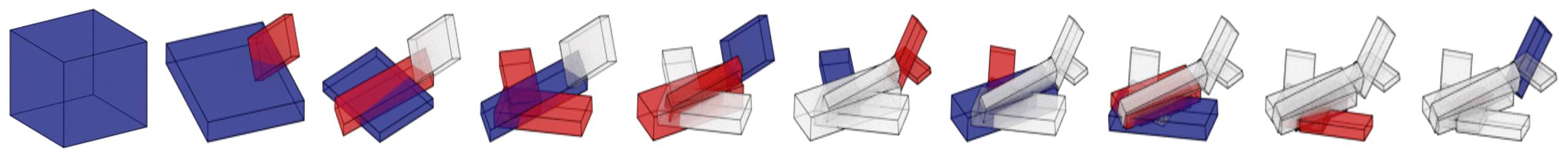}
\caption{\textbf{The evolution of box splits by ours.} In each abstraction, new generated boxes are marked in red, and the pivot to be split in the next split is in blue.}
\label{fig:sequence_visualization}
}

\end{figure*}

Figure~\ref{fig:sequence_visualization} shows the evolution of bounding box splitting by our method over multiple splits. Starting from an initial unit cube, the boxes are gradually split to capture finer details, transitioning from a coarse and simple structure to a more refined and complex 3D form. See the first and last rows of Figure~\ref{fig:sequence_visualization} that gradually produce the detail of the lamp shade and the airplane tail, respectively.

\begin{center}
\textbf{Sections for more qualitative results are\\in the following pages.}
\end{center}

\clearpage
\newpage
\subsection{More Qualitative Results of Box-Splitting Generation}
\label{sec:suppl_more_box_splitting_qualitative_results}
In Figure~\ref{fig:stage_1_comparison_more}, we present more qualitative results of box splitting generation and box-to-shape generation.

\begingroup
    \centering
\begin{figure*}[p!]
\centering
{
\scriptsize
\setlength{\tabcolsep}{0em}
\renewcommand\tabularxcolumn[1]{m{#1}}
\begin{tabularx}{\linewidth}{Y Y Y |  Y Y Y | Y Y Y | Y Y Y}
 \rotatebox{0}{\makecell{Token Pred.\\Model}} & \rotatebox{0}{\makecell{Unconod.\\Diffusion}} & \rotatebox{0}{\makecell{Cond.\\Diffusion}} & \rotatebox{0}{\makecell{Token Pred.\\Model}} & \rotatebox{0}{\makecell{Unconod.\\Diffusion}} & \rotatebox{0}{\makecell{Cond.\\Diffusion}} & \rotatebox{0}{\makecell{Token Pred.\\Model}} & \rotatebox{0}{\makecell{Unconod.\\Diffusion}} & \rotatebox{0}{\makecell{Cond.\\Diffusion}} & \rotatebox{0}{\makecell{Token Pred.\\Model}} & \rotatebox{0}{\makecell{Unconod.\\Diffusion}} & \rotatebox{0}{\makecell{Cond.\\Diffusion}} \\
\midrule

\multicolumn{3}{c|}{\includegraphics[width=.25\textwidth]{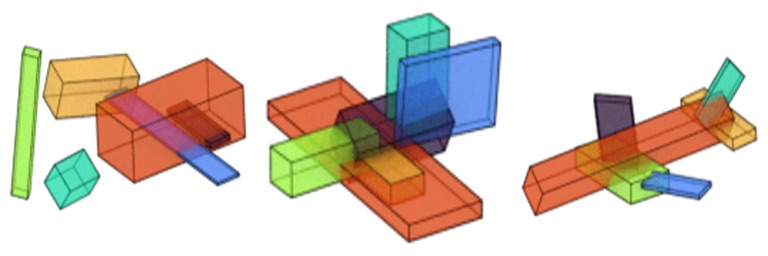}} &
\multicolumn{3}{c|}{\includegraphics[width=.25\textwidth]{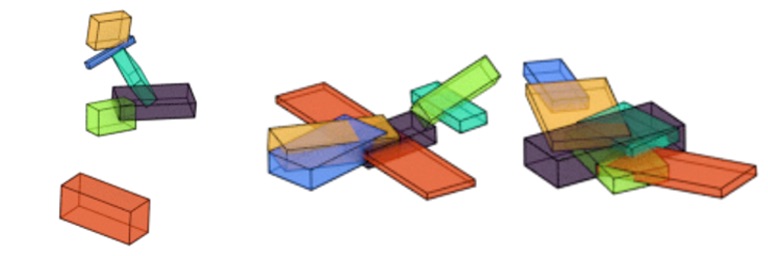}} &
\multicolumn{3}{c|}{\includegraphics[width=.25\textwidth]{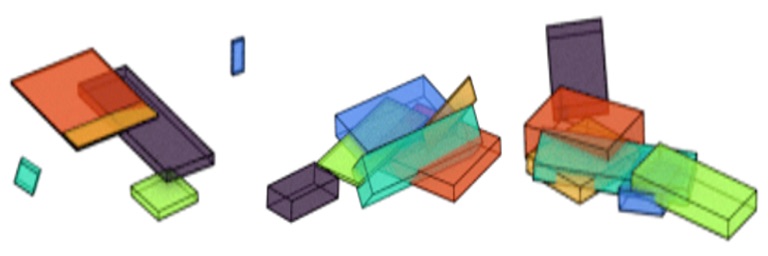}} &
\multicolumn{3}{c}{\includegraphics[width=.25\textwidth]{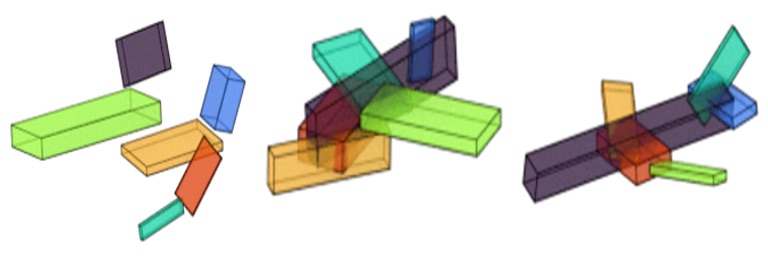}} \\
\multicolumn{3}{c|}{\includegraphics[width=.25\textwidth]{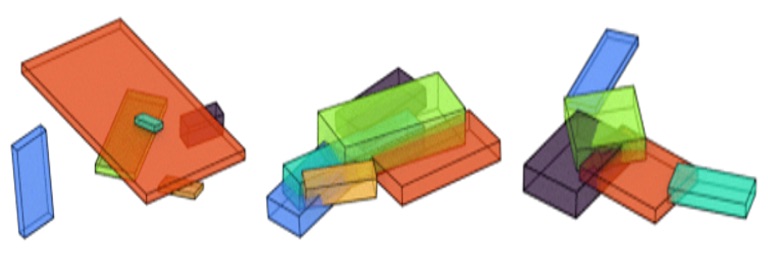}} &
\multicolumn{3}{c|}{\includegraphics[width=.25\textwidth]{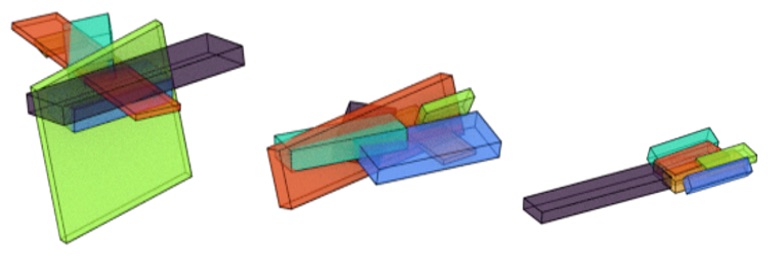}} &
\multicolumn{3}{c|}{\includegraphics[width=.25\textwidth]{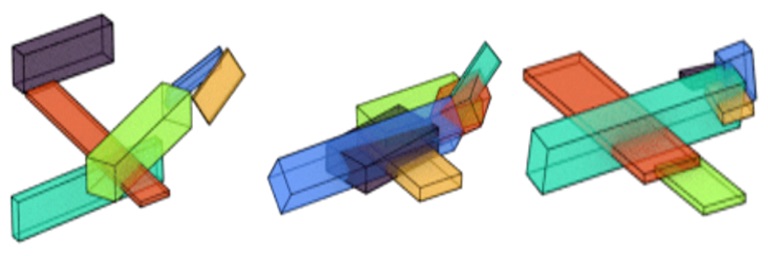}} &
\multicolumn{3}{c}{\includegraphics[width=.25\textwidth]{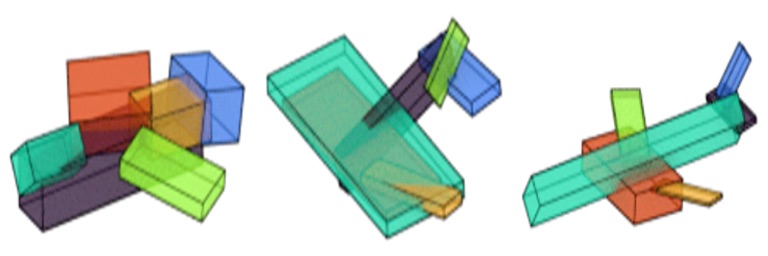}} \\
\multicolumn{3}{c|}{\includegraphics[width=.25\textwidth]{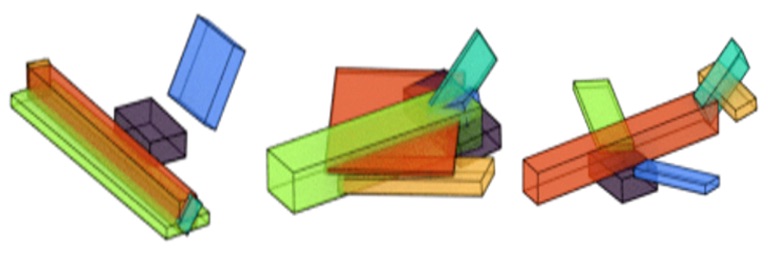}} &
\multicolumn{3}{c|}{\includegraphics[width=.25\textwidth]{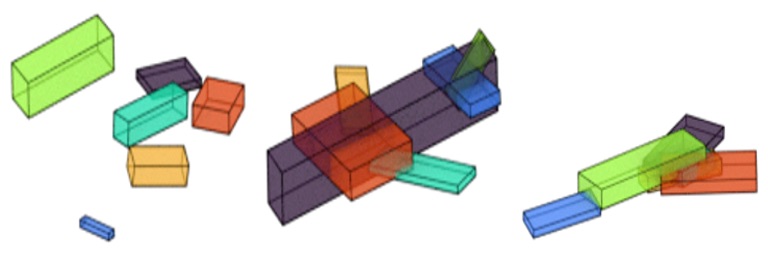}} &
\multicolumn{3}{c|}{\includegraphics[width=.25\textwidth]{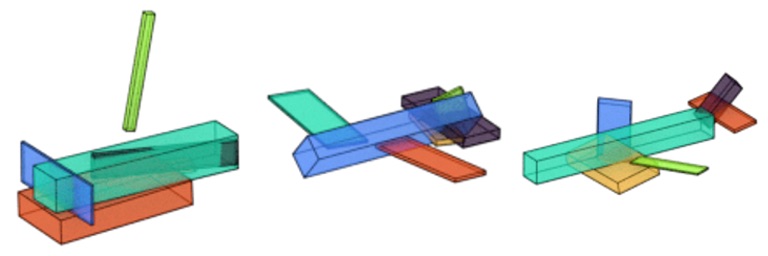}} &
\multicolumn{3}{c}{\includegraphics[width=.25\textwidth]{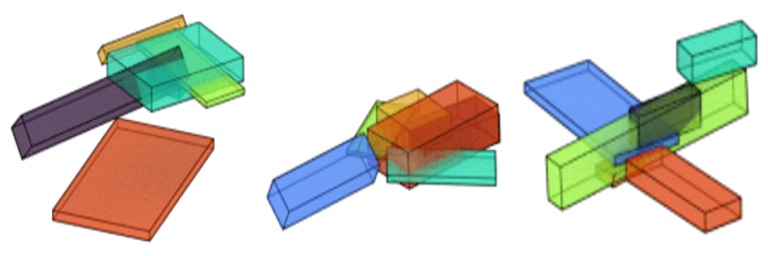}} \\
\multicolumn{3}{c|}{\includegraphics[width=.25\textwidth]{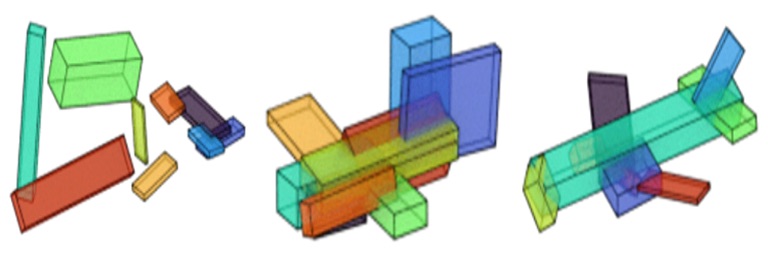}} &
\multicolumn{3}{c|}{\includegraphics[width=.25\textwidth]{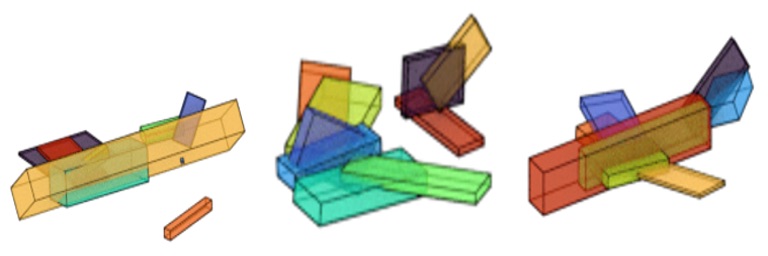}} &
\multicolumn{3}{c|}{\includegraphics[width=.25\textwidth]{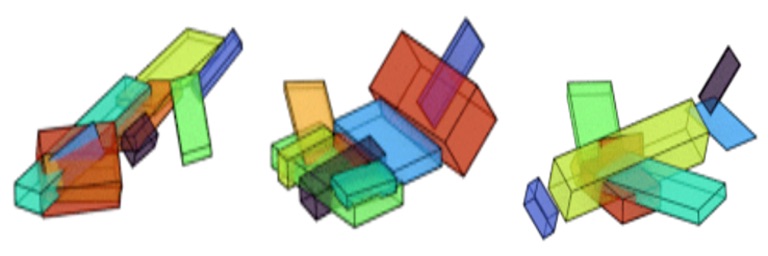}} &
\multicolumn{3}{c}{\includegraphics[width=.25\textwidth]{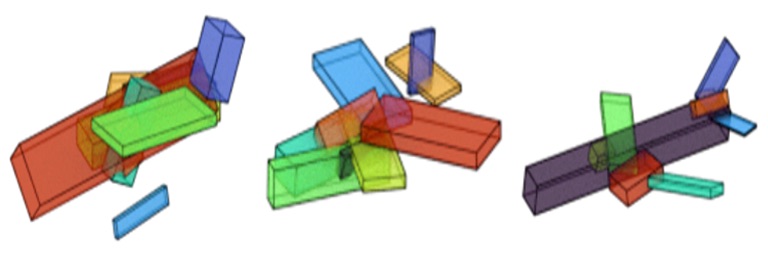}} \\
\multicolumn{3}{c|}{\includegraphics[width=.25\textwidth]{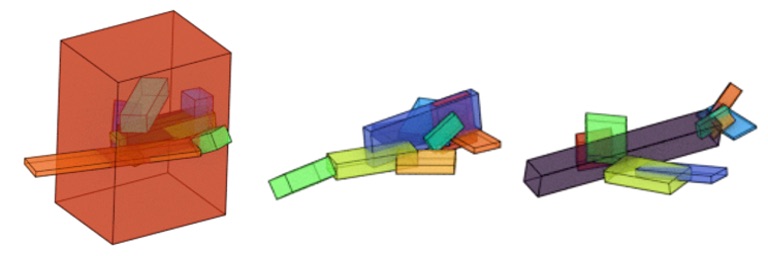}} &
\multicolumn{3}{c|}{\includegraphics[width=.25\textwidth]{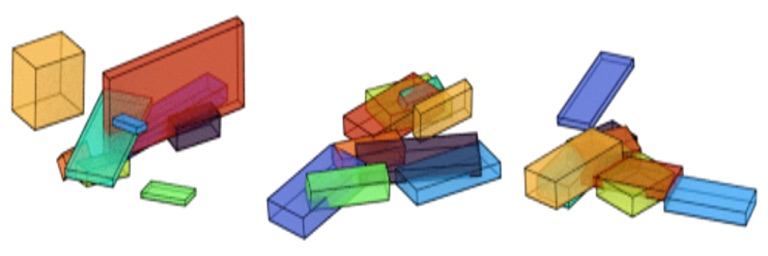}} &
\multicolumn{3}{c|}{\includegraphics[width=.25\textwidth]{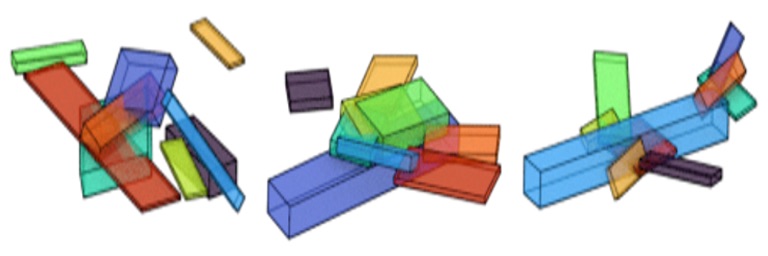}} &
\multicolumn{3}{c}{\includegraphics[width=.25\textwidth]{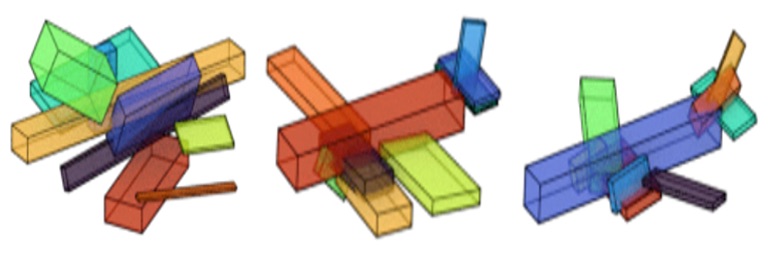}} \\
\multicolumn{3}{c|}{\includegraphics[width=.25\textwidth]{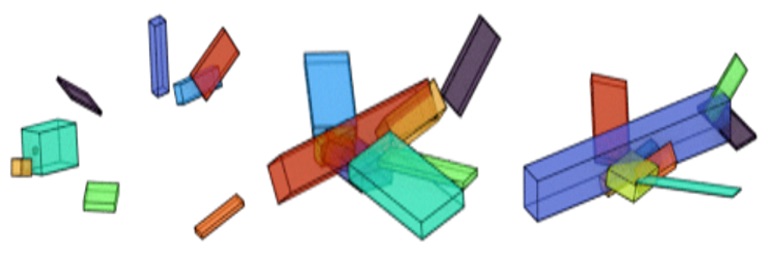}} &
\multicolumn{3}{c|}{\includegraphics[width=.25\textwidth]{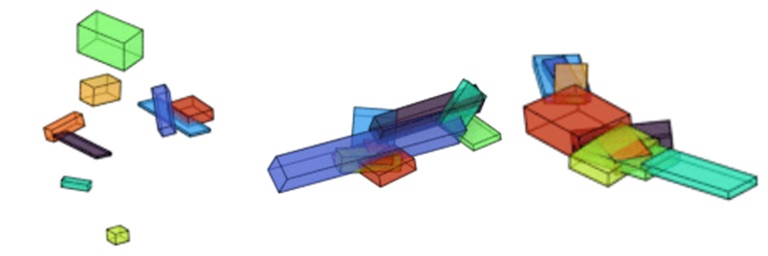}} &
\multicolumn{3}{c|}{\includegraphics[width=.25\textwidth]{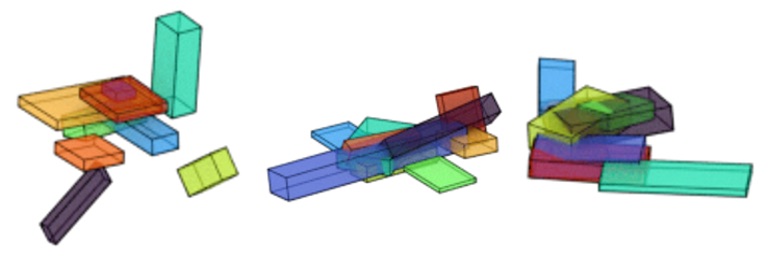}} &
\multicolumn{3}{c}{\includegraphics[width=.25\textwidth]{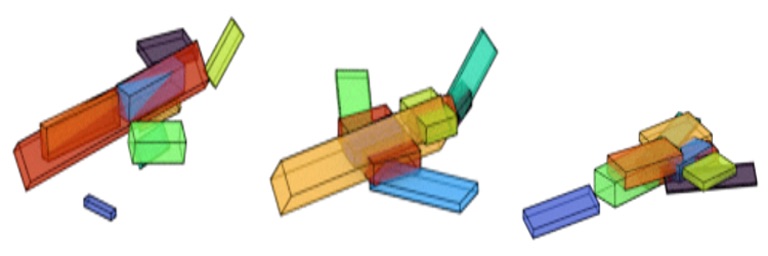}} \\
\multicolumn{3}{c|}{\includegraphics[width=.25\textwidth]{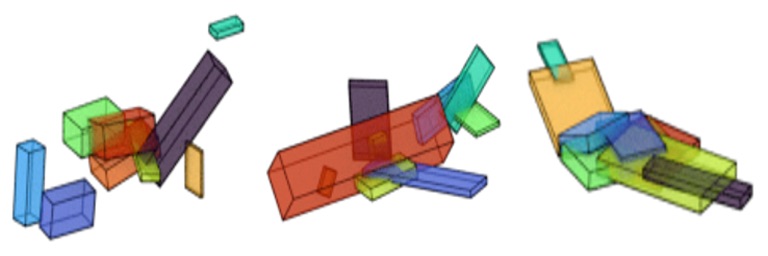}} &
\multicolumn{3}{c|}{\includegraphics[width=.25\textwidth]{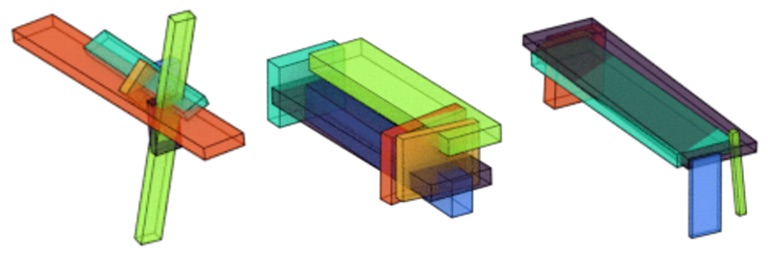}} &
\multicolumn{3}{c|}{\includegraphics[width=.25\textwidth]{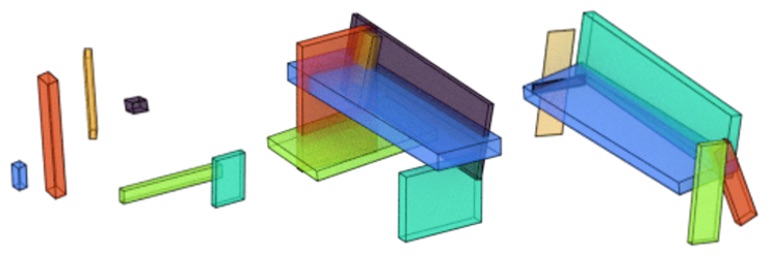}} &
\multicolumn{3}{c}{\includegraphics[width=.25\textwidth]{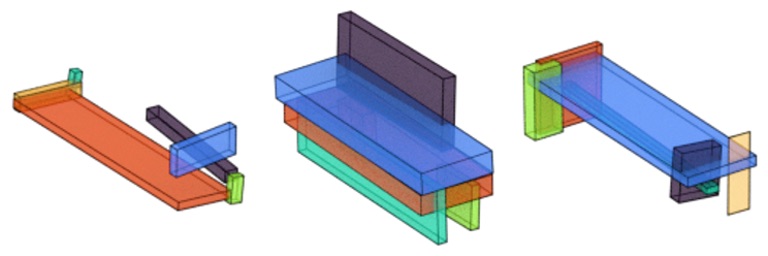}} \\
\multicolumn{3}{c|}{\includegraphics[width=.25\textwidth]{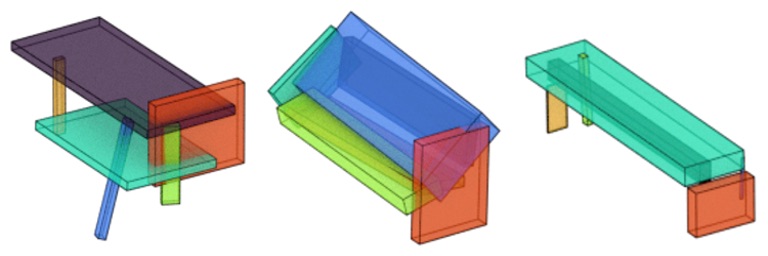}} &
\multicolumn{3}{c|}{\includegraphics[width=.25\textwidth]{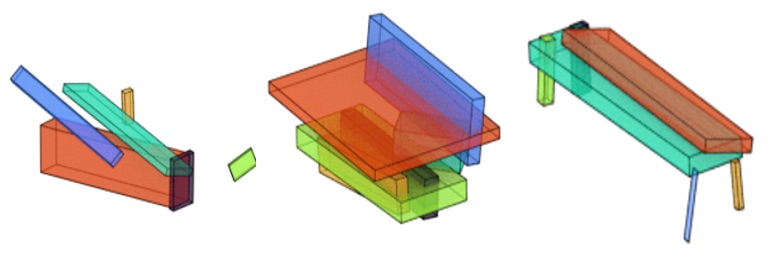}} &
\multicolumn{3}{c|}{\includegraphics[width=.25\textwidth]{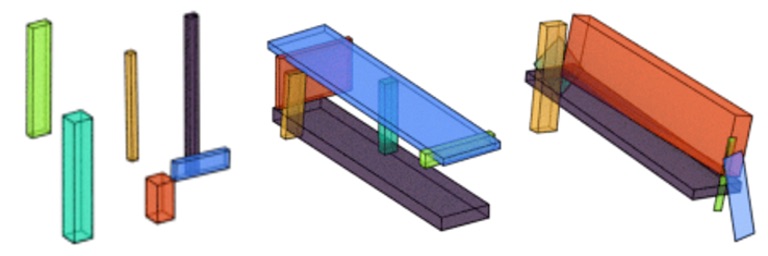}} &
\multicolumn{3}{c}{\includegraphics[width=.25\textwidth]{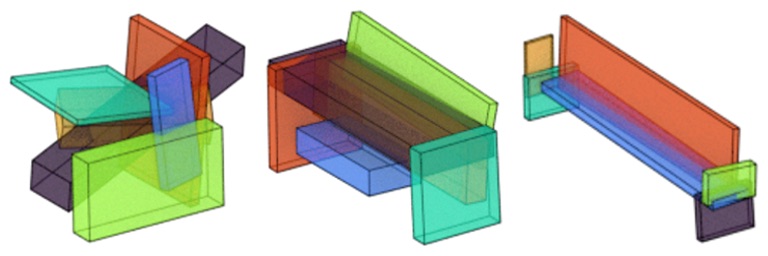}} \\
\multicolumn{3}{c|}{\includegraphics[width=.25\textwidth]{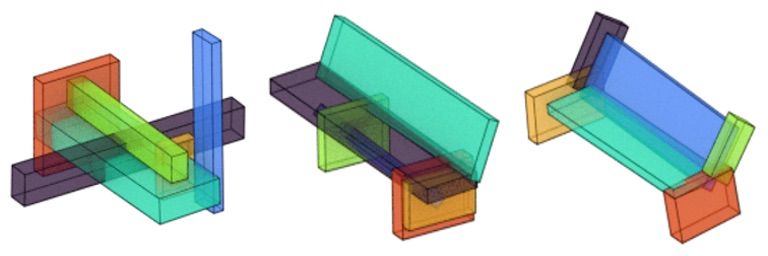}} &
\multicolumn{3}{c|}{\includegraphics[width=.25\textwidth]{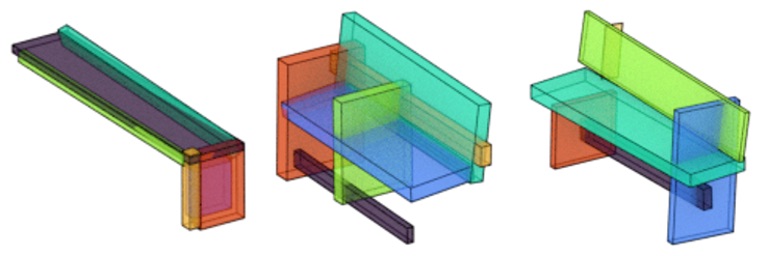}} &
\multicolumn{3}{c|}{\includegraphics[width=.25\textwidth]{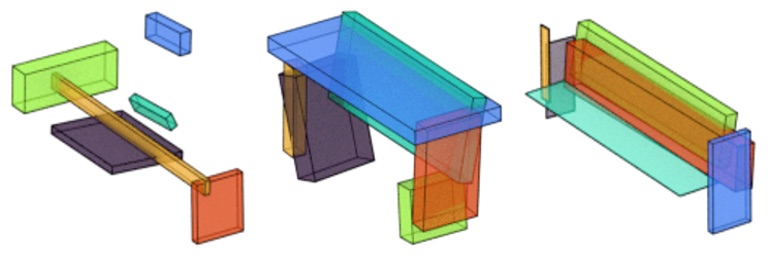}} &
\multicolumn{3}{c}{\includegraphics[width=.25\textwidth]{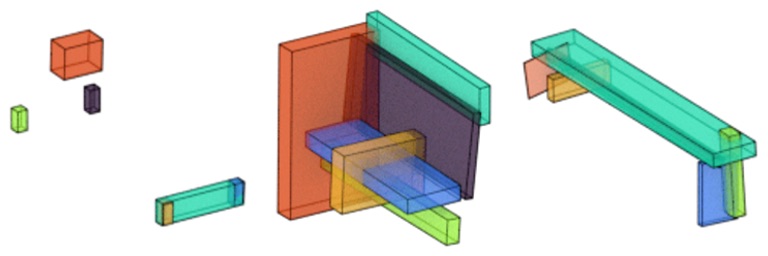}} \\
\multicolumn{3}{c|}{\includegraphics[width=.25\textwidth]{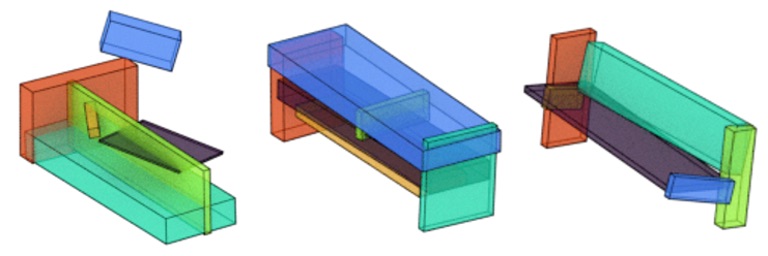}} &
\multicolumn{3}{c|}{\includegraphics[width=.25\textwidth]{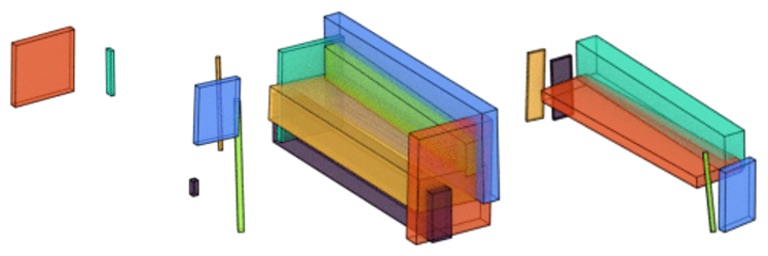}} &
\multicolumn{3}{c|}{\includegraphics[width=.25\textwidth]{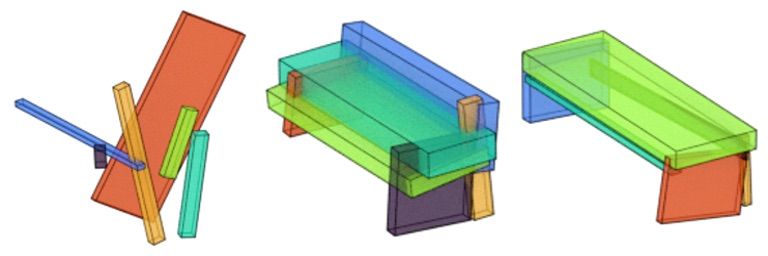}} &
\multicolumn{3}{c}{\includegraphics[width=.25\textwidth]{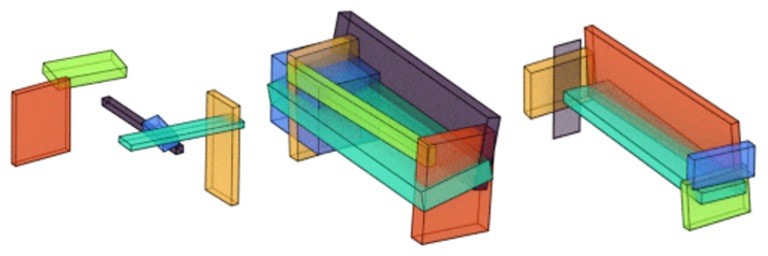}} \\
\multicolumn{3}{c|}{\includegraphics[width=.25\textwidth]{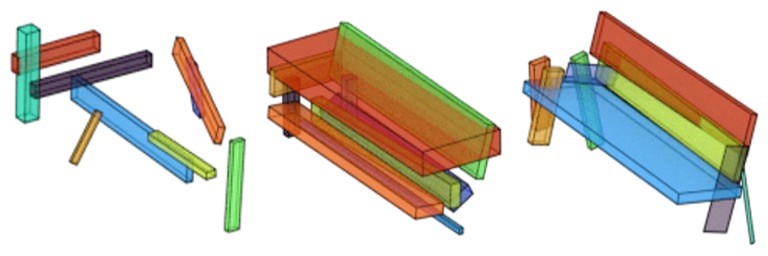}} &
\multicolumn{3}{c|}{\includegraphics[width=.25\textwidth]{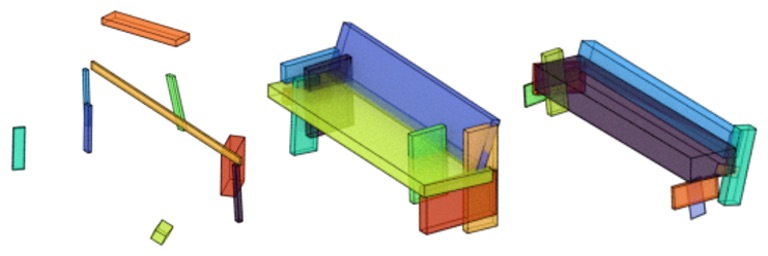}} &
\multicolumn{3}{c|}{\includegraphics[width=.25\textwidth]{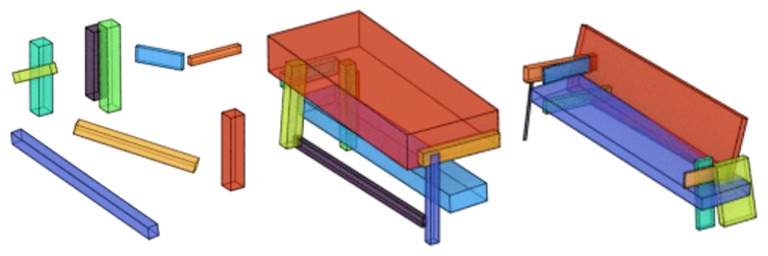}} &
\multicolumn{3}{c}{\includegraphics[width=.25\textwidth]{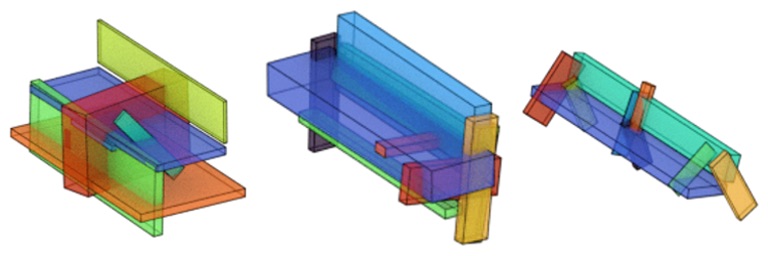}} \\
\multicolumn{3}{c|}{\includegraphics[width=.25\textwidth]{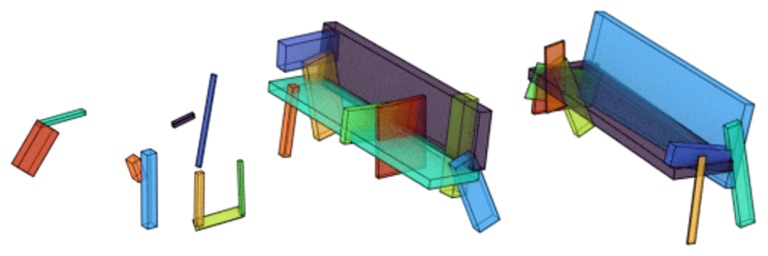}} &
\multicolumn{3}{c|}{\includegraphics[width=.25\textwidth]{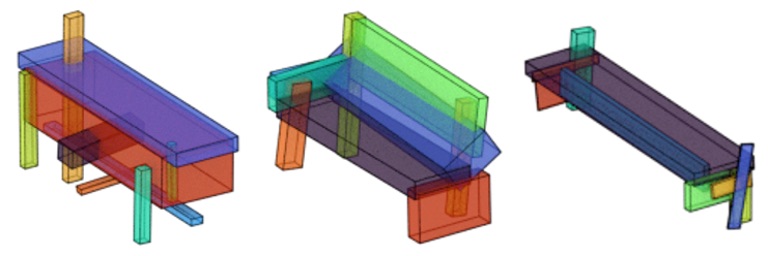}} &
\multicolumn{3}{c|}{\includegraphics[width=.25\textwidth]{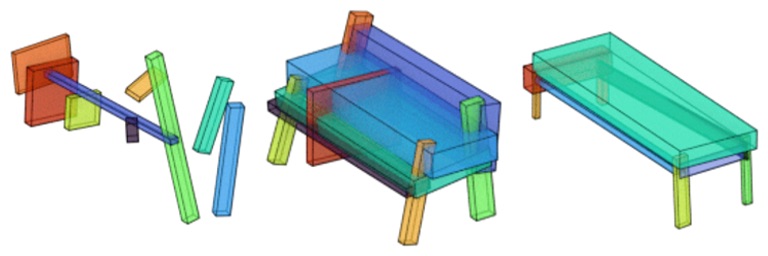}} &
\multicolumn{3}{c}{\includegraphics[width=.25\textwidth]{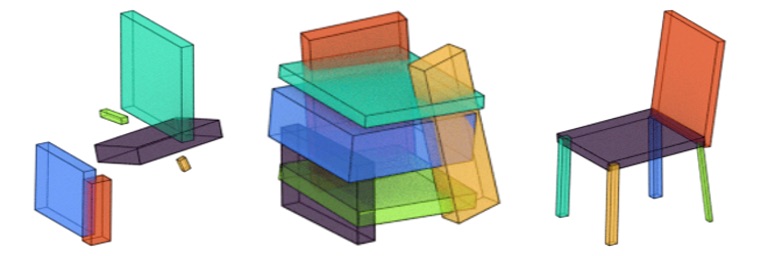}} \\
\multicolumn{3}{c|}{\includegraphics[width=.25\textwidth]{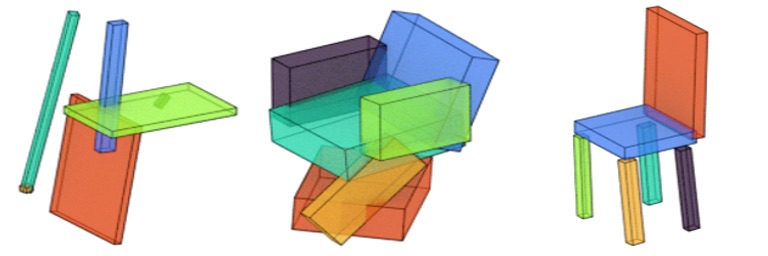}} &
\multicolumn{3}{c|}{\includegraphics[width=.25\textwidth]{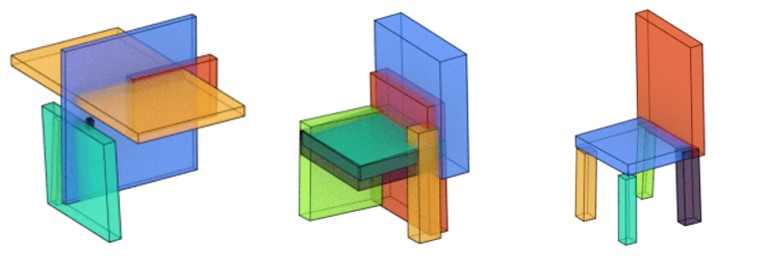}} &
\multicolumn{3}{c|}{\includegraphics[width=.25\textwidth]{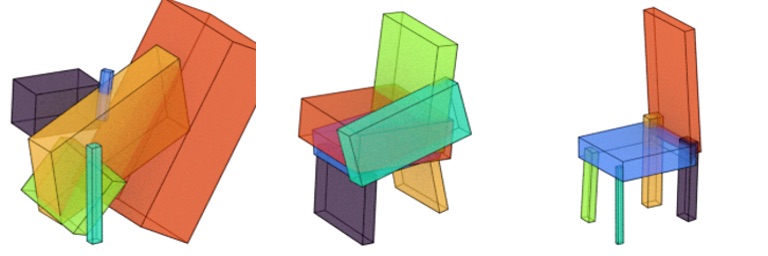}} &
\multicolumn{3}{c}{\includegraphics[width=.25\textwidth]{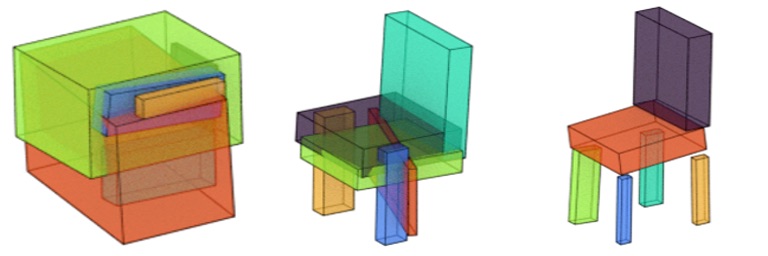}} \\

\end{tabularx}
\caption{\textbf{Qualitative comparison of shape abstraction generation. } For each pair of columns, we query the ground truth shape and retrieve the closest generated boxes measured with chamfer distance. Our method demonstrates higher-fidelity boxes.}
\label{fig:stage_1_comparison_more}
}
\end{figure*}
\begin{figure*}[p!]
\ContinuedFloat
\centering
{
\scriptsize
\setlength{\tabcolsep}{0em}
\renewcommand\tabularxcolumn[1]{m{#1}}
\begin{tabularx}{\linewidth}{Y Y Y |  Y Y Y | Y Y Y | Y Y Y}
 \rotatebox{0}{\makecell{Token Pred.\\Model}} & \rotatebox{0}{\makecell{Unconod.\\Diffusion}} & \rotatebox{0}{\makecell{Cond.\\Diffusion}} & \rotatebox{0}{\makecell{Token Pred.\\Model}} & \rotatebox{0}{\makecell{Unconod.\\Diffusion}} & \rotatebox{0}{\makecell{Cond.\\Diffusion}} & \rotatebox{0}{\makecell{Token Pred.\\Model}} & \rotatebox{0}{\makecell{Unconod.\\Diffusion}} & \rotatebox{0}{\makecell{Cond.\\Diffusion}} & \rotatebox{0}{\makecell{Token Pred.\\Model}} & \rotatebox{0}{\makecell{Unconod.\\Diffusion}} & \rotatebox{0}{\makecell{Cond.\\Diffusion}} \\
\midrule
\multicolumn{3}{c|}{\includegraphics[width=.25\textwidth]{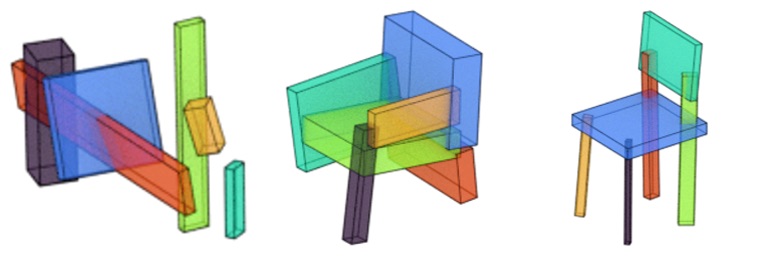}} &
\multicolumn{3}{c|}{\includegraphics[width=.25\textwidth]{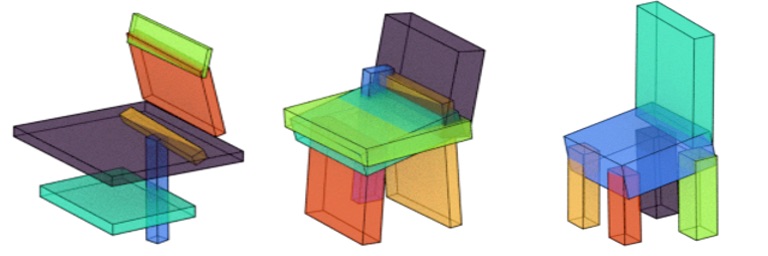}} &
\multicolumn{3}{c|}{\includegraphics[width=.25\textwidth]{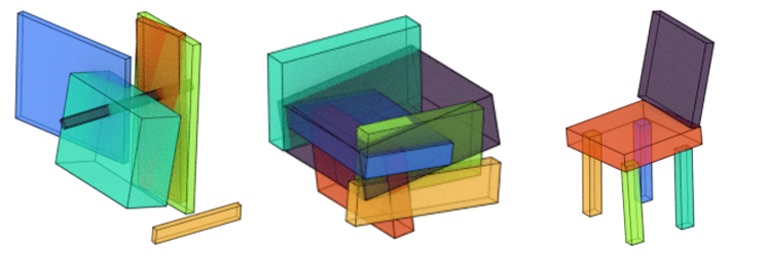}} &
\multicolumn{3}{c}{\includegraphics[width=.25\textwidth]{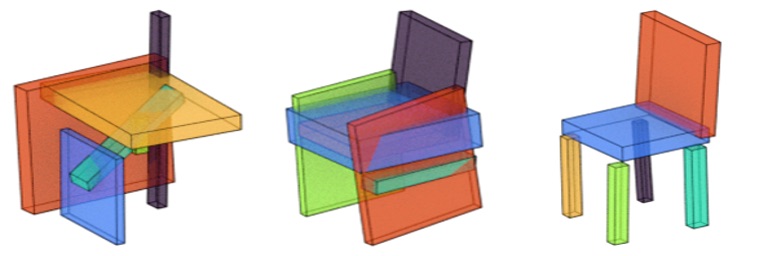}} \\
\multicolumn{3}{c|}{\includegraphics[width=.25\textwidth]{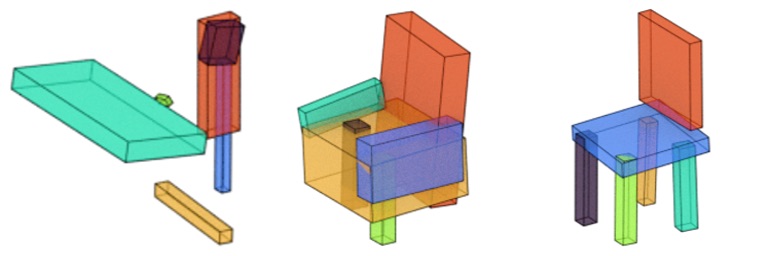}} &
\multicolumn{3}{c|}{\includegraphics[width=.25\textwidth]{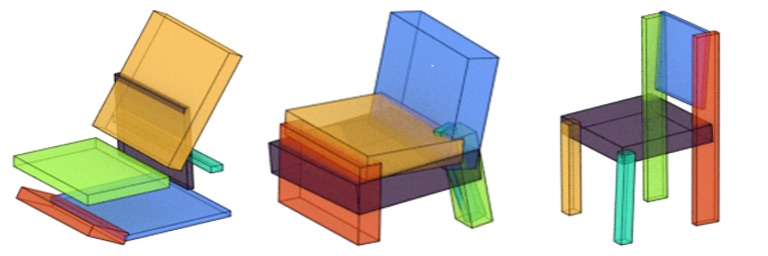}} &
\multicolumn{3}{c|}{\includegraphics[width=.25\textwidth]{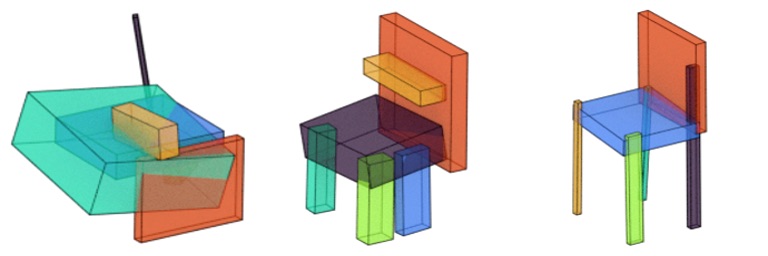}} &
\multicolumn{3}{c}{\includegraphics[width=.25\textwidth]{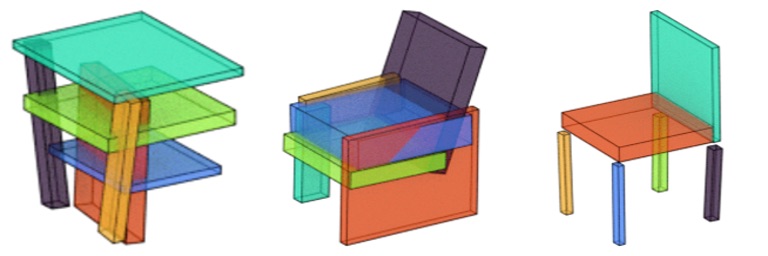}} \\
\multicolumn{3}{c|}{\includegraphics[width=.25\textwidth]{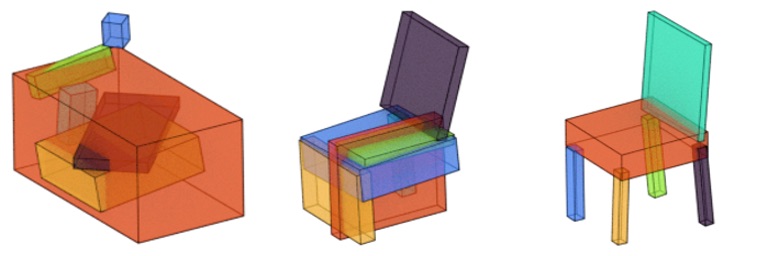}} &
\multicolumn{3}{c|}{\includegraphics[width=.25\textwidth]{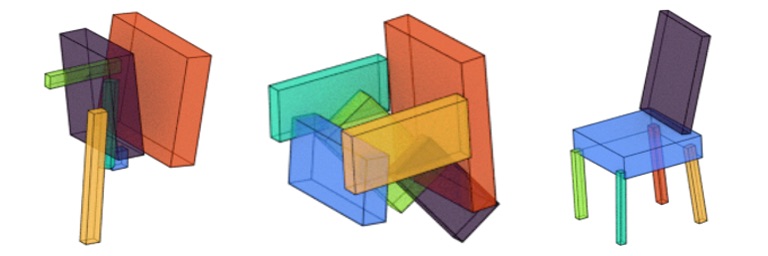}} &
\multicolumn{3}{c|}{\includegraphics[width=.25\textwidth]{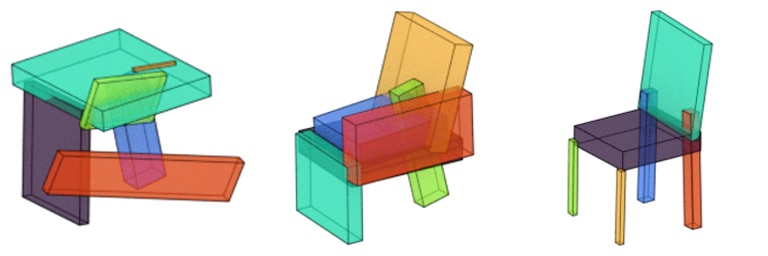}} &
\multicolumn{3}{c}{\includegraphics[width=.25\textwidth]{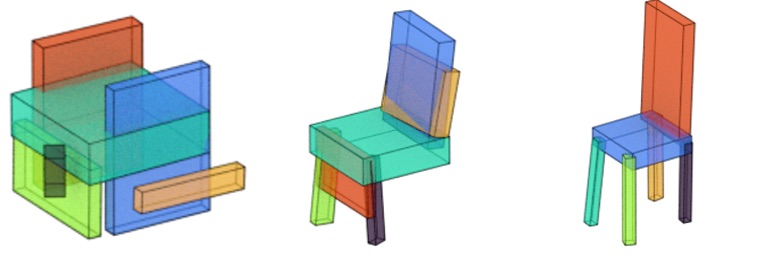}} \\
\multicolumn{3}{c|}{\includegraphics[width=.25\textwidth]{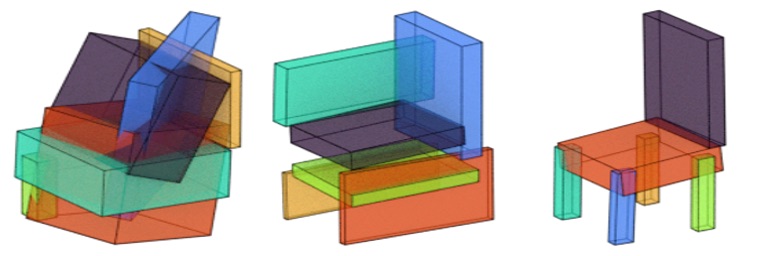}} &
\multicolumn{3}{c|}{\includegraphics[width=.25\textwidth]{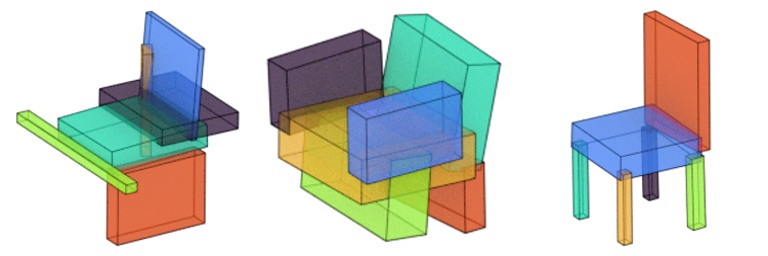}} &
\multicolumn{3}{c|}{\includegraphics[width=.25\textwidth]{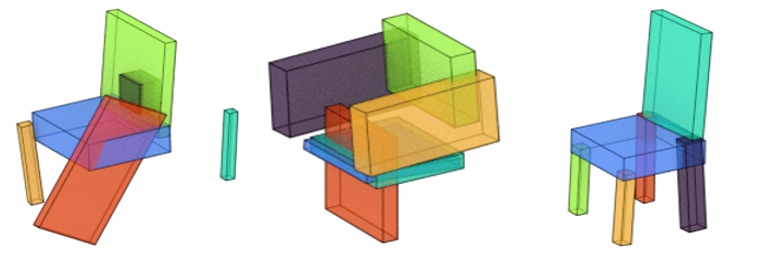}} &
\multicolumn{3}{c}{\includegraphics[width=.25\textwidth]{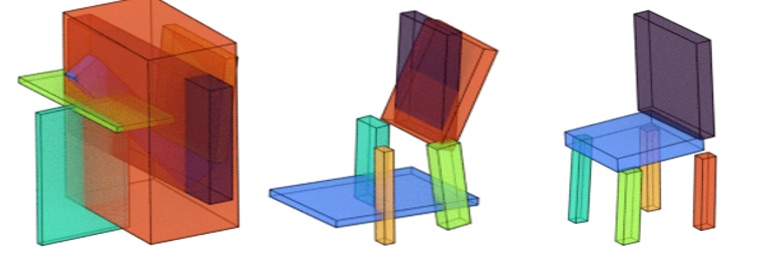}} \\
\multicolumn{3}{c|}{\includegraphics[width=.25\textwidth]{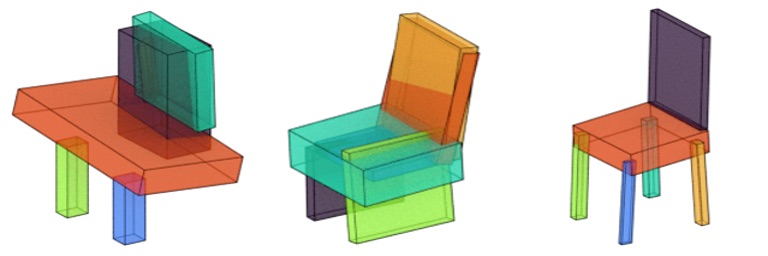}} &
\multicolumn{3}{c|}{\includegraphics[width=.25\textwidth]{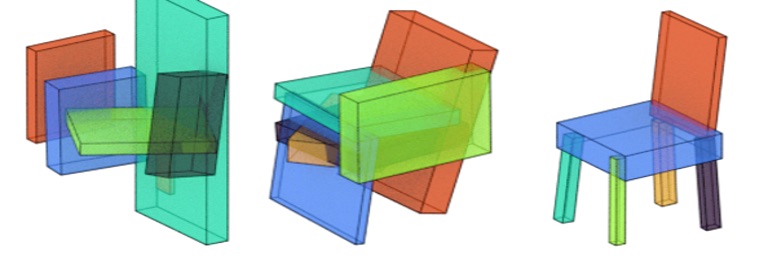}} &
\multicolumn{3}{c|}{\includegraphics[width=.25\textwidth]{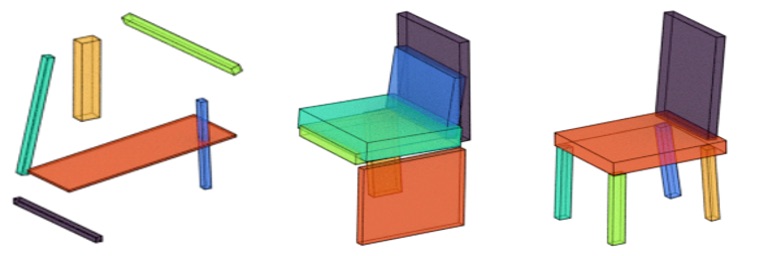}} &
\multicolumn{3}{c}{\includegraphics[width=.25\textwidth]{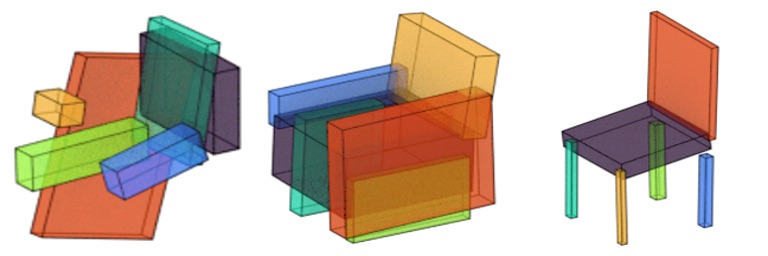}} \\
\multicolumn{3}{c|}{\includegraphics[width=.25\textwidth]{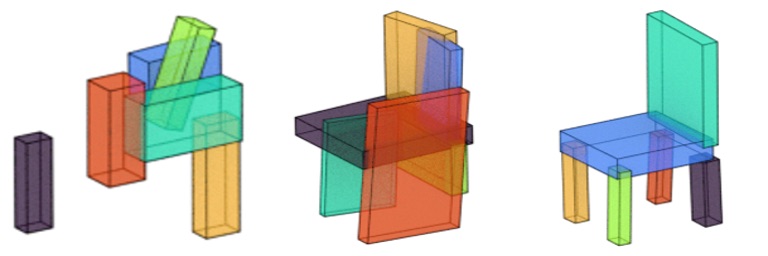}} &
\multicolumn{3}{c|}{\includegraphics[width=.25\textwidth]{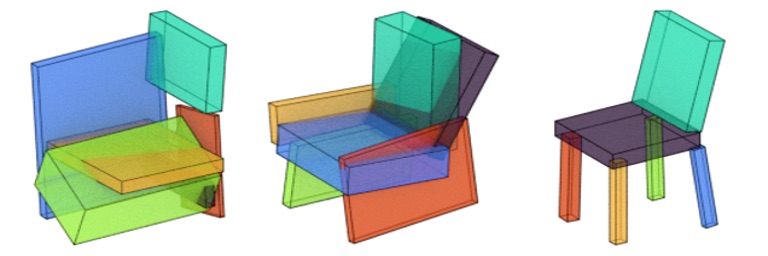}} &
\multicolumn{3}{c|}{\includegraphics[width=.25\textwidth]{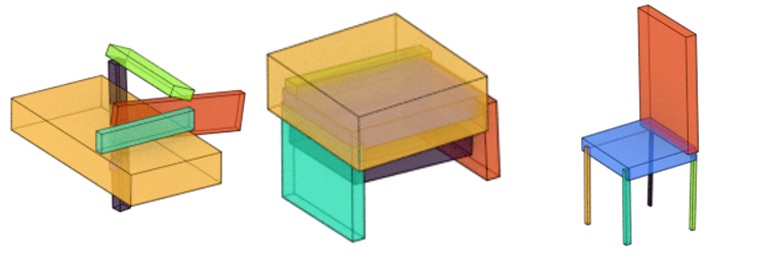}} &
\multicolumn{3}{c}{\includegraphics[width=.25\textwidth]{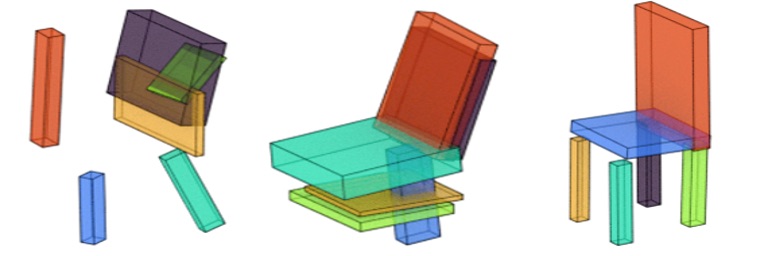}} \\
\multicolumn{3}{c|}{\includegraphics[width=.25\textwidth]{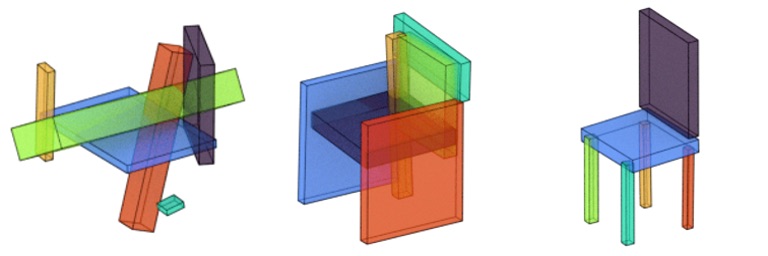}} &
\multicolumn{3}{c|}{\includegraphics[width=.25\textwidth]{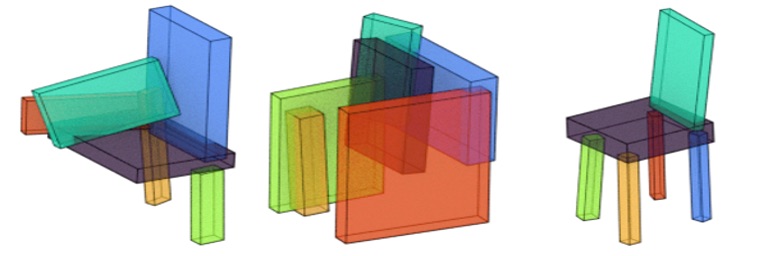}} &
\multicolumn{3}{c|}{\includegraphics[width=.25\textwidth]{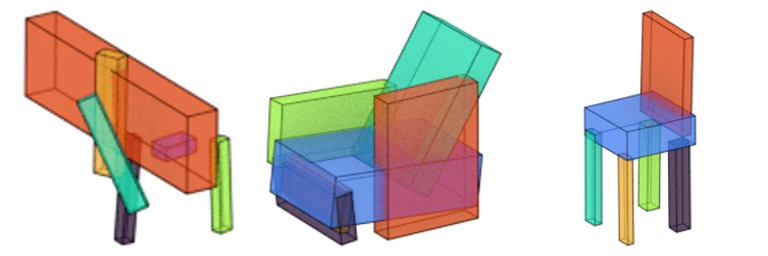}} &
\multicolumn{3}{c}{\includegraphics[width=.25\textwidth]{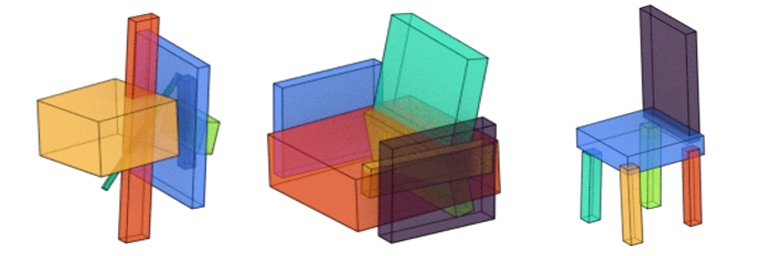}} \\
\multicolumn{3}{c|}{\includegraphics[width=.25\textwidth]{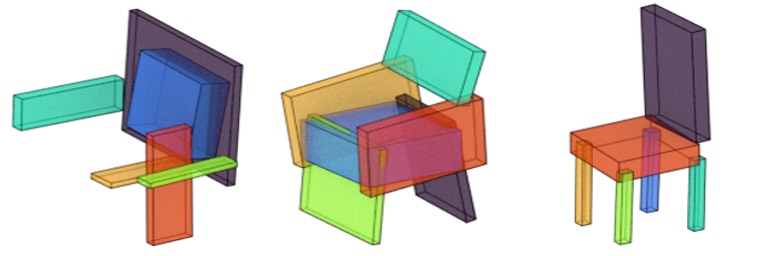}} &
\multicolumn{3}{c|}{\includegraphics[width=.25\textwidth]{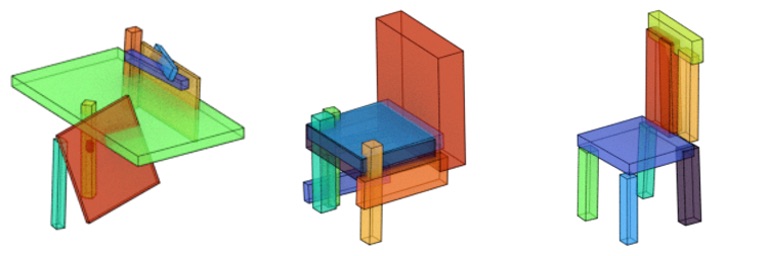}} &
\multicolumn{3}{c|}{\includegraphics[width=.25\textwidth]{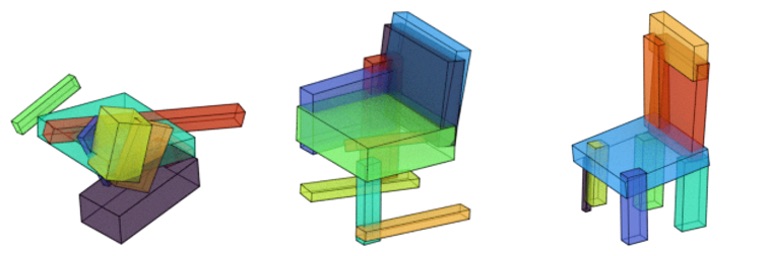}} &
\multicolumn{3}{c}{\includegraphics[width=.25\textwidth]{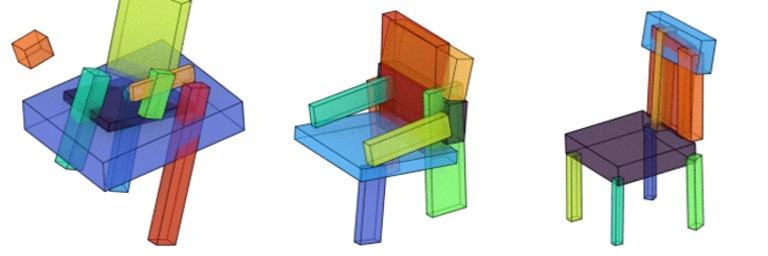}} \\
\multicolumn{3}{c|}{\includegraphics[width=.25\textwidth]{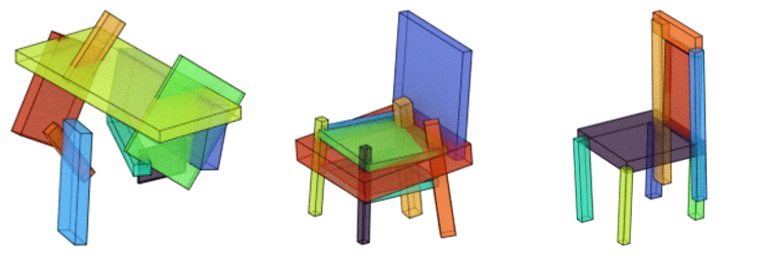}} &
\multicolumn{3}{c|}{\includegraphics[width=.25\textwidth]{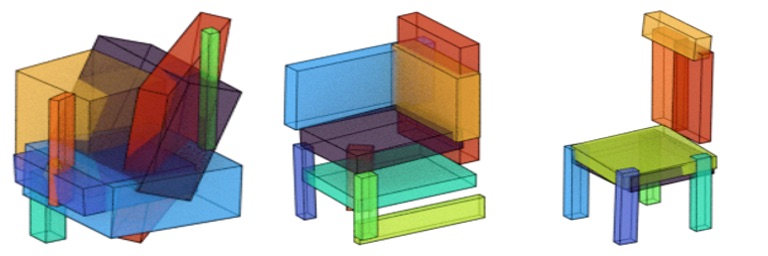}} &
\multicolumn{3}{c|}{\includegraphics[width=.25\textwidth]{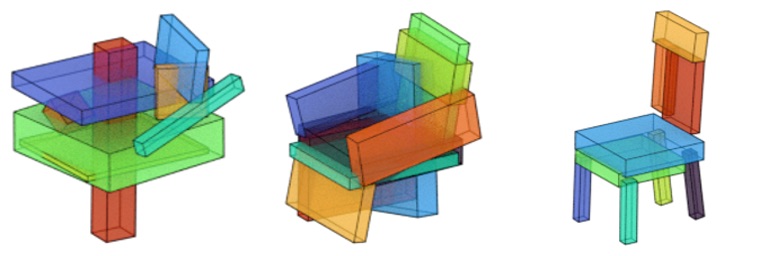}} &
\multicolumn{3}{c}{\includegraphics[width=.25\textwidth]{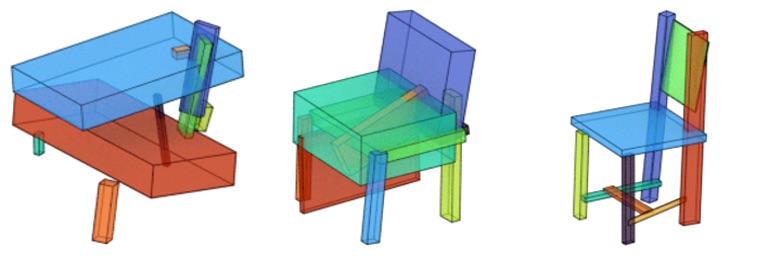}} \\
\multicolumn{3}{c|}{\includegraphics[width=.25\textwidth]{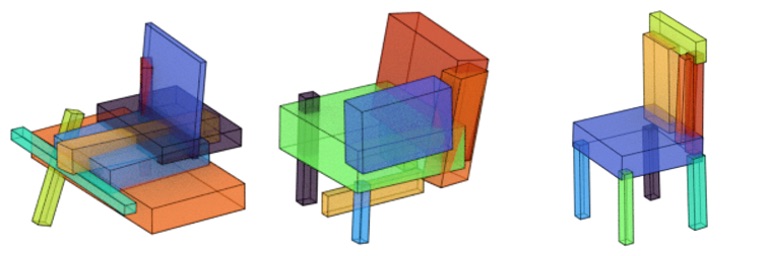}} &
\multicolumn{3}{c|}{\includegraphics[width=.25\textwidth]{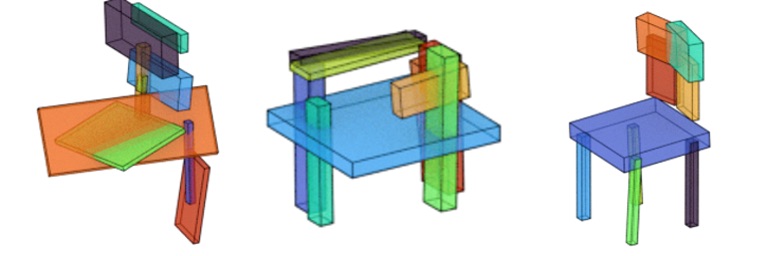}} &
\multicolumn{3}{c|}{\includegraphics[width=.25\textwidth]{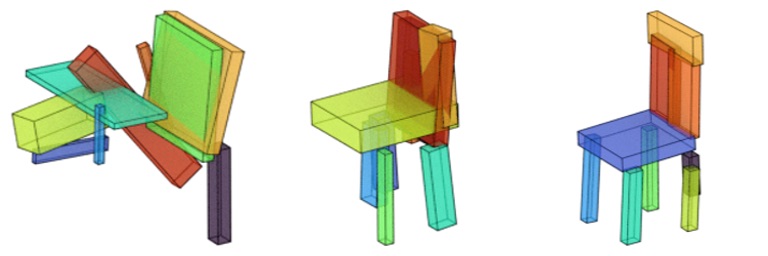}} &
\multicolumn{3}{c}{\includegraphics[width=.25\textwidth]{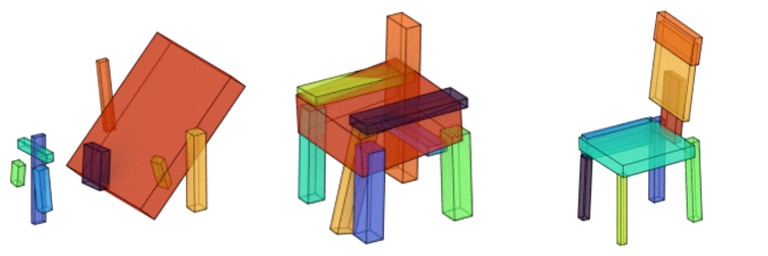}} \\
\multicolumn{3}{c|}{\includegraphics[width=.25\textwidth]{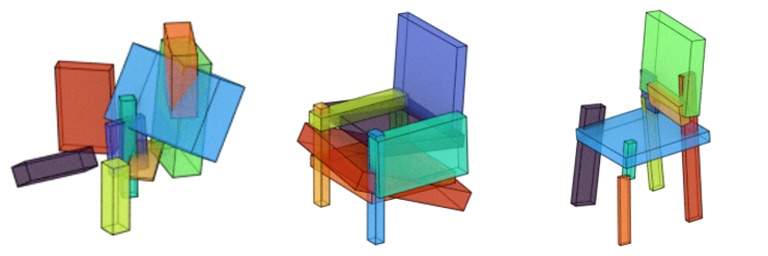}} &
\multicolumn{3}{c|}{\includegraphics[width=.25\textwidth]{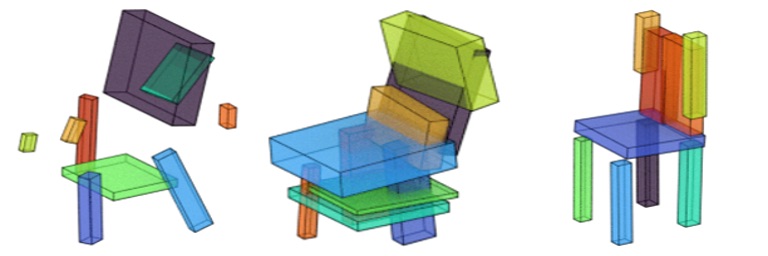}} &
\multicolumn{3}{c|}{\includegraphics[width=.25\textwidth]{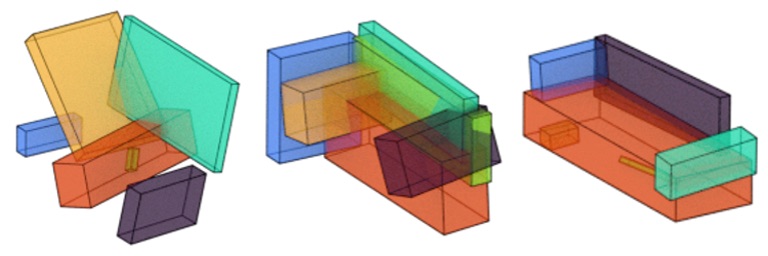}} &
\multicolumn{3}{c}{\includegraphics[width=.25\textwidth]{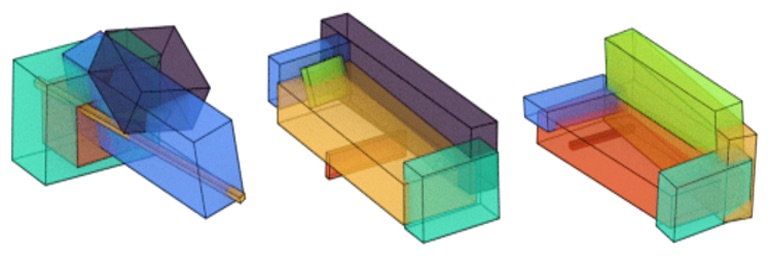}} \\
\multicolumn{3}{c|}{\includegraphics[width=.25\textwidth]{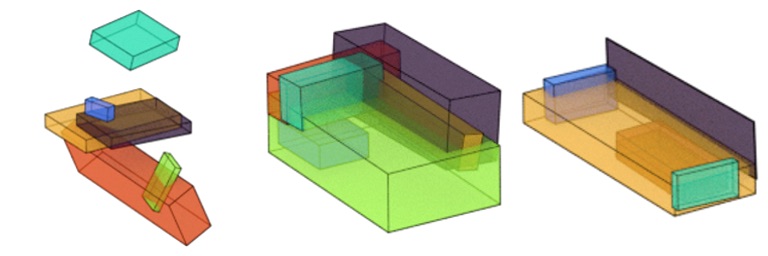}} &
\multicolumn{3}{c|}{\includegraphics[width=.25\textwidth]{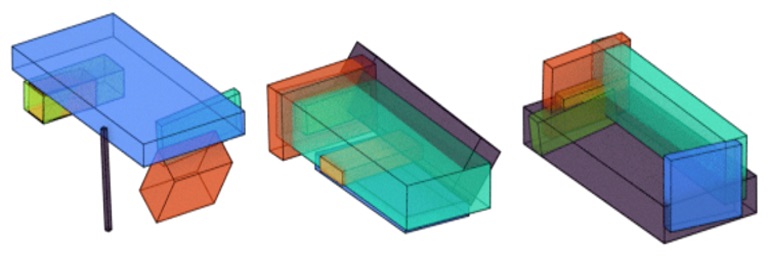}} &
\multicolumn{3}{c|}{\includegraphics[width=.25\textwidth]{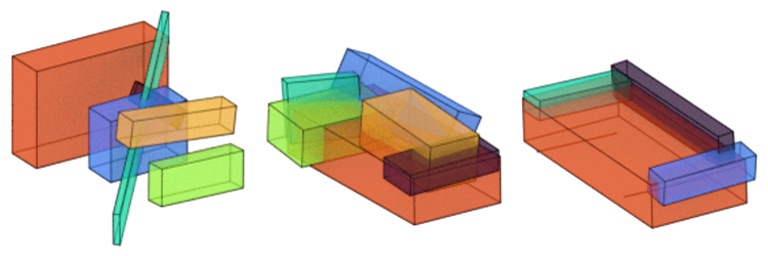}} &
\multicolumn{3}{c}{\includegraphics[width=.25\textwidth]{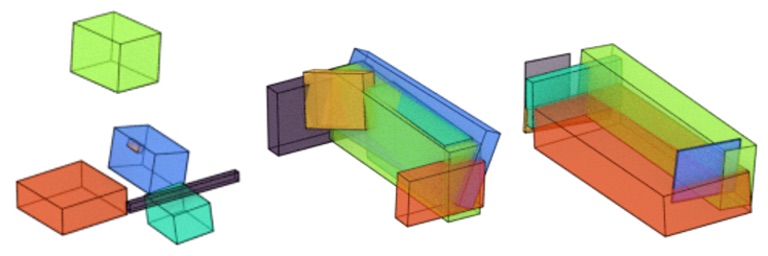}} \\
\multicolumn{3}{c|}{\includegraphics[width=.25\textwidth]{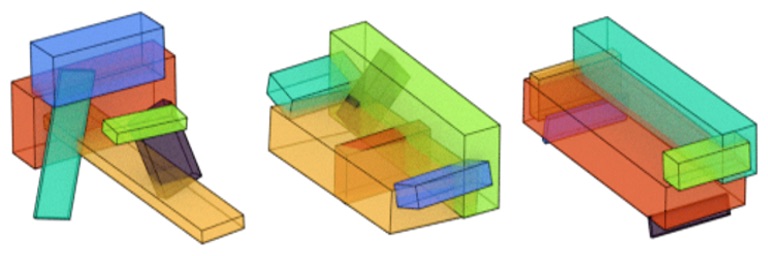}} &
\multicolumn{3}{c|}{\includegraphics[width=.25\textwidth]{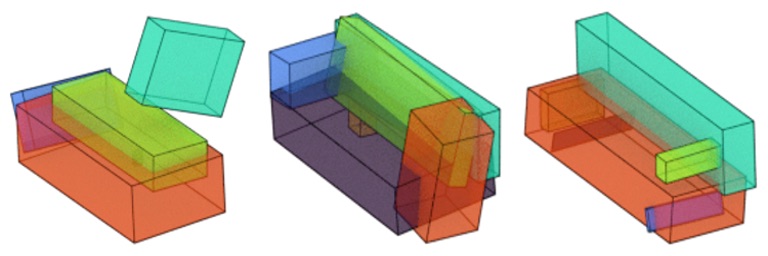}} &
\multicolumn{3}{c|}{\includegraphics[width=.25\textwidth]{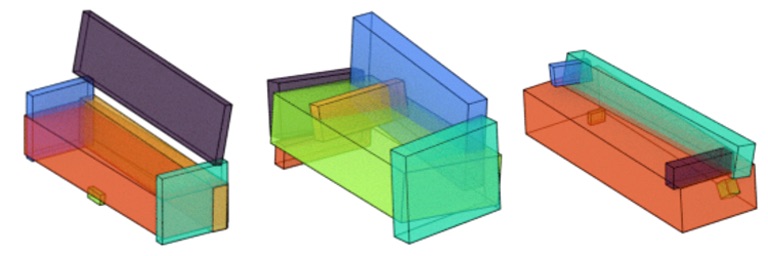}} &
\multicolumn{3}{c}{\includegraphics[width=.25\textwidth]{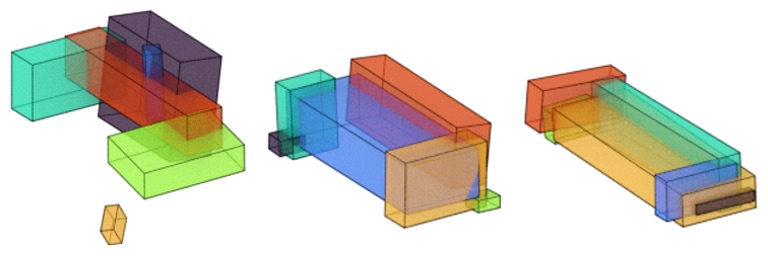}} \\

\end{tabularx}
\caption{\textbf{Qualitative comparison of shape abstraction generation. } For each pair of columns, we query the ground truth shape and retrieve the closest generated boxes measured with chamfer distance. Our method demonstrates higher-fidelity boxes.}
}
\end{figure*}
\begin{figure*}[p!]
\ContinuedFloat
\centering
{
\scriptsize
\setlength{\tabcolsep}{0em}
\renewcommand\tabularxcolumn[1]{m{#1}}
\begin{tabularx}{\linewidth}{Y Y Y |  Y Y Y | Y Y Y | Y Y Y}
 \rotatebox{0}{\makecell{Token Pred.\\Model}} & \rotatebox{0}{\makecell{Unconod.\\Diffusion}} & \rotatebox{0}{\makecell{Cond.\\Diffusion}} & \rotatebox{0}{\makecell{Token Pred.\\Model}} & \rotatebox{0}{\makecell{Unconod.\\Diffusion}} & \rotatebox{0}{\makecell{Cond.\\Diffusion}} & \rotatebox{0}{\makecell{Token Pred.\\Model}} & \rotatebox{0}{\makecell{Unconod.\\Diffusion}} & \rotatebox{0}{\makecell{Cond.\\Diffusion}} & \rotatebox{0}{\makecell{Token Pred.\\Model}} & \rotatebox{0}{\makecell{Unconod.\\Diffusion}} & \rotatebox{0}{\makecell{Cond.\\Diffusion}} \\
\midrule

\multicolumn{3}{c|}{\includegraphics[width=.25\textwidth]{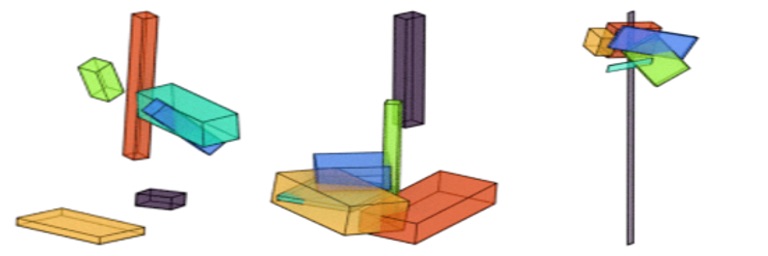}} &
\multicolumn{3}{c|}{\includegraphics[width=.25\textwidth]{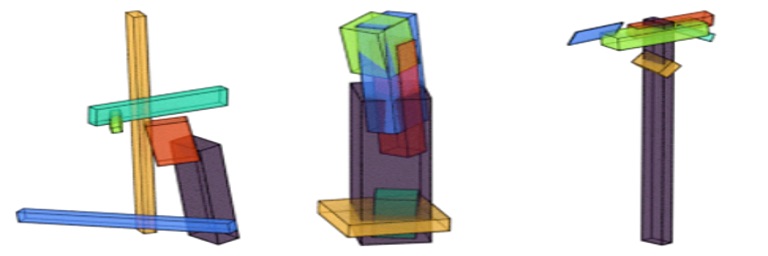}} &
\multicolumn{3}{c|}{\includegraphics[width=.25\textwidth]{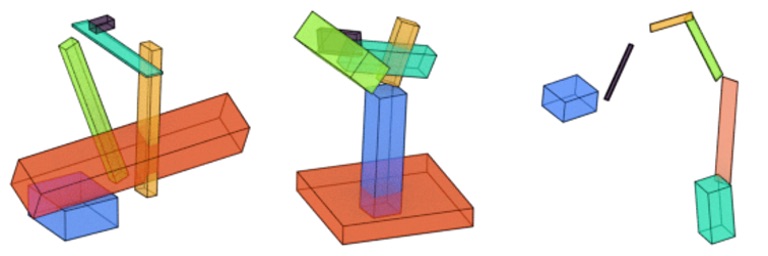}} &
\multicolumn{3}{c}{\includegraphics[width=.25\textwidth]{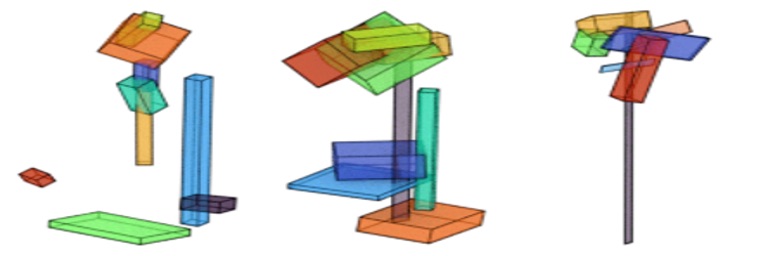}} \\
\multicolumn{3}{c|}{\includegraphics[width=.25\textwidth]{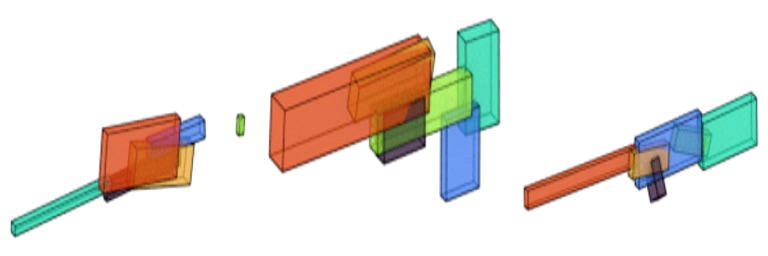}} &
\multicolumn{3}{c|}{\includegraphics[width=.25\textwidth]{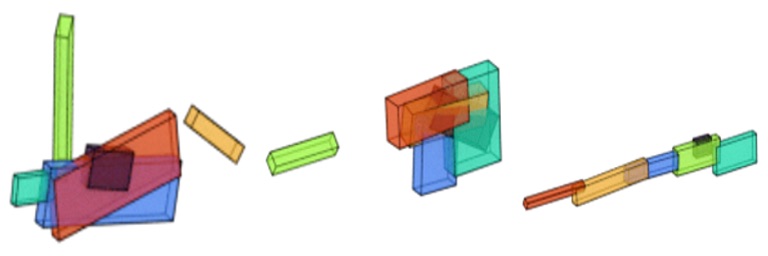}} &
\multicolumn{3}{c|}{\includegraphics[width=.25\textwidth]{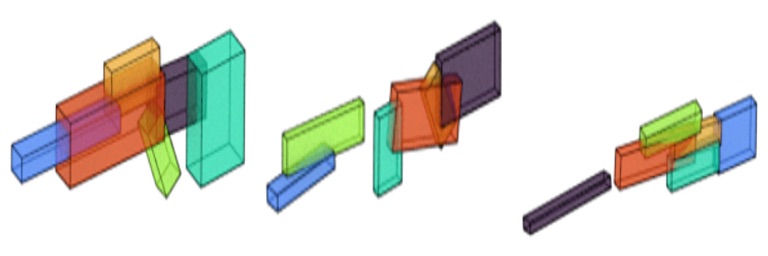}} &
\multicolumn{3}{c}{\includegraphics[width=.25\textwidth]{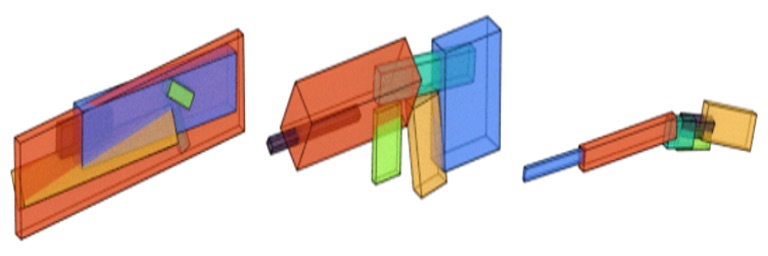}} \\
\multicolumn{3}{c|}{\includegraphics[width=.25\textwidth]{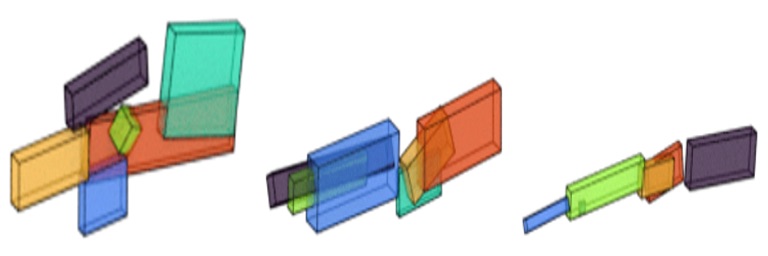}} &
\multicolumn{3}{c|}{\includegraphics[width=.25\textwidth]{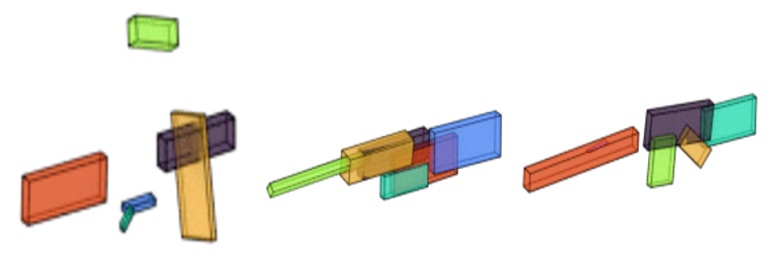}} &
\multicolumn{3}{c|}{\includegraphics[width=.25\textwidth]{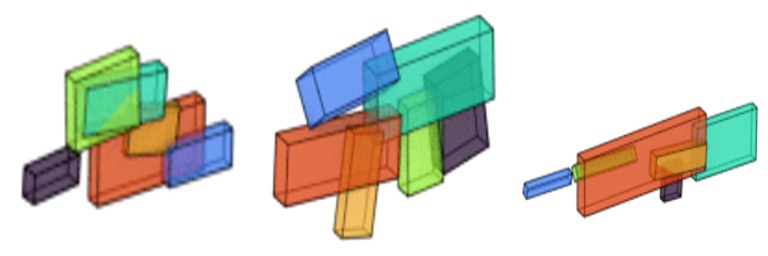}} &
\multicolumn{3}{c}{\includegraphics[width=.25\textwidth]{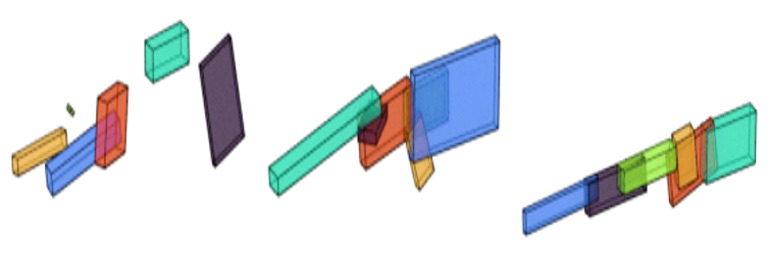}} \\
\multicolumn{3}{c|}{\includegraphics[width=.25\textwidth]{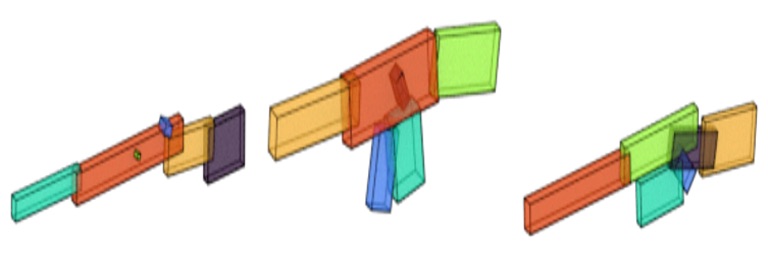}} &
\multicolumn{3}{c|}{\includegraphics[width=.25\textwidth]{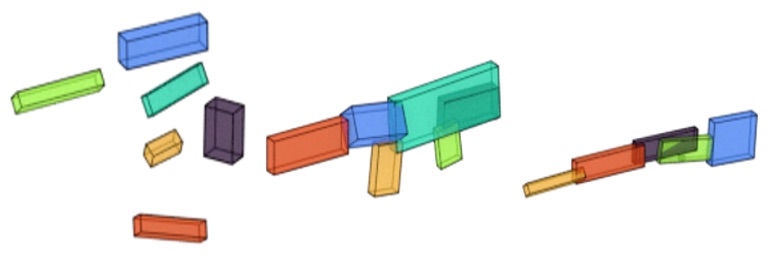}} &
\multicolumn{3}{c|}{\includegraphics[width=.25\textwidth]{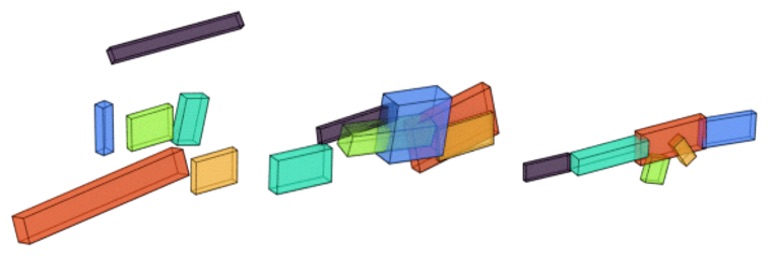}} &
\multicolumn{3}{c}{\includegraphics[width=.25\textwidth]{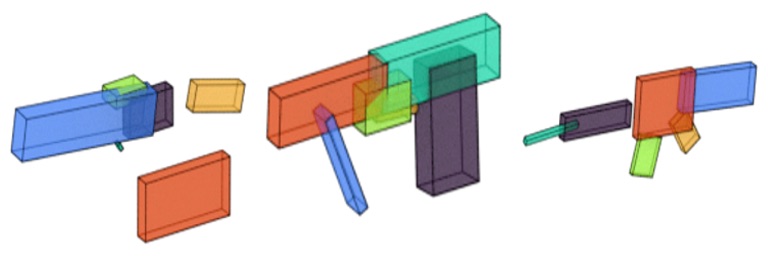}} \\
\multicolumn{3}{c|}{\includegraphics[width=.25\textwidth]{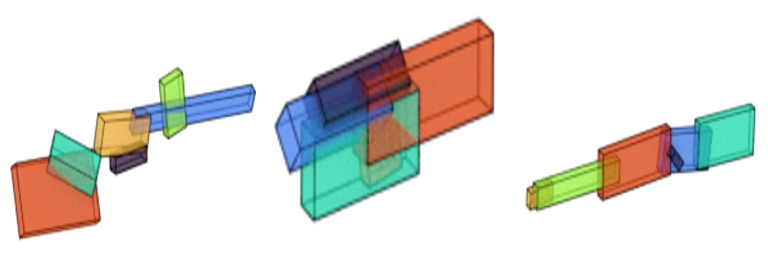}} &
\multicolumn{3}{c|}{\includegraphics[width=.25\textwidth]{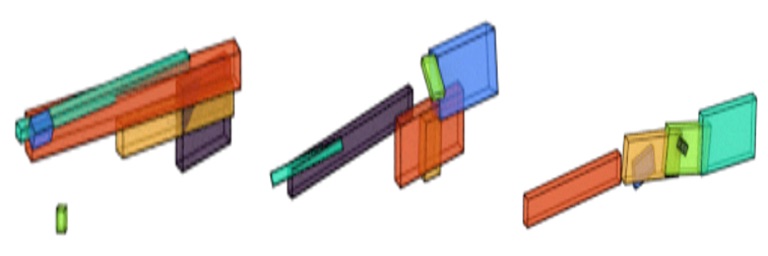}} &
\multicolumn{3}{c|}{\includegraphics[width=.25\textwidth]{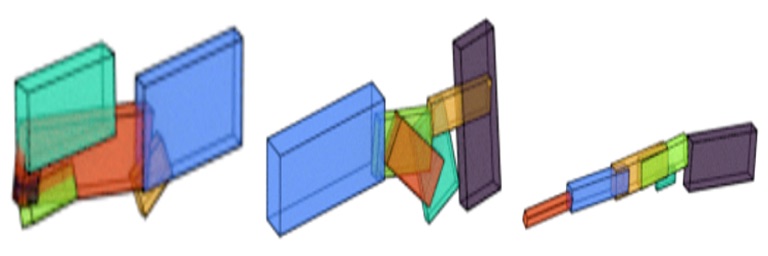}} &
\multicolumn{3}{c}{\includegraphics[width=.25\textwidth]{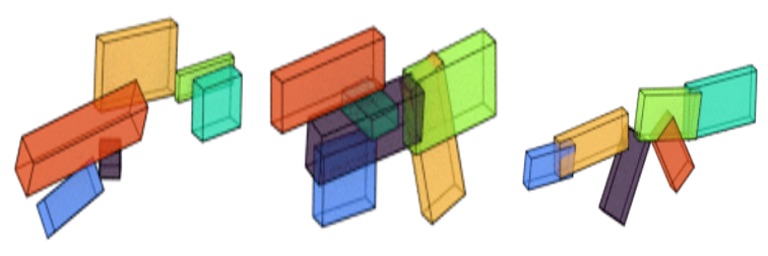}} \\
\multicolumn{3}{c|}{\includegraphics[width=.25\textwidth]{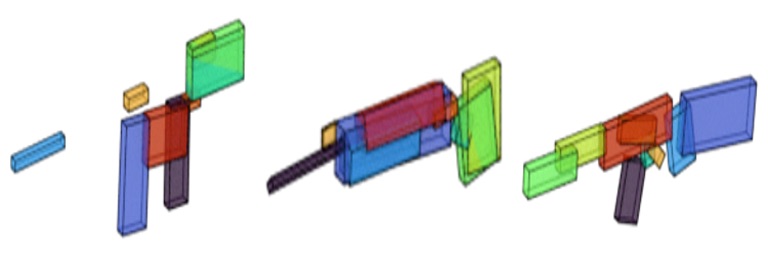}} &
\multicolumn{3}{c|}{\includegraphics[width=.25\textwidth]{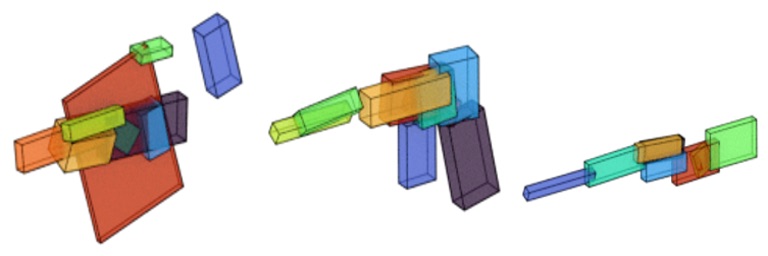}} &
\multicolumn{3}{c|}{\includegraphics[width=.25\textwidth]{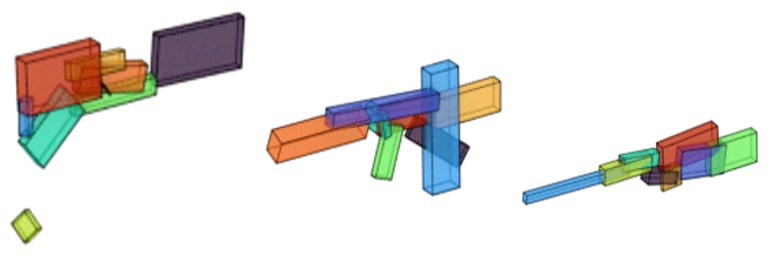}} &
\multicolumn{3}{c}{\includegraphics[width=.25\textwidth]{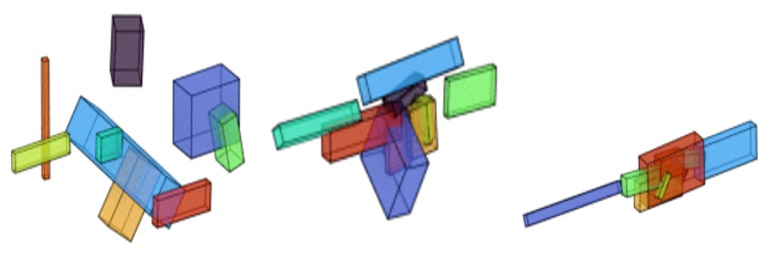}} \\
\multicolumn{3}{c|}{\includegraphics[width=.25\textwidth]{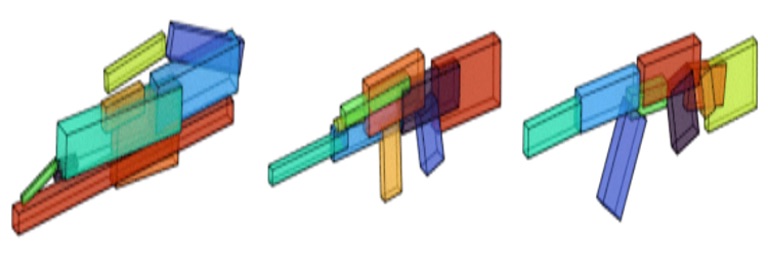}} &
\multicolumn{3}{c|}{\includegraphics[width=.25\textwidth]{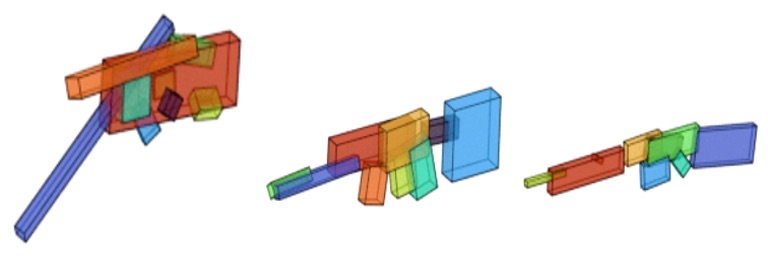}} &
\multicolumn{3}{c|}{\includegraphics[width=.25\textwidth]{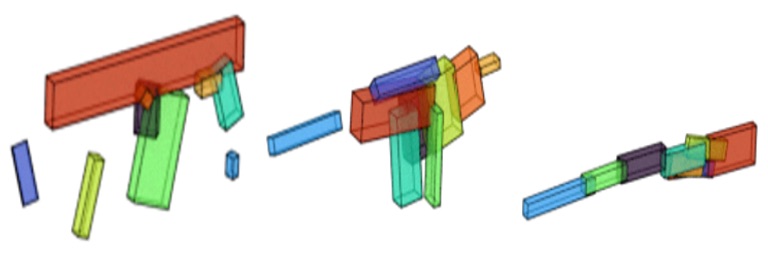}} &
\multicolumn{3}{c}{\includegraphics[width=.25\textwidth]{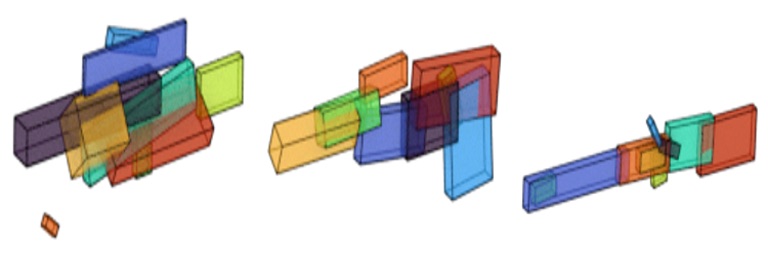}} \\
\multicolumn{3}{c|}{\includegraphics[width=.25\textwidth]{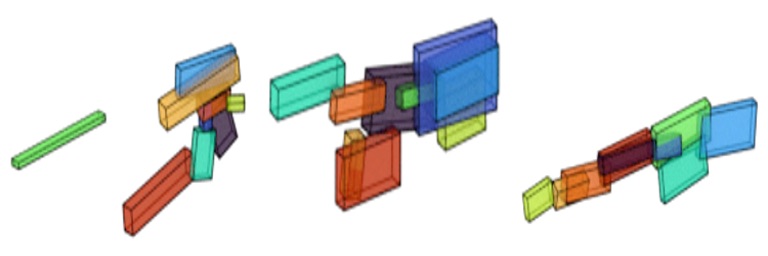}} &
\multicolumn{3}{c|}{\includegraphics[width=.25\textwidth]{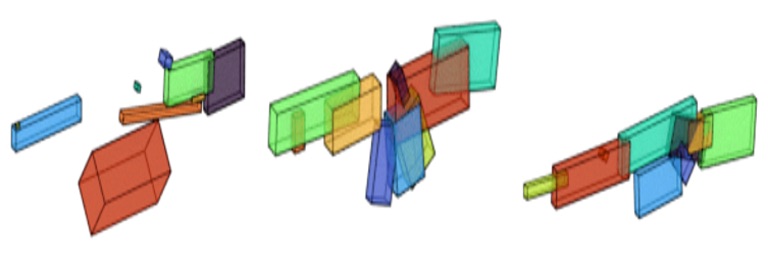}} &
\multicolumn{3}{c|}{\includegraphics[width=.25\textwidth]{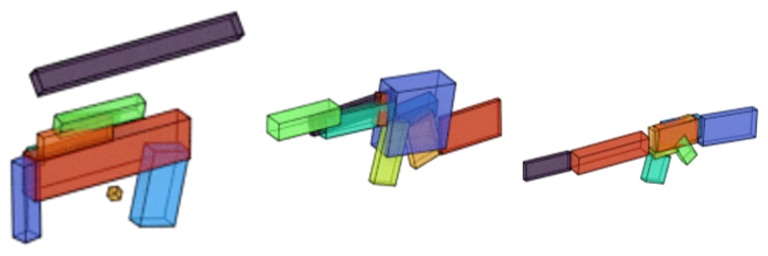}} &
\multicolumn{3}{c}{\includegraphics[width=.25\textwidth]{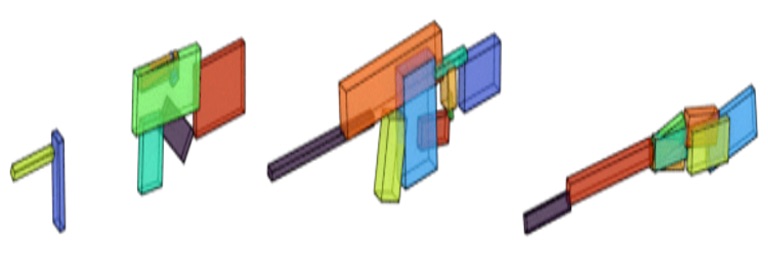}} \\
\multicolumn{3}{c|}{\includegraphics[width=.25\textwidth]{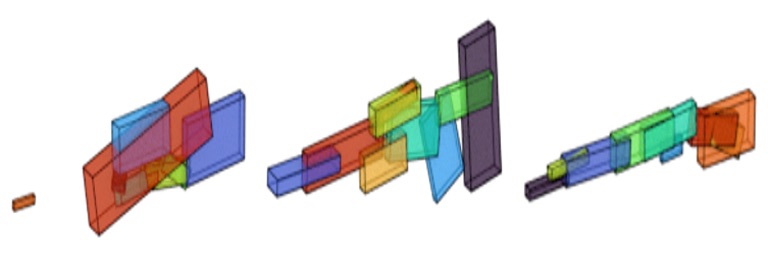}} &
\multicolumn{3}{c|}{\includegraphics[width=.25\textwidth]{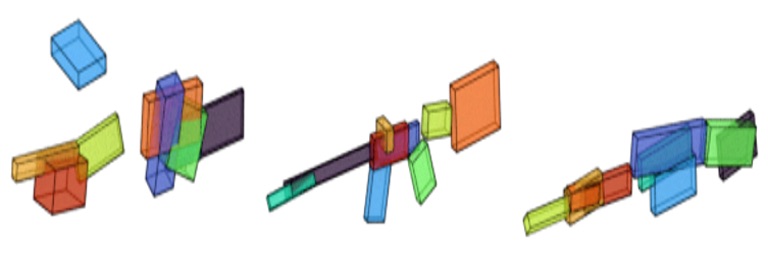}} &
\multicolumn{3}{c|}{\includegraphics[width=.25\textwidth]{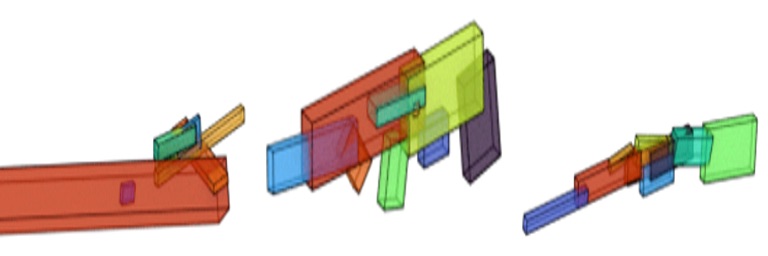}} &
\multicolumn{3}{c}{\includegraphics[width=.25\textwidth]{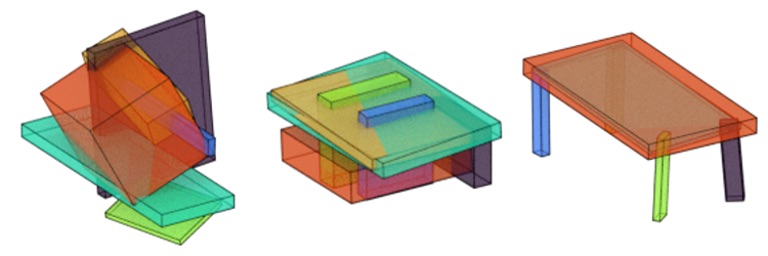}} \\
\multicolumn{3}{c|}{\includegraphics[width=.25\textwidth]{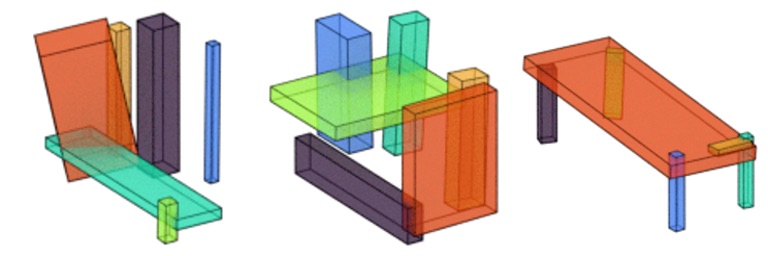}} &
\multicolumn{3}{c|}{\includegraphics[width=.25\textwidth]{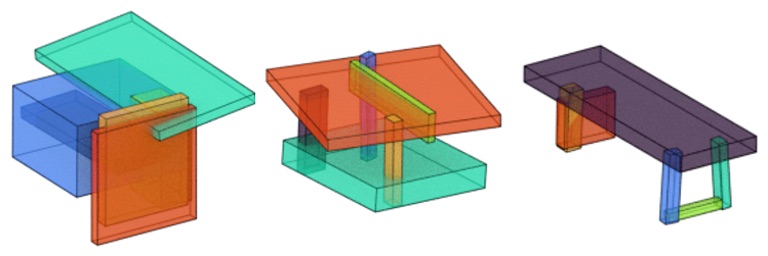}} &
\multicolumn{3}{c|}{\includegraphics[width=.25\textwidth]{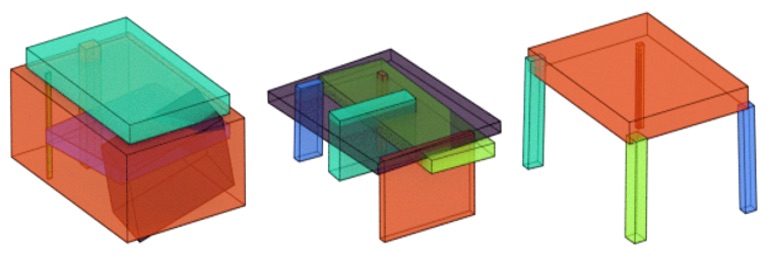}} &
\multicolumn{3}{c}{\includegraphics[width=.25\textwidth]{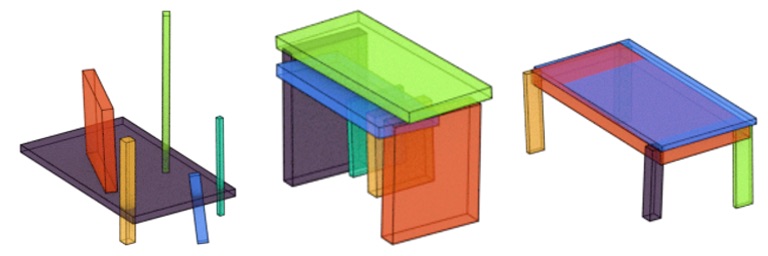}} \\
\multicolumn{3}{c|}{\includegraphics[width=.25\textwidth]{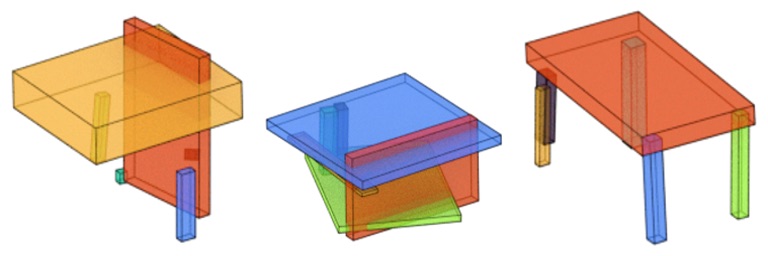}} &
\multicolumn{3}{c|}{\includegraphics[width=.25\textwidth]{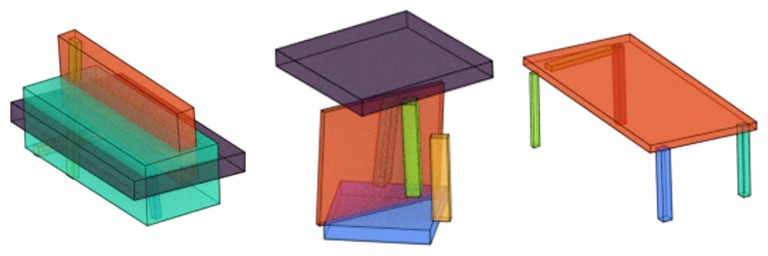}} &
\multicolumn{3}{c|}{\includegraphics[width=.25\textwidth]{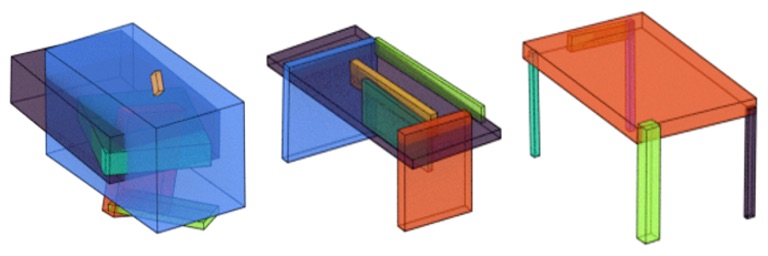}} &
\multicolumn{3}{c}{\includegraphics[width=.25\textwidth]{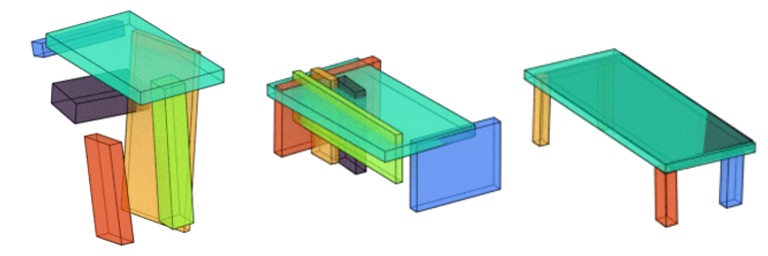}} \\
\multicolumn{3}{c|}{\includegraphics[width=.25\textwidth]{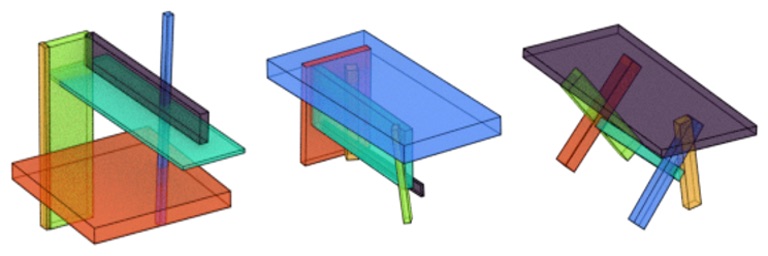}} &
\multicolumn{3}{c|}{\includegraphics[width=.25\textwidth]{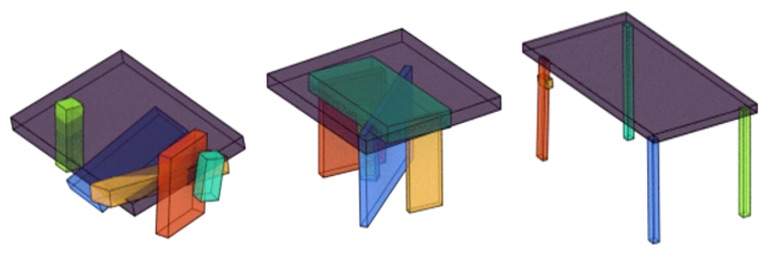}} &
\multicolumn{3}{c|}{\includegraphics[width=.25\textwidth]{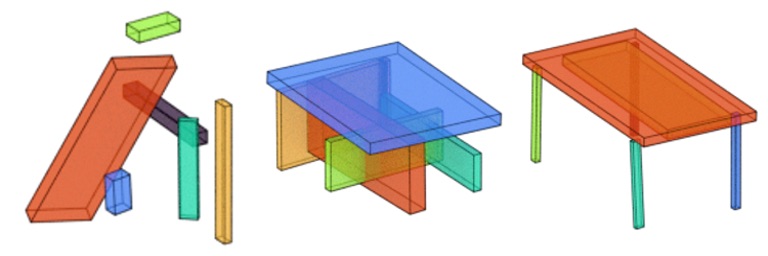}} &
\multicolumn{3}{c}{\includegraphics[width=.25\textwidth]{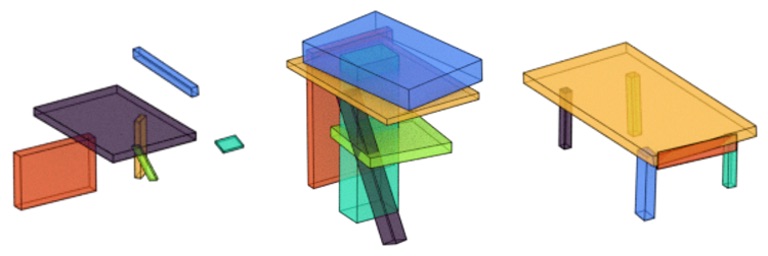}} \\
\multicolumn{3}{c|}{\includegraphics[width=.25\textwidth]{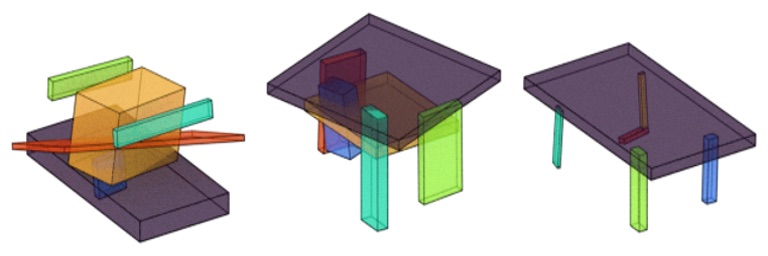}} &
\multicolumn{3}{c|}{\includegraphics[width=.25\textwidth]{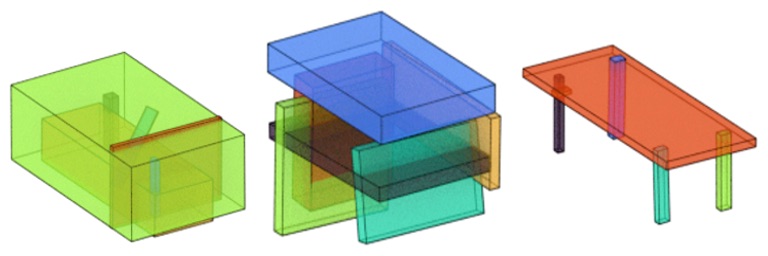}} &
\multicolumn{3}{c|}{\includegraphics[width=.25\textwidth]{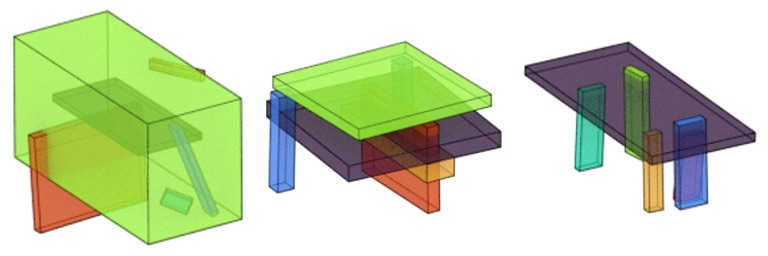}} &
\multicolumn{3}{c}{\includegraphics[width=.25\textwidth]{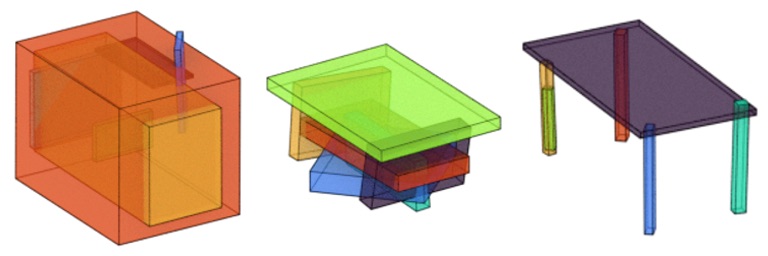}} \\

\end{tabularx}
\caption{\textbf{Qualitative comparison of shape abstraction generation. } For each pair of columns, we query the ground truth shape and retrieve the closest generated boxes measured with chamfer distance. Our method demonstrates higher-fidelity boxes.}
}
\end{figure*}

\endgroup
\clearpage
\newpage
\subsection{More Qualitative Results of Box-Conditioned Shape Generation}
\label{sec:suppl_more_shape_generation_qualitative_results}
Figure~\ref{fig:stage_2_comparison_more} presents more qualitative results of box-conditioned shape generation.

\begingroup
    \centering
\begin{figure*}[p!]

\centering
\scriptsize
\setlength{\tabcolsep}{0em}
\begin{tabularx}{\linewidth}{YYYY | YYYY | YYYY | YYYY}
\rotatebox{0}{\makecell{Input\\Boxes}} & \rotatebox{0}{\makecell{Spice-E\\\cite{Sella:2023SpicE}}} & \rotatebox{0}{\makecell{Gated\\3DS2V~\cite{Zhang:2023Shape2Vec}}} & \rotatebox{0}{Ours} & \rotatebox{0}{\makecell{Input\\Boxes}} & \rotatebox{0}{\makecell{Spice-E\\\cite{Sella:2023SpicE}}} & \rotatebox{0}{\makecell{Gated\\3DS2V~\cite{Zhang:2023Shape2Vec}}} & \rotatebox{0}{Ours} & \rotatebox{0}{\makecell{Input\\Boxes}} & \rotatebox{0}{\makecell{Spice-E\\\cite{Sella:2023SpicE}}} & \rotatebox{0}{\makecell{Gated\\3DS2V~\cite{Zhang:2023Shape2Vec}}} & \rotatebox{0}{Ours} & \rotatebox{0}{\makecell{Input\\Boxes}} & \rotatebox{0}{\makecell{Spice-E\\\cite{Sella:2023SpicE}}} & \rotatebox{0}{\makecell{Gated\\3DS2V~\cite{Zhang:2023Shape2Vec}}} & \rotatebox{0}{Ours}  \\ 

\midrule

\multicolumn{4}{c|}{\includegraphics[width=.25\textwidth]{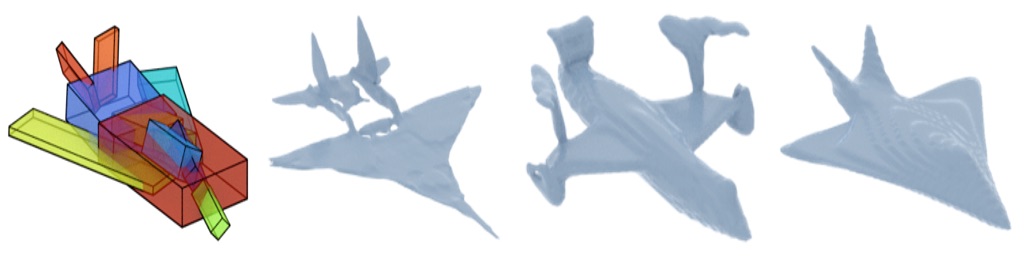}} &
\multicolumn{4}{c|}{\includegraphics[width=.25\textwidth]{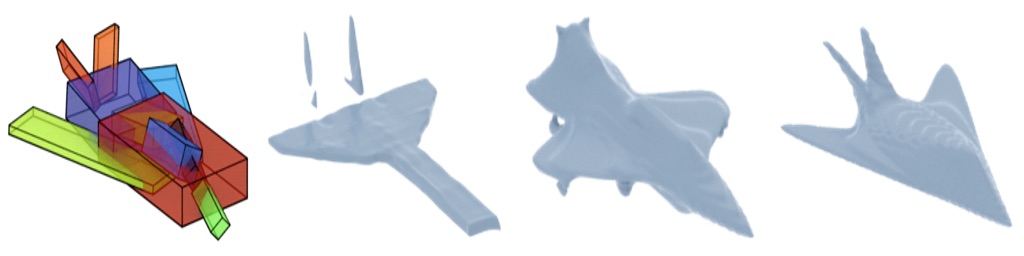}} &
\multicolumn{4}{c|}{\includegraphics[width=.25\textwidth]{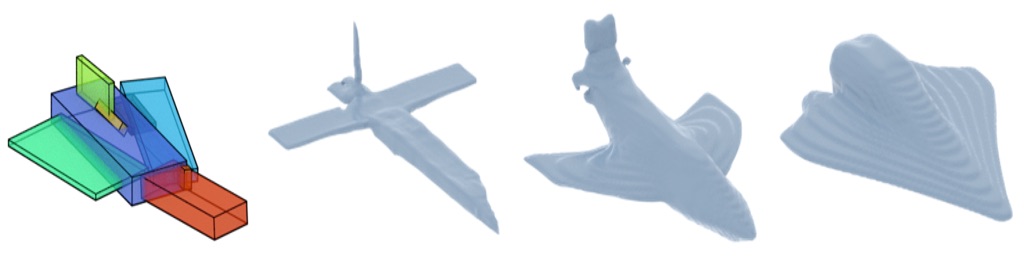}} &
\multicolumn{4}{c}{\includegraphics[width=.25\textwidth]{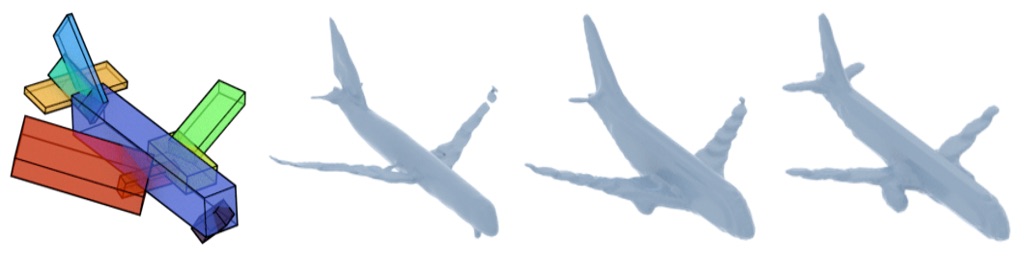}} \\
\multicolumn{4}{c|}{\includegraphics[width=.25\textwidth]{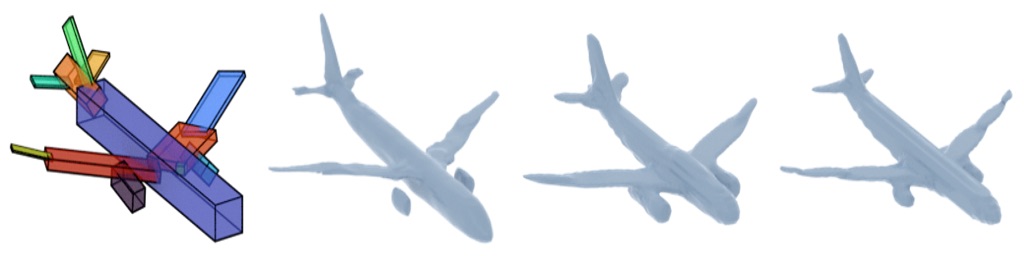}} &
\multicolumn{4}{c|}{\includegraphics[width=.25\textwidth]{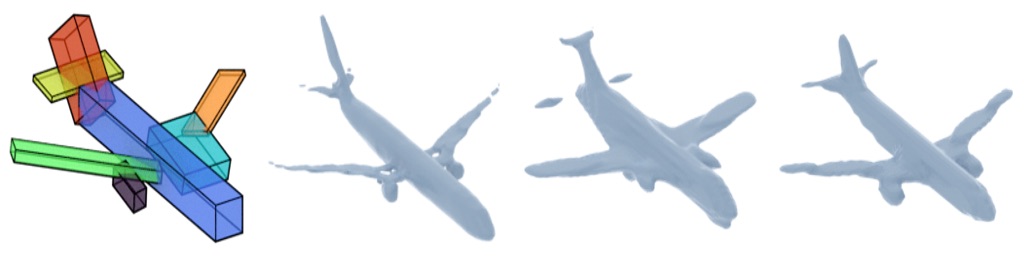}} &
\multicolumn{4}{c|}{\includegraphics[width=.25\textwidth]{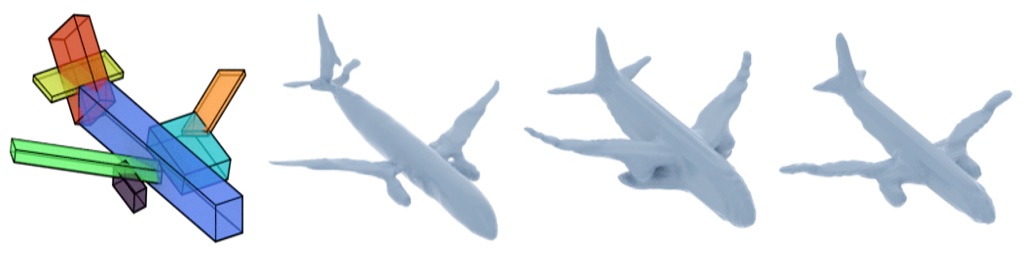}} &
\multicolumn{4}{c}{\includegraphics[width=.25\textwidth]{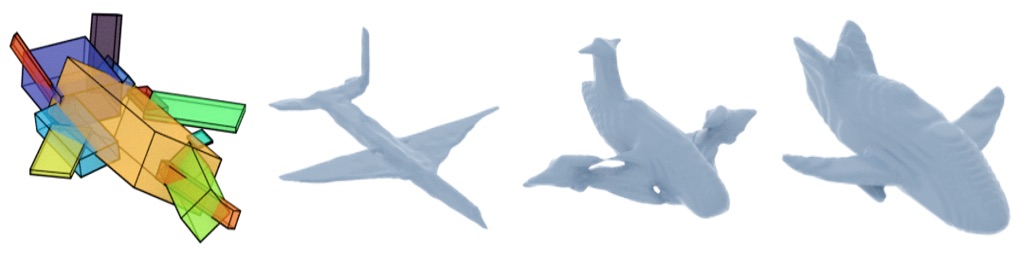}} \\
\multicolumn{4}{c|}{\includegraphics[width=.25\textwidth]{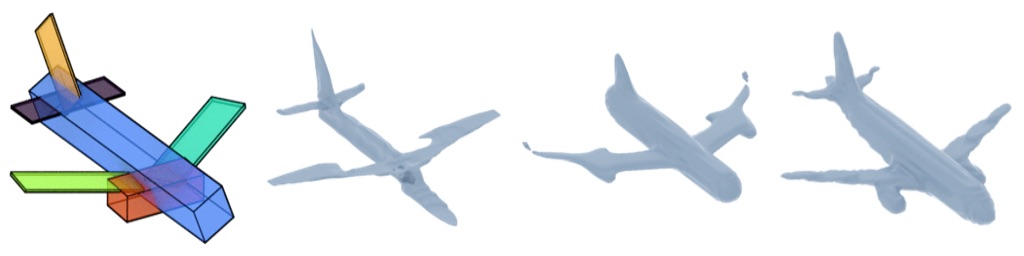}} &
\multicolumn{4}{c|}{\includegraphics[width=.25\textwidth]{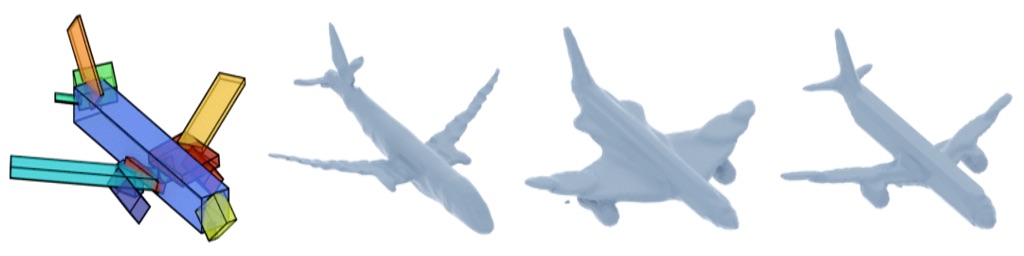}} &
\multicolumn{4}{c|}{\includegraphics[width=.25\textwidth]{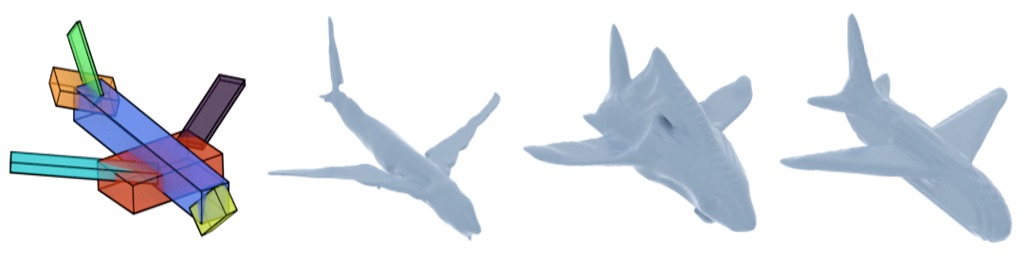}} &
\multicolumn{4}{c}{\includegraphics[width=.25\textwidth]{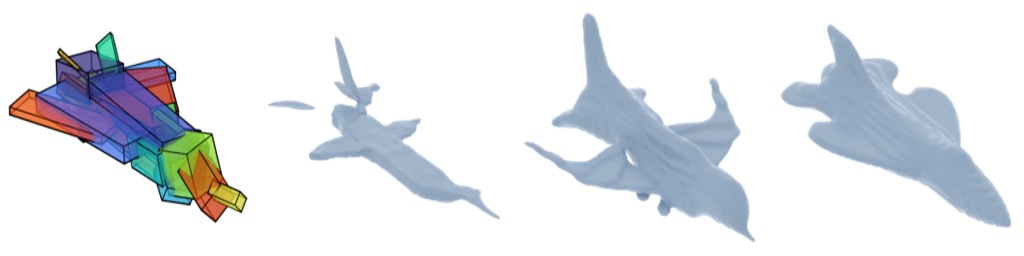}} \\
\multicolumn{4}{c|}{\includegraphics[width=.25\textwidth]{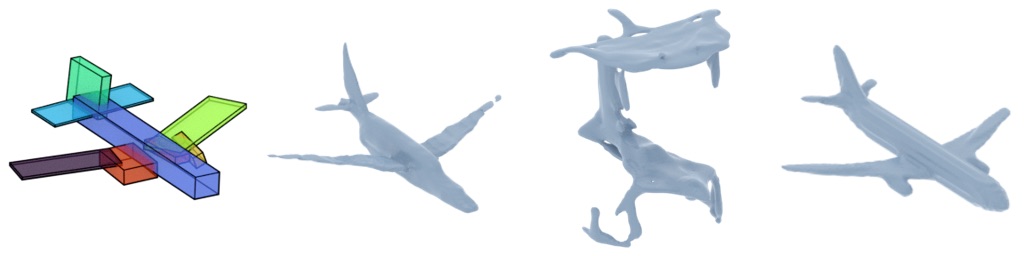}} &
\multicolumn{4}{c|}{\includegraphics[width=.25\textwidth]{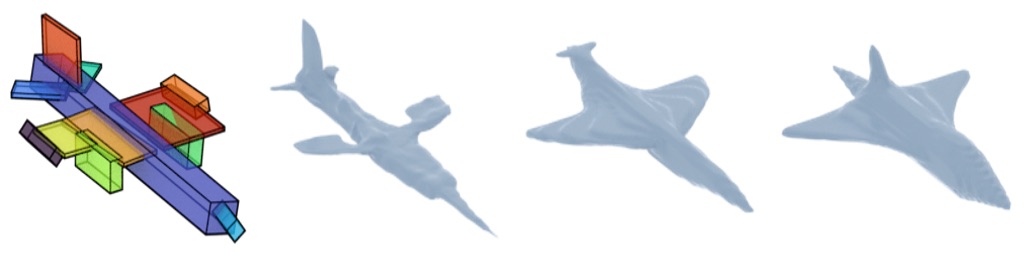}} &
\multicolumn{4}{c|}{\includegraphics[width=.25\textwidth]{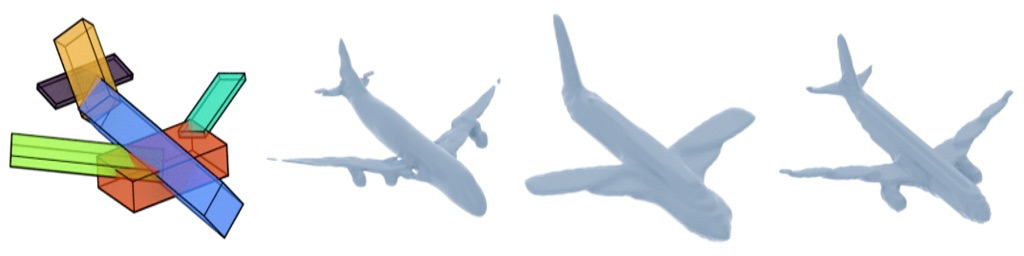}} &
\multicolumn{4}{c}{\includegraphics[width=.25\textwidth]{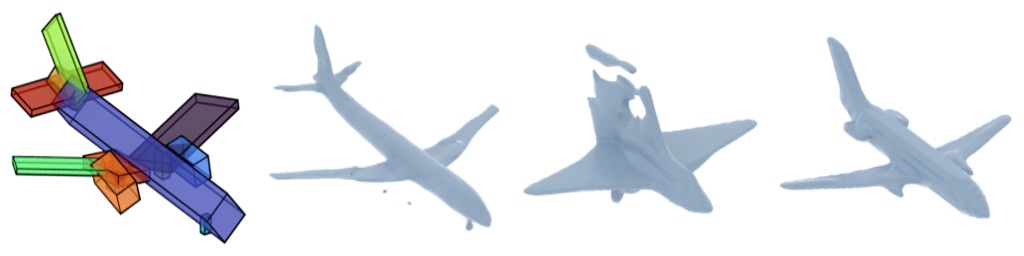}} \\
\multicolumn{4}{c|}{\includegraphics[width=.25\textwidth]{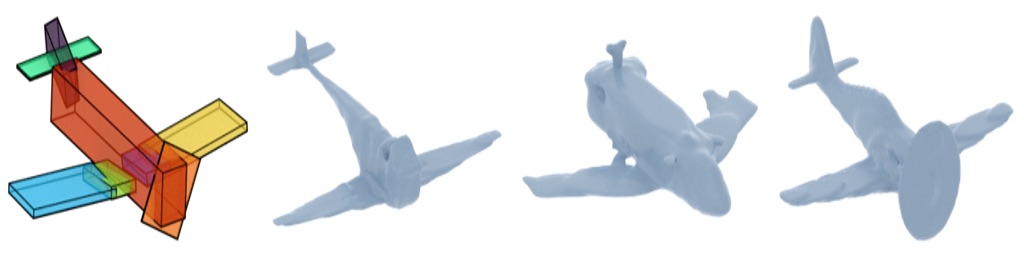}} &
\multicolumn{4}{c|}{\includegraphics[width=.25\textwidth]{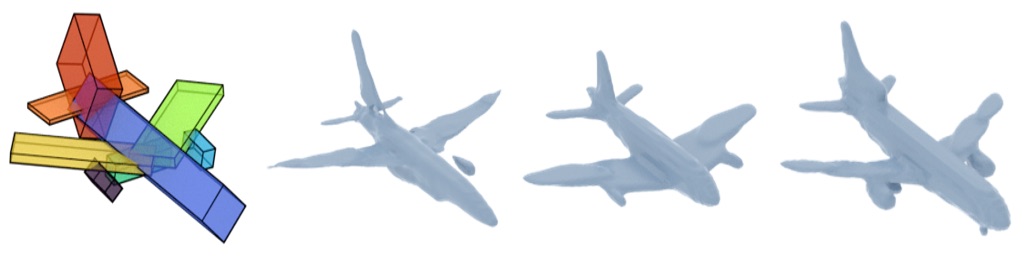}} &
\multicolumn{4}{c|}{\includegraphics[width=.25\textwidth]{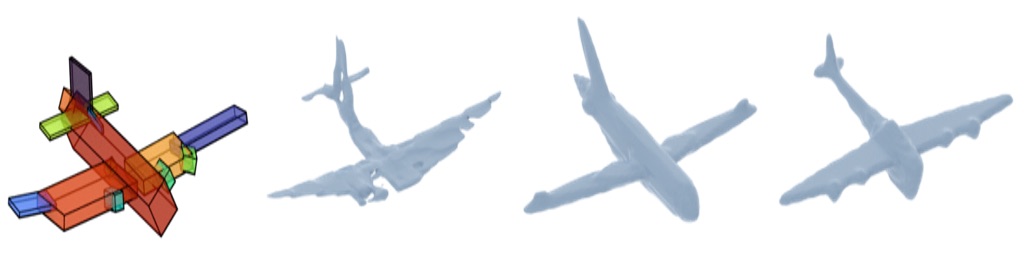}} &
\multicolumn{4}{c}{\includegraphics[width=.25\textwidth]{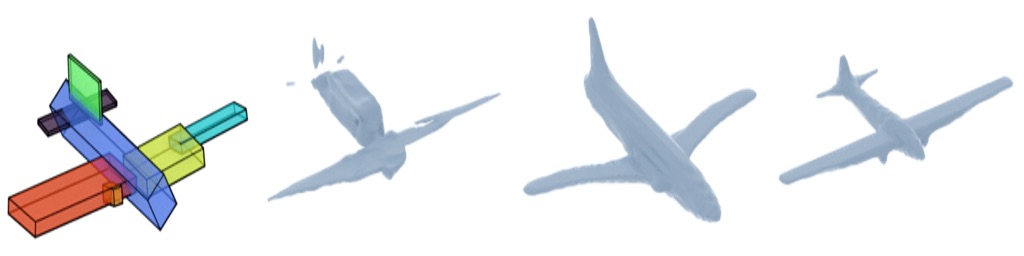}} \\
\multicolumn{4}{c|}{\includegraphics[width=.25\textwidth]{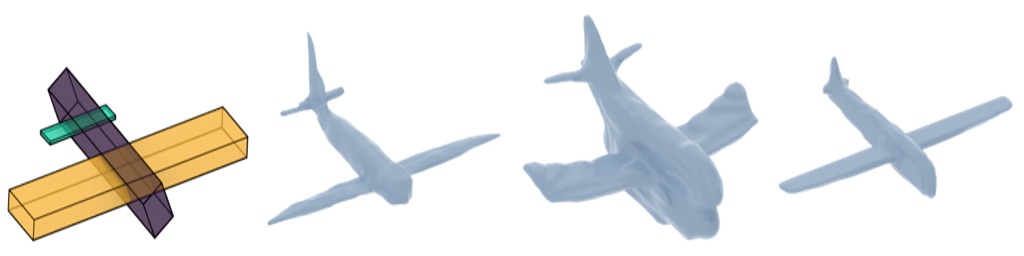}} &
\multicolumn{4}{c|}{\includegraphics[width=.25\textwidth]{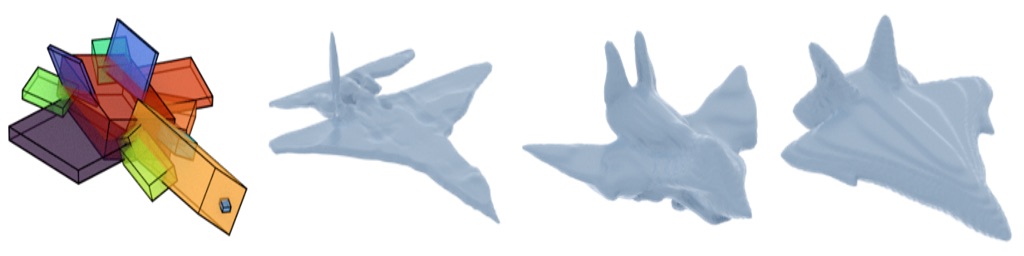}} &
\multicolumn{4}{c|}{\includegraphics[width=.25\textwidth]{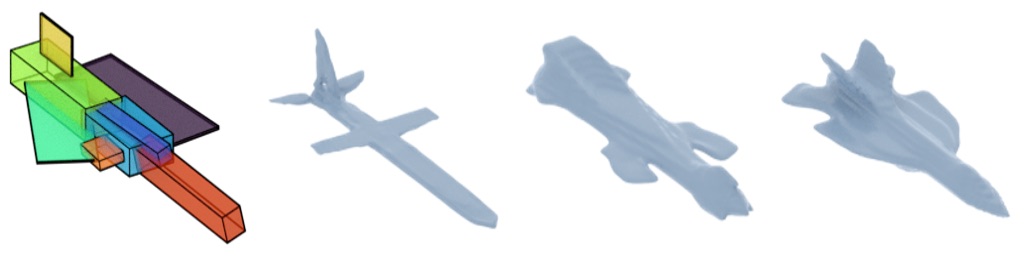}} &
\multicolumn{4}{c}{\includegraphics[width=.25\textwidth]{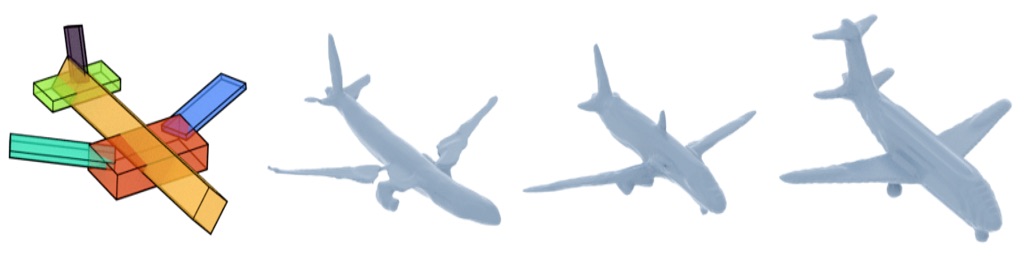}} \\
\multicolumn{4}{c|}{\includegraphics[width=.25\textwidth]{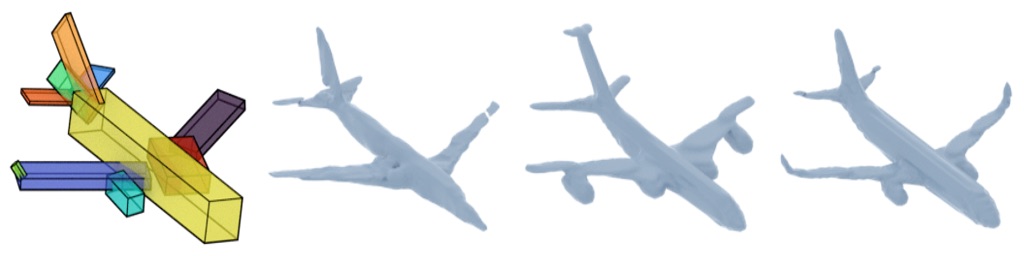}} &
\multicolumn{4}{c|}{\includegraphics[width=.25\textwidth]{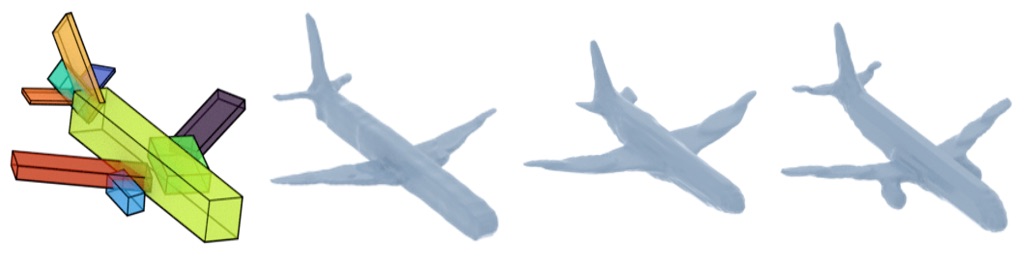}} &
\multicolumn{4}{c|}{\includegraphics[width=.25\textwidth]{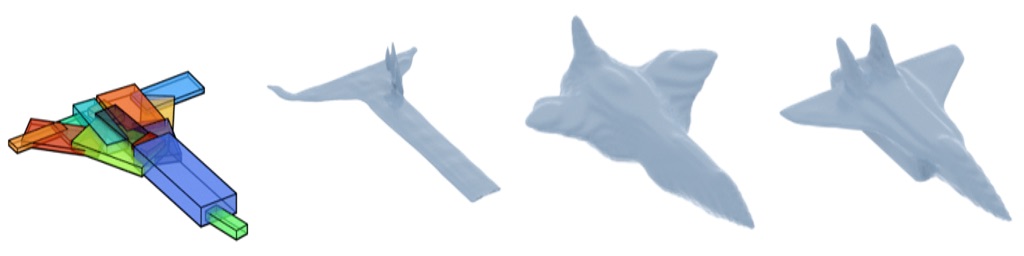}} &
\multicolumn{4}{c}{\includegraphics[width=.25\textwidth]{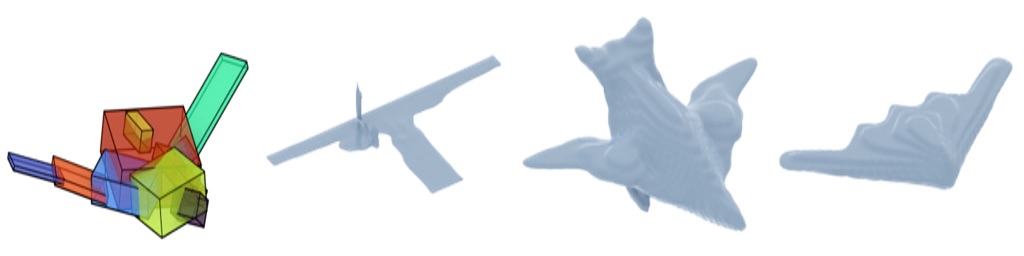}} \\
\multicolumn{4}{c|}{\includegraphics[width=.25\textwidth]{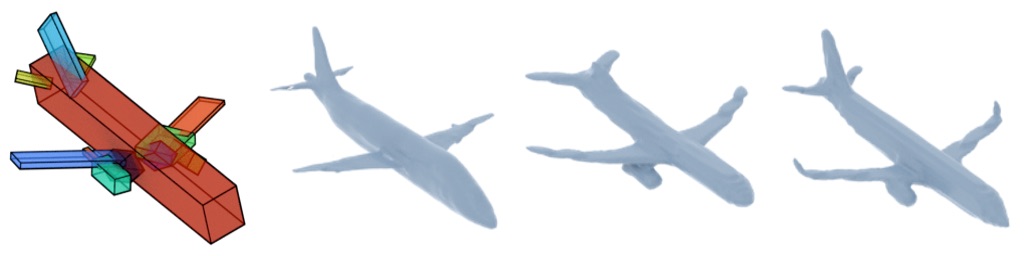}} &
\multicolumn{4}{c|}{\includegraphics[width=.25\textwidth]{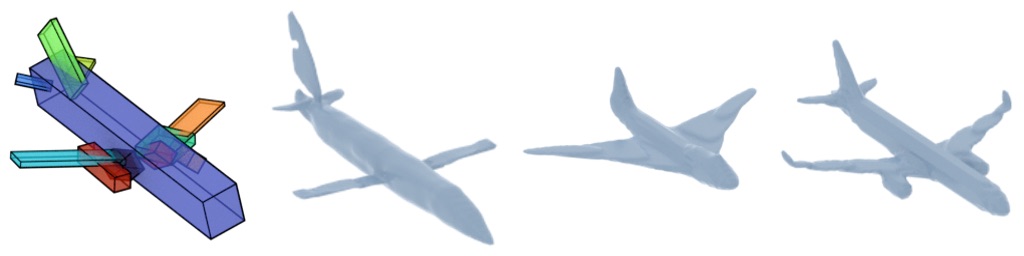}} &
\multicolumn{4}{c|}{\includegraphics[width=.25\textwidth]{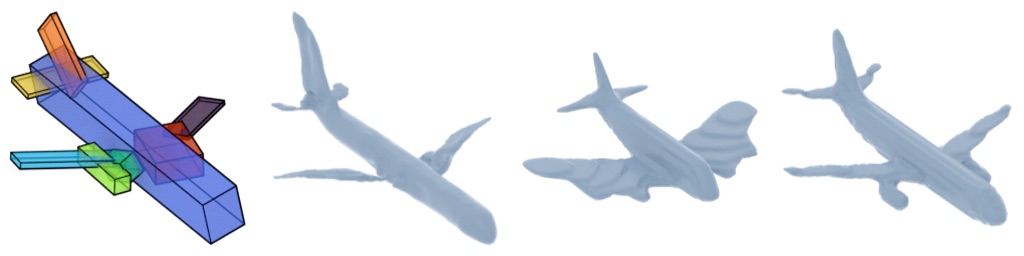}} &
\multicolumn{4}{c}{\includegraphics[width=.25\textwidth]{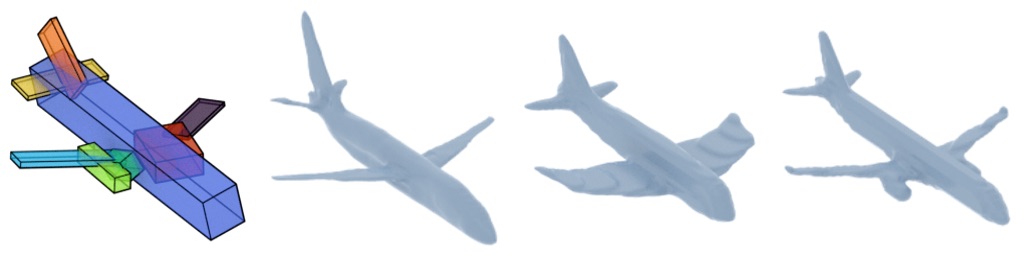}} \\
\multicolumn{4}{c|}{\includegraphics[width=.25\textwidth]{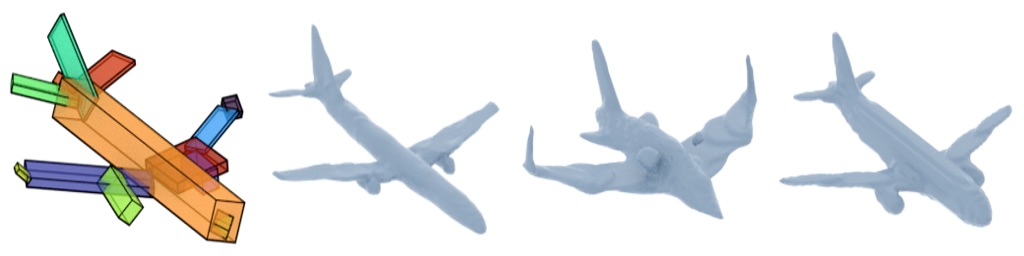}} &
\multicolumn{4}{c|}{\includegraphics[width=.25\textwidth]{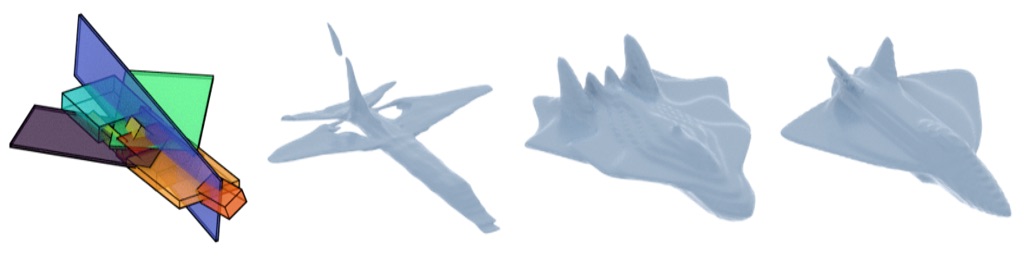}} &
\multicolumn{4}{c|}{\includegraphics[width=.25\textwidth]{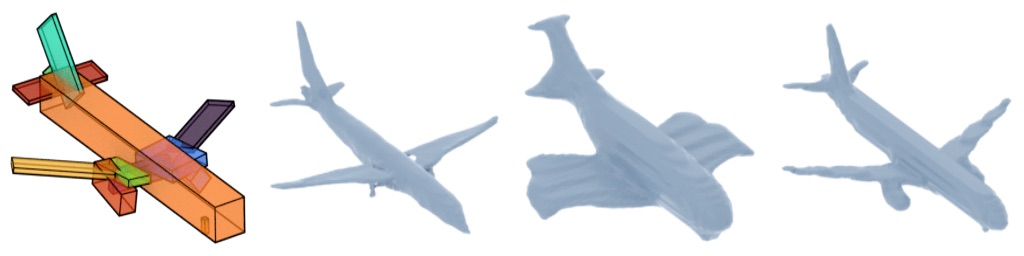}} &
\multicolumn{4}{c}{\includegraphics[width=.25\textwidth]{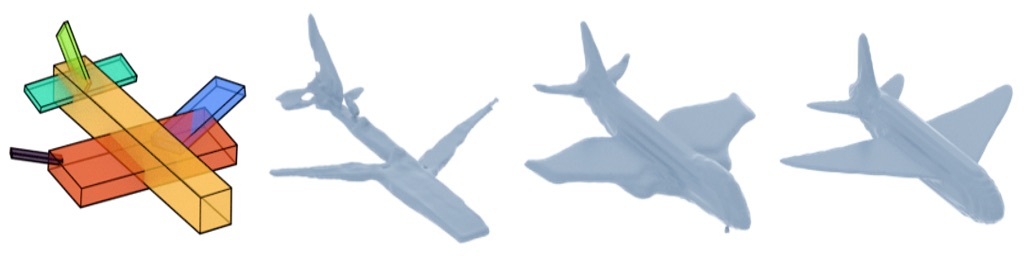}} \\
\multicolumn{4}{c|}{\includegraphics[width=.25\textwidth]{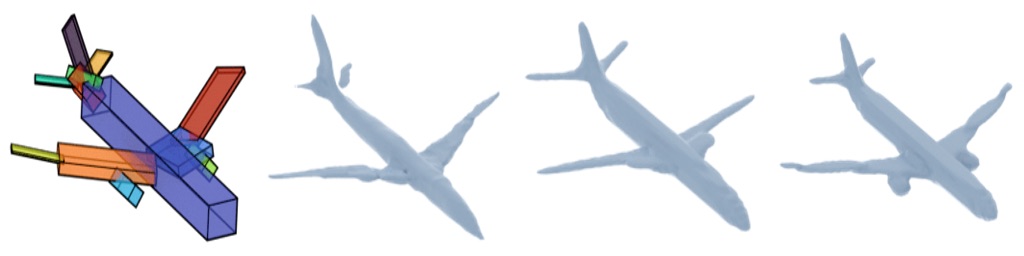}} &
\multicolumn{4}{c|}{\includegraphics[width=.25\textwidth]{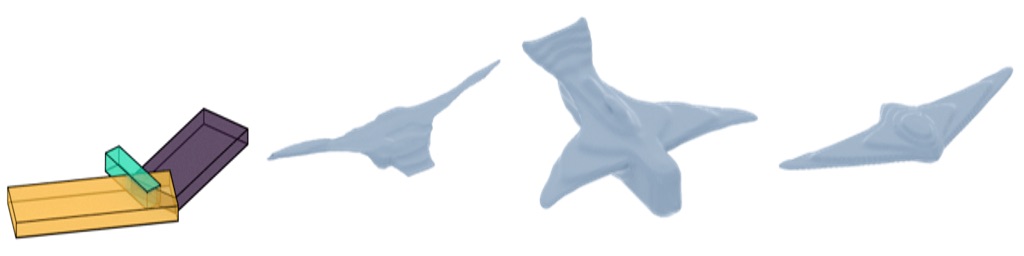}} &
\multicolumn{4}{c|}{\includegraphics[width=.25\textwidth]{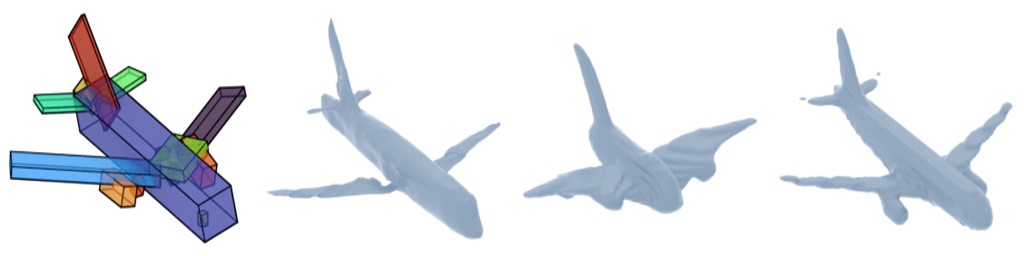}} &
\multicolumn{4}{c}{\includegraphics[width=.25\textwidth]{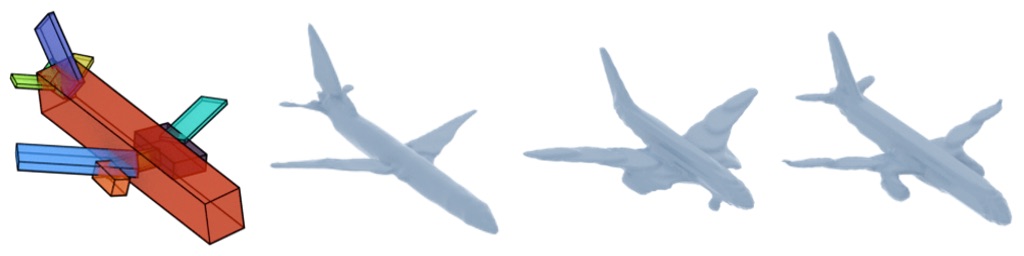}} \\
\multicolumn{4}{c|}{\includegraphics[width=.25\textwidth]{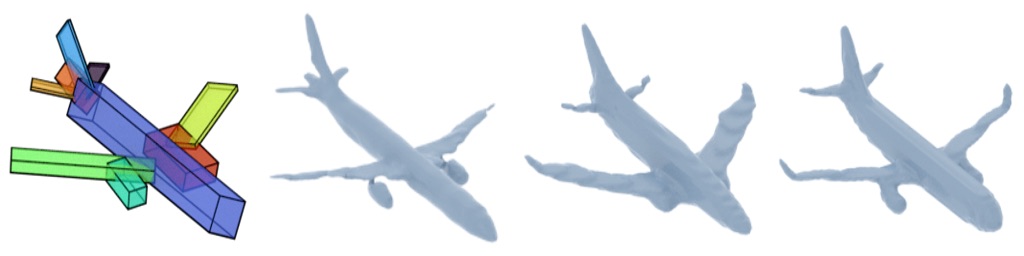}} &
\multicolumn{4}{c|}{\includegraphics[width=.25\textwidth]{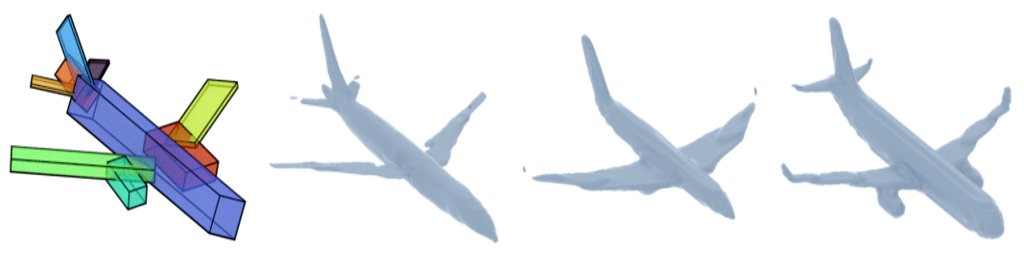}} &
\multicolumn{4}{c|}{\includegraphics[width=.25\textwidth]{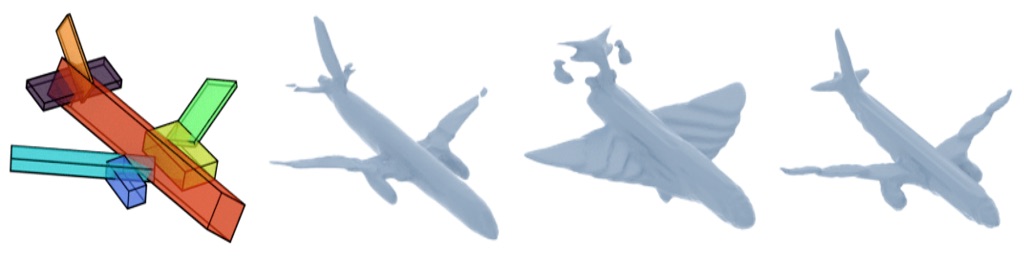}} &
\multicolumn{4}{c}{\includegraphics[width=.25\textwidth]{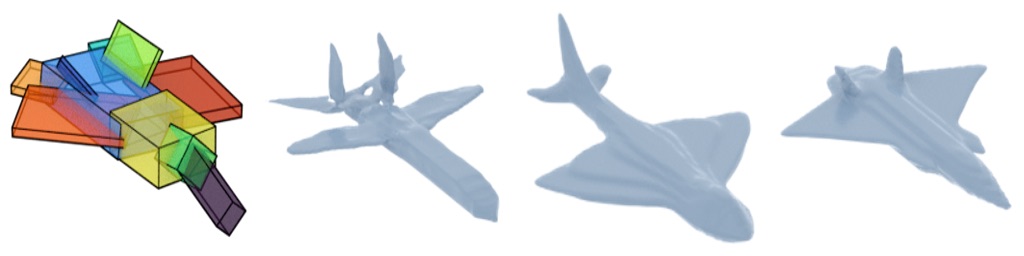}} \\
\multicolumn{4}{c|}{\includegraphics[width=.25\textwidth]{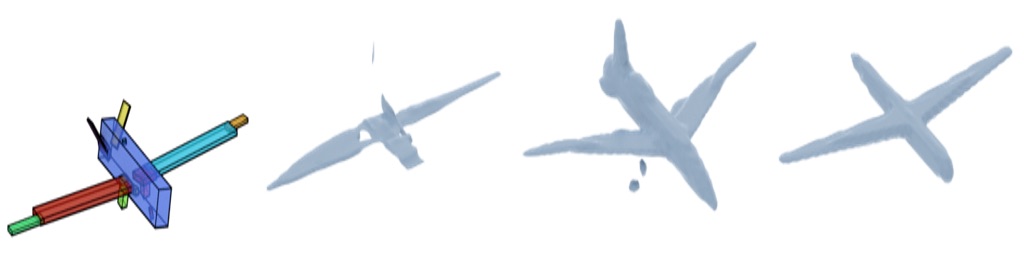}} &
\multicolumn{4}{c|}{\includegraphics[width=.25\textwidth]{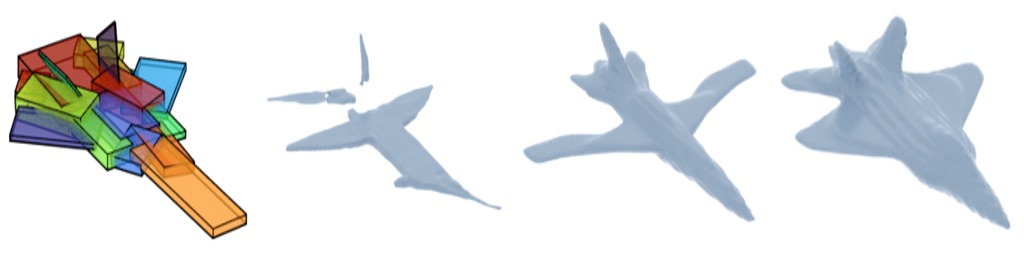}} &
\multicolumn{4}{c|}{\includegraphics[width=.25\textwidth]{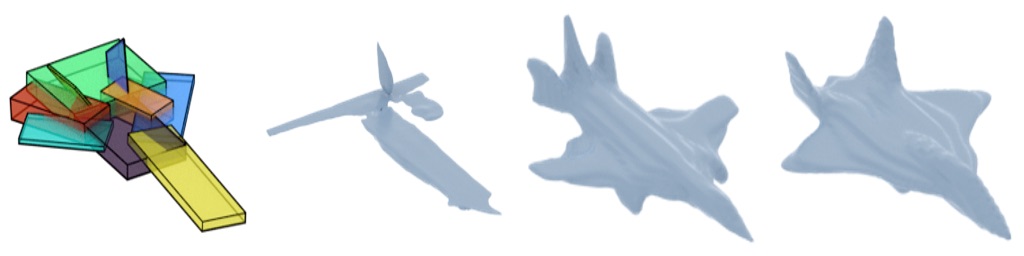}} &
\multicolumn{4}{c}{\includegraphics[width=.25\textwidth]{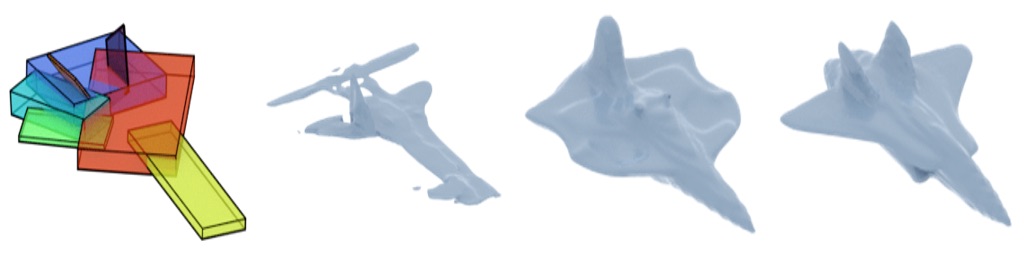}} \\
\multicolumn{4}{c|}{\includegraphics[width=.25\textwidth]{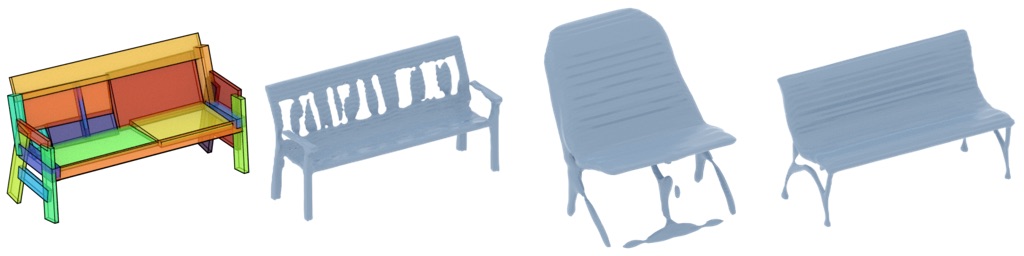}} &
\multicolumn{4}{c|}{\includegraphics[width=.25\textwidth]{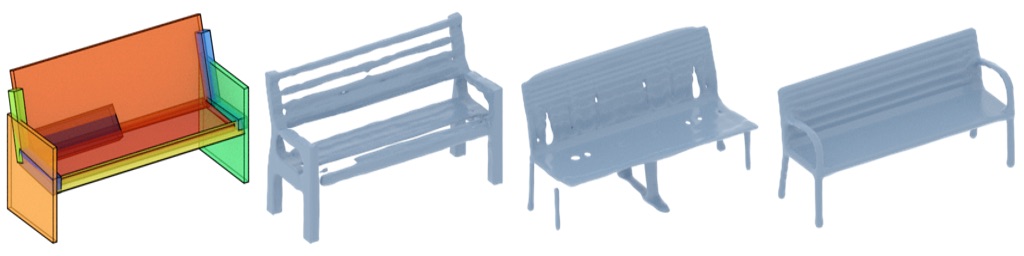}} &
\multicolumn{4}{c|}{\includegraphics[width=.25\textwidth]{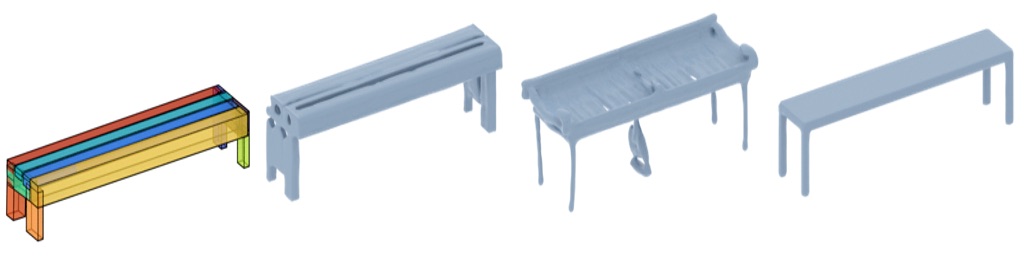}} &
\multicolumn{4}{c}{\includegraphics[width=.25\textwidth]{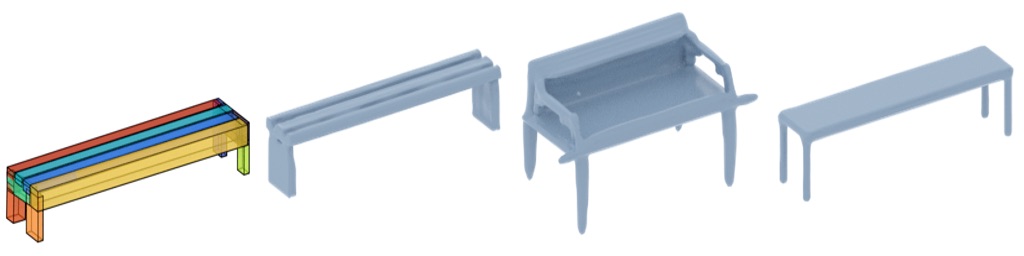}} \\
\multicolumn{4}{c|}{\includegraphics[width=.25\textwidth]{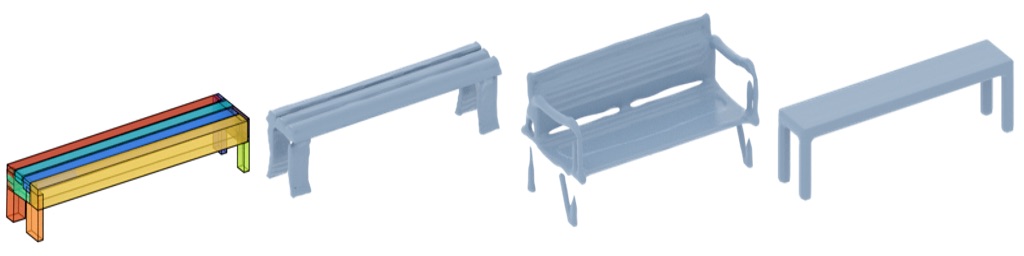}} &
\multicolumn{4}{c|}{\includegraphics[width=.25\textwidth]{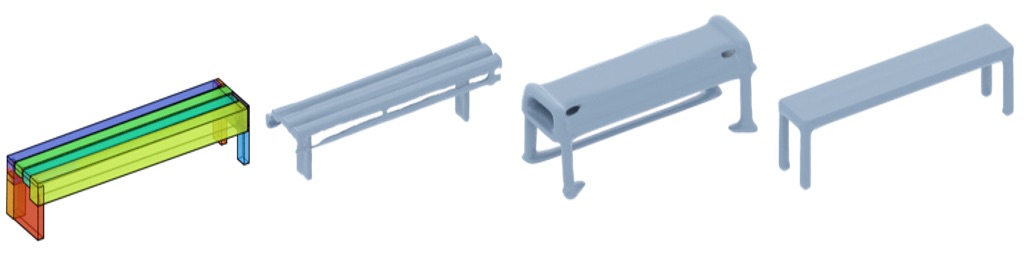}} &
\multicolumn{4}{c|}{\includegraphics[width=.25\textwidth]{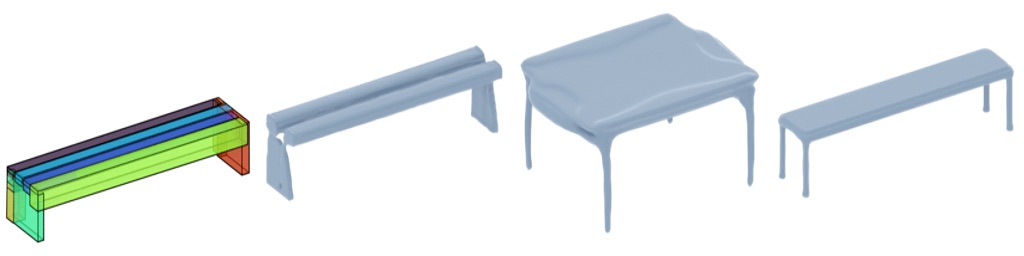}} &
\multicolumn{4}{c}{\includegraphics[width=.25\textwidth]{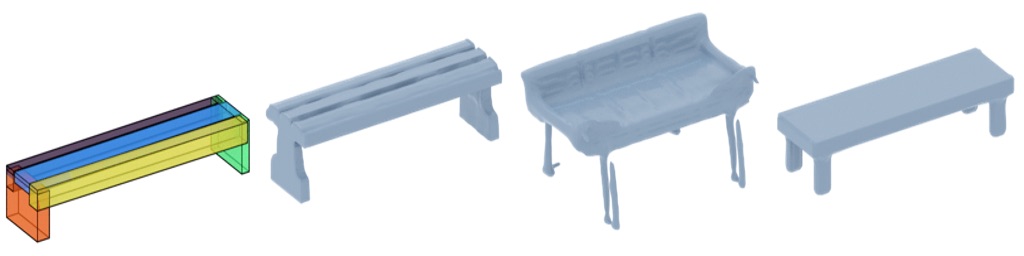}} \\
\multicolumn{4}{c|}{\includegraphics[width=.25\textwidth]{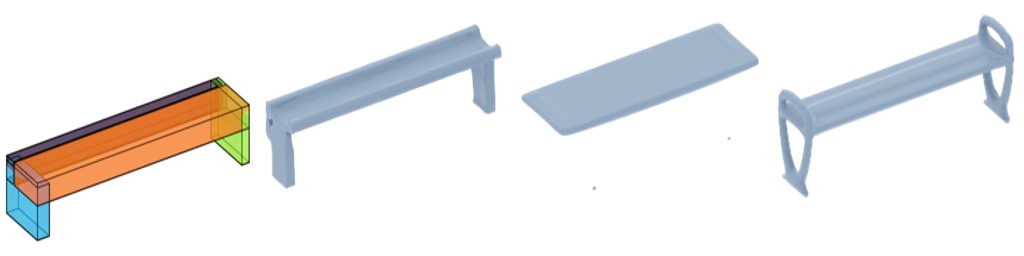}} &
\multicolumn{4}{c|}{\includegraphics[width=.25\textwidth]{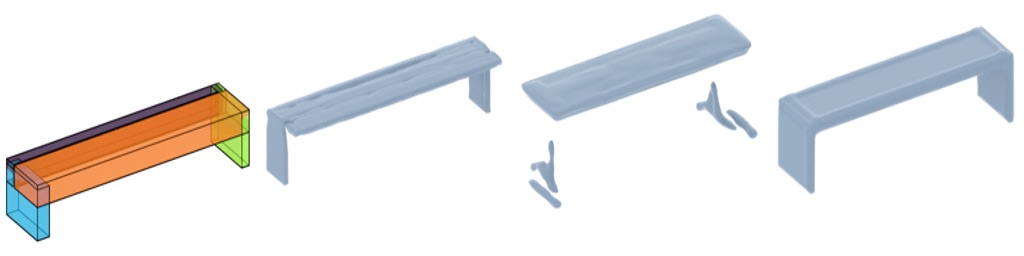}} &
\multicolumn{4}{c|}{\includegraphics[width=.25\textwidth]{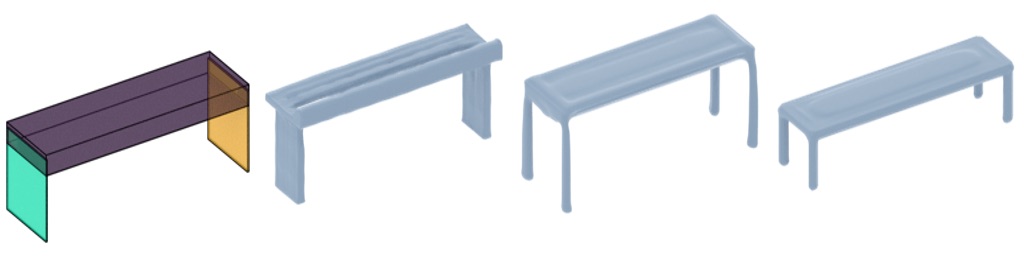}} &
\multicolumn{4}{c}{\includegraphics[width=.25\textwidth]{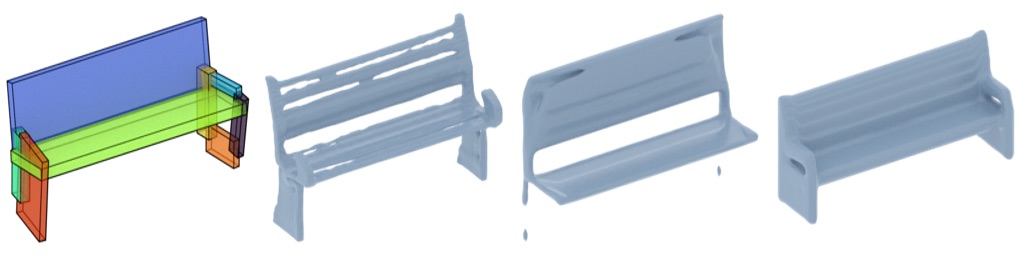}} \\
\multicolumn{4}{c|}{\includegraphics[width=.25\textwidth]{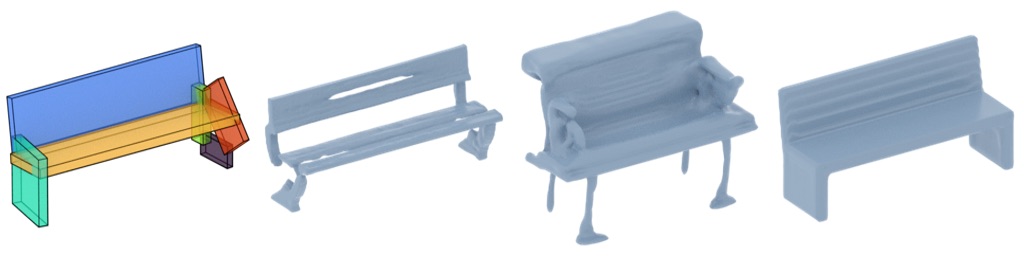}} &
\multicolumn{4}{c|}{\includegraphics[width=.25\textwidth]{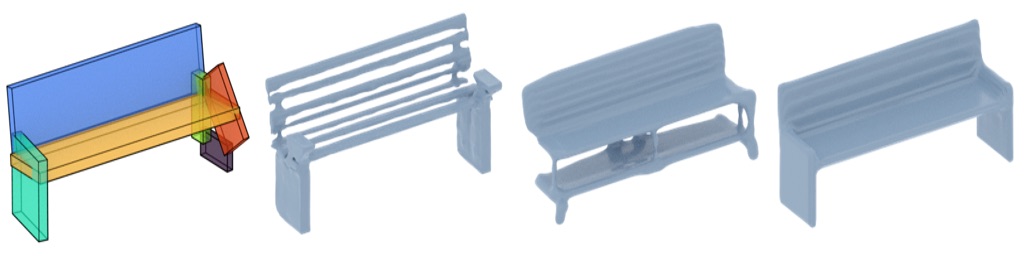}} &
\multicolumn{4}{c|}{\includegraphics[width=.25\textwidth]{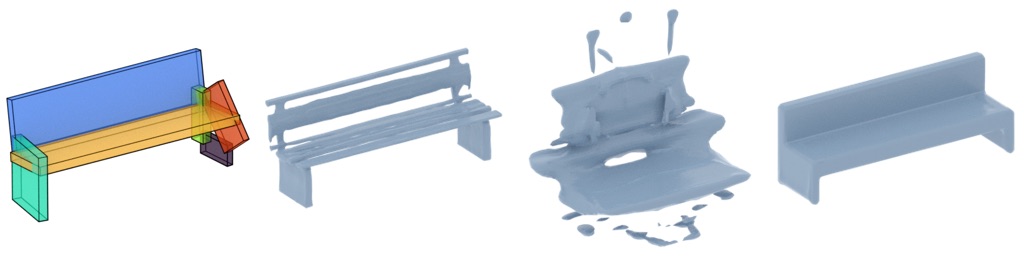}} &
\multicolumn{4}{c}{\includegraphics[width=.25\textwidth]{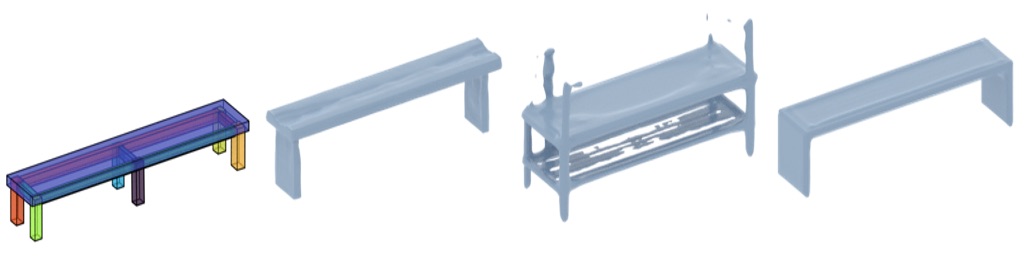}} \\
\multicolumn{4}{c|}{\includegraphics[width=.25\textwidth]{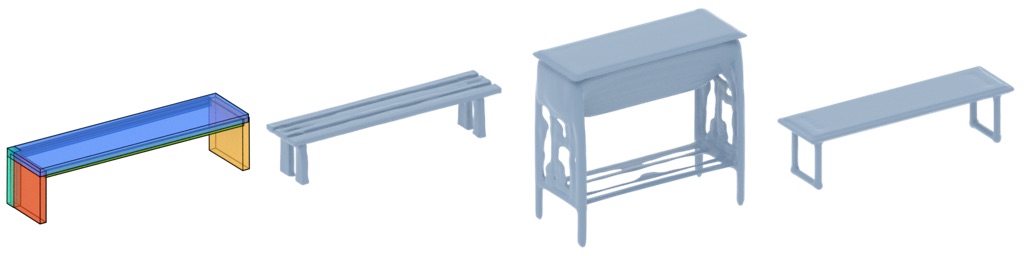}} &
\multicolumn{4}{c|}{\includegraphics[width=.25\textwidth]{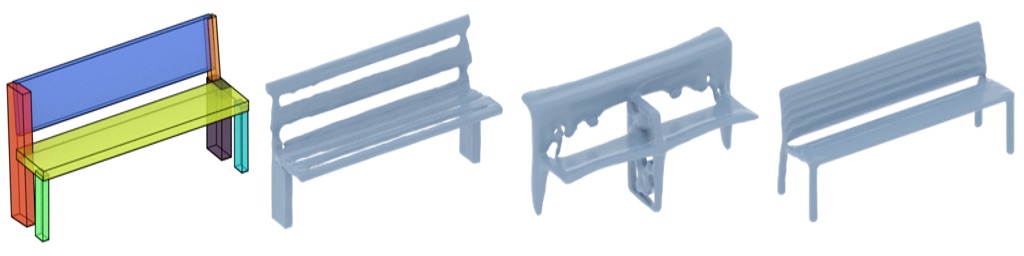}} &
\multicolumn{4}{c|}{\includegraphics[width=.25\textwidth]{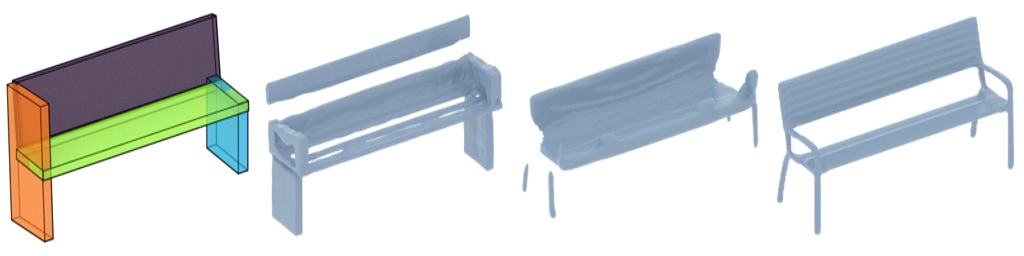}} &
\multicolumn{4}{c}{\includegraphics[width=.25\textwidth]{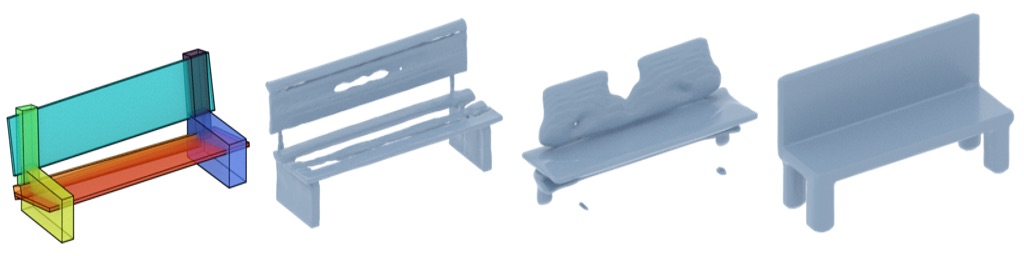}} \\

\end{tabularx}

\caption{\textbf{Gallery of our generated bounding boxes and their final decoded 3D shapes by box-conditioned shape generation network.} Each pair of columns shows the input condition bounding box (left) and its corresponding decoded 3D shape (right).}
\label{fig:stage_2_comparison_more}
\end{figure*}

\begin{figure*}[p!]
\ContinuedFloat
\centering
\scriptsize
\setlength{\tabcolsep}{0em}
\begin{tabularx}{\linewidth}{YYYY | YYYY | YYYY | YYYY}
\rotatebox{0}{\makecell{Input\\Boxes}} & \rotatebox{0}{\makecell{Spice-E\\\cite{Sella:2023SpicE}}} & \rotatebox{0}{\makecell{Gated\\3DS2V~\cite{Zhang:2023Shape2Vec}}} & \rotatebox{0}{Ours} & \rotatebox{0}{\makecell{Input\\Boxes}} & \rotatebox{0}{\makecell{Spice-E\\\cite{Sella:2023SpicE}}} & \rotatebox{0}{\makecell{Gated\\3DS2V~\cite{Zhang:2023Shape2Vec}}} & \rotatebox{0}{Ours} & \rotatebox{0}{\makecell{Input\\Boxes}} & \rotatebox{0}{\makecell{Spice-E\\\cite{Sella:2023SpicE}}} & \rotatebox{0}{\makecell{Gated\\3DS2V~\cite{Zhang:2023Shape2Vec}}} & \rotatebox{0}{Ours} & \rotatebox{0}{\makecell{Input\\Boxes}} & \rotatebox{0}{\makecell{Spice-E\\\cite{Sella:2023SpicE}}} & \rotatebox{0}{\makecell{Gated\\3DS2V~\cite{Zhang:2023Shape2Vec}}} & \rotatebox{0}{Ours}  \\ 

\midrule

\multicolumn{4}{c|}{\includegraphics[width=.25\textwidth]{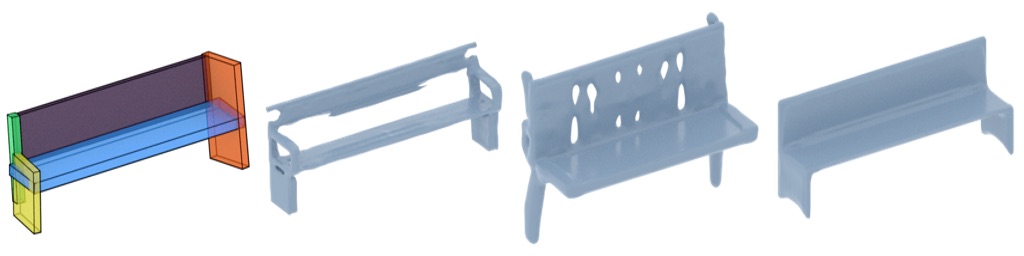}} &
\multicolumn{4}{c|}{\includegraphics[width=.25\textwidth]{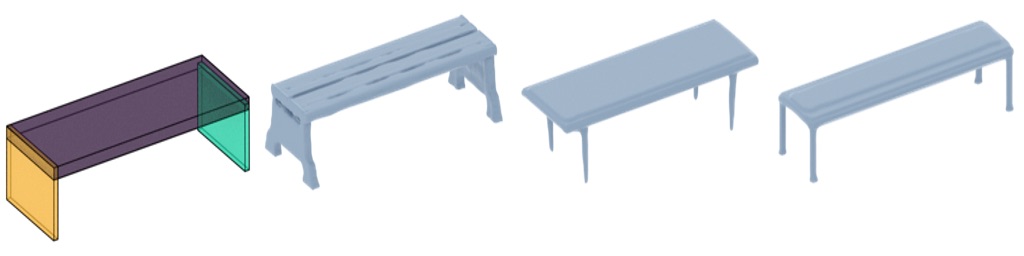}} &
\multicolumn{4}{c|}{\includegraphics[width=.25\textwidth]{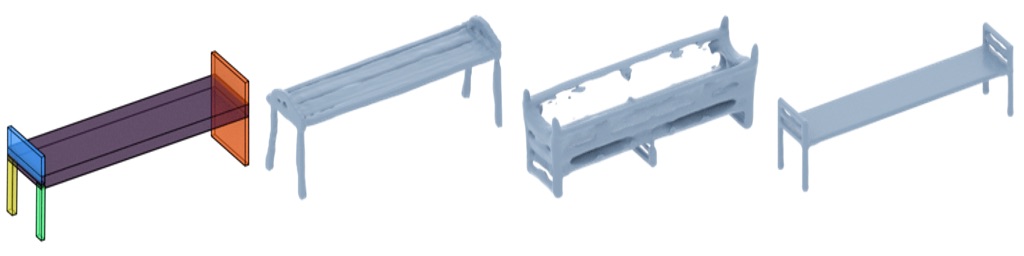}} &
\multicolumn{4}{c}{\includegraphics[width=.25\textwidth]{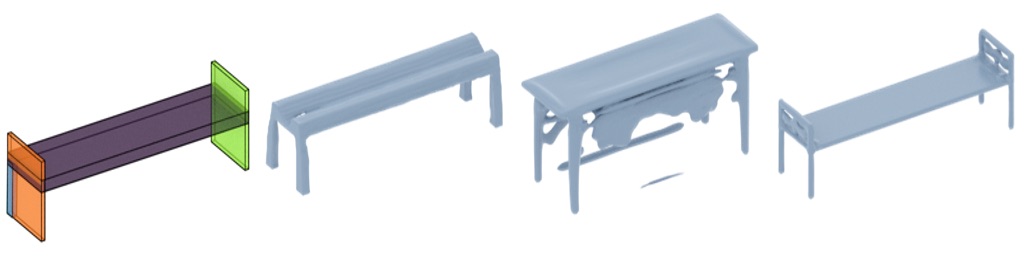}} \\
\multicolumn{4}{c|}{\includegraphics[width=.25\textwidth]{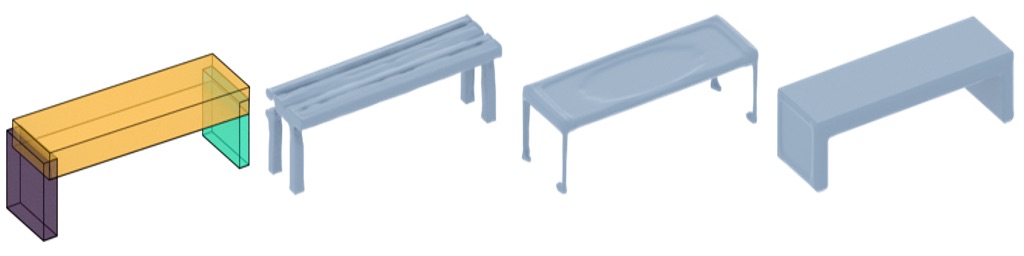}} &
\multicolumn{4}{c|}{\includegraphics[width=.25\textwidth]{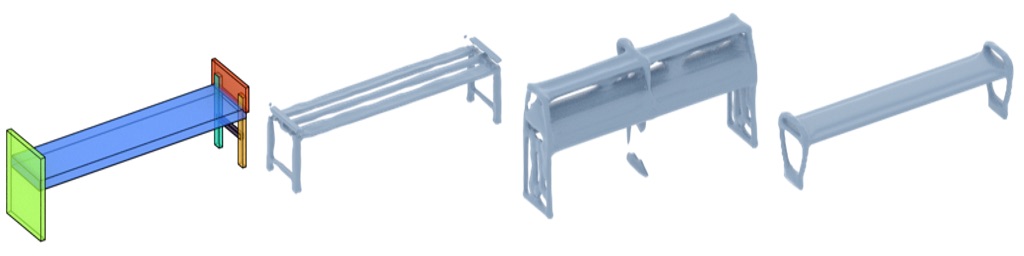}} &
\multicolumn{4}{c|}{\includegraphics[width=.25\textwidth]{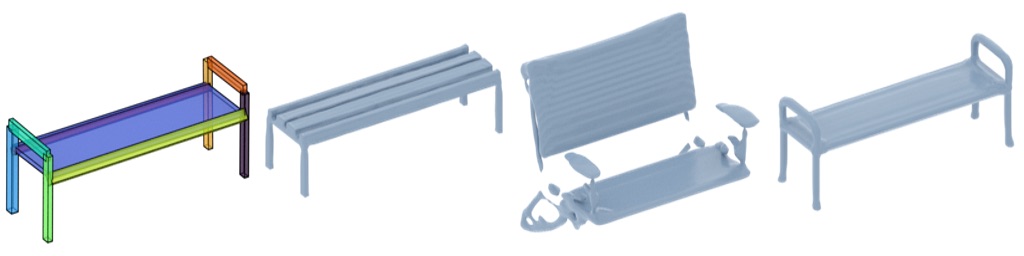}} &
\multicolumn{4}{c}{\includegraphics[width=.25\textwidth]{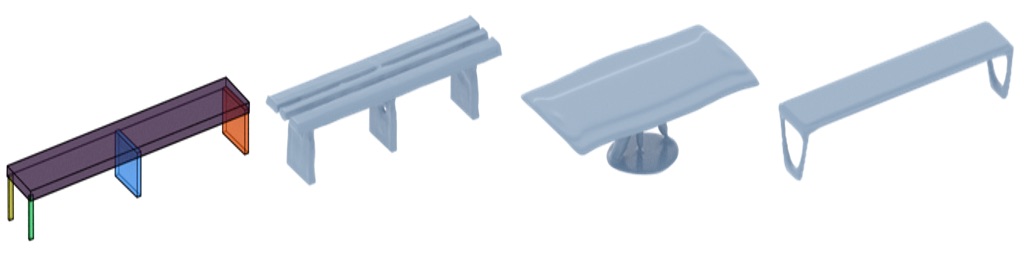}} \\
\multicolumn{4}{c|}{\includegraphics[width=.25\textwidth]{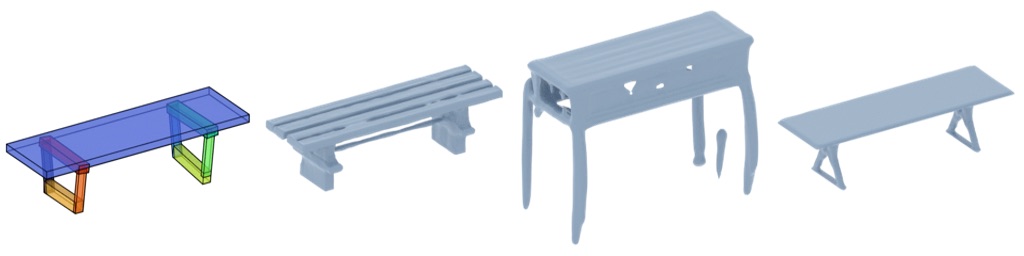}} &
\multicolumn{4}{c|}{\includegraphics[width=.25\textwidth]{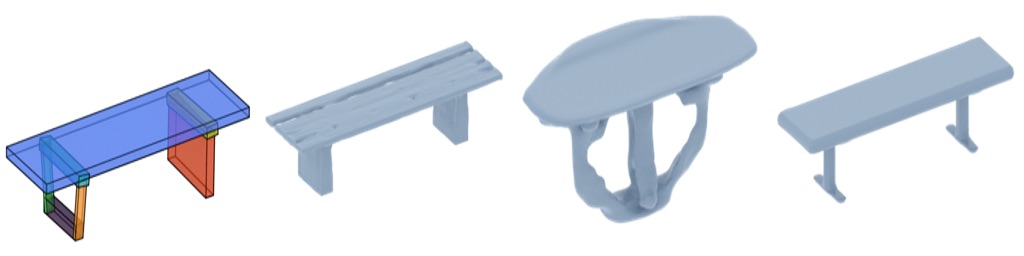}} &
\multicolumn{4}{c|}{\includegraphics[width=.25\textwidth]{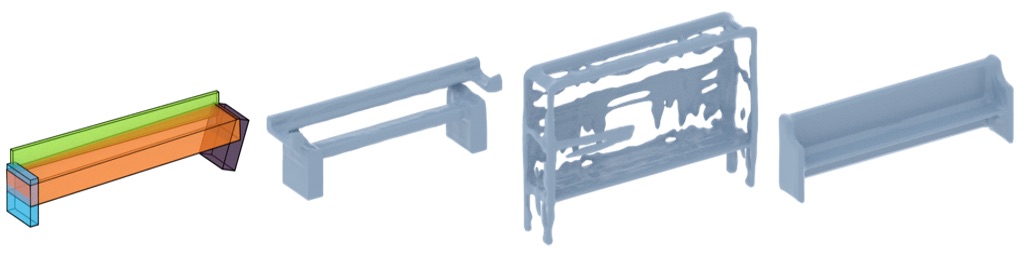}} &
\multicolumn{4}{c}{\includegraphics[width=.25\textwidth]{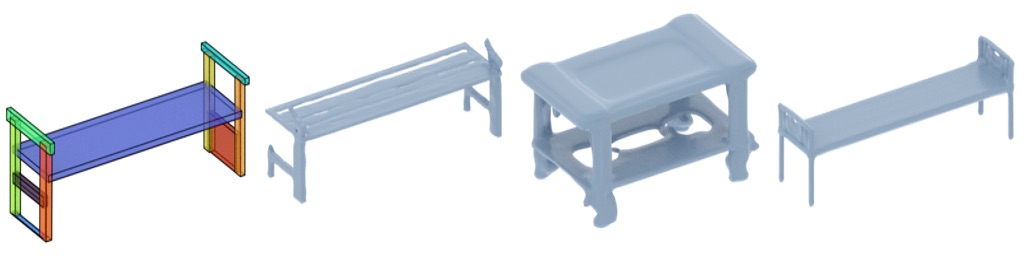}} \\
\multicolumn{4}{c|}{\includegraphics[width=.25\textwidth]{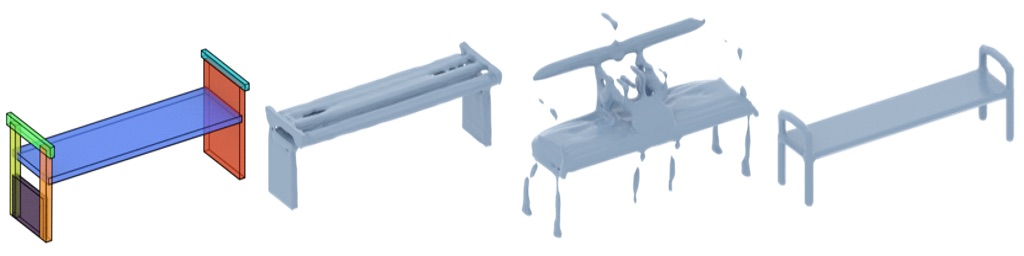}} &
\multicolumn{4}{c|}{\includegraphics[width=.25\textwidth]{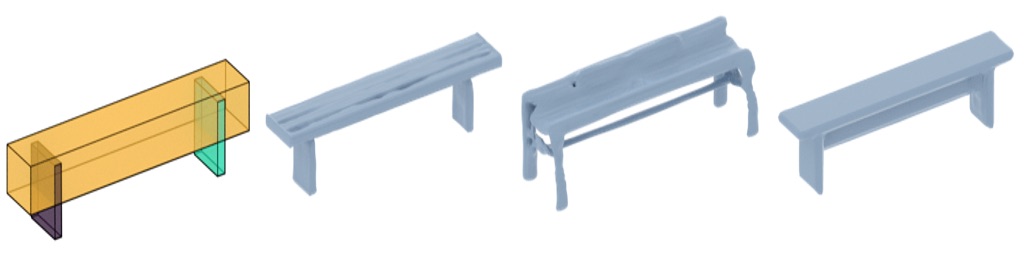}} &
\multicolumn{4}{c|}{\includegraphics[width=.25\textwidth]{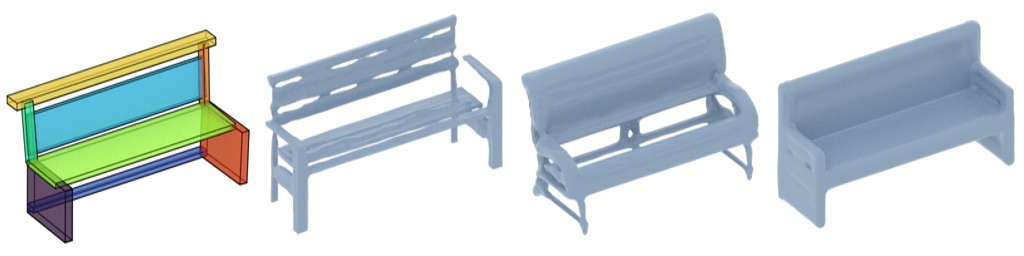}} &
\multicolumn{4}{c}{\includegraphics[width=.25\textwidth]{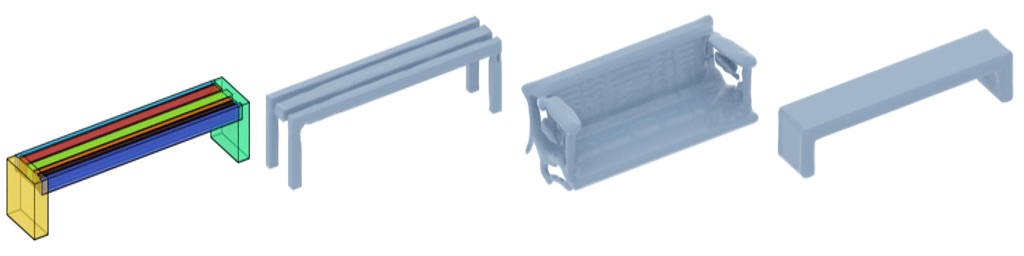}} \\
\multicolumn{4}{c|}{\includegraphics[width=.25\textwidth]{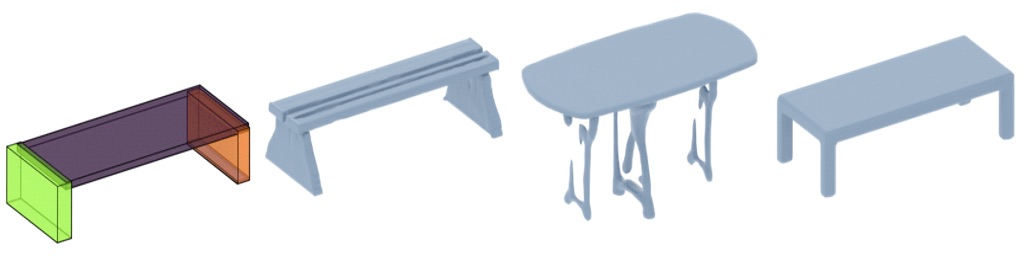}} &
\multicolumn{4}{c|}{\includegraphics[width=.25\textwidth]{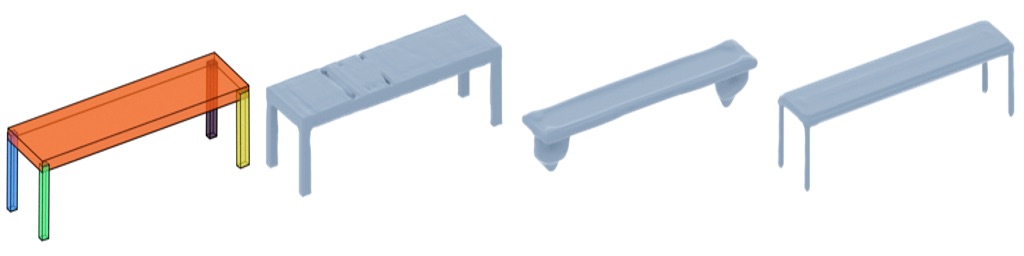}} &
\multicolumn{4}{c|}{\includegraphics[width=.25\textwidth]{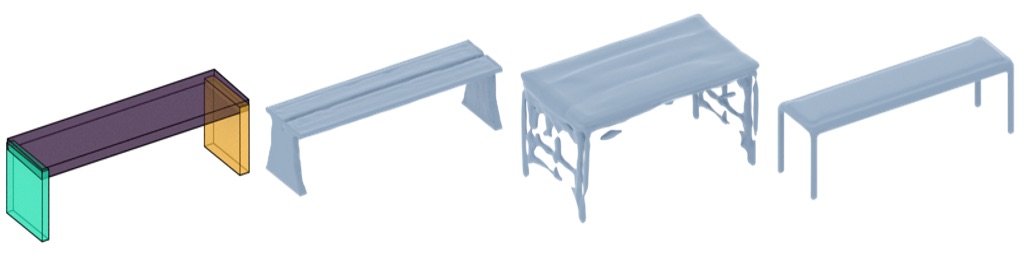}} &
\multicolumn{4}{c}{\includegraphics[width=.25\textwidth]{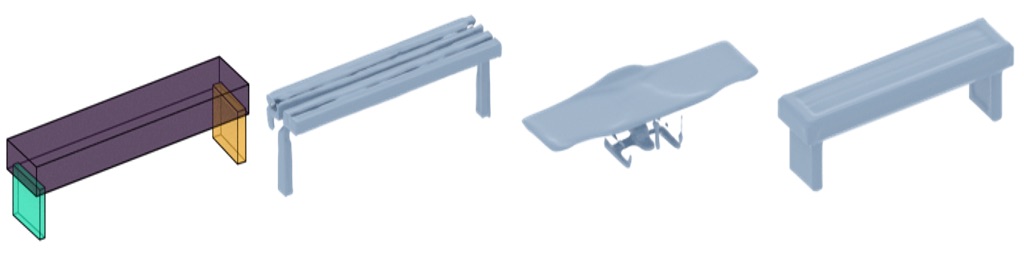}} \\
\multicolumn{4}{c|}{\includegraphics[width=.25\textwidth]{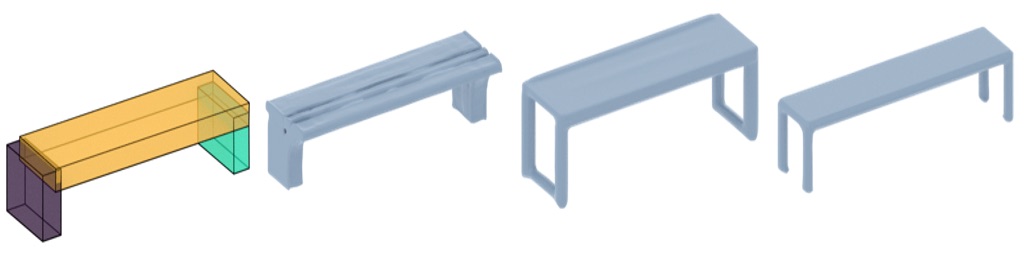}} &
\multicolumn{4}{c|}{\includegraphics[width=.25\textwidth]{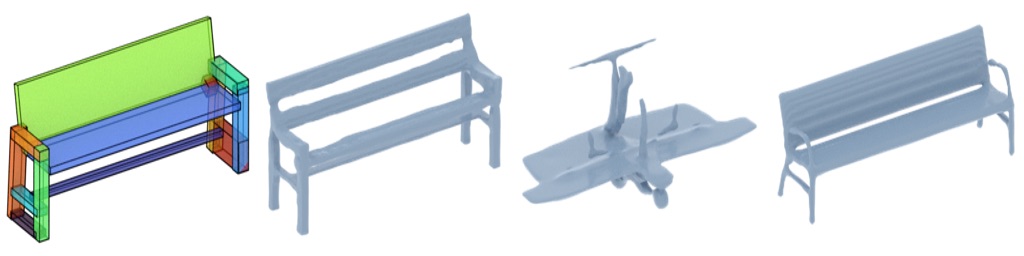}} &
\multicolumn{4}{c|}{\includegraphics[width=.25\textwidth]{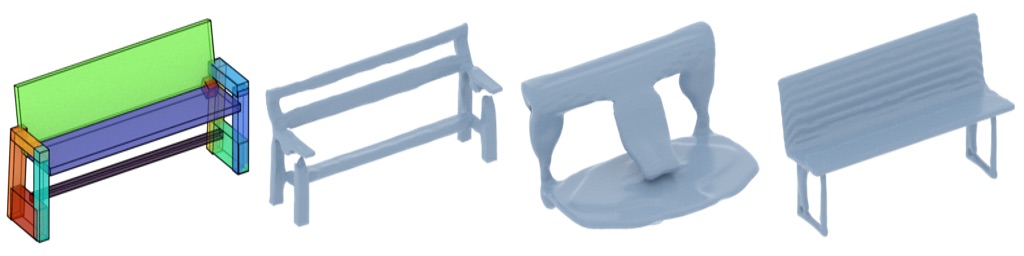}} &
\multicolumn{4}{c}{\includegraphics[width=.25\textwidth]{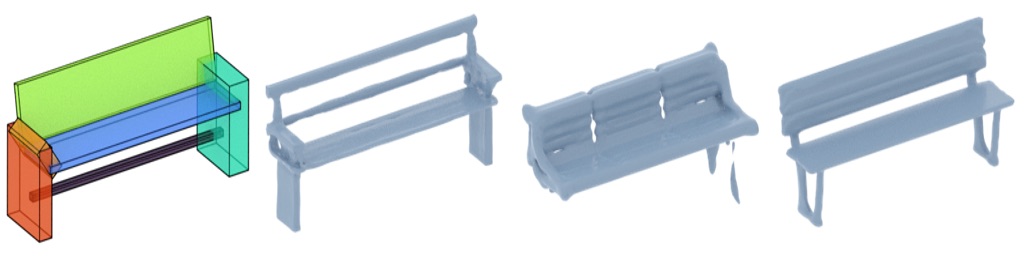}} \\
\multicolumn{4}{c|}{\includegraphics[width=.25\textwidth]{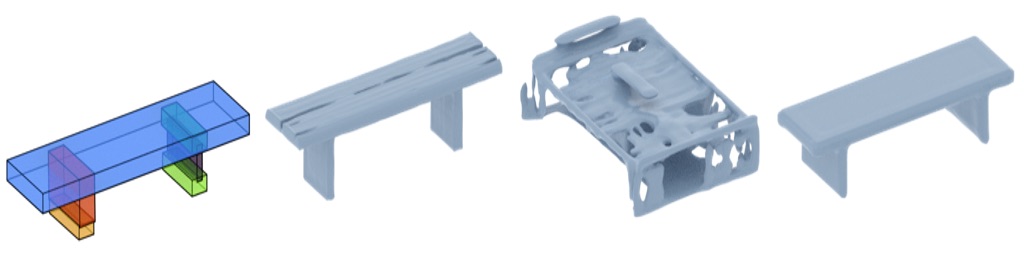}} &
\multicolumn{4}{c|}{\includegraphics[width=.25\textwidth]{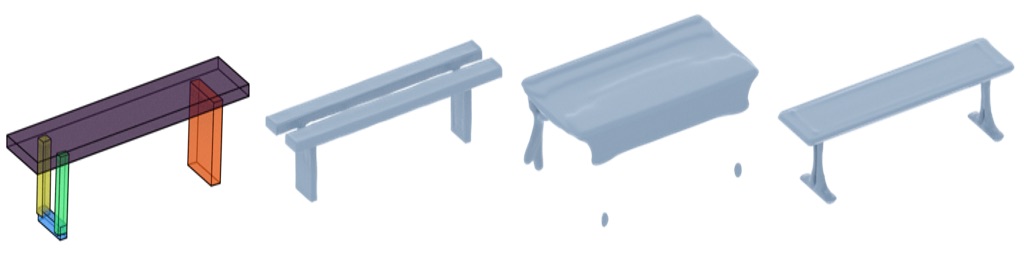}} &
\multicolumn{4}{c|}{\includegraphics[width=.25\textwidth]{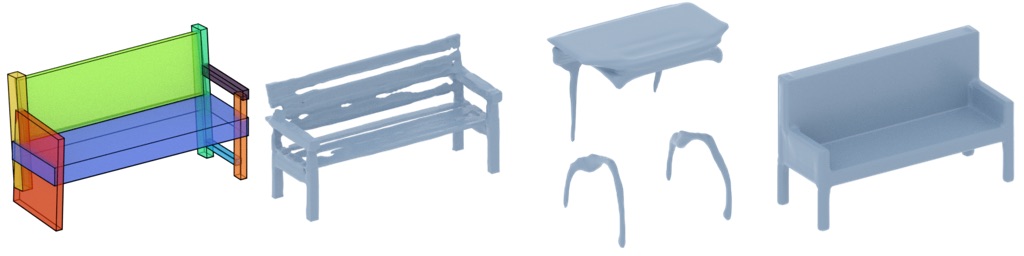}} &
\multicolumn{4}{c}{\includegraphics[width=.25\textwidth]{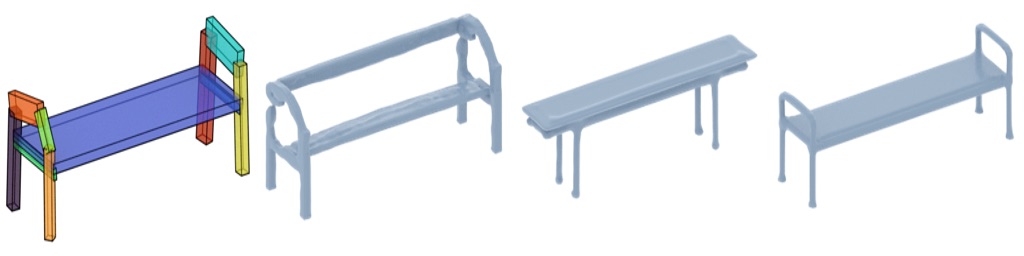}} \\
\multicolumn{4}{c|}{\includegraphics[width=.25\textwidth]{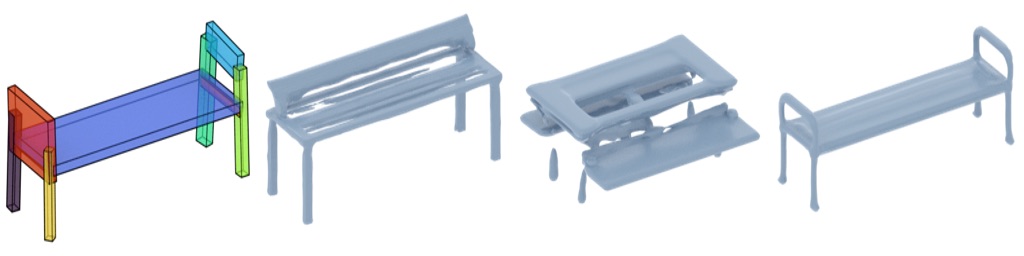}} &
\multicolumn{4}{c|}{\includegraphics[width=.25\textwidth]{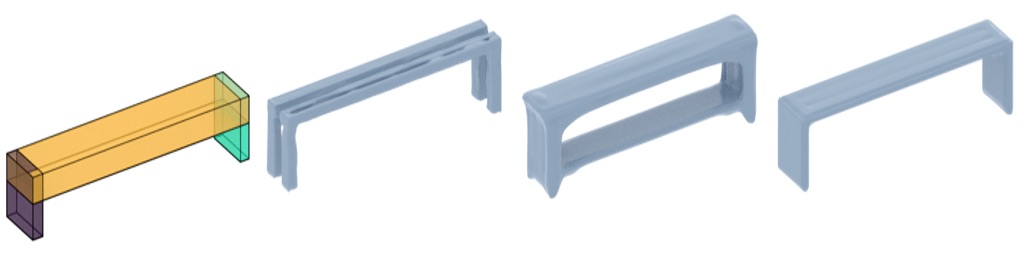}} &
\multicolumn{4}{c|}{\includegraphics[width=.25\textwidth]{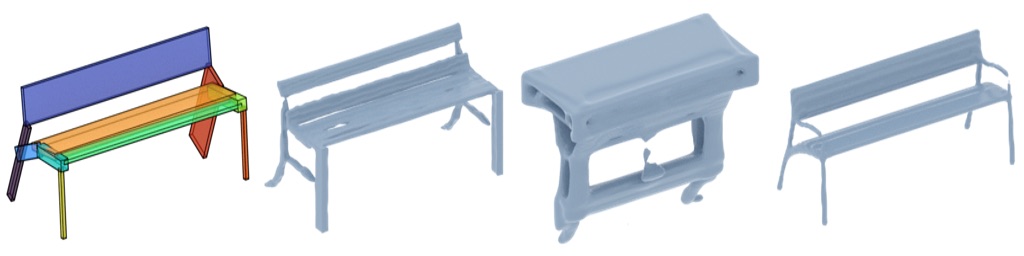}} &
\multicolumn{4}{c}{\includegraphics[width=.25\textwidth]{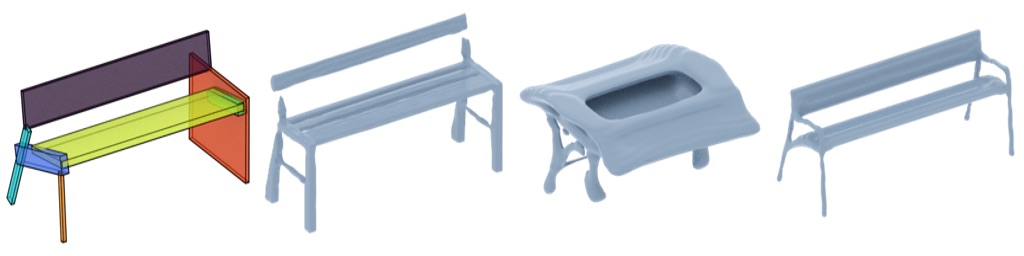}} \\
\multicolumn{4}{c|}{\includegraphics[width=.25\textwidth]{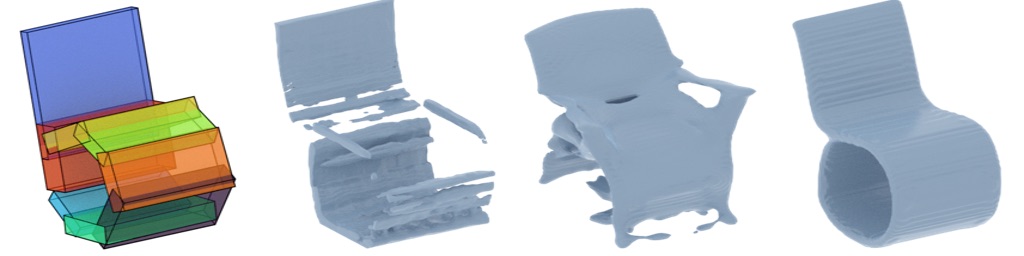}} &
\multicolumn{4}{c|}{\includegraphics[width=.25\textwidth]{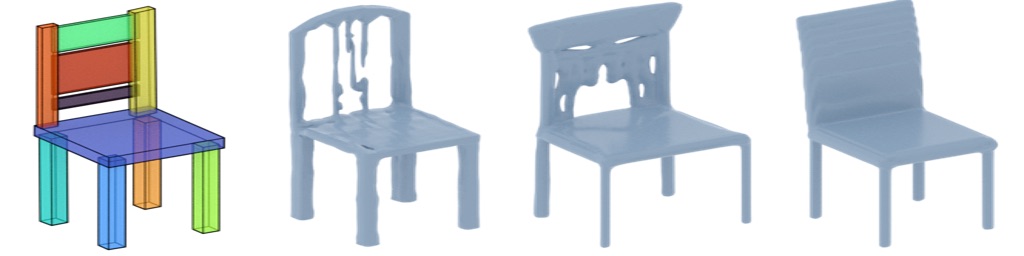}} &
\multicolumn{4}{c|}{\includegraphics[width=.25\textwidth]{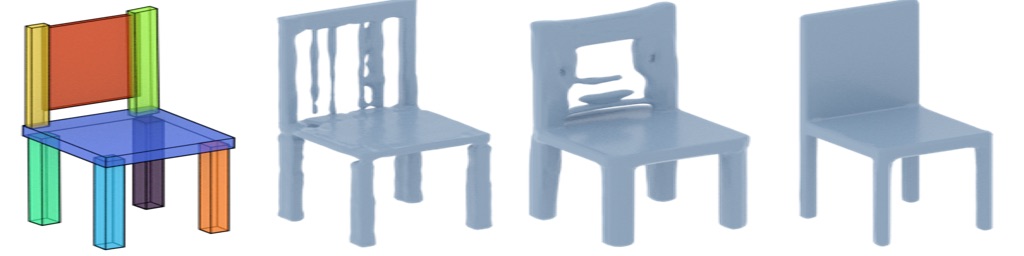}} &
\multicolumn{4}{c}{\includegraphics[width=.25\textwidth]{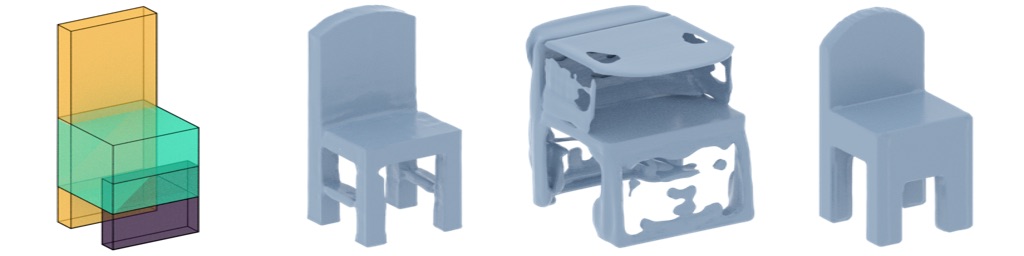}} \\
\multicolumn{4}{c|}{\includegraphics[width=.25\textwidth]{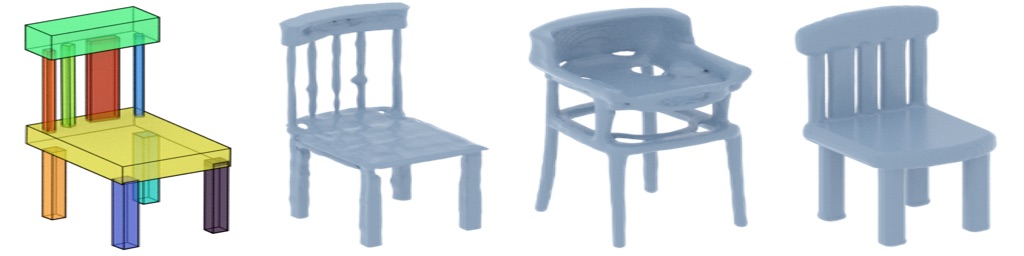}} &
\multicolumn{4}{c|}{\includegraphics[width=.25\textwidth]{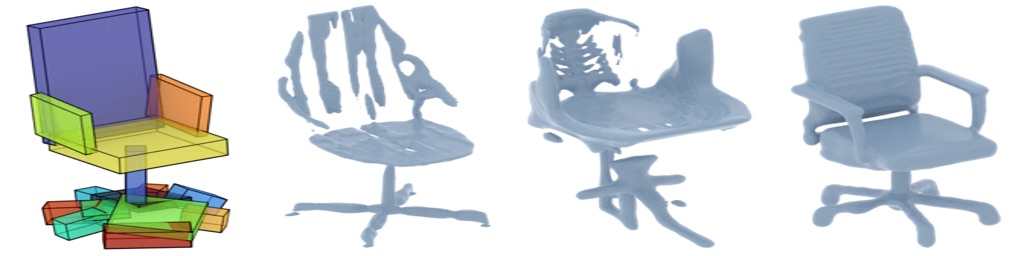}} &
\multicolumn{4}{c|}{\includegraphics[width=.25\textwidth]{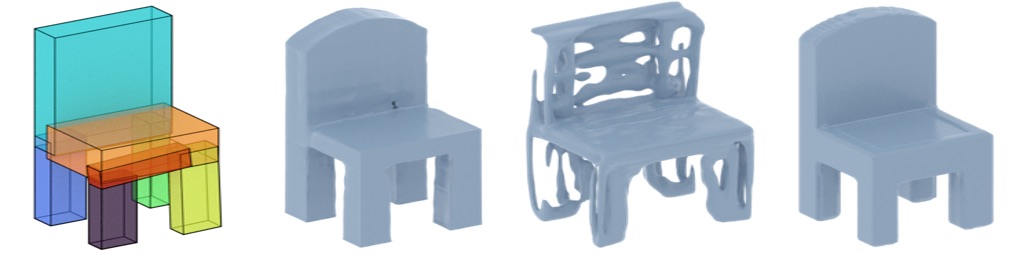}} &
\multicolumn{4}{c}{\includegraphics[width=.25\textwidth]{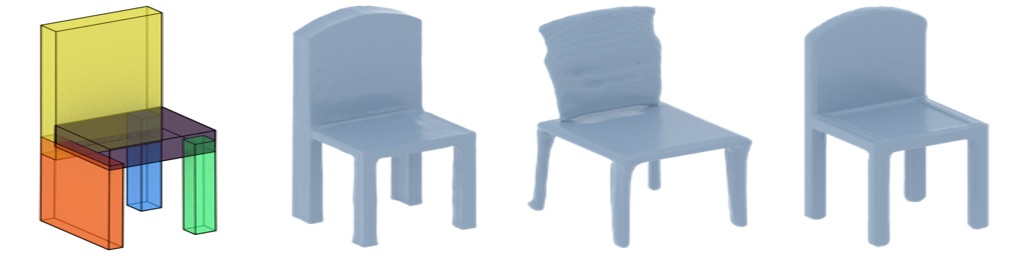}} \\
\multicolumn{4}{c|}{\includegraphics[width=.25\textwidth]{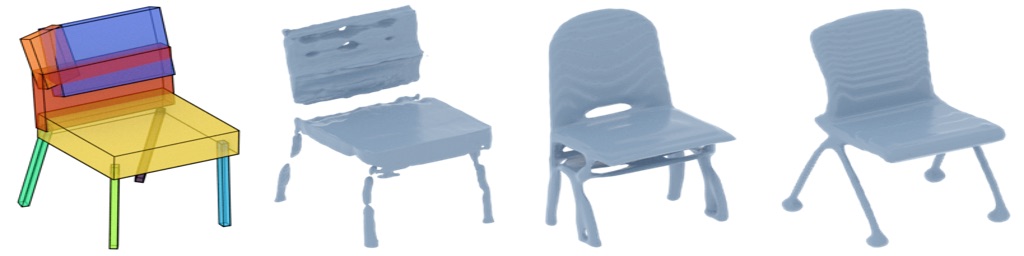}} &
\multicolumn{4}{c|}{\includegraphics[width=.25\textwidth]{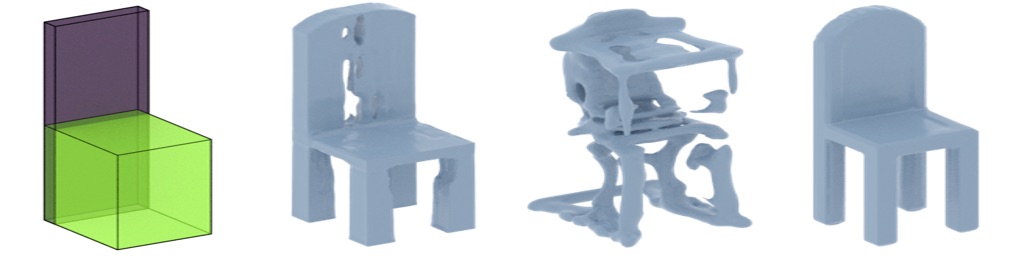}} &
\multicolumn{4}{c|}{\includegraphics[width=.25\textwidth]{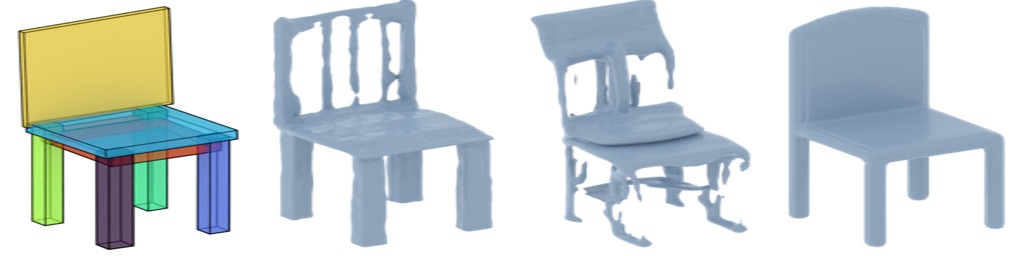}} &
\multicolumn{4}{c}{\includegraphics[width=.25\textwidth]{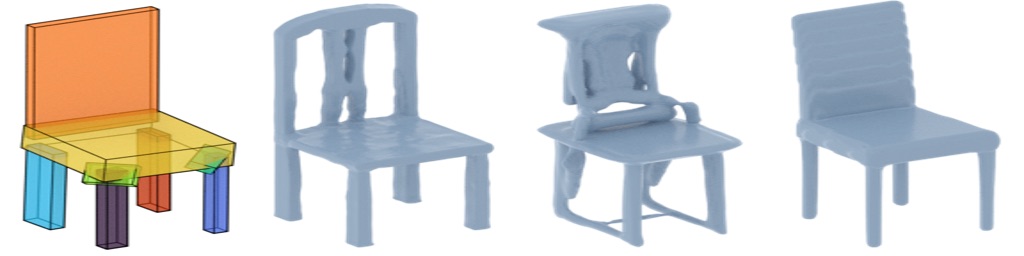}} \\
\multicolumn{4}{c|}{\includegraphics[width=.25\textwidth]{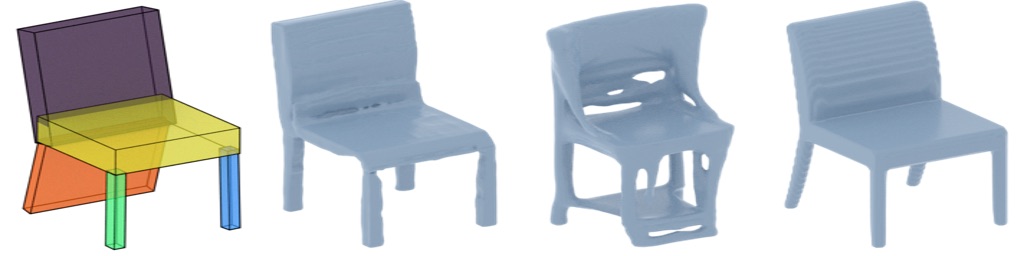}} &
\multicolumn{4}{c|}{\includegraphics[width=.25\textwidth]{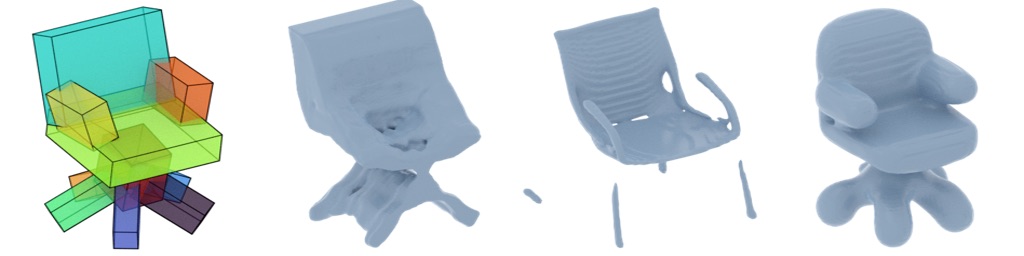}} &
\multicolumn{4}{c|}{\includegraphics[width=.25\textwidth]{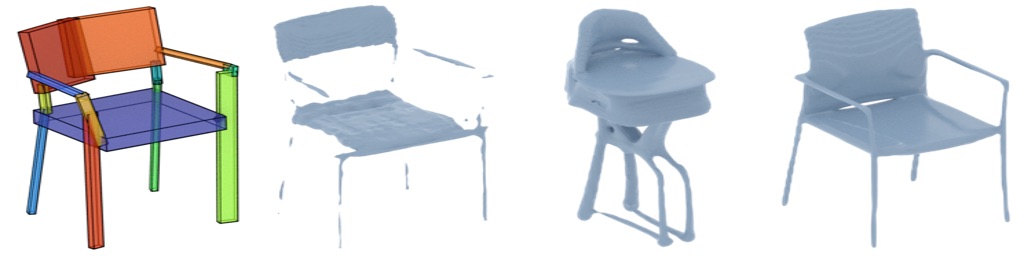}} &
\multicolumn{4}{c}{\includegraphics[width=.25\textwidth]{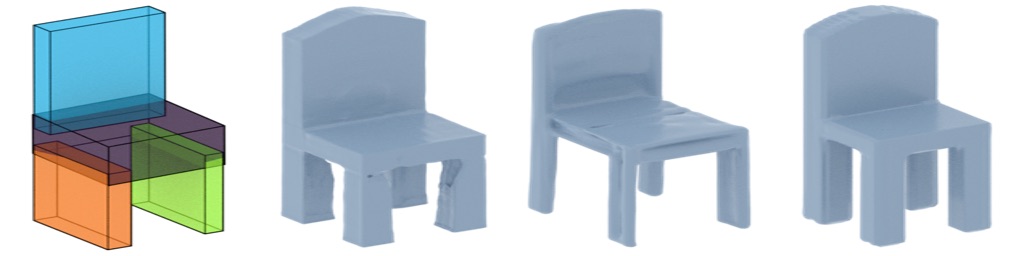}} \\
\multicolumn{4}{c|}{\includegraphics[width=.25\textwidth]{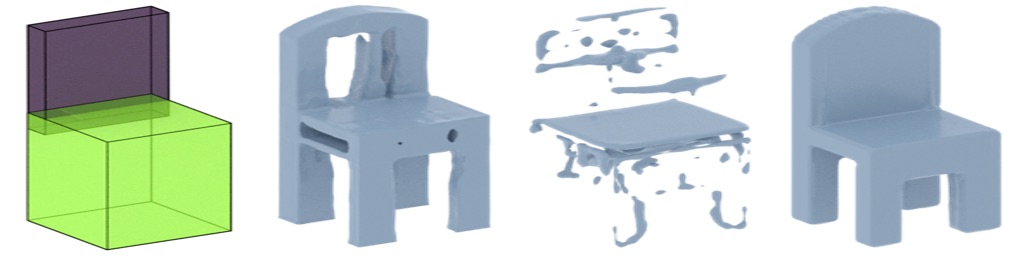}} &
\multicolumn{4}{c|}{\includegraphics[width=.25\textwidth]{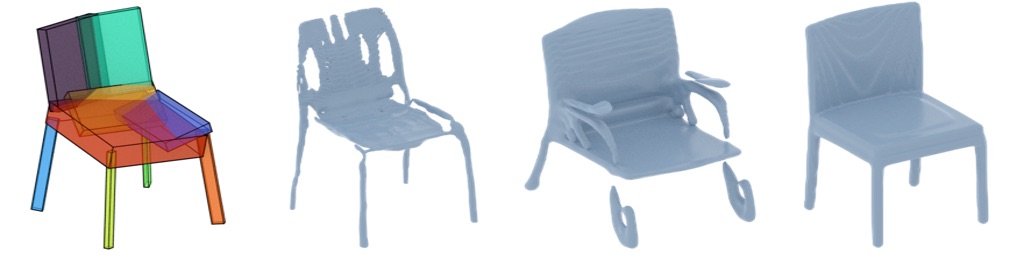}} &
\multicolumn{4}{c|}{\includegraphics[width=.25\textwidth]{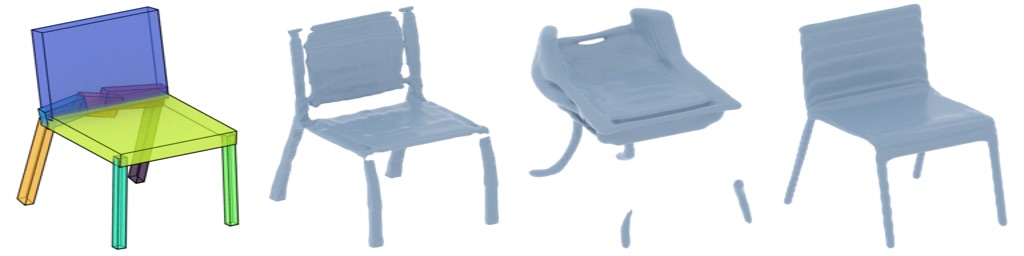}} &
\multicolumn{4}{c}{\includegraphics[width=.25\textwidth]{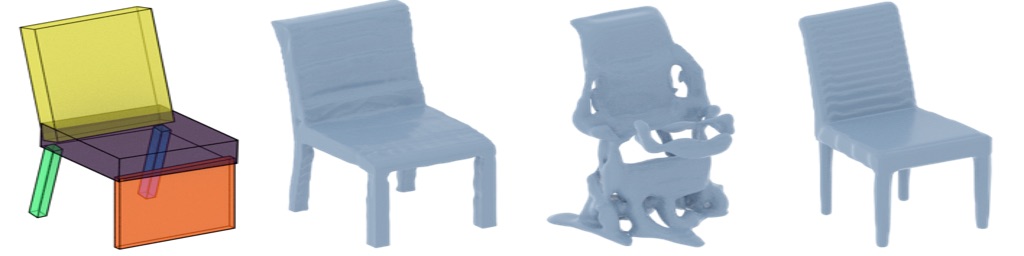}} \\
\multicolumn{4}{c|}{\includegraphics[width=.25\textwidth]{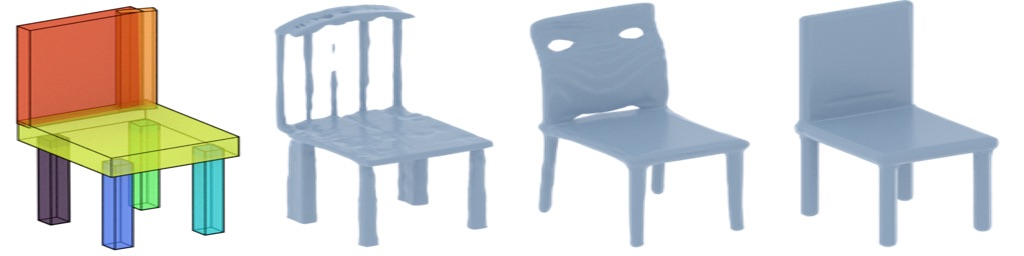}} &
\multicolumn{4}{c|}{\includegraphics[width=.25\textwidth]{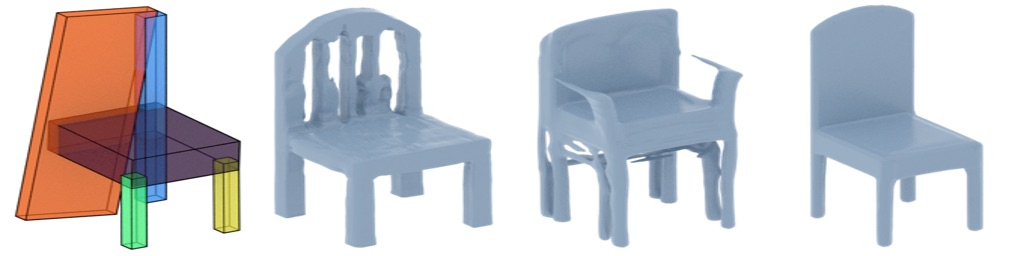}} &
\multicolumn{4}{c|}{\includegraphics[width=.25\textwidth]{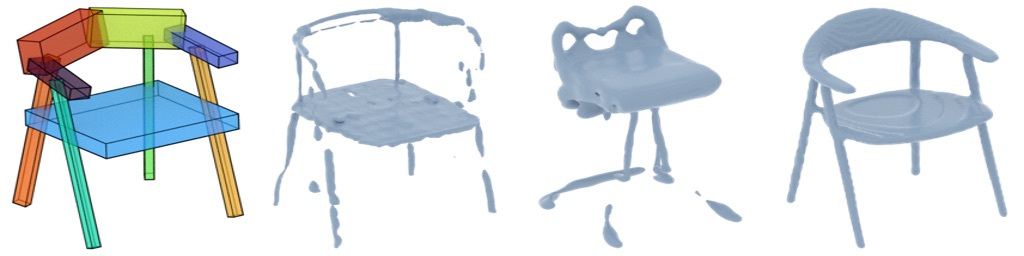}} &
\multicolumn{4}{c}{\includegraphics[width=.25\textwidth]{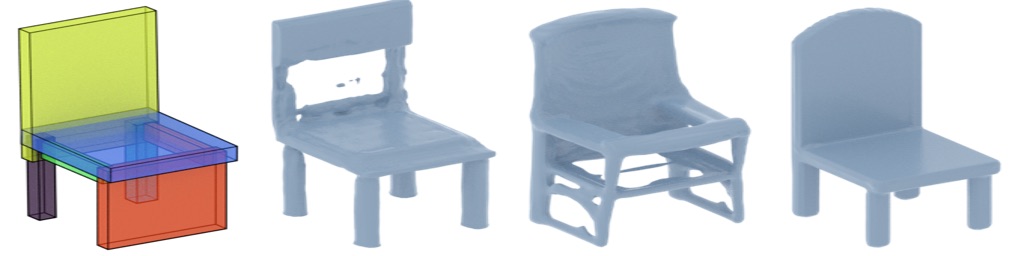}} \\
\multicolumn{4}{c|}{\includegraphics[width=.25\textwidth]{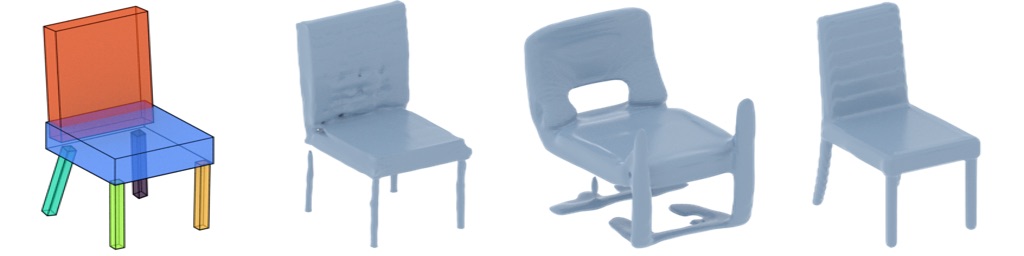}} &
\multicolumn{4}{c|}{\includegraphics[width=.25\textwidth]{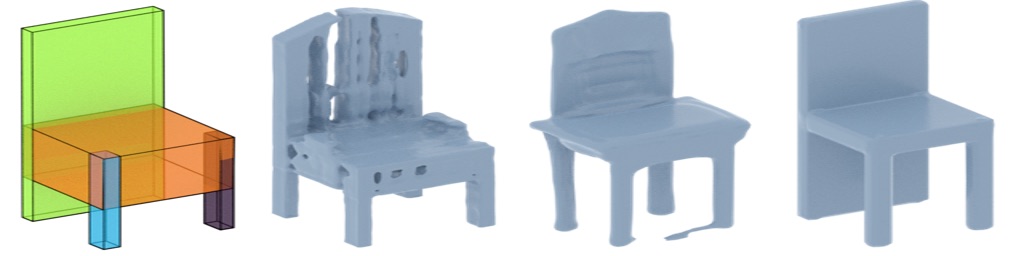}} &
\multicolumn{4}{c|}{\includegraphics[width=.25\textwidth]{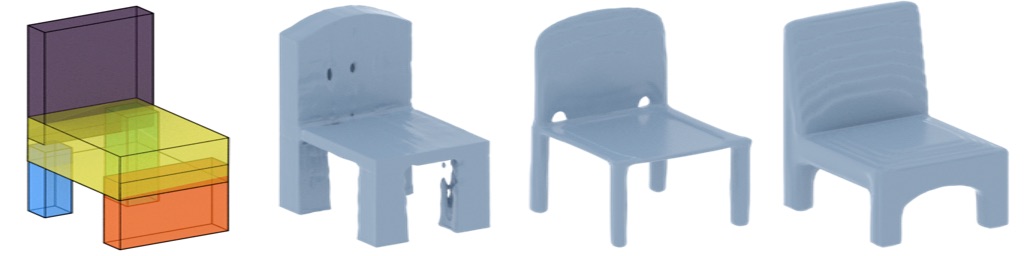}} &
\multicolumn{4}{c}{\includegraphics[width=.25\textwidth]{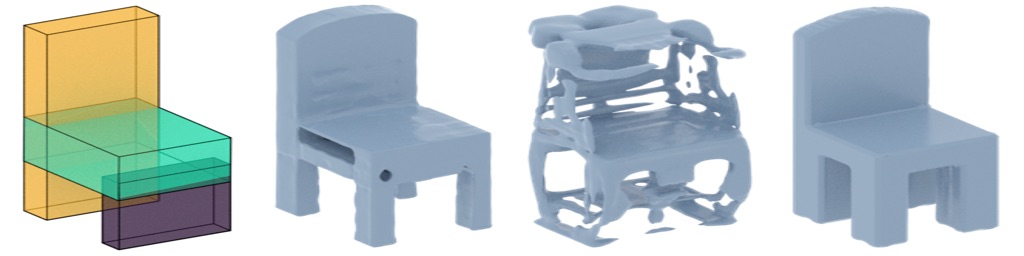}} \\
\multicolumn{4}{c|}{\includegraphics[width=.25\textwidth]{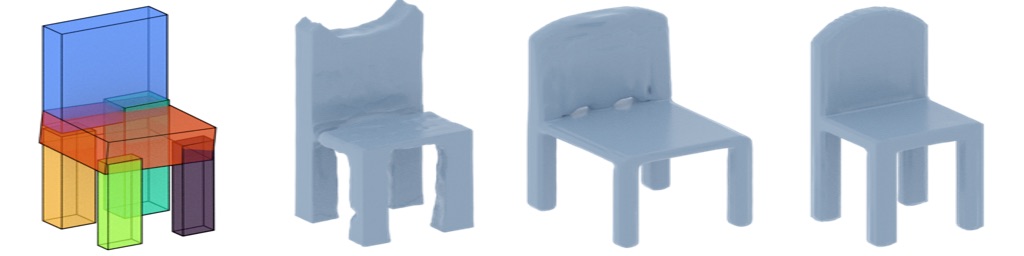}} &
\multicolumn{4}{c|}{\includegraphics[width=.25\textwidth]{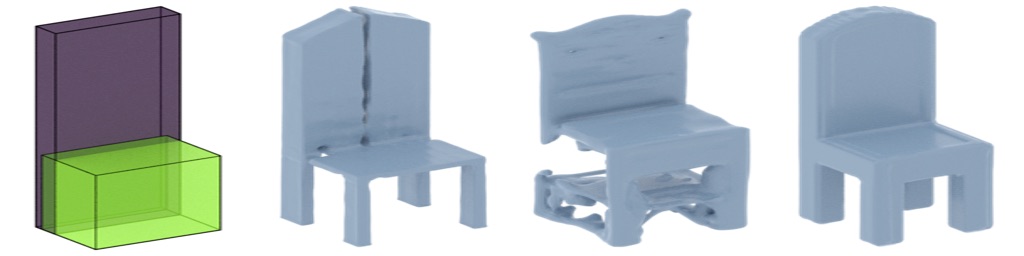}} &
\multicolumn{4}{c|}{\includegraphics[width=.25\textwidth]{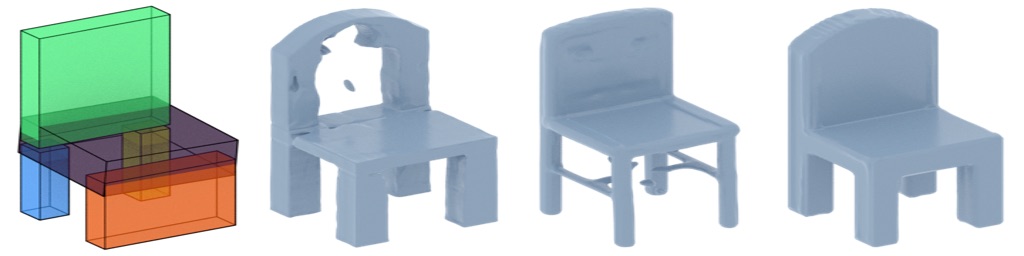}} &
\multicolumn{4}{c}{\includegraphics[width=.25\textwidth]{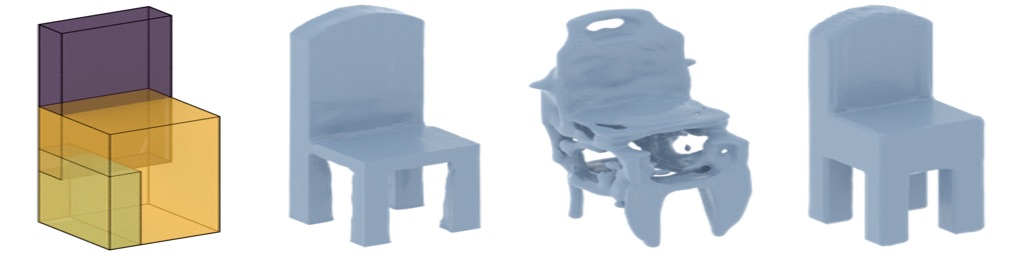}} \\
\multicolumn{4}{c|}{\includegraphics[width=.25\textwidth]{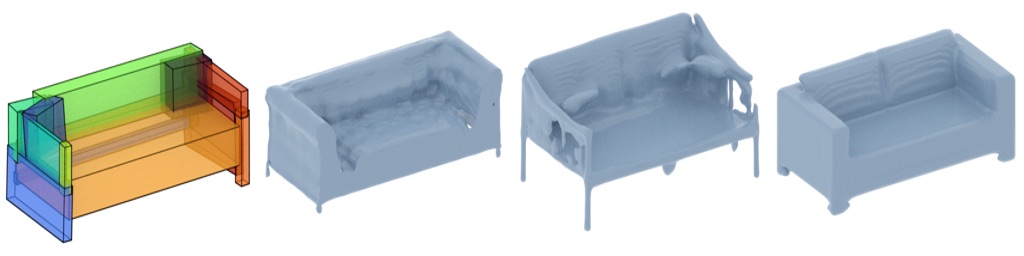}} &
\multicolumn{4}{c|}{\includegraphics[width=.25\textwidth]{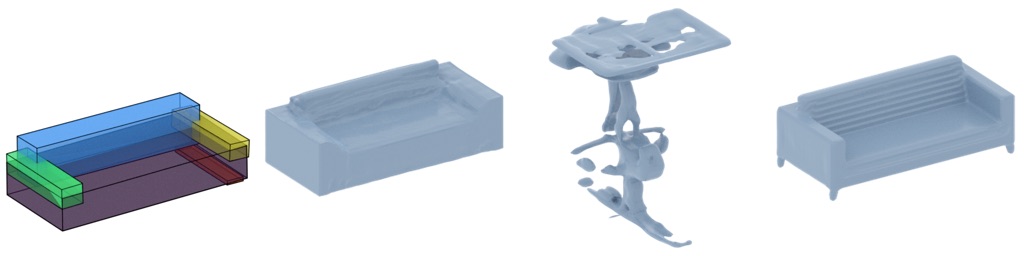}} &
\multicolumn{4}{c|}{\includegraphics[width=.25\textwidth]{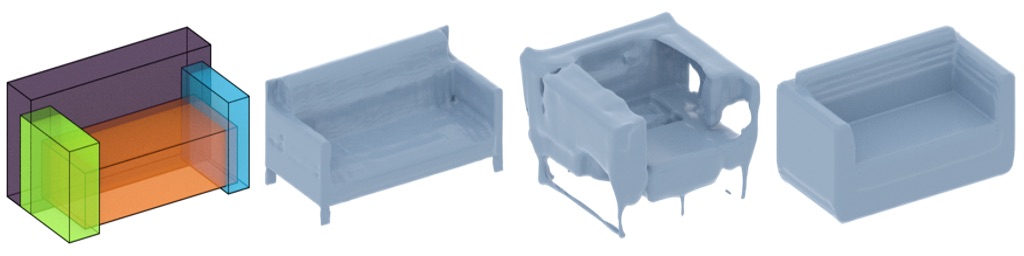}} &
\multicolumn{4}{c}{\includegraphics[width=.25\textwidth]{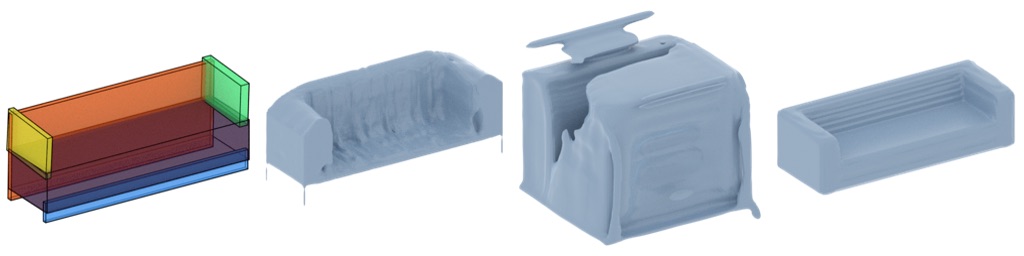}} \\

\end{tabularx}

\caption{\textbf{Gallery of our generated bounding boxes and their final decoded 3D shapes by box-conditioned shape generation network.} Each pair of columns shows the input condition bounding box (left) and its corresponding decoded 3D shape (right).}
\end{figure*}

\begin{figure*}[p!]
\ContinuedFloat
\centering
\scriptsize
\setlength{\tabcolsep}{0em}
\begin{tabularx}{\linewidth}{YYYY | YYYY | YYYY | YYYY}
\rotatebox{0}{\makecell{Input\\Boxes}} & \rotatebox{0}{\makecell{Spice-E\\\cite{Sella:2023SpicE}}} & \rotatebox{0}{\makecell{Gated\\3DS2V~\cite{Zhang:2023Shape2Vec}}} & \rotatebox{0}{Ours} & \rotatebox{0}{\makecell{Input\\Boxes}} & \rotatebox{0}{\makecell{Spice-E\\\cite{Sella:2023SpicE}}} & \rotatebox{0}{\makecell{Gated\\3DS2V~\cite{Zhang:2023Shape2Vec}}} & \rotatebox{0}{Ours} & \rotatebox{0}{\makecell{Input\\Boxes}} & \rotatebox{0}{\makecell{Spice-E\\\cite{Sella:2023SpicE}}} & \rotatebox{0}{\makecell{Gated\\3DS2V~\cite{Zhang:2023Shape2Vec}}} & \rotatebox{0}{Ours} & \rotatebox{0}{\makecell{Input\\Boxes}} & \rotatebox{0}{\makecell{Spice-E\\\cite{Sella:2023SpicE}}} & \rotatebox{0}{\makecell{Gated\\3DS2V~\cite{Zhang:2023Shape2Vec}}} & \rotatebox{0}{Ours}  \\ 

\midrule

\multicolumn{4}{c|}{\includegraphics[width=.25\textwidth]{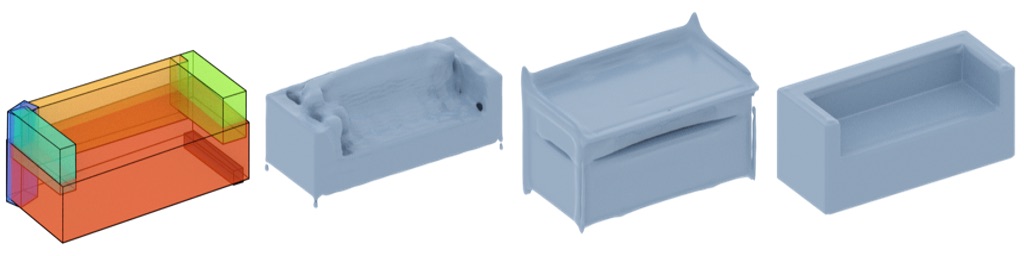}} &
\multicolumn{4}{c|}{\includegraphics[width=.25\textwidth]{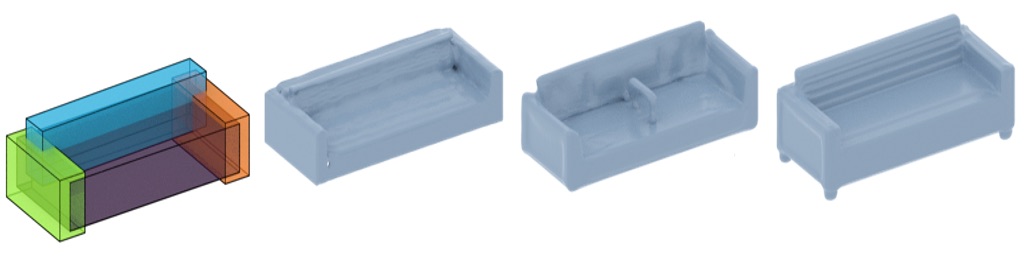}} &
\multicolumn{4}{c|}{\includegraphics[width=.25\textwidth]{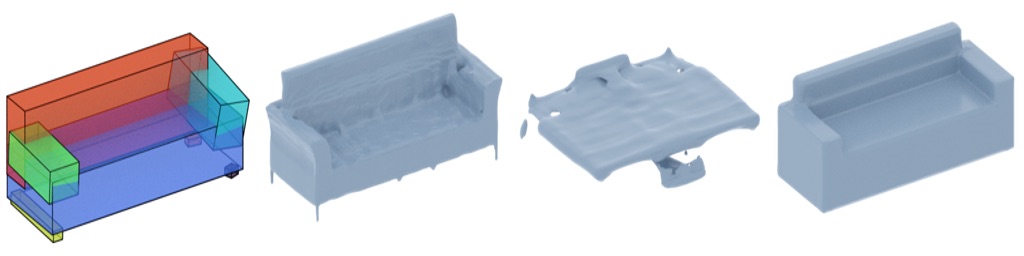}} &
\multicolumn{4}{c}{\includegraphics[width=.25\textwidth]{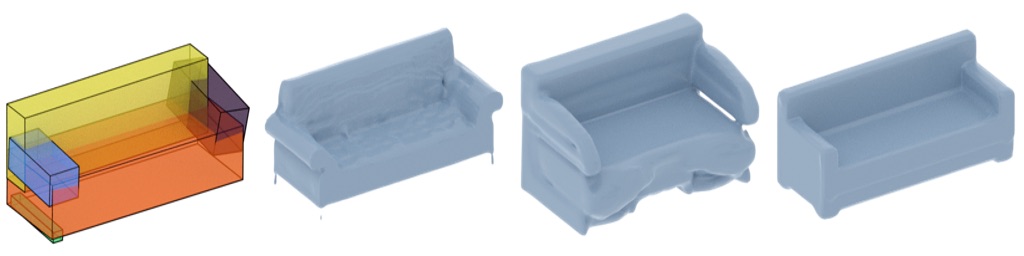}} \\
\multicolumn{4}{c|}{\includegraphics[width=.25\textwidth]{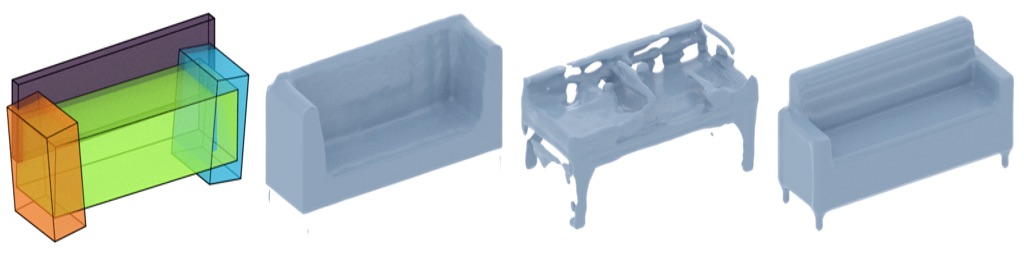}} &
\multicolumn{4}{c|}{\includegraphics[width=.25\textwidth]{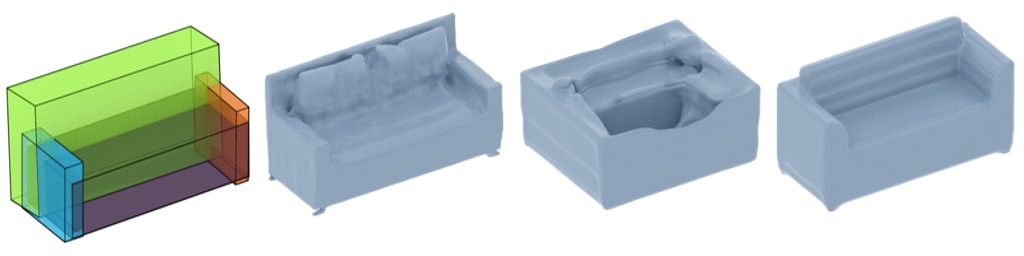}} &
\multicolumn{4}{c|}{\includegraphics[width=.25\textwidth]{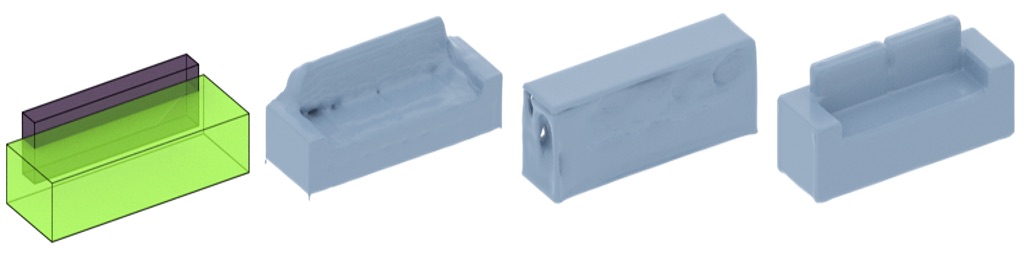}} &
\multicolumn{4}{c}{\includegraphics[width=.25\textwidth]{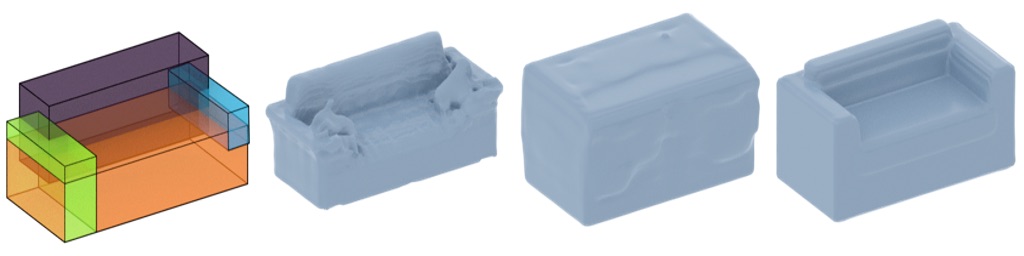}} \\
\multicolumn{4}{c|}{\includegraphics[width=.25\textwidth]{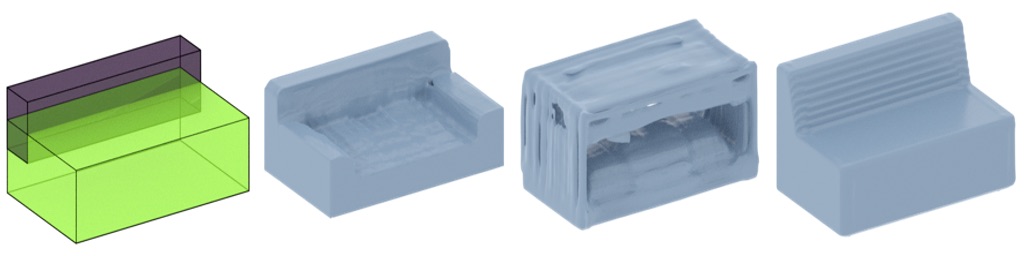}} &
\multicolumn{4}{c|}{\includegraphics[width=.25\textwidth]{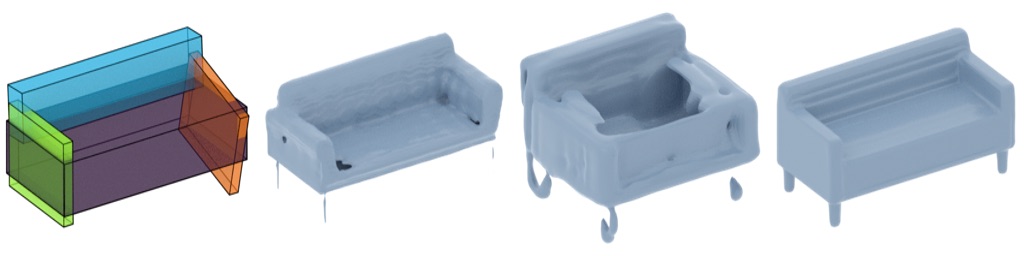}} &
\multicolumn{4}{c|}{\includegraphics[width=.25\textwidth]{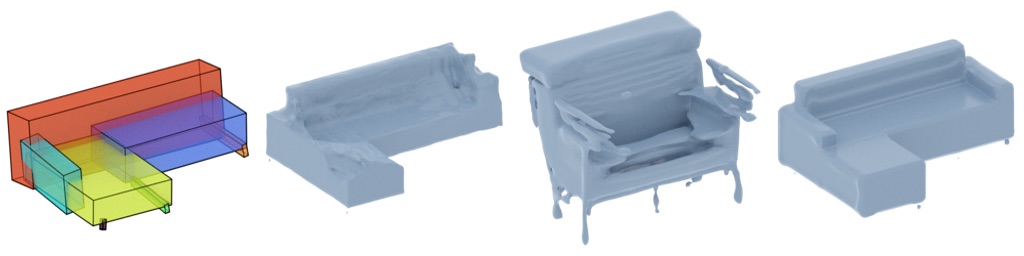}} &
\multicolumn{4}{c}{\includegraphics[width=.25\textwidth]{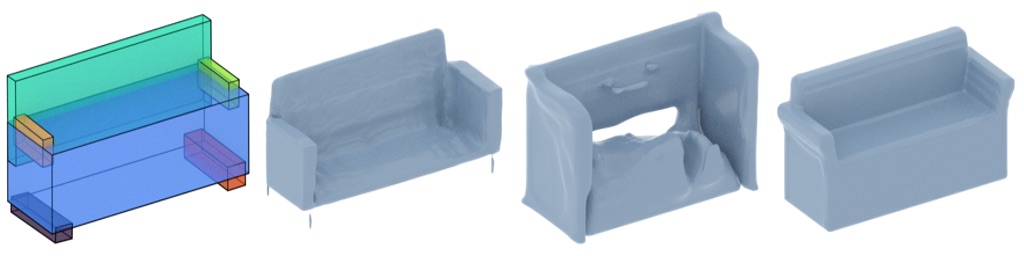}} \\
\multicolumn{4}{c|}{\includegraphics[width=.25\textwidth]{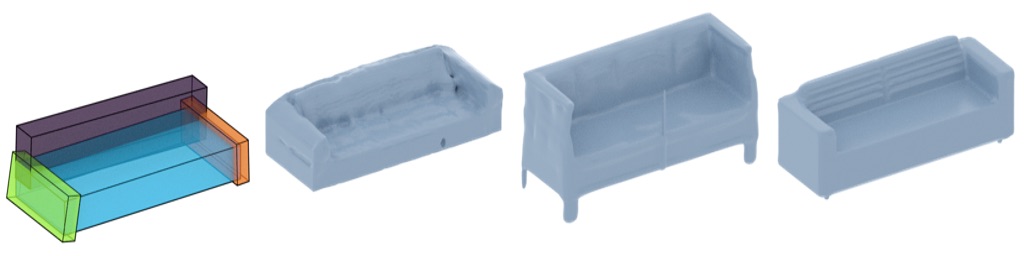}} &
\multicolumn{4}{c|}{\includegraphics[width=.25\textwidth]{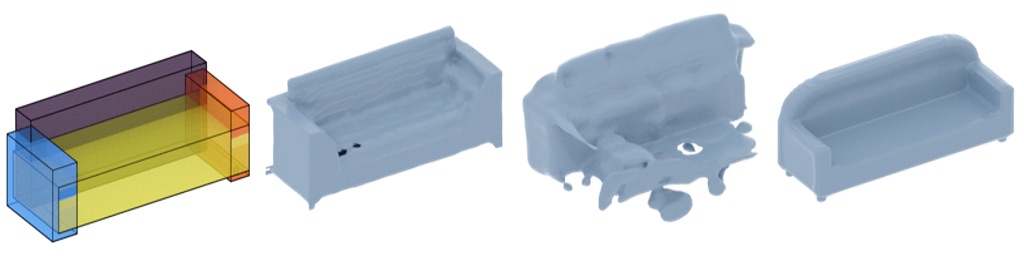}} &
\multicolumn{4}{c|}{\includegraphics[width=.25\textwidth]{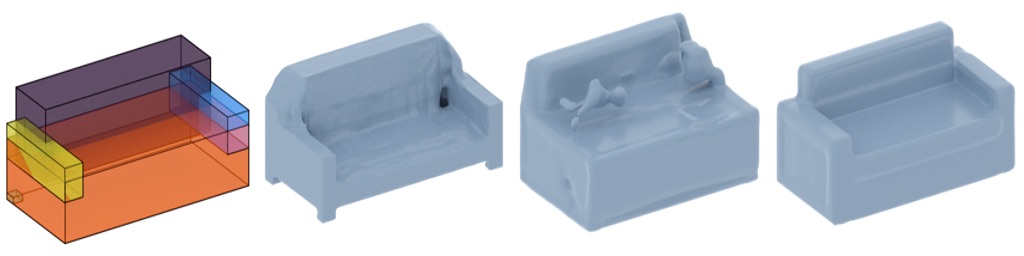}} &
\multicolumn{4}{c}{\includegraphics[width=.25\textwidth]{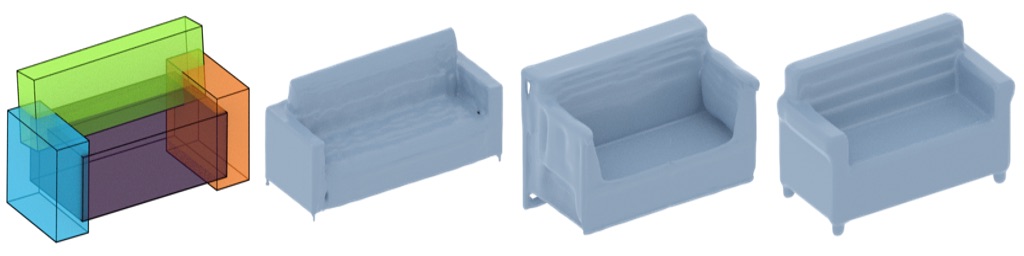}} \\
\multicolumn{4}{c|}{\includegraphics[width=.25\textwidth]{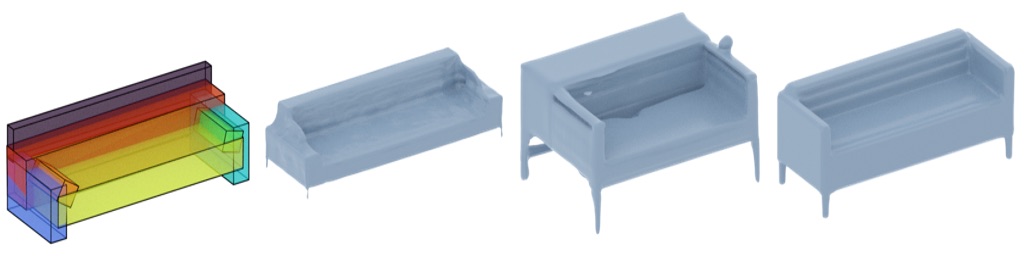}} &
\multicolumn{4}{c|}{\includegraphics[width=.25\textwidth]{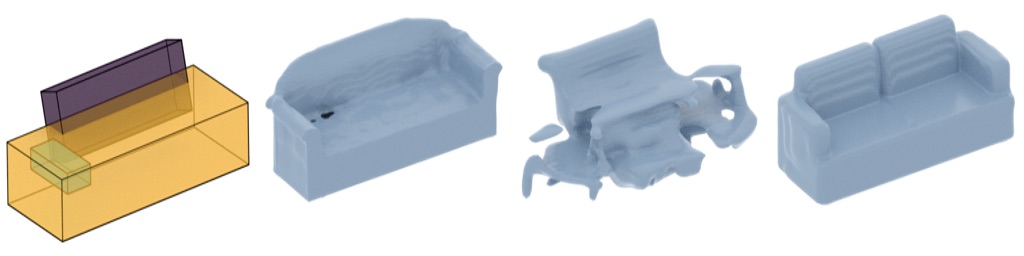}} &
\multicolumn{4}{c|}{\includegraphics[width=.25\textwidth]{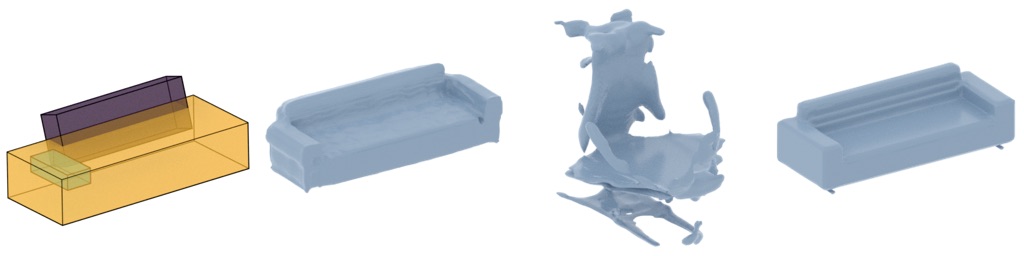}} &
\multicolumn{4}{c}{\includegraphics[width=.25\textwidth]{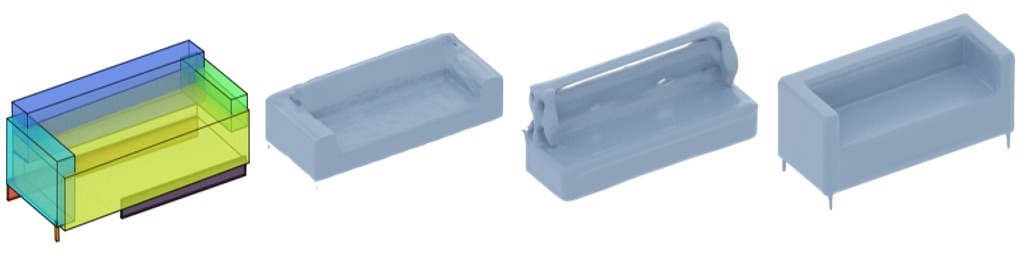}} \\
\multicolumn{4}{c|}{\includegraphics[width=.25\textwidth]{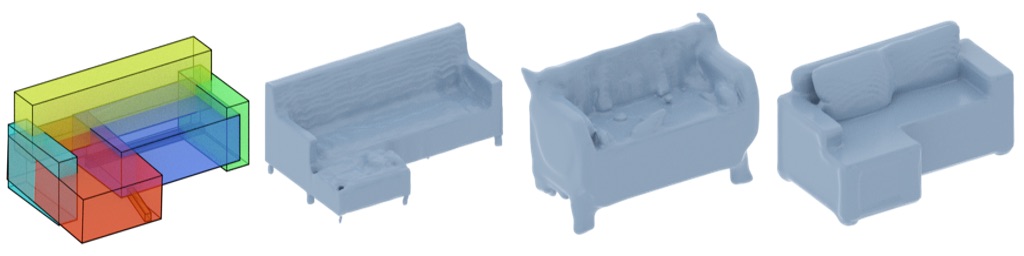}} &
\multicolumn{4}{c|}{\includegraphics[width=.25\textwidth]{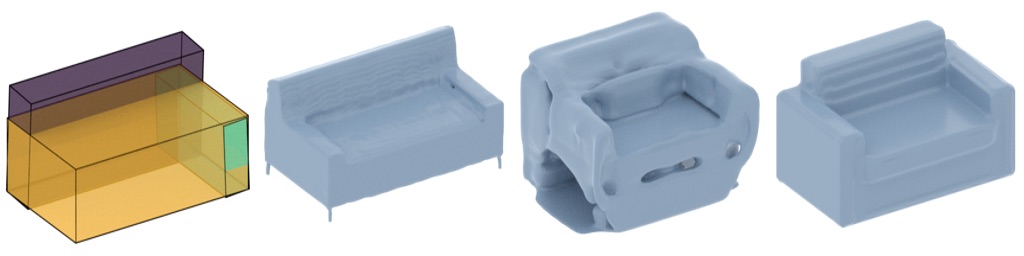}} &
\multicolumn{4}{c|}{\includegraphics[width=.25\textwidth]{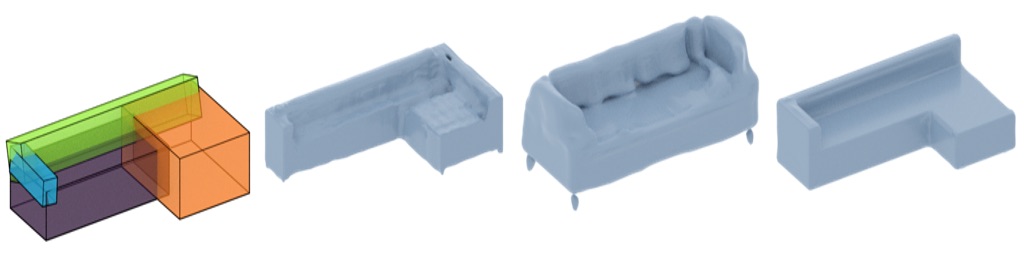}} &
\multicolumn{4}{c}{\includegraphics[width=.25\textwidth]{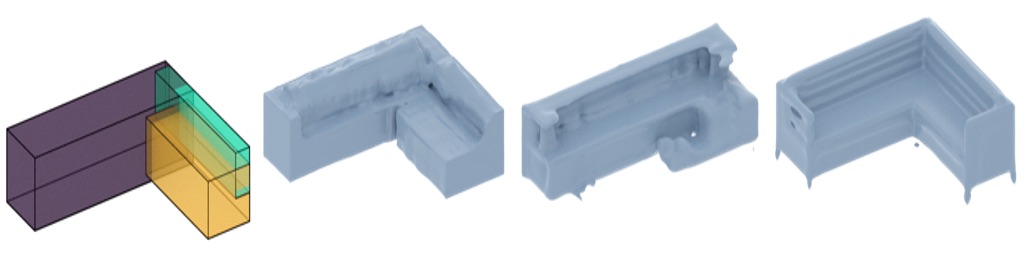}} \\
\multicolumn{4}{c|}{\includegraphics[width=.25\textwidth]{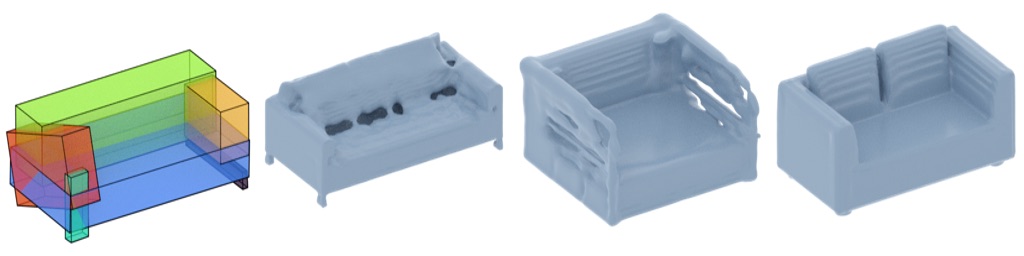}} &
\multicolumn{4}{c|}{\includegraphics[width=.25\textwidth]{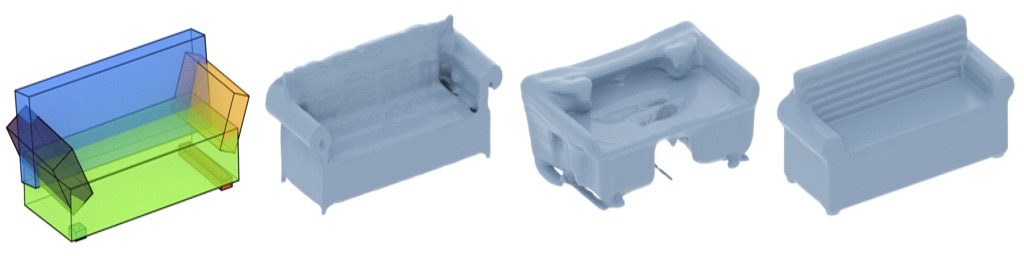}} &
\multicolumn{4}{c|}{\includegraphics[width=.25\textwidth]{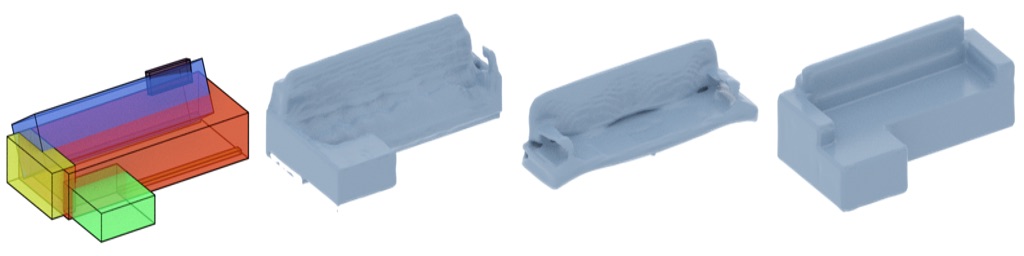}} &
\multicolumn{4}{c}{\includegraphics[width=.25\textwidth]{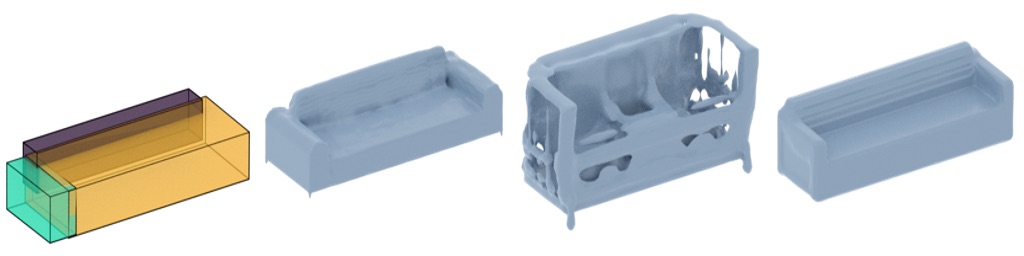}} \\
\multicolumn{4}{c|}{\includegraphics[width=.25\textwidth]{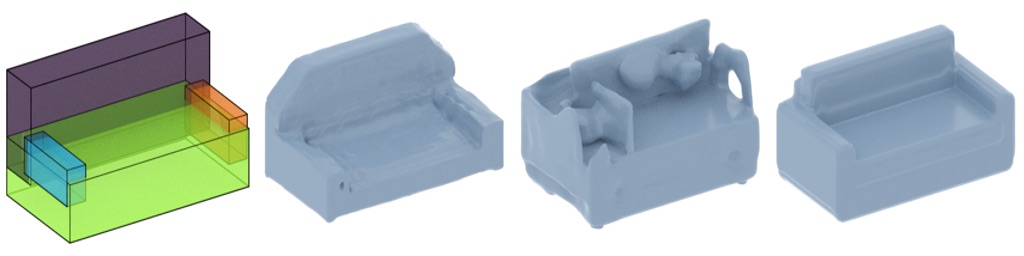}} &
\multicolumn{4}{c|}{\includegraphics[width=.25\textwidth]{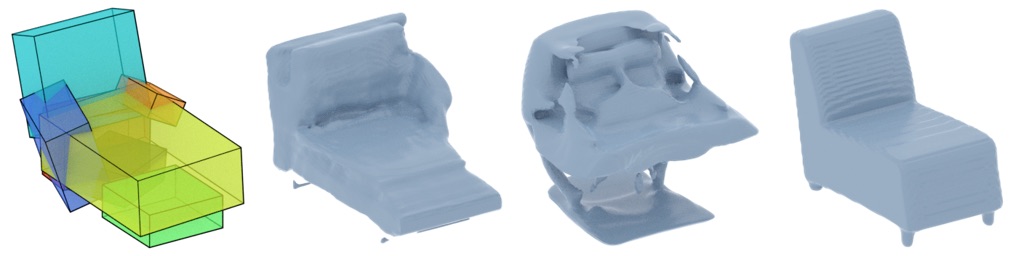}} &
\multicolumn{4}{c|}{\includegraphics[width=.25\textwidth]{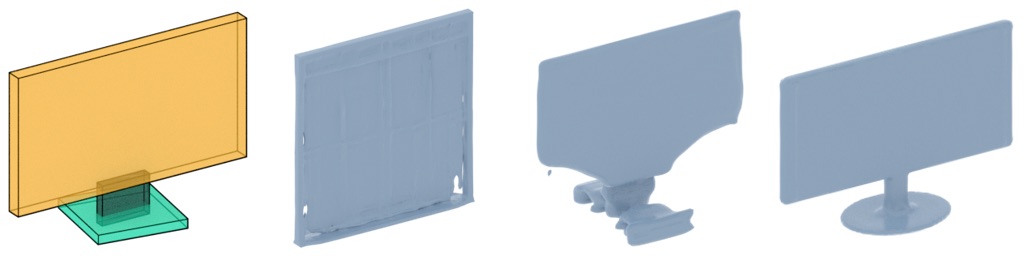}} &
\multicolumn{4}{c}{\includegraphics[width=.25\textwidth]{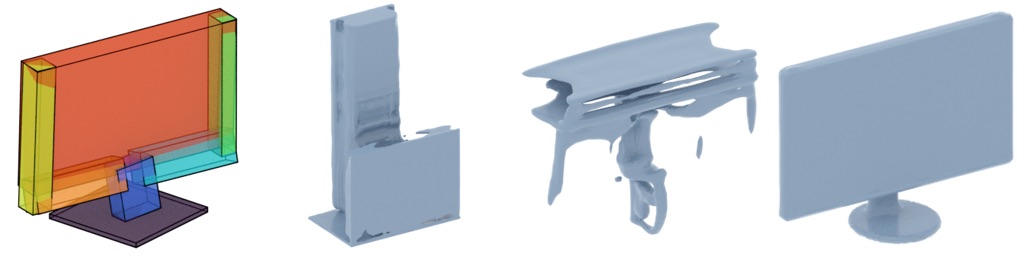}} \\
\multicolumn{4}{c|}{\includegraphics[width=.25\textwidth]{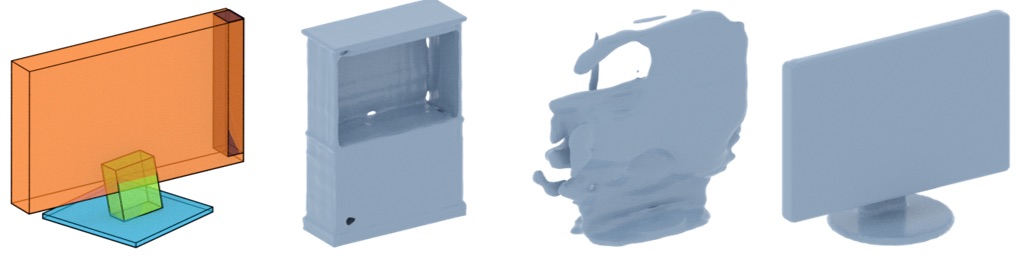}} &
\multicolumn{4}{c|}{\includegraphics[width=.25\textwidth]{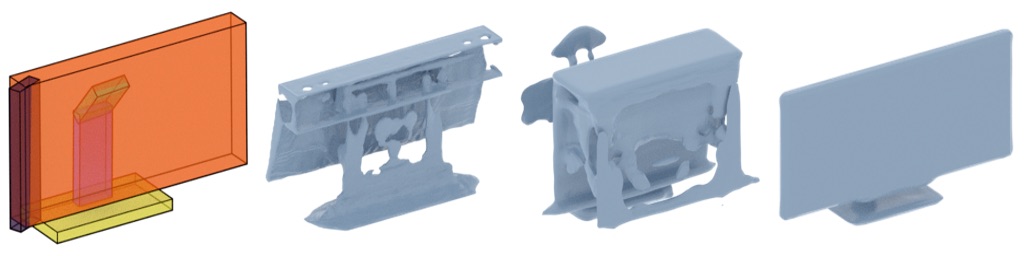}} &
\multicolumn{4}{c|}{\includegraphics[width=.25\textwidth]{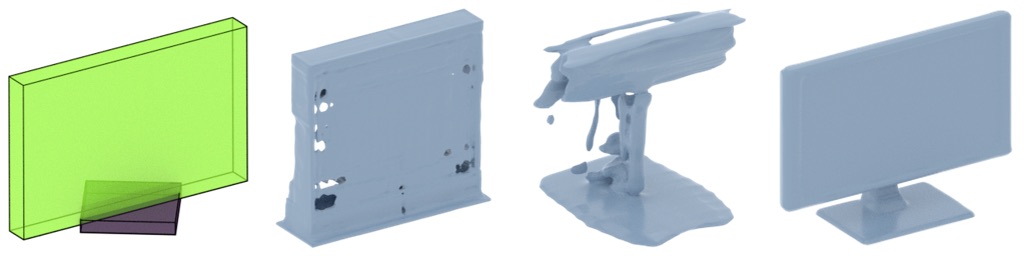}} &
\multicolumn{4}{c}{\includegraphics[width=.25\textwidth]{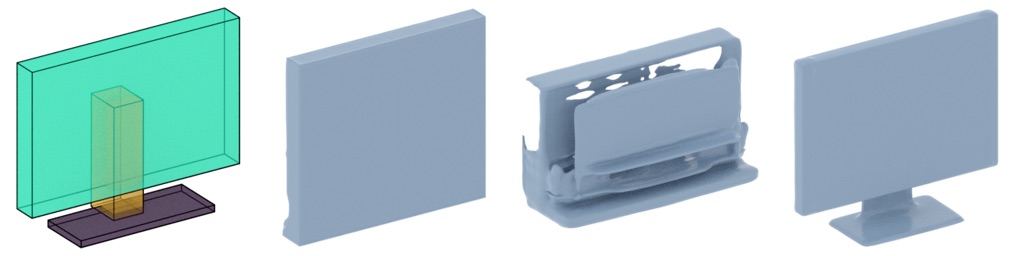}} \\
\multicolumn{4}{c|}{\includegraphics[width=.25\textwidth]{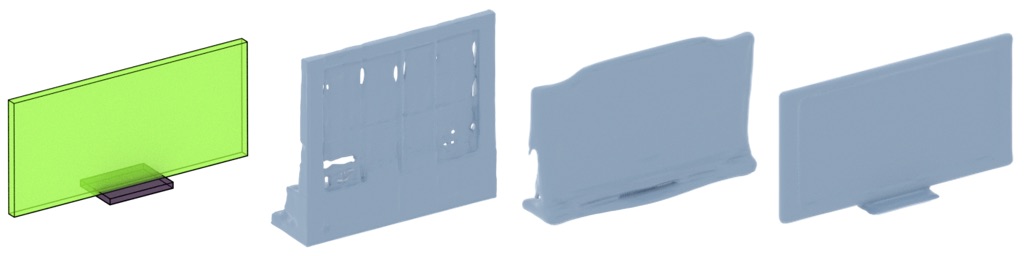}} &
\multicolumn{4}{c|}{\includegraphics[width=.25\textwidth]{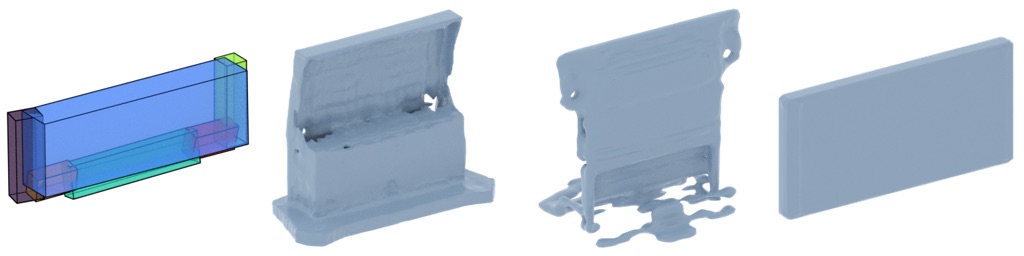}} &
\multicolumn{4}{c|}{\includegraphics[width=.25\textwidth]{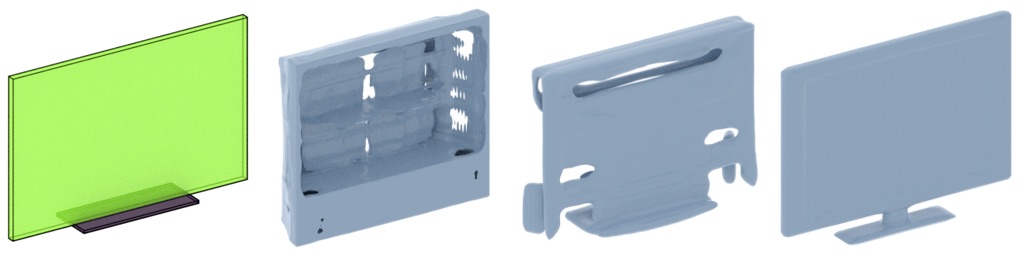}} &
\multicolumn{4}{c}{\includegraphics[width=.25\textwidth]{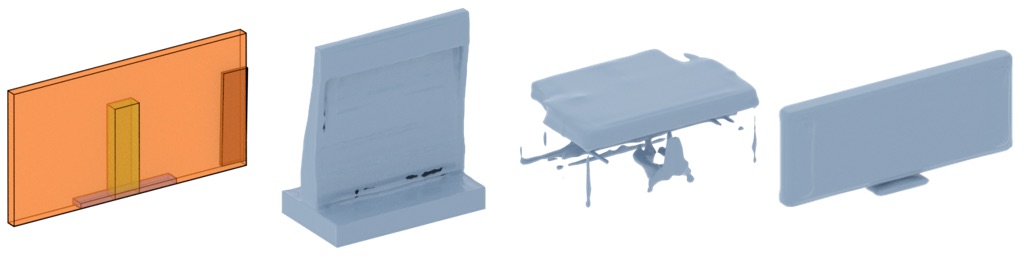}} \\
\multicolumn{4}{c|}{\includegraphics[width=.25\textwidth]{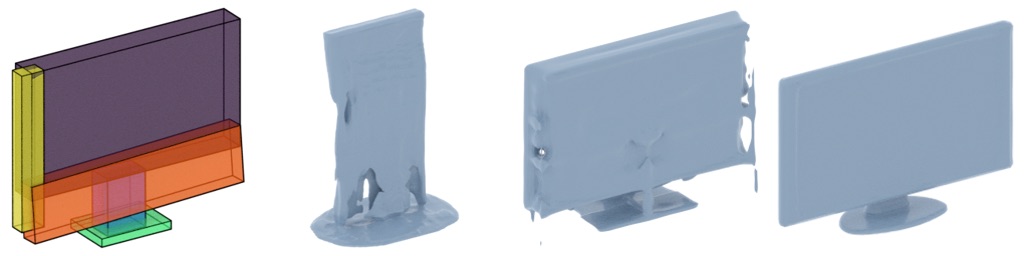}} &
\multicolumn{4}{c|}{\includegraphics[width=.25\textwidth]{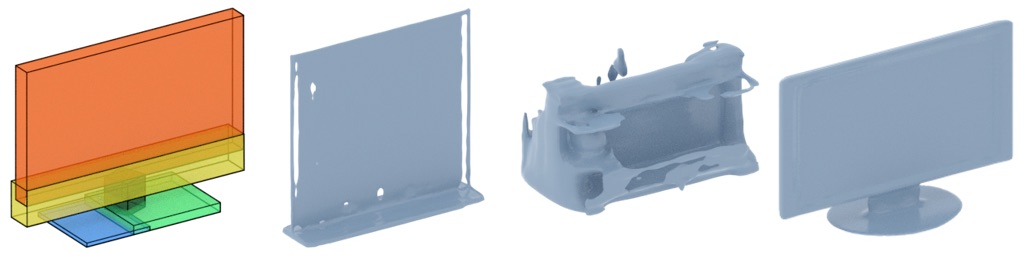}} &
\multicolumn{4}{c|}{\includegraphics[width=.25\textwidth]{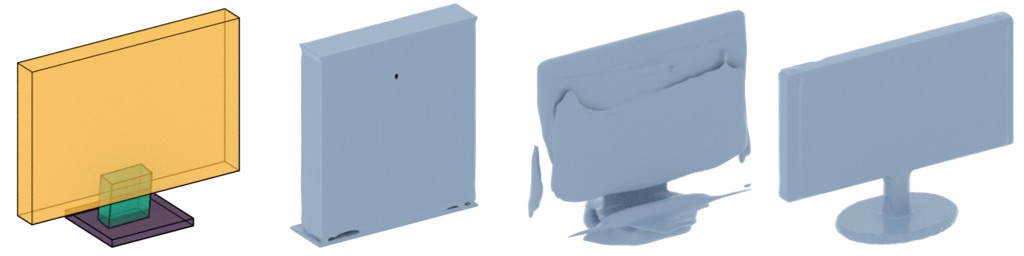}} &
\multicolumn{4}{c}{\includegraphics[width=.25\textwidth]{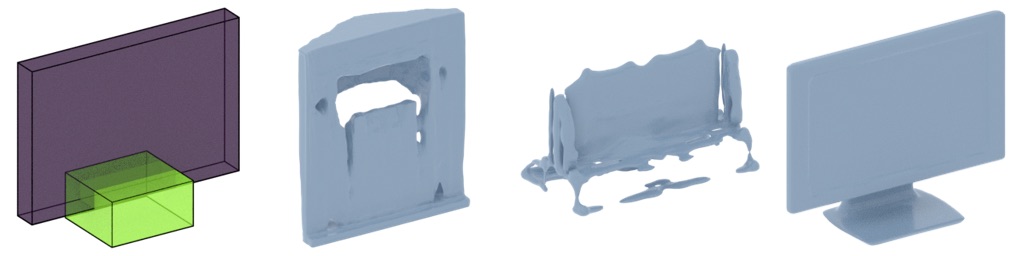}} \\
\multicolumn{4}{c|}{\includegraphics[width=.25\textwidth]{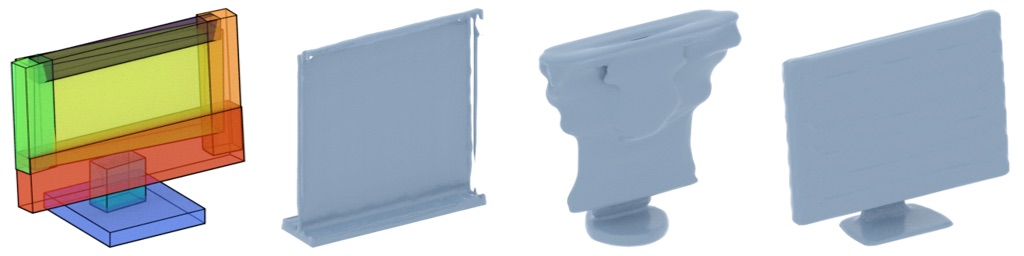}} &
\multicolumn{4}{c|}{\includegraphics[width=.25\textwidth]{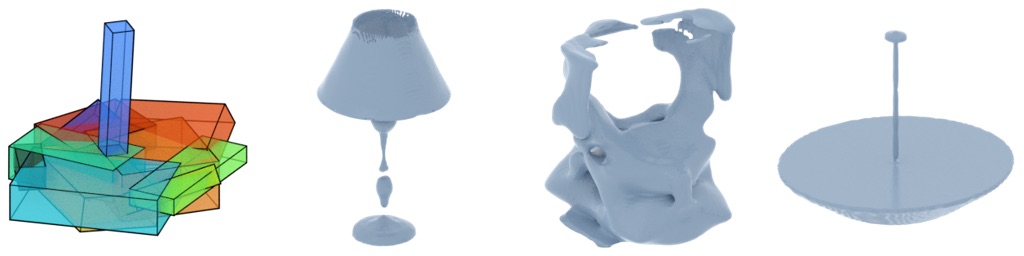}} &
\multicolumn{4}{c|}{\includegraphics[width=.25\textwidth]{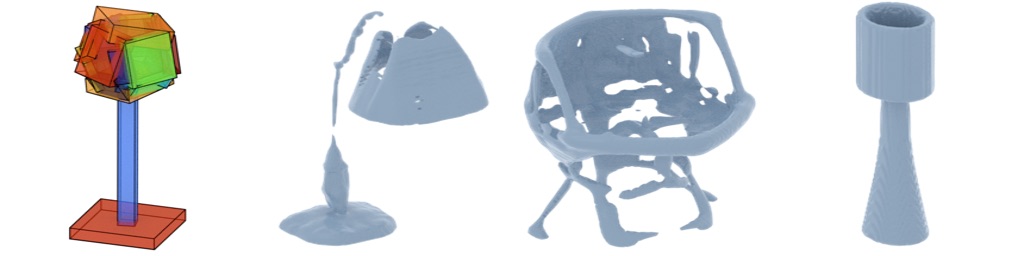}} &
\multicolumn{4}{c}{\includegraphics[width=.25\textwidth]{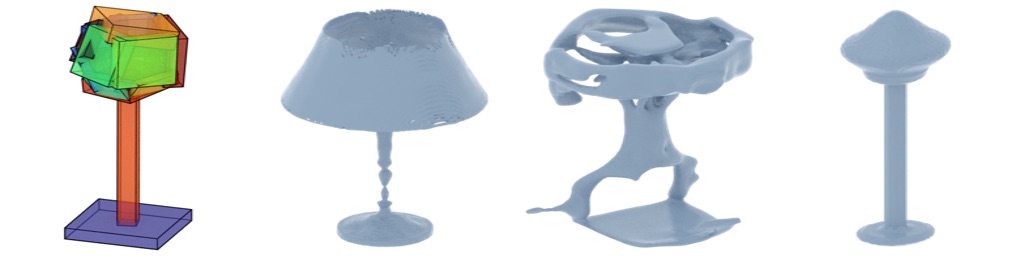}} \\
\multicolumn{4}{c|}{\includegraphics[width=.25\textwidth]{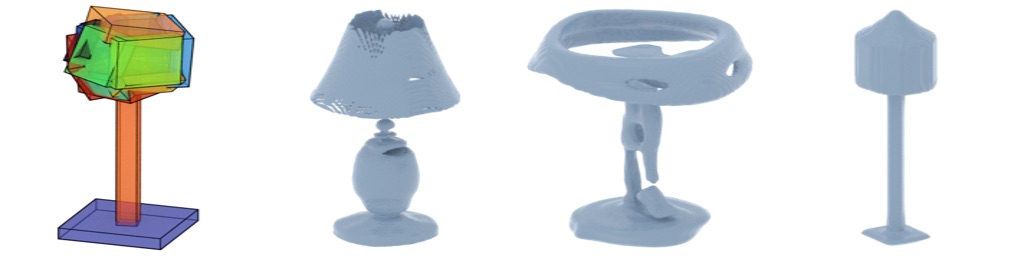}} &
\multicolumn{4}{c|}{\includegraphics[width=.25\textwidth]{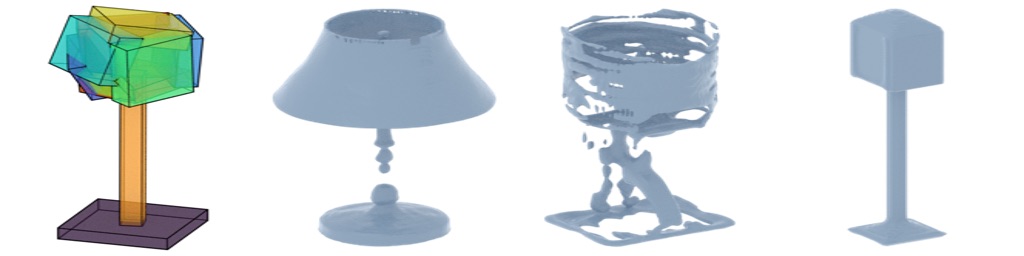}} &
\multicolumn{4}{c|}{\includegraphics[width=.25\textwidth]{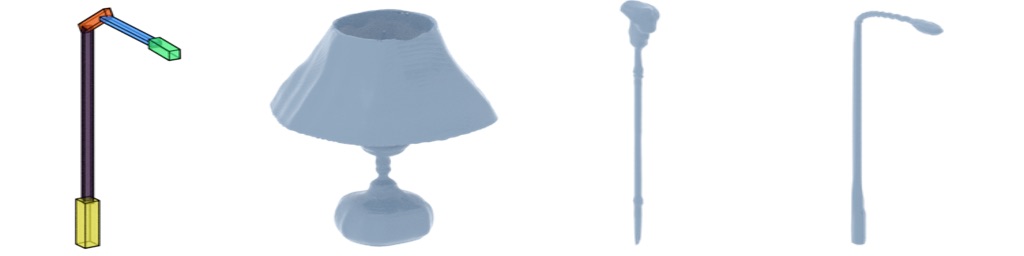}} &
\multicolumn{4}{c}{\includegraphics[width=.25\textwidth]{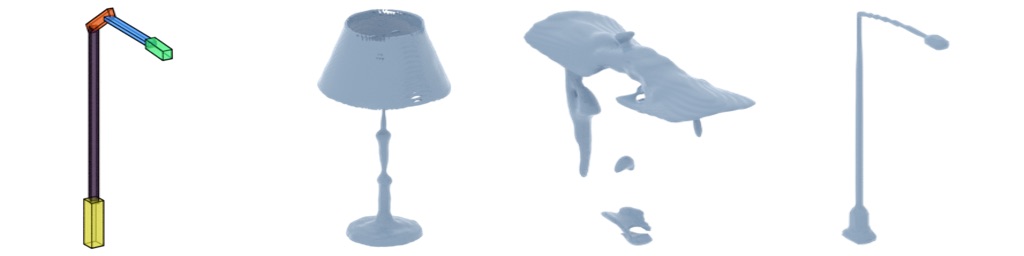}} \\
\multicolumn{4}{c|}{\includegraphics[width=.25\textwidth]{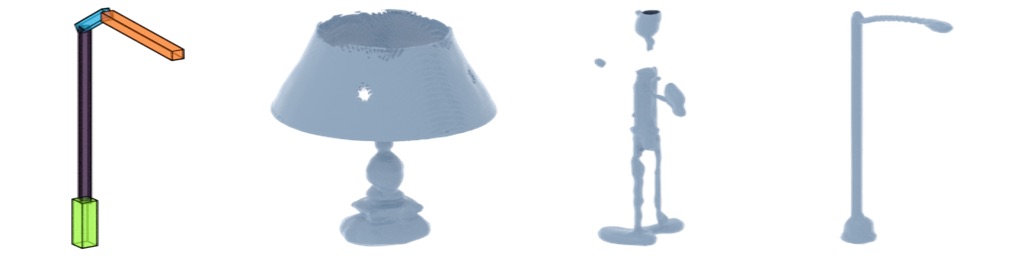}} &
\multicolumn{4}{c|}{\includegraphics[width=.25\textwidth]{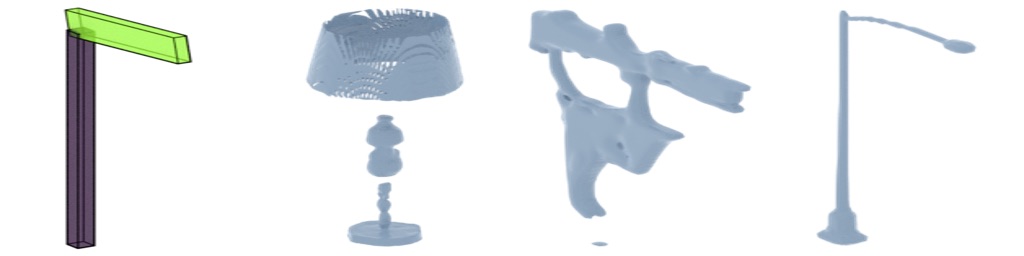}} &
\multicolumn{4}{c|}{\includegraphics[width=.25\textwidth]{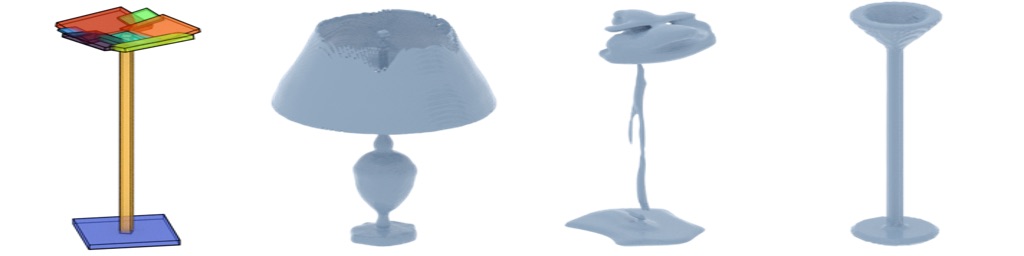}} &
\multicolumn{4}{c}{\includegraphics[width=.25\textwidth]{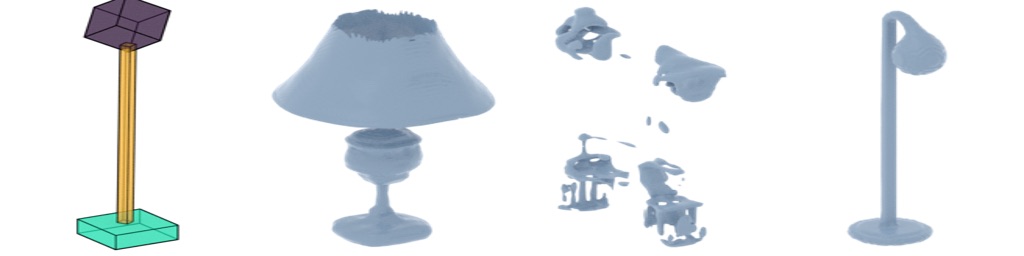}} \\
\multicolumn{4}{c|}{\includegraphics[width=.25\textwidth]{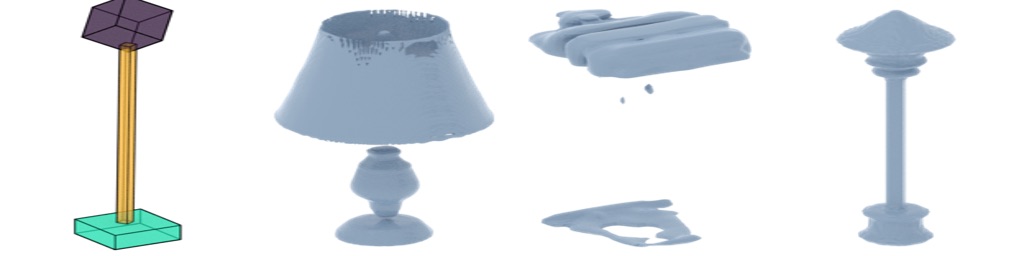}} &
\multicolumn{4}{c|}{\includegraphics[width=.25\textwidth]{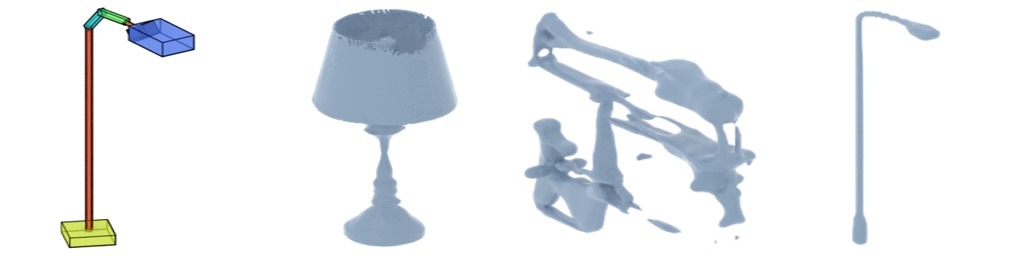}} &
\multicolumn{4}{c|}{\includegraphics[width=.25\textwidth]{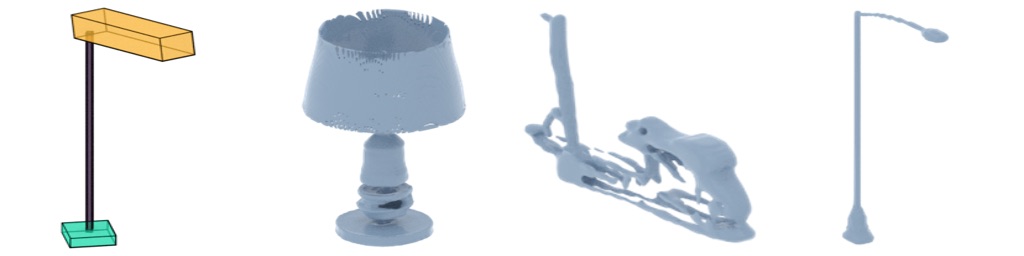}} &
\multicolumn{4}{c}{\includegraphics[width=.25\textwidth]{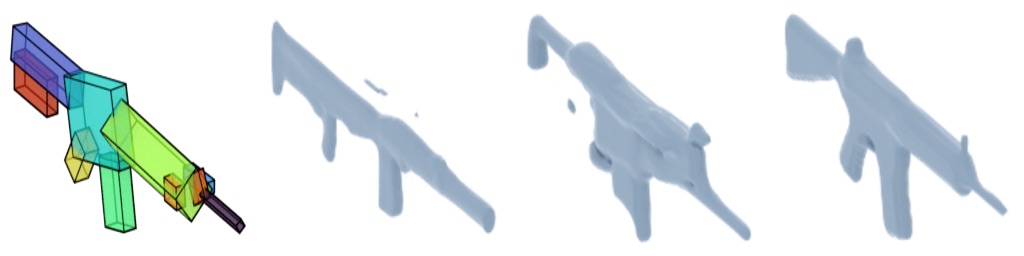}} \\
\multicolumn{4}{c|}{\includegraphics[width=.25\textwidth]{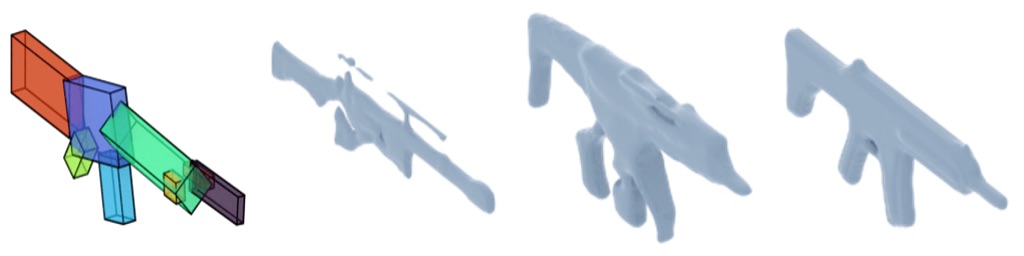}} &
\multicolumn{4}{c|}{\includegraphics[width=.25\textwidth]{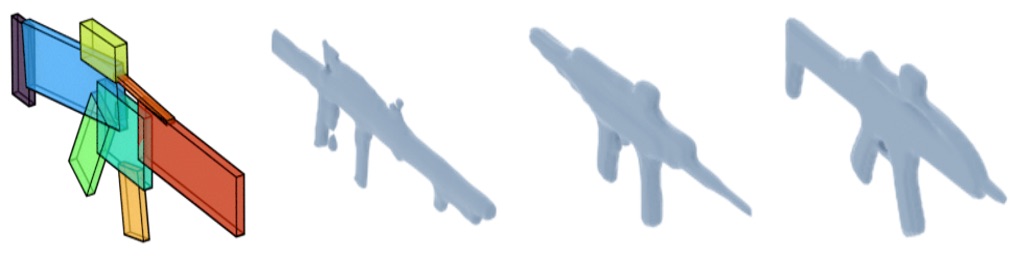}} &
\multicolumn{4}{c|}{\includegraphics[width=.25\textwidth]{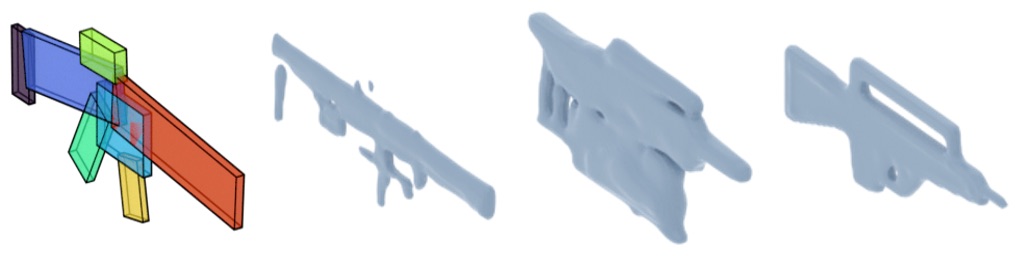}} &
\multicolumn{4}{c}{\includegraphics[width=.25\textwidth]{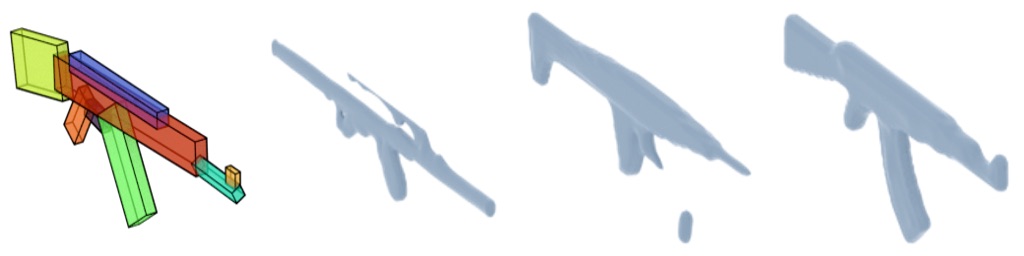}} \\
\multicolumn{4}{c|}{\includegraphics[width=.25\textwidth]{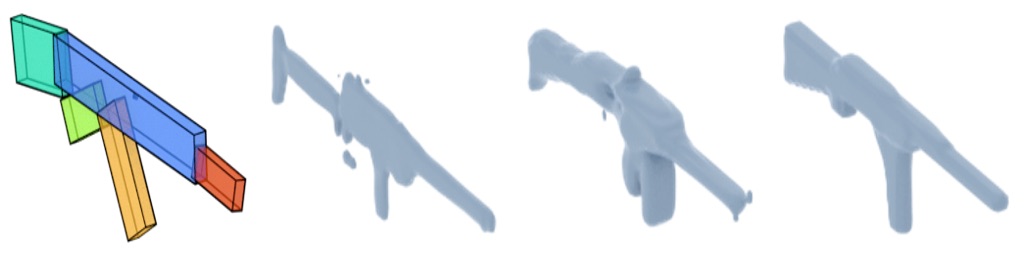}} &
\multicolumn{4}{c|}{\includegraphics[width=.25\textwidth]{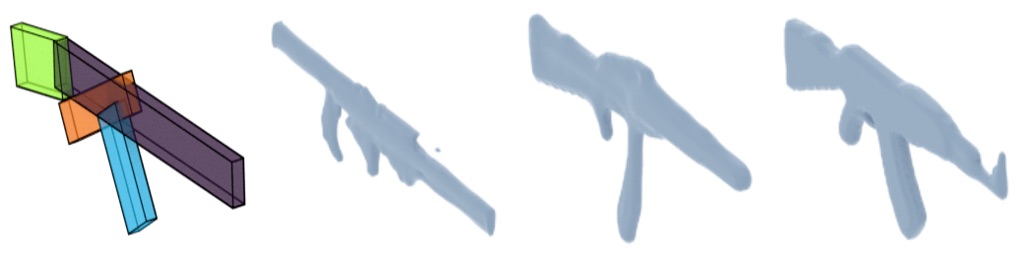}} &
\multicolumn{4}{c|}{\includegraphics[width=.25\textwidth]{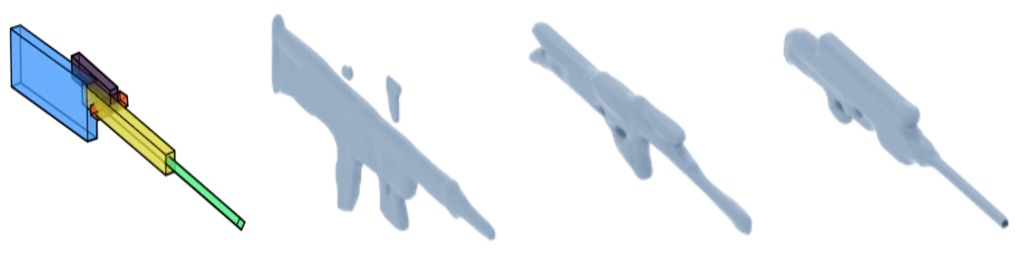}} &
\multicolumn{4}{c}{\includegraphics[width=.25\textwidth]{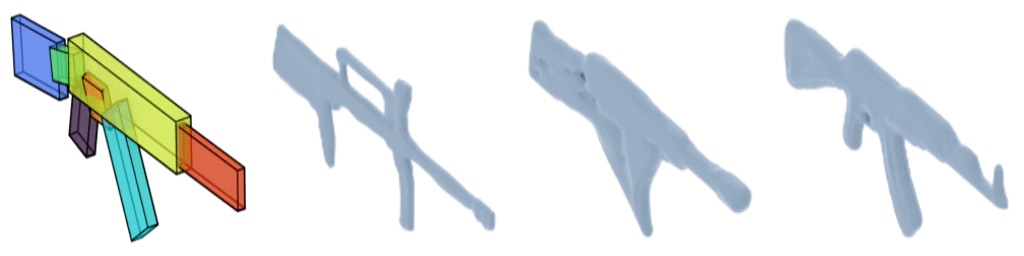}} \\

\end{tabularx}

\caption{\textbf{Gallery of our generated bounding boxes and their final decoded 3D shapes by box-conditioned shape generation network.} Each pair of columns shows the input condition bounding box (left) and its corresponding decoded 3D shape (right).}
\end{figure*}

\begin{figure*}[p!]
\ContinuedFloat
\centering
\scriptsize
\setlength{\tabcolsep}{0em}
\begin{tabularx}{\linewidth}{YYYY | YYYY | YYYY | YYYY}
\rotatebox{0}{\makecell{Input\\Boxes}} & \rotatebox{0}{\makecell{Spice-E\\\cite{Sella:2023SpicE}}} & \rotatebox{0}{\makecell{Gated\\3DS2V~\cite{Zhang:2023Shape2Vec}}} & \rotatebox{0}{Ours} & \rotatebox{0}{\makecell{Input\\Boxes}} & \rotatebox{0}{\makecell{Spice-E\\\cite{Sella:2023SpicE}}} & \rotatebox{0}{\makecell{Gated\\3DS2V~\cite{Zhang:2023Shape2Vec}}} & \rotatebox{0}{Ours} & \rotatebox{0}{\makecell{Input\\Boxes}} & \rotatebox{0}{\makecell{Spice-E\\\cite{Sella:2023SpicE}}} & \rotatebox{0}{\makecell{Gated\\3DS2V~\cite{Zhang:2023Shape2Vec}}} & \rotatebox{0}{Ours} & \rotatebox{0}{\makecell{Input\\Boxes}} & \rotatebox{0}{\makecell{Spice-E\\\cite{Sella:2023SpicE}}} & \rotatebox{0}{\makecell{Gated\\3DS2V~\cite{Zhang:2023Shape2Vec}}} & \rotatebox{0}{Ours}  \\ 

\midrule

\multicolumn{4}{c|}{\includegraphics[width=.25\textwidth]{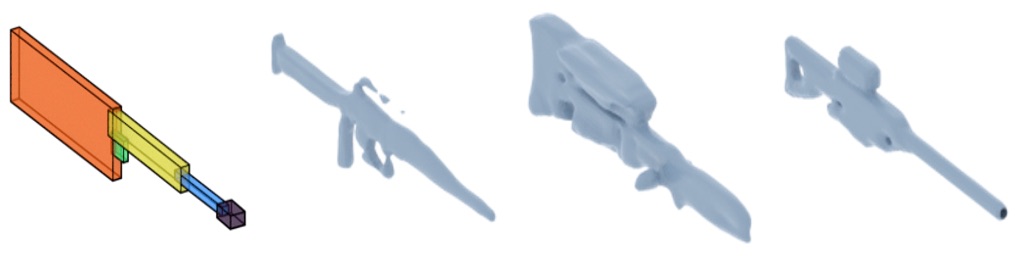}} &
\multicolumn{4}{c|}{\includegraphics[width=.25\textwidth]{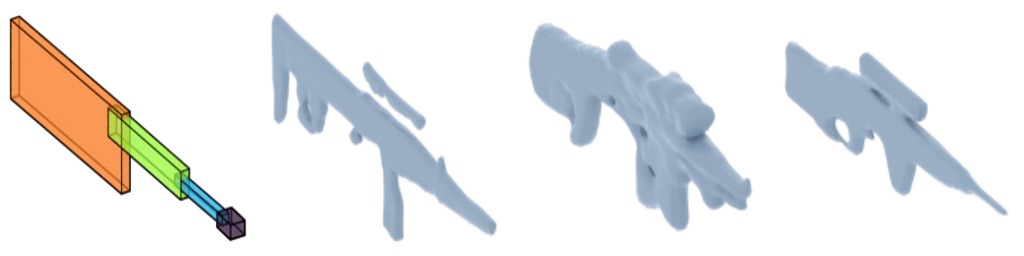}} &
\multicolumn{4}{c|}{\includegraphics[width=.25\textwidth]{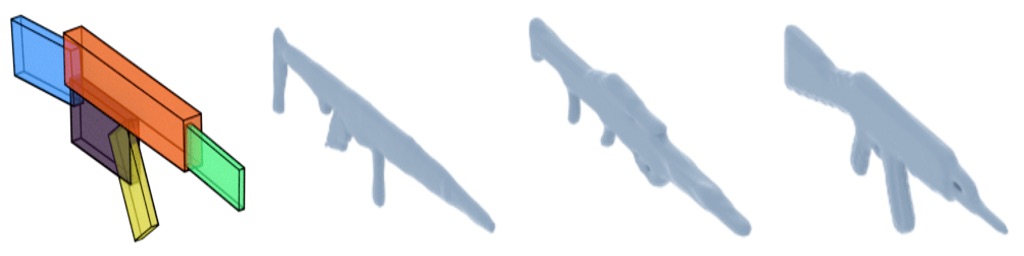}} &
\multicolumn{4}{c}{\includegraphics[width=.25\textwidth]{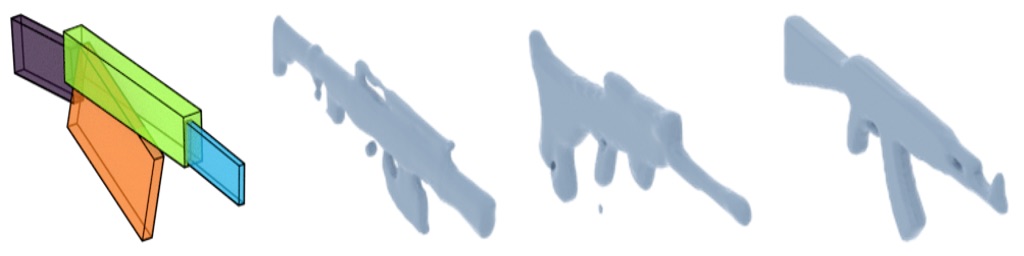}} \\
\multicolumn{4}{c|}{\includegraphics[width=.25\textwidth]{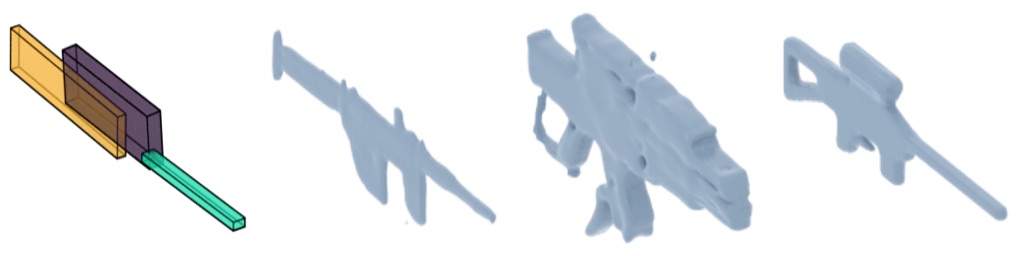}} &
\multicolumn{4}{c|}{\includegraphics[width=.25\textwidth]{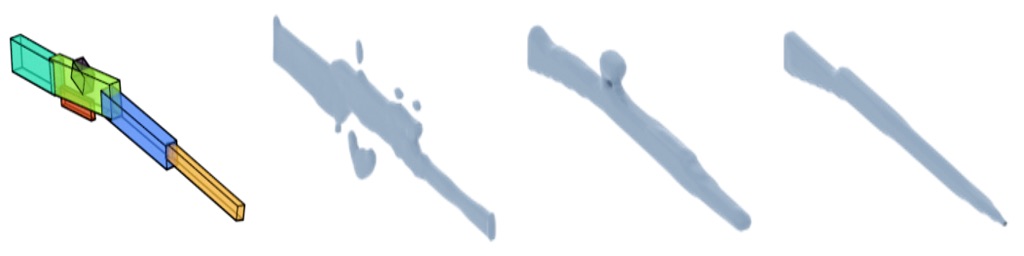}} &
\multicolumn{4}{c|}{\includegraphics[width=.25\textwidth]{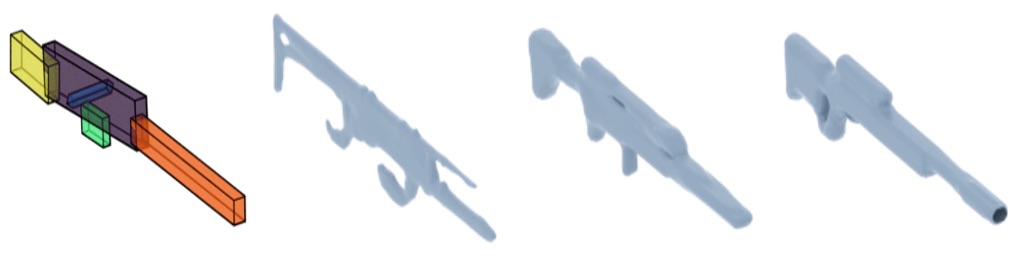}} &
\multicolumn{4}{c}{\includegraphics[width=.25\textwidth]{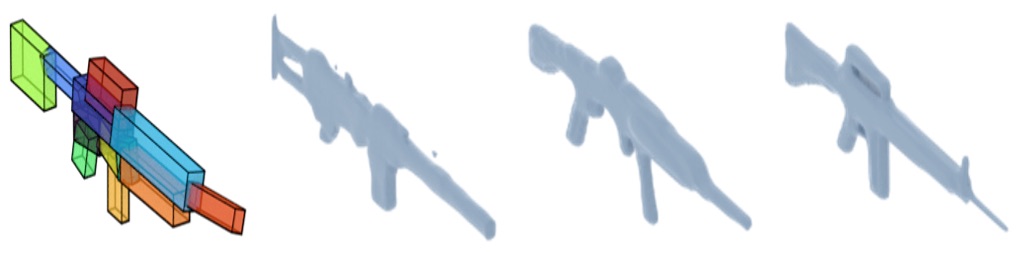}} \\
\multicolumn{4}{c|}{\includegraphics[width=.25\textwidth]{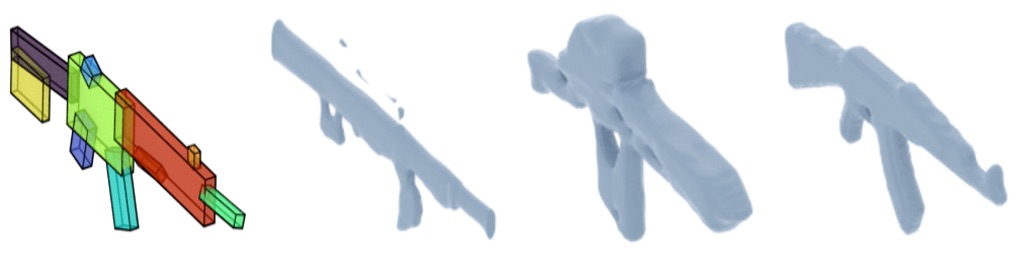}} &
\multicolumn{4}{c|}{\includegraphics[width=.25\textwidth]{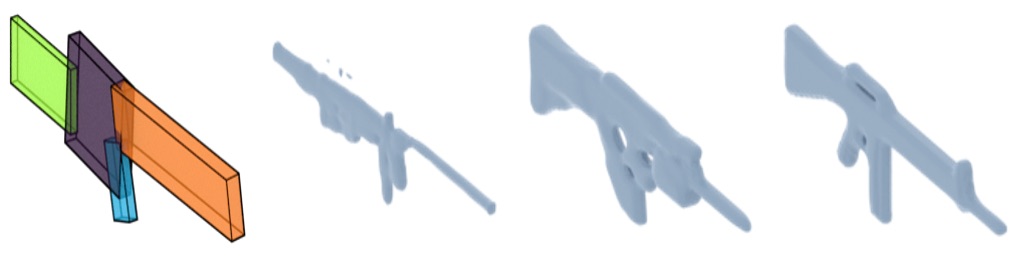}} &
\multicolumn{4}{c|}{\includegraphics[width=.25\textwidth]{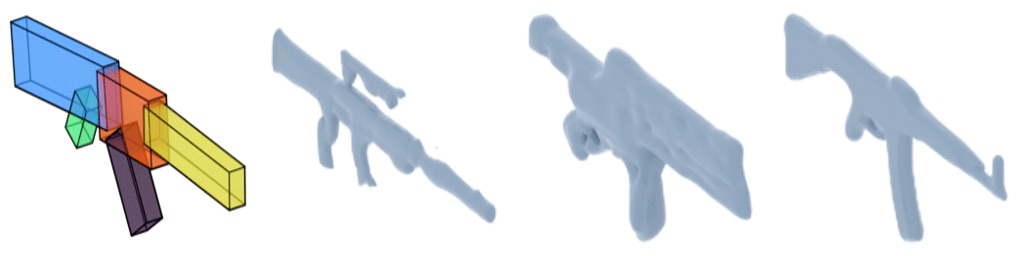}} &
\multicolumn{4}{c}{\includegraphics[width=.25\textwidth]{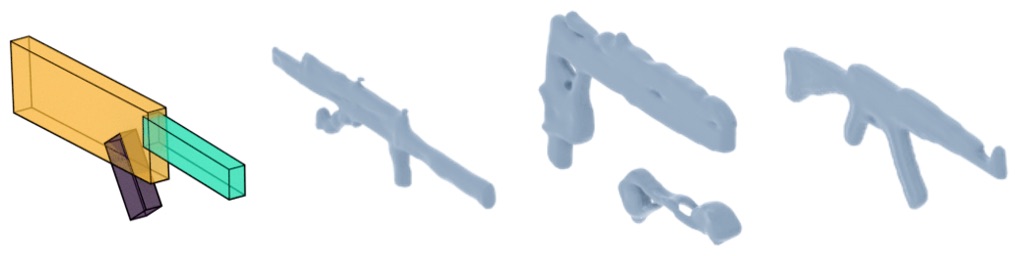}} \\
\multicolumn{4}{c|}{\includegraphics[width=.25\textwidth]{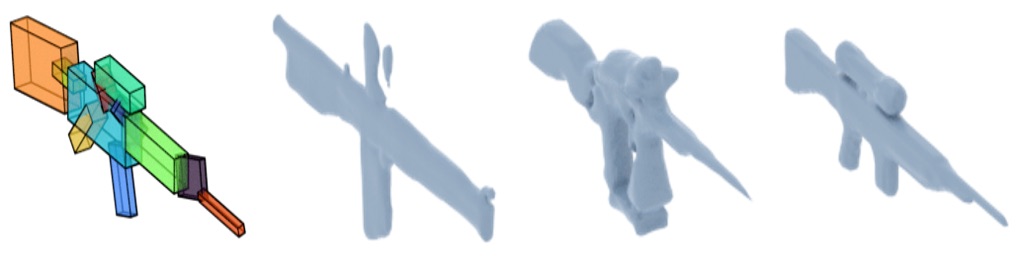}} &
\multicolumn{4}{c|}{\includegraphics[width=.25\textwidth]{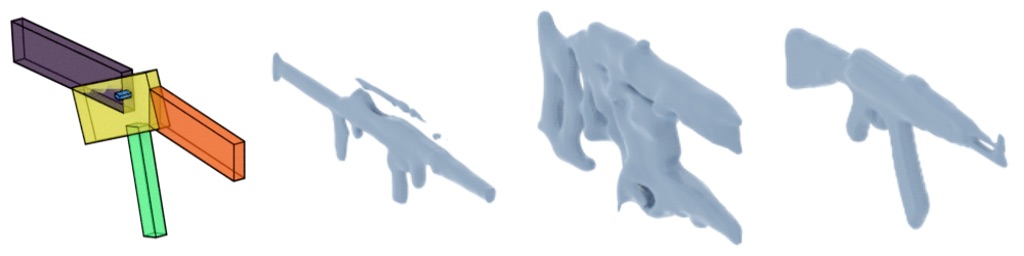}} &
\multicolumn{4}{c|}{\includegraphics[width=.25\textwidth]{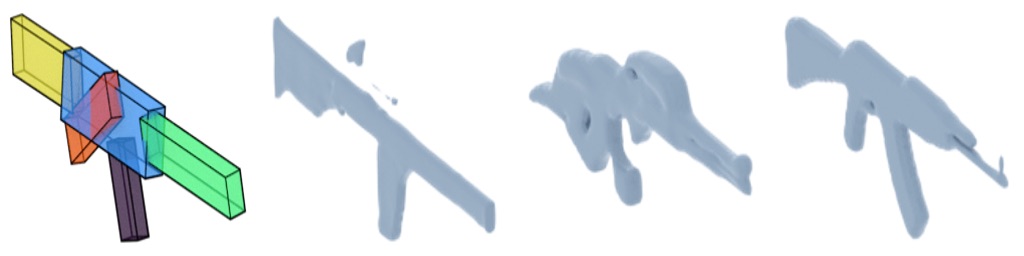}} &
\multicolumn{4}{c}{\includegraphics[width=.25\textwidth]{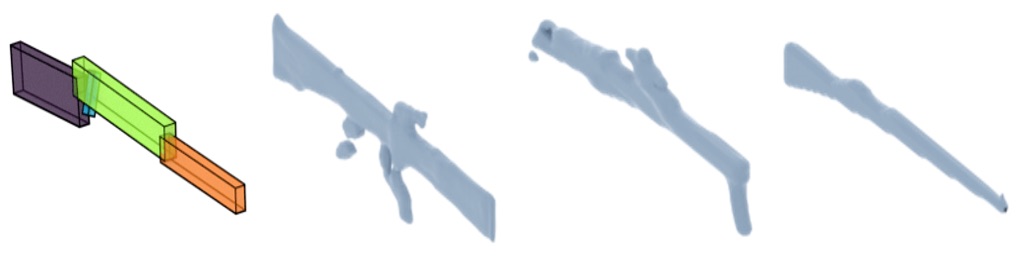}} \\
\multicolumn{4}{c|}{\includegraphics[width=.25\textwidth]{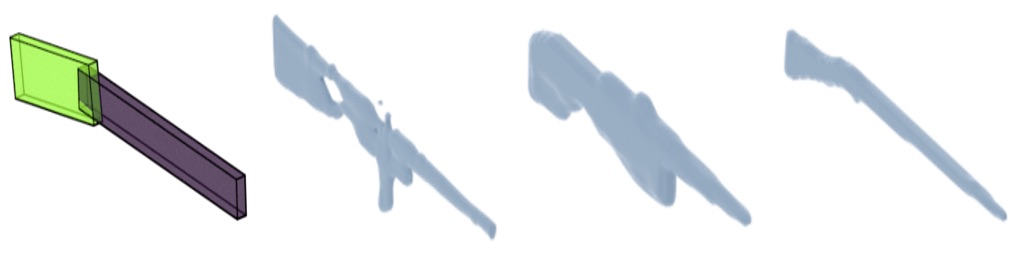}} &
\multicolumn{4}{c|}{\includegraphics[width=.25\textwidth]{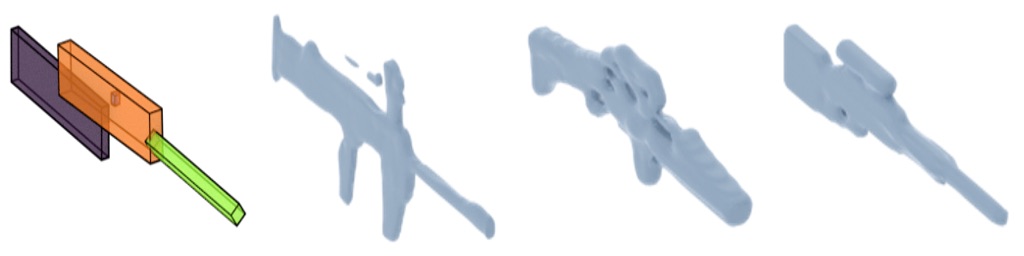}} &
\multicolumn{4}{c|}{\includegraphics[width=.25\textwidth]{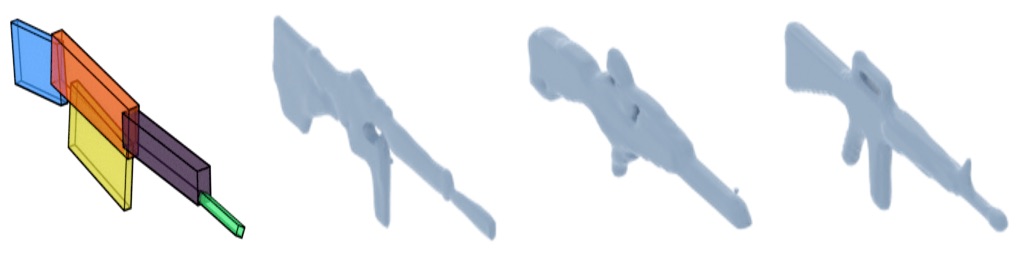}} &
\multicolumn{4}{c}{\includegraphics[width=.25\textwidth]{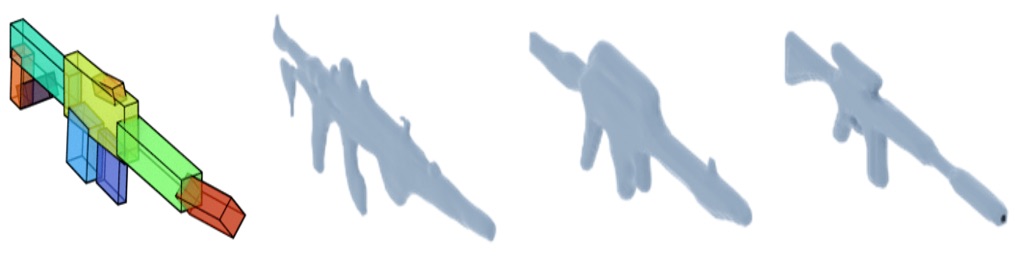}} \\
\multicolumn{4}{c|}{\includegraphics[width=.25\textwidth]{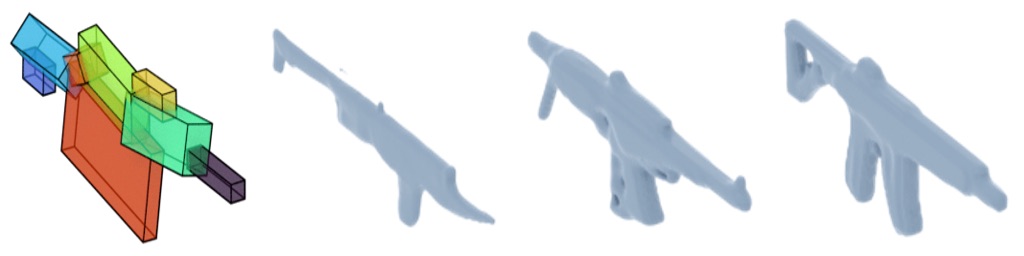}} &
\multicolumn{4}{c|}{\includegraphics[width=.25\textwidth]{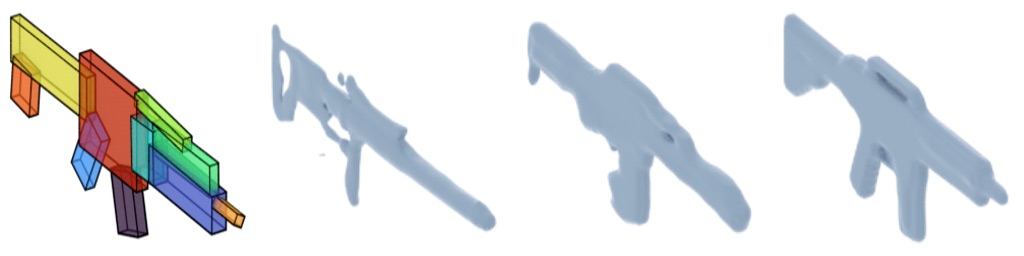}} &
\multicolumn{4}{c|}{\includegraphics[width=.25\textwidth]{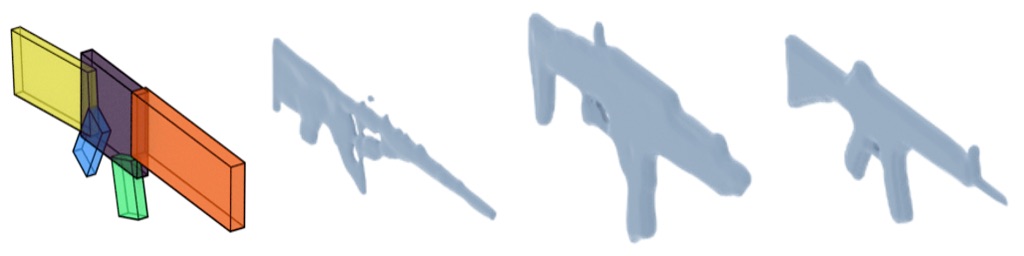}} &
\multicolumn{4}{c}{\includegraphics[width=.25\textwidth]{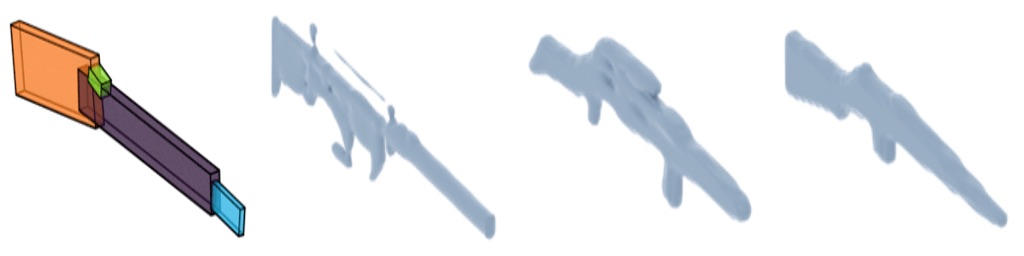}} \\
\multicolumn{4}{c|}{\includegraphics[width=.25\textwidth]{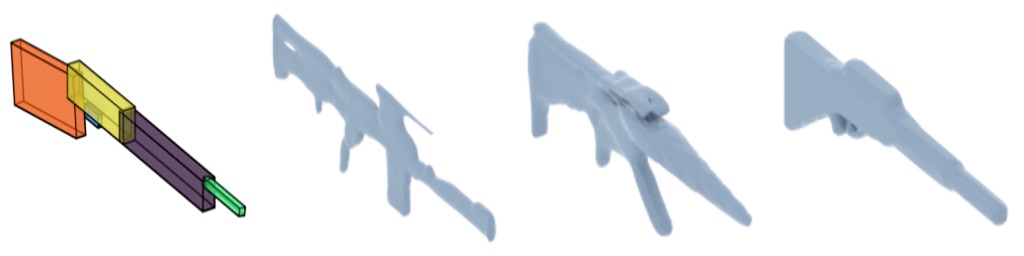}} &
\multicolumn{4}{c|}{\includegraphics[width=.25\textwidth]{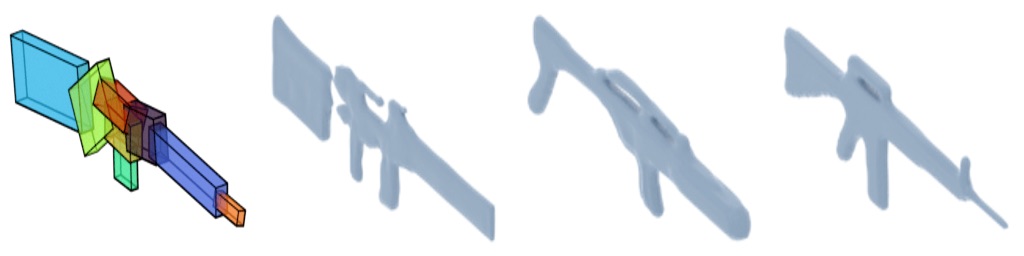}} &
\multicolumn{4}{c|}{\includegraphics[width=.25\textwidth]{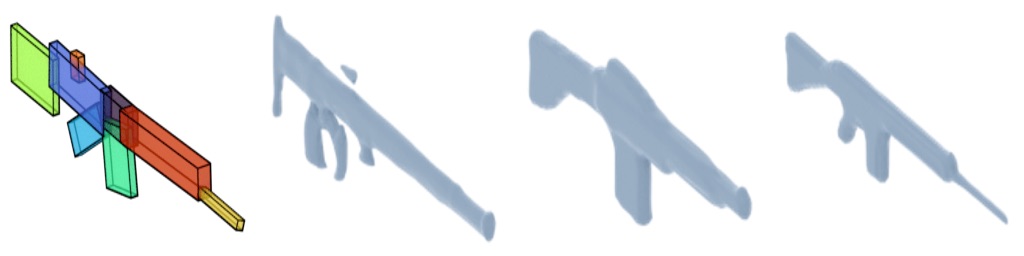}} &
\multicolumn{4}{c}{\includegraphics[width=.25\textwidth]{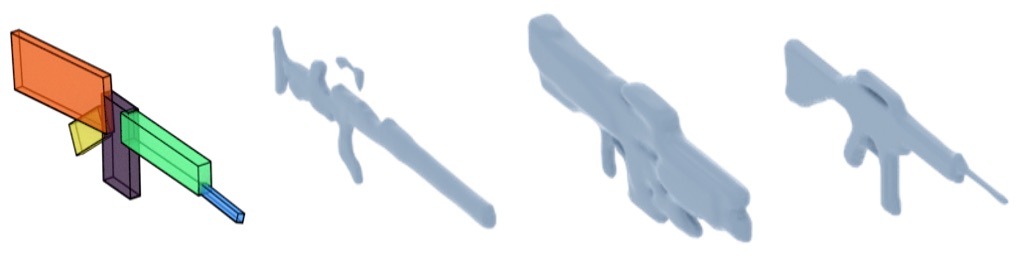}} \\
\multicolumn{4}{c|}{\includegraphics[width=.25\textwidth]{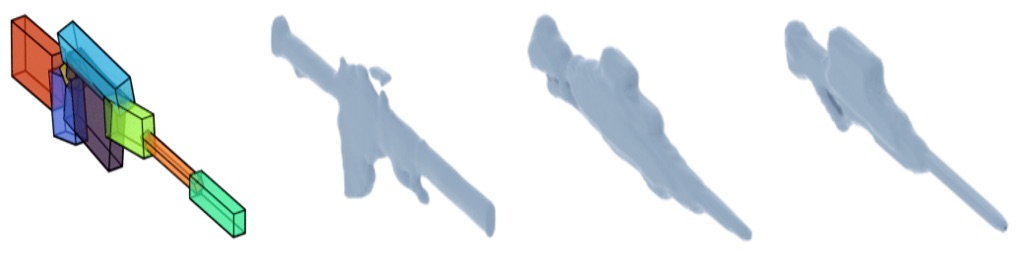}} &
\multicolumn{4}{c|}{\includegraphics[width=.25\textwidth]{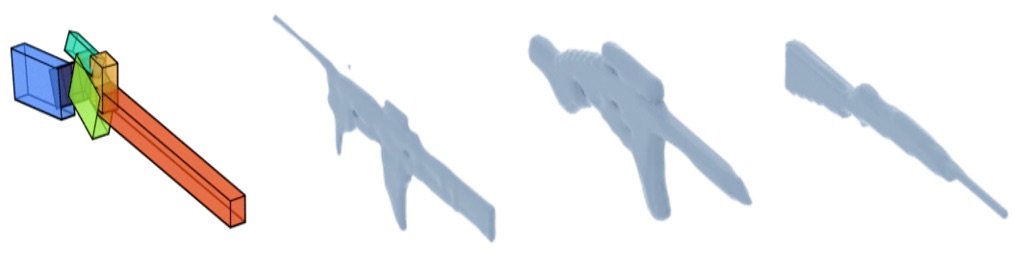}} &
\multicolumn{4}{c|}{\includegraphics[width=.25\textwidth]{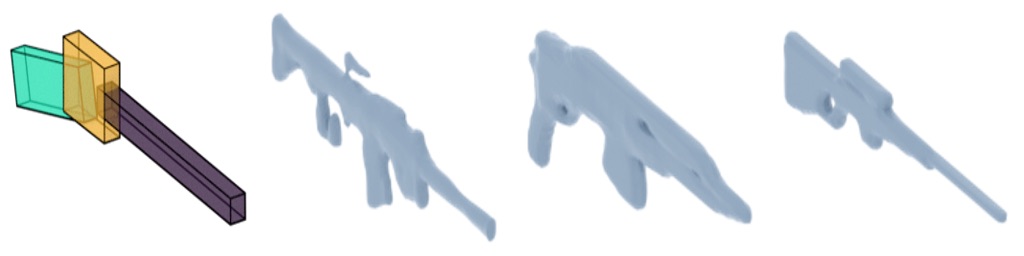}} &
\multicolumn{4}{c}{\includegraphics[width=.25\textwidth]{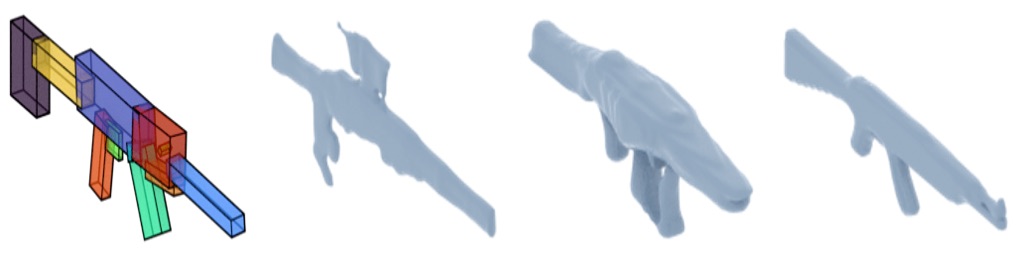}} \\
\multicolumn{4}{c|}{\includegraphics[width=.25\textwidth]{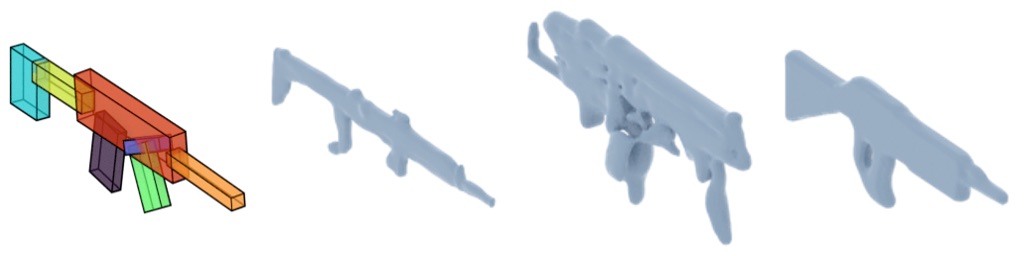}} &
\multicolumn{4}{c|}{\includegraphics[width=.25\textwidth]{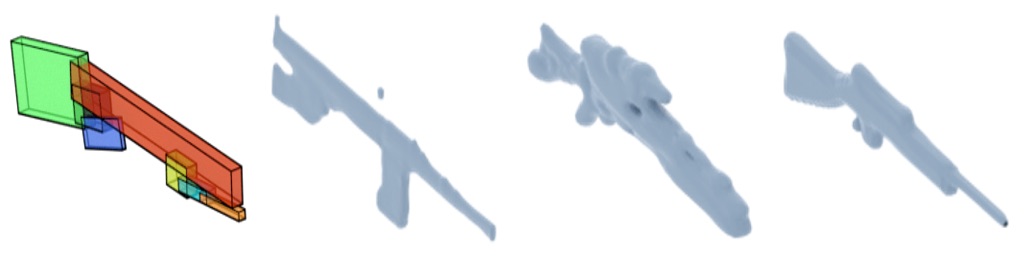}} &
\multicolumn{4}{c|}{\includegraphics[width=.25\textwidth]{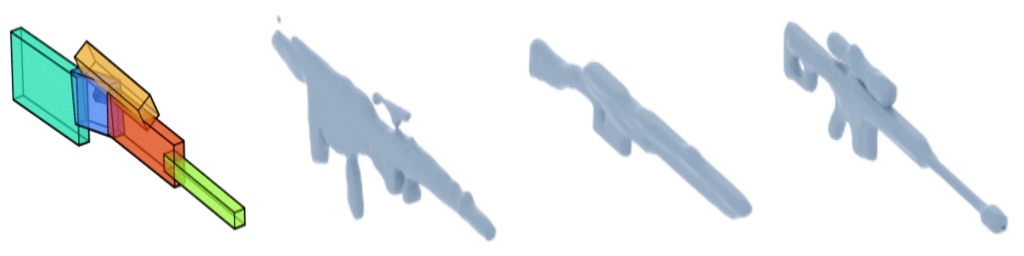}} &
\multicolumn{4}{c}{\includegraphics[width=.25\textwidth]{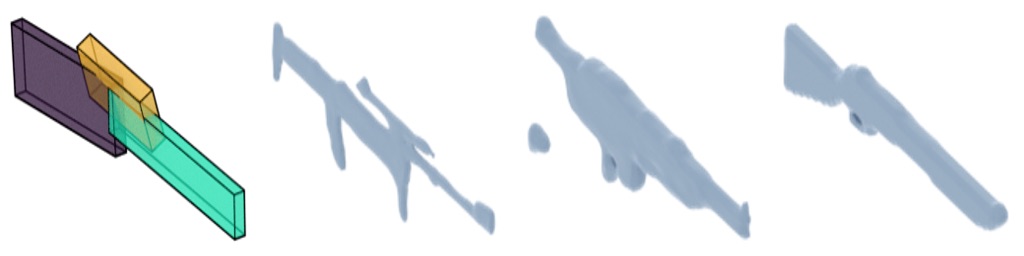}} \\
\multicolumn{4}{c|}{\includegraphics[width=.25\textwidth]{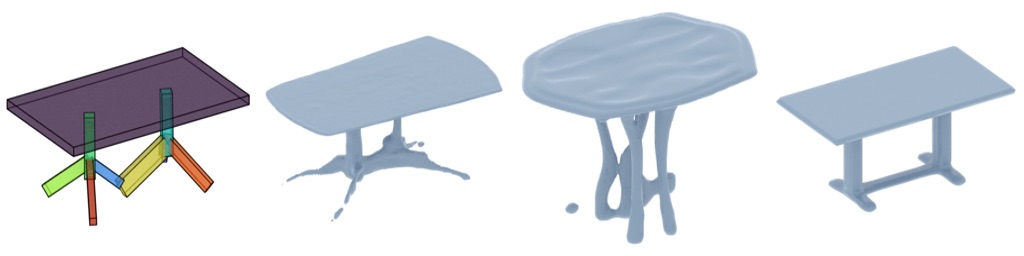}} &
\multicolumn{4}{c|}{\includegraphics[width=.25\textwidth]{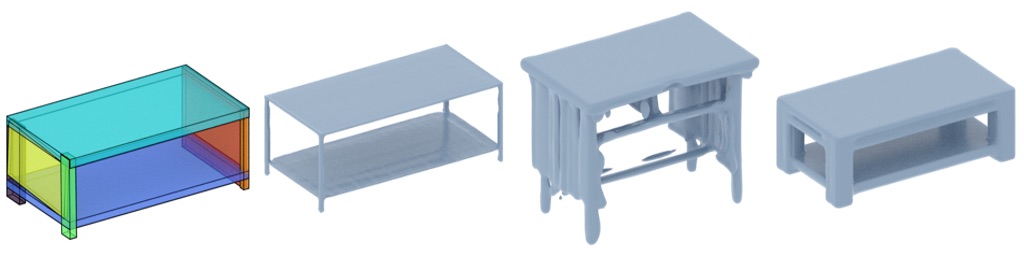}} &
\multicolumn{4}{c|}{\includegraphics[width=.25\textwidth]{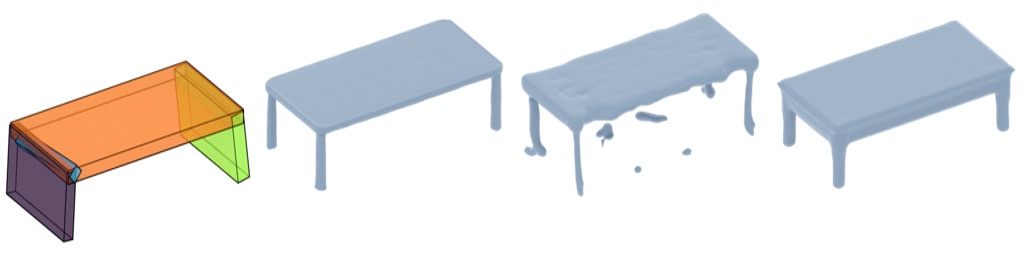}} &
\multicolumn{4}{c}{\includegraphics[width=.25\textwidth]{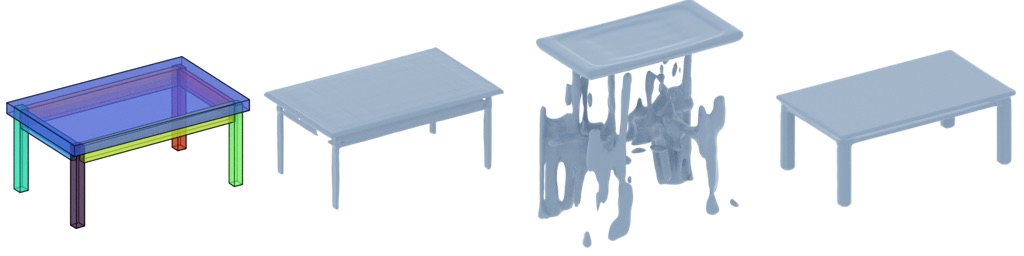}} \\
\multicolumn{4}{c|}{\includegraphics[width=.25\textwidth]{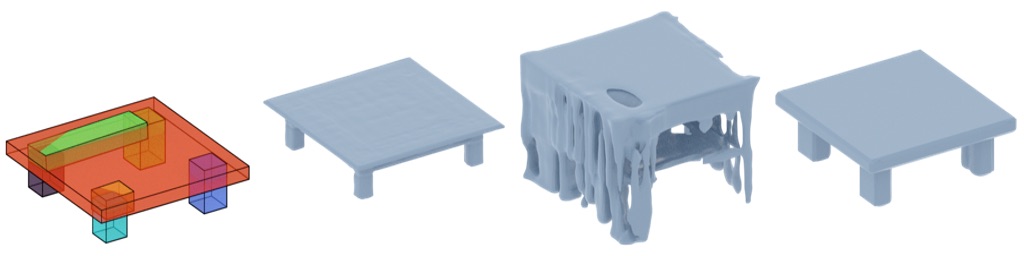}} &
\multicolumn{4}{c|}{\includegraphics[width=.25\textwidth]{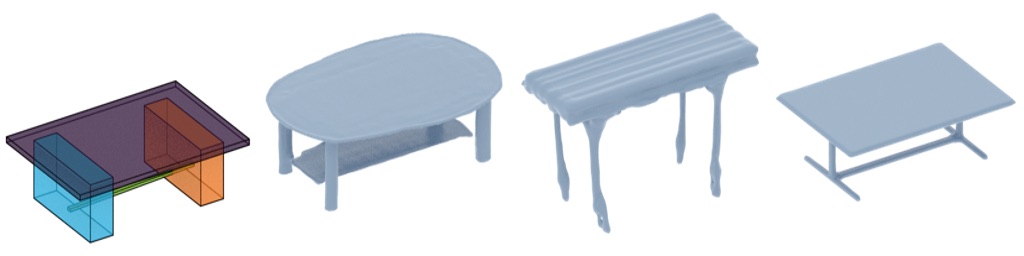}} &
\multicolumn{4}{c|}{\includegraphics[width=.25\textwidth]{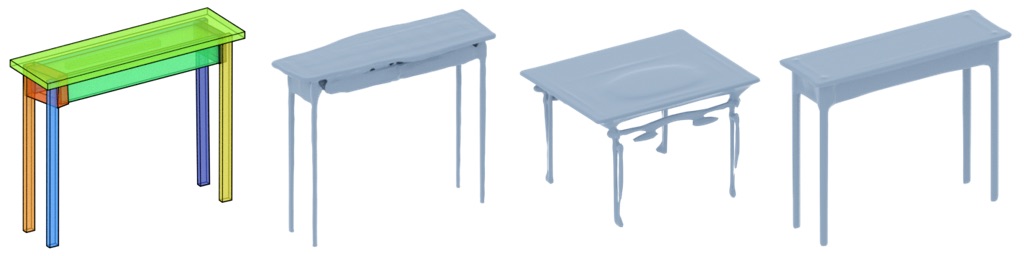}} &
\multicolumn{4}{c}{\includegraphics[width=.25\textwidth]{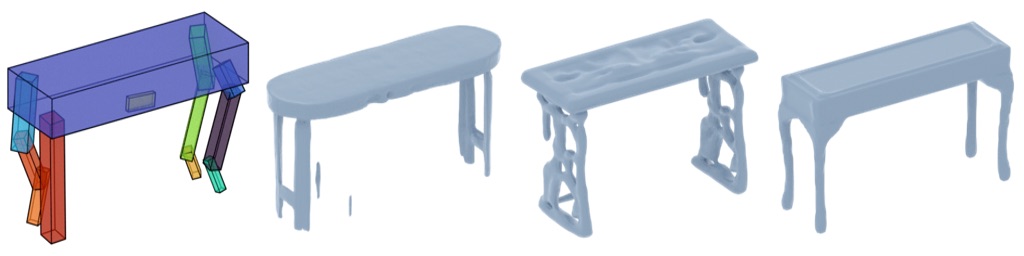}} \\
\multicolumn{4}{c|}{\includegraphics[width=.25\textwidth]{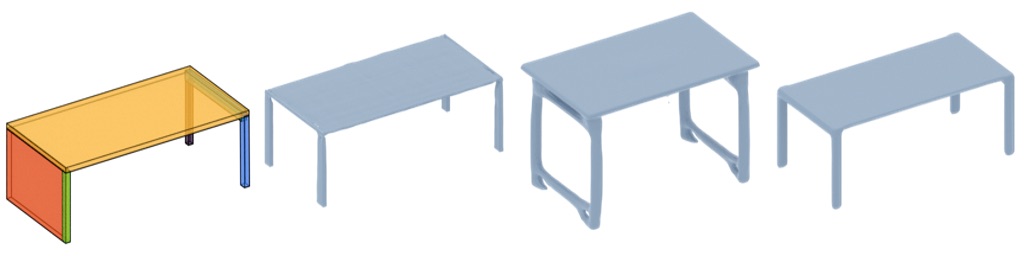}} &
\multicolumn{4}{c|}{\includegraphics[width=.25\textwidth]{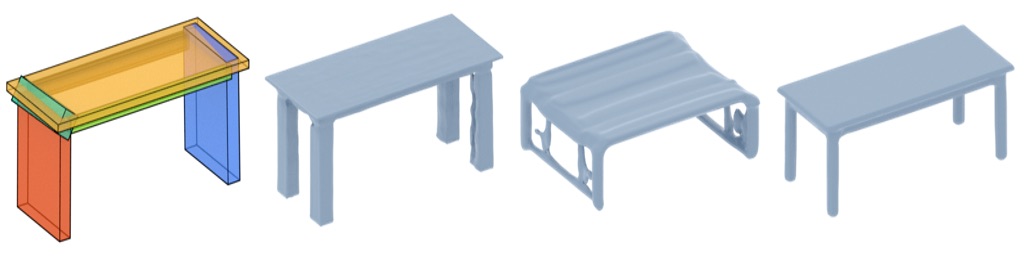}} &
\multicolumn{4}{c|}{\includegraphics[width=.25\textwidth]{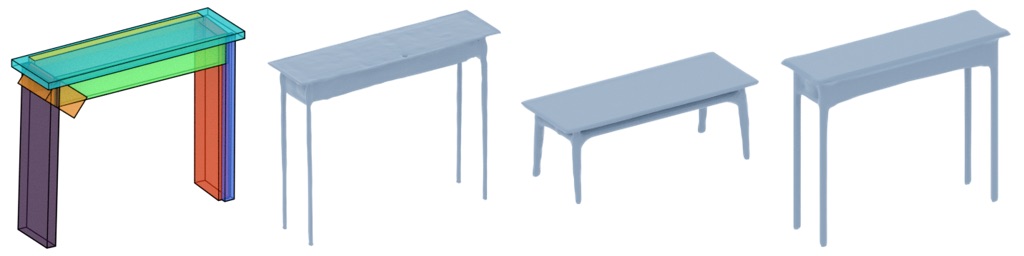}} &
\multicolumn{4}{c}{\includegraphics[width=.25\textwidth]{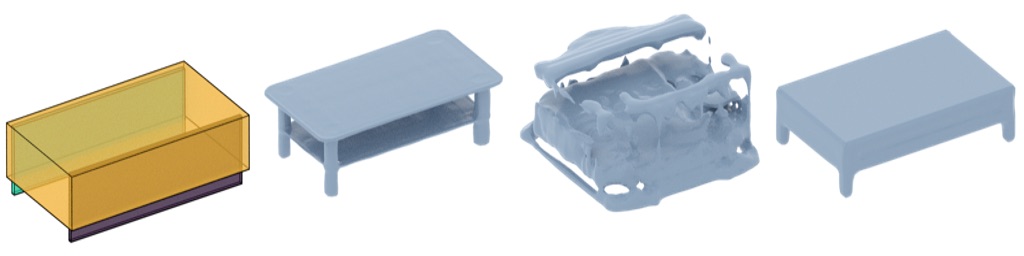}} \\
\multicolumn{4}{c|}{\includegraphics[width=.25\textwidth]{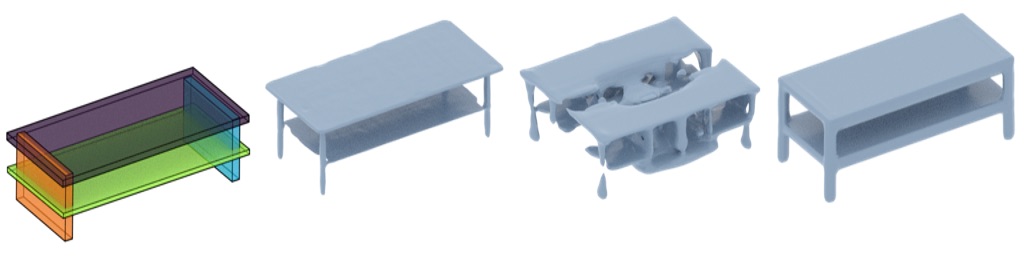}} &
\multicolumn{4}{c|}{\includegraphics[width=.25\textwidth]{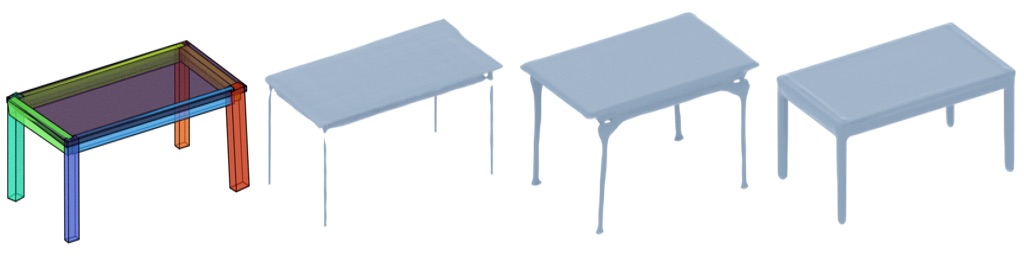}} &
\multicolumn{4}{c|}{\includegraphics[width=.25\textwidth]{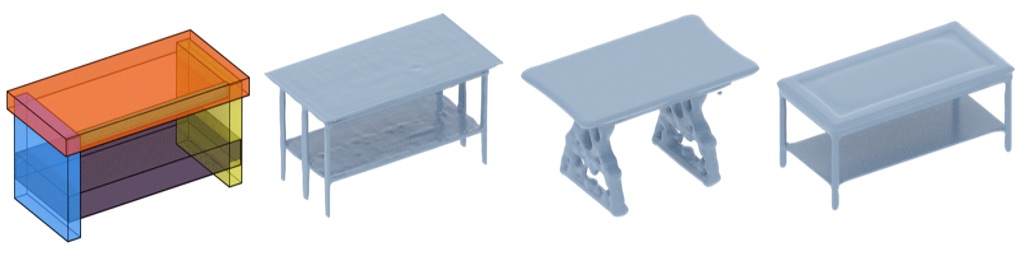}} &
\multicolumn{4}{c}{\includegraphics[width=.25\textwidth]{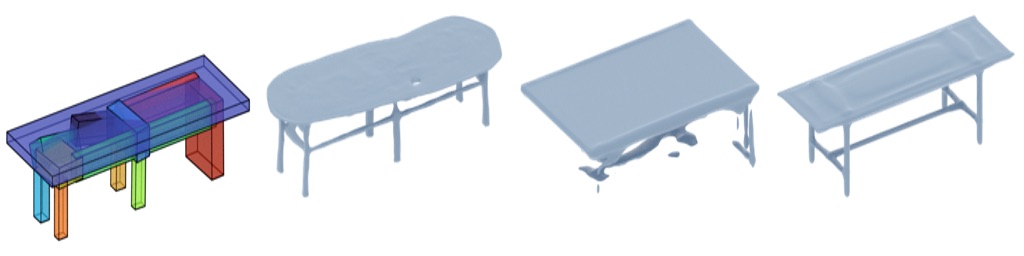}} \\
\multicolumn{4}{c|}{\includegraphics[width=.25\textwidth]{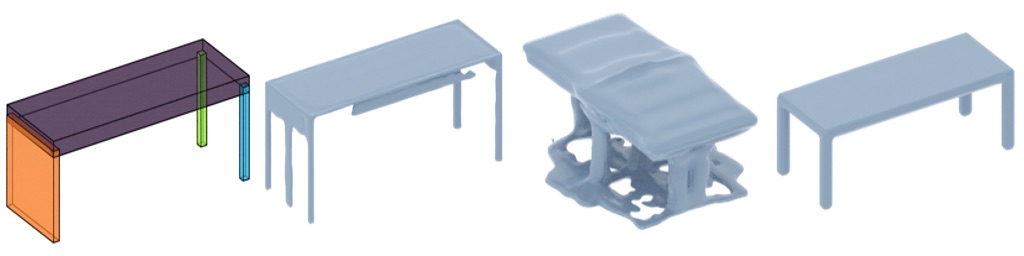}} &
\multicolumn{4}{c|}{\includegraphics[width=.25\textwidth]{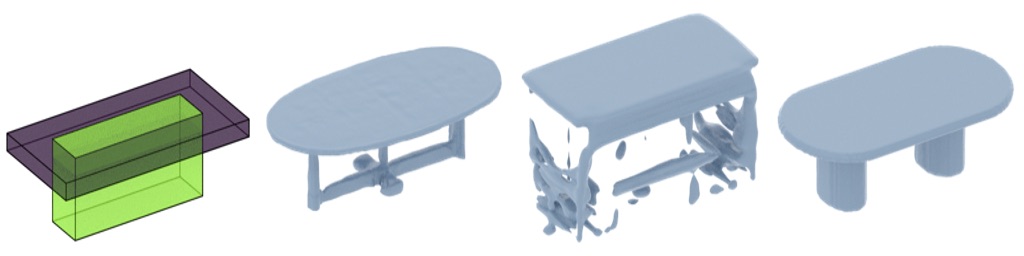}} &
\multicolumn{4}{c|}{\includegraphics[width=.25\textwidth]{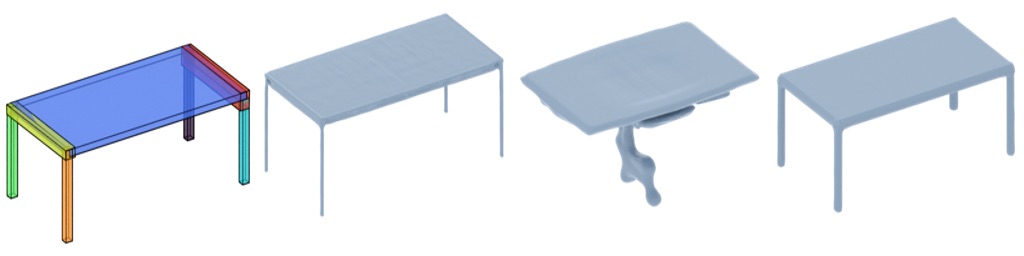}} &
\multicolumn{4}{c}{\includegraphics[width=.25\textwidth]{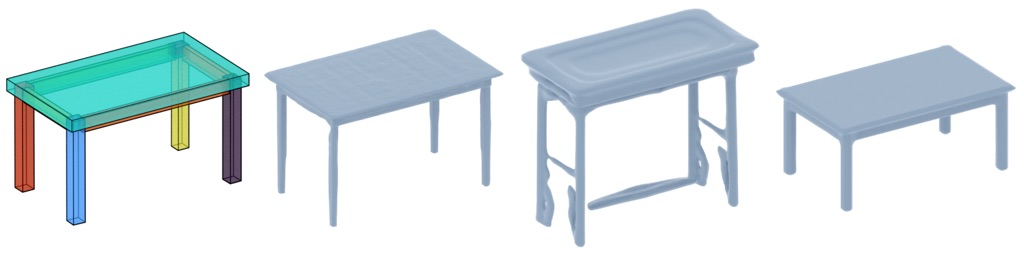}} \\
\multicolumn{4}{c|}{\includegraphics[width=.25\textwidth]{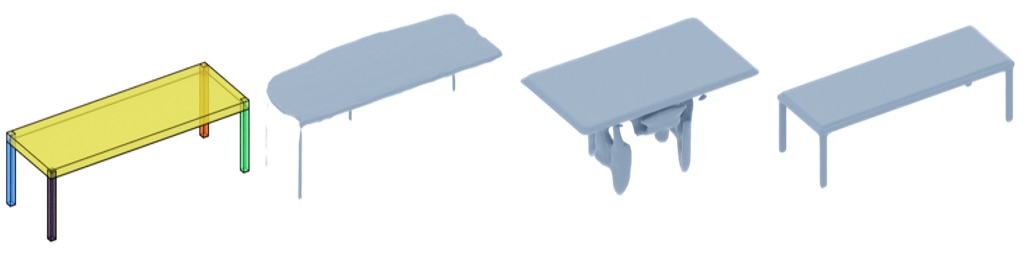}} &
\multicolumn{4}{c|}{\includegraphics[width=.25\textwidth]{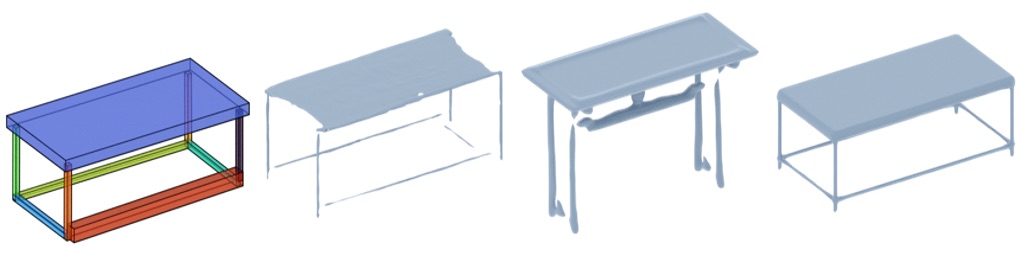}} &
\multicolumn{4}{c|}{\includegraphics[width=.25\textwidth]{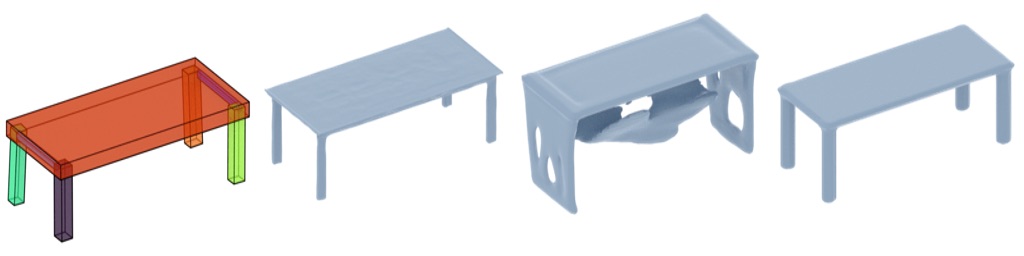}} &
\multicolumn{4}{c}{\includegraphics[width=.25\textwidth]{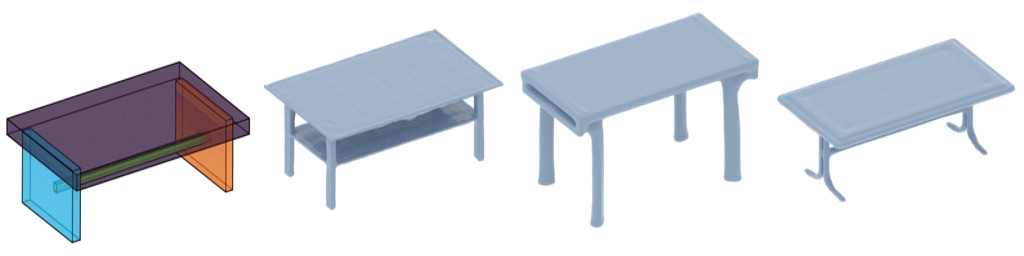}} \\
\multicolumn{4}{c|}{\includegraphics[width=.25\textwidth]{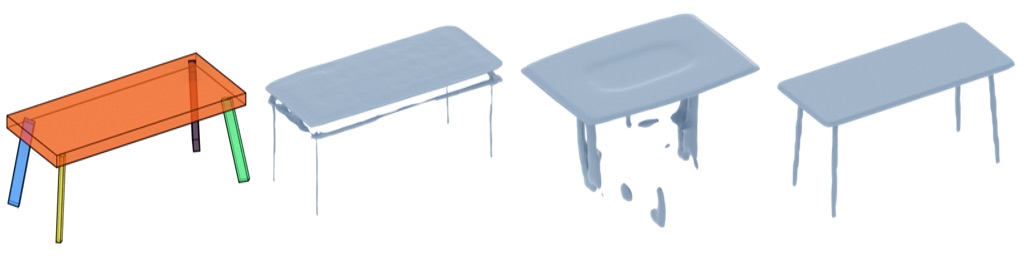}} &
\multicolumn{4}{c|}{\includegraphics[width=.25\textwidth]{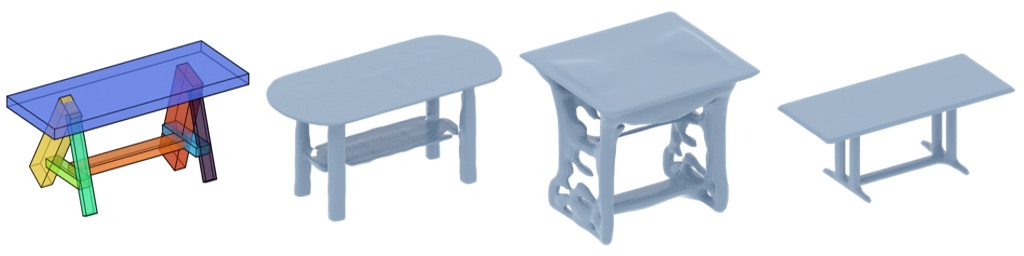}} &
\multicolumn{4}{c|}{\includegraphics[width=.25\textwidth]{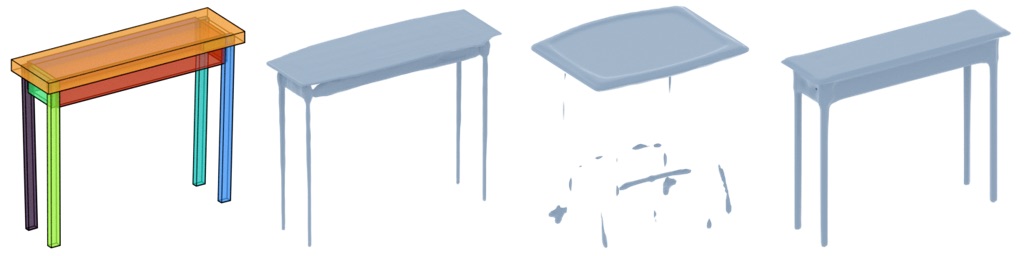}} &
\multicolumn{4}{c}{\includegraphics[width=.25\textwidth]{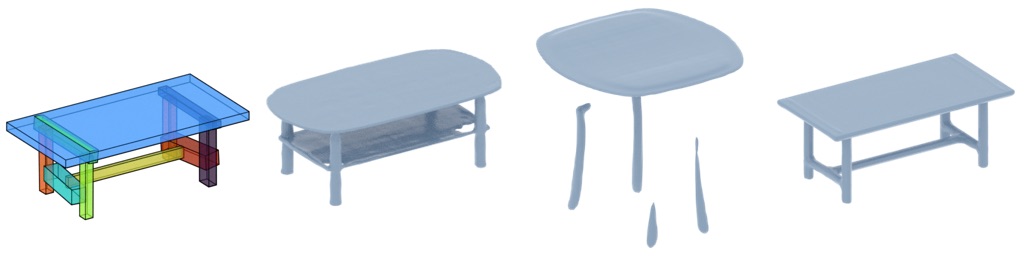}} \\
\multicolumn{4}{c|}{\includegraphics[width=.25\textwidth]{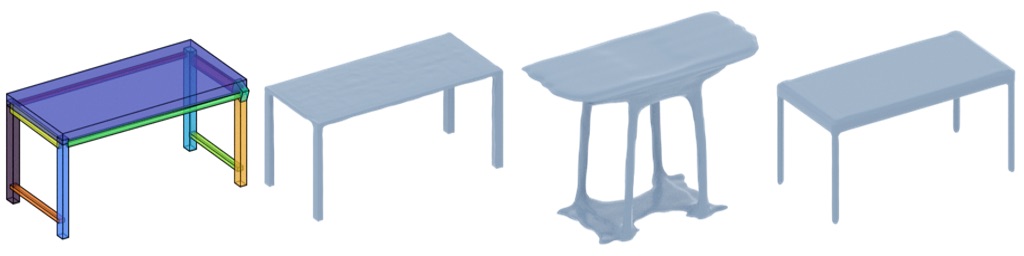}} &
\multicolumn{4}{c|}{\includegraphics[width=.25\textwidth]{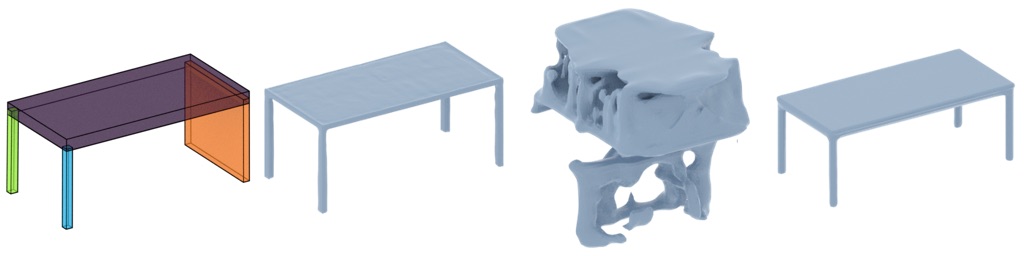}} &
\multicolumn{4}{c|}{\includegraphics[width=.25\textwidth]{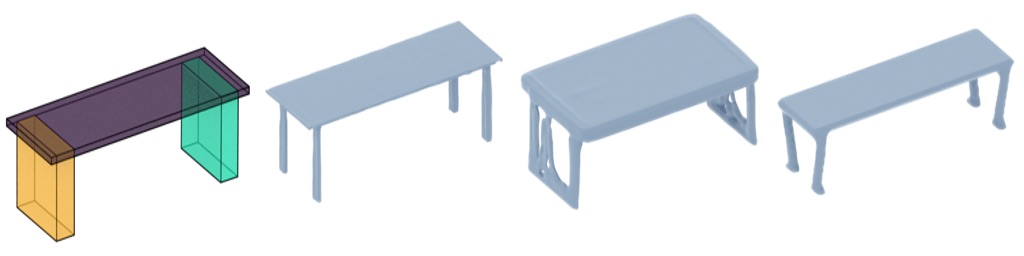}} &
\multicolumn{4}{c}{\includegraphics[width=.25\textwidth]{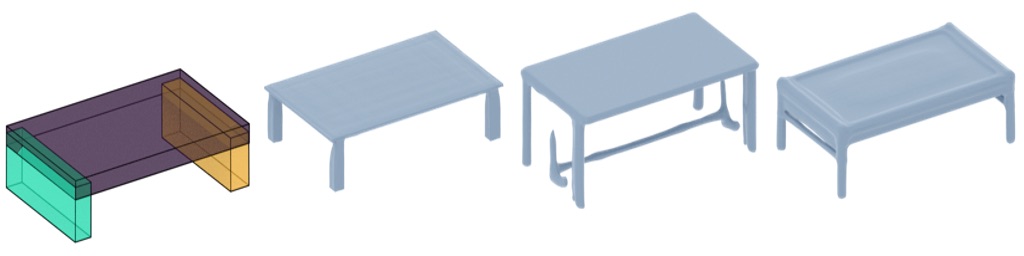}} \\

\end{tabularx}

\caption{\textbf{Gallery of our generated bounding boxes and their final decoded 3D shapes by box-conditioned shape generation network.} Each pair of columns shows the input condition bounding box (left) and its corresponding decoded 3D shape (right).}
\end{figure*}

\endgroup

\end{document}